\documentclass[a4paper]{scrartcl}
       
\usepackage[utf8]{inputenc}
\usepackage[T1]{fontenc}
\usepackage[english]{babel} 
\usepackage{lmodern}
\usepackage[onehalfspacing]{setspace}
\usepackage{amsmath,amssymb,amsthm,amsfonts,amsbsy,latexsym}
\usepackage{graphicx}
\usepackage{natbib}
\usepackage{geometry}
\usepackage{float}
\usepackage[]{algorithm2e}
\usepackage{csquotes}
\usepackage{pgf,tikz}
\usepackage[pdfpagelabels=true]{hyperref}
\usepackage[affil-it]{authblk}

\newcommand{\R}{\mathbb{R}}

\DeclareMathOperator*{\argmin}{argmin}

\DeclareMathOperator*{\SNR}{SNR}

\numberwithin{equation}{section}

\newtheorem{Def}{Definition}[section]

\newtheorem{rem}{Remark}[section]

\begin{document}

\author{Tino Werner}
\affil{Institute for Mathematics, Carl von Ossietzky University Oldenburg, Carl-von-Ossietzky-Straße 9-11, 26129 Oldenburg, Lower Saxony, Germany, \texttt{tino.werner1@uni-oldenburg.de}}
\date{\today}
\title{Double descent for least-squares interpolation on contaminated data: A simulation study} 
\maketitle

\begin{abstract} \begin{small}
Overparametrized models can exhibit an excellent generalization performance, although they should be prone to overfitting according to classical statistical theory. The discovery of the \enquote{double descent}, indicating that the generalization error decreases after a certain model complexity has been reached, opened a new line of research. Robust statistics considers statistical estimation on contaminated data, which, due to assumptions that do not hold on real data, let data points appear as outliers w.r.t. the assumed \enquote{ideal} distribution, potentially severely distorting any classical estimator. We address the question whether a double descent phenomenon can be observed in a linear regression setting with contaminated training data. We compare the performance of the highly non-robust least-squares interpolation estimator with several robust alternatives. It turns out that large overparametrization indeed allows for a double descent phenomenon, resulting in a very good generalization performance of the least-squares interpolator, surpassing that of the robust alternatives. \end{small}
\end{abstract}

\section{Introduction}

Classical statistical theory quantifies the generalization ability of machine learning algorithms in terms of the complexity of the considered model class, quantified by the Rademacher complexity, the Vapnik-Chervonenkis dimension \cite{vapnik71} or the pseudo-dimension \cite{pol90} (e.g., \cite{kolt01, kolt02, kolt06, bart02b, bart05b}). These results motivate to find an optimal complexity of the model class by a bias-variance tradeoff, which decourages too sparse models due to a high bias but also too complex models due to a high variance. As the complexity of neural network classes grows with their number of parameters (\cite{bart03}, \cite{maass}), generalization bounds for large neural networks are very loose and hence cannot explain their often excellent generalization ability. 

In a recent line of research, starting with the empirical work \cite{belkin19} and the theoretical work \cite{belkin}, the \enquote{double descent} phenomenon has been discovered, indicating that the generalization loss of a model class grows with increasing complexity, but drops as the model complexity further increases. While the increase in the generalization loss is referred to as overfitting, the descent behavior has been coined as \enquote{interpolation regime}. While the first works on this topics considered least-squares interpolation  \cite{belkin, neal, muthu20}, results for overparametrized classification \cite{muthu21,dar21} and neural network classifiers \cite{frei,frei23,george,zhu23} followed.

Robust statistics (\cite{huber,hampel,rieder,maronna}) considers the analysis of contaminated data. Such contamination arises from model misspecification, letting realizations from the unknown, real data distribution appear as outliers w.r.t. the assumed \enquote{ideal} distribution. Since classical, non-robust estimators are prone to be distorted when applied on contaminated data, leading to a bad generalization ability, robust statistics provides methods in order to perform estimation on contaminated data while maintaining a good generalization ability, for the price of reduced efficiency. 

In this work, we purely empirically study the generalization performance of overparametrized models that have been trained on contaminated data and are evaluated on clean data. We consider the least-squares interpolator and different robust counterparts, namely regression with the Huber loss, with the Tukey loss, the sparse least trimmed squares (SLTS; \cite{alfons13}) and robust Boosting (RRBoost; \cite{ju21}). We further provide an idea that combines minimum $l_2$-norm interpolation with clean subset selection by first applying SLTS or RRBoost, respectively, and by training a minimum $l_2$-norm interpolator solely the identified clean subset. 

The paper is organized as follows. In Sec. \ref{sec:relwork}, we list related work on the double descent phenomenon. Sec. \ref{sec:prelim} compiles basic concepts of robust statistics, overparametrized regression as well as our approaches where we first use a robust algorithm for identifying a clean subset and apply a minimum $l_2$-norm interpolator on this subset. Sec. \ref{sec:sim} is devoted to a description of our simulation setting, the applied algorithms, $\mathsf{R}$-packages and the evaluation. The results concerning test errors, training errors, coefficients and the number of iterations are graphically presented in Sec. \ref{sec:testerr}, Sec. \ref{sec:trainerr}, Sec. \ref{sec:l1diff} and Sec. \ref{sec:iter}, respectively. We discuss the results and conclude in Sec. \ref{sec:concl}.

\section{Related work} \label{sec:relwork}

The experimental work \cite{belkin19} where the double descent phenomenon was discovered has been the starting point for some theoretical works on least-squares interpolation. In \cite{belkin}, the prediction risk for the least-squares estimator is computed, confirming a theoretical double descent provided a sufficiently high signal-to-noise ratio (SNR). A double descent can also occur w.r.t. the number of epochs in neural network training (\cite{nakkiran}). 

In some works (e.g., \cite{belkin, neal, muthu20}), the test MSE for the minimum $l_2$-norm interpolator in a Gaussian setting is computed, showing that the MSE decreases for $p>n$ with growing $p$. Moreover, the test MSE can be decomposed into a noise-dependent component, which depends on the noise variance and hence vanishes on noise-free data, and a signal-dependent component which depends on the norm of the coefficient vector. Noise-dependent and signal-dependent error bounds have been computed in \cite{bartlett20,hastie22,muthu,dar21,tsigler}.

It has been shown in \cite{dar21,bartlett21,tsigler} that the double descent requires a low-dimensional signal, i.e., it must be aligned with the $s<<n$ highest eigenvalues, as well as a low effective data dimension. Further, it has been shown in \cite{hastie22} that for growing $p$, the $l_2$-norm of the minimum $l_2$-norm solution decreases due to more degrees of freedom, i.e., a larger $p$ can lead to a more regularized solution, reducing the variance. 

A decomposition of the entire model has been considered in \cite{bartlett21} who show that for linear regression,  the minimum-norm interpolant can be decomposed into a prediction (in the subspace of  the first $s$ eigenvectors) and an interpolation component, which they describe as not being useful but also not being harmful for the prediction, more precisely, whose contribution to the test error is small provided that the eigenvalues of the data matrix decay slowly.

Overparametrized classification has been considered for example in \cite{deng22,montanari,kini,chatterji,muthu21}. As for neural networks, the double descent for shallow classification networks has been studied in \cite{frei,frei23,george,karhadkar}, while deep ReLU neural networks have been considered in \cite{zhu23}.

As for noisy or contaminated data, \cite{liu22d} point out that overfitting on contaminated data (label noise) leads to a poor generalization ability. They argue that even robust losses like the absolute error do not solve the problem as over-parametrized models would still interpolate and hence overfit the contamination component. \cite{rahimi24} experimentally study the multiple descent in unsupervised learning and the presence of Gaussian (and hence not heavy-tailed) noise and domain shift, revealing that the multiple descent occurs even in the presence of heavy noise. They use auto-encoders and the double descent occurs both w.r.t. the layer size as well as the epoch number.
\cite{park25} consider overparametrized linear regression in the presence of noise, missing data and $Y$-outliers. They generalize the risk bounds from \cite{bartlett20} in the setting of additive $X$-contamination (from a centered distribution) and cells missing completely at random, and experimentally show the double descent in this setting.
\cite{singh22} express the risk for NNs in terms of influence curves. The resulting theorem \cite[Thm. 5]{singh22} characterizes the MSE in terms of $n$ and $p$, confirming the peak at $p \approx n$ and a decrease as $p>n$. \cite{vilucchio} study regression with the Huber loss and mention in the outlook that one could study the interplay of contamination with benign overfitting. \cite{kausik} consider noisy inputs, i.e., that one only has access to $X+A$ for a noise matrix $A$ that stems from a rotationally bi-invariant distribution whose eigenvalues still allow for the Marchenko-Pastur law. They show that the double descent also occurs for the minimum $l_2$-norm interpolator.  \cite{tripuraneni} consider covariate shifts in the form of a power shift law that can suppress the largest eigendirections. They confirm robustness against such shifts for overparametrized linear models.

\section{Preliminaries} \label{sec:prelim}

\subsection{Robust regression}

Let $n$ and $p$ denote the number of instances and predictors, respectively. In a linear regression setting, let $D=(X,Y) \in \R^{n \times (p+1)}$ be a data set where $Y \in \R^n$ is the response vector and where $X \in \R^{n \times p}$ is the predictor matrix. We assume the linear structure \begin{equation} \label{lsmodel} Y_i=X_i\beta+\epsilon_i \end{equation} for a coefficient vector $\beta \in \R^p$ and for a noise term $\epsilon_i$. In the least-squares setting, one assumes $\epsilon_i \sim \mathcal{N}(0,\sigma^2)$ i.i.d.. 

The least-squares estimator is, in the classical setting where $p<n$, given by \begin{equation}\label{lssol} \hat \beta^{\text{LS}}=\argmin_{\beta \in \R^p}\left(\frac{1}{n}\sum_{i=1}^n (Y_i-X_i\beta)^2\right)=(X^TX)^{-1}X^TY.  \end{equation} This estimator is the maximum likelihood estimator in the Gaussian setting and therefore highly efficient, but also highly non-robust on contaminated data. See \cite[Sec. 4.2]{rieder} for the following definition.

\begin{Def} Let $(\Omega, \mathcal{A})$ be a measurable space and let $\mathcal{P}:=\{P_{\theta} \ \vert \ \theta \in \Theta\}$ be a parametric model, with the parameter space $\Theta \subset \mathbb{R}^p$. Each $P_{\theta}$ denotes a distribution on $(\Omega, \mathcal{A})$. Let $P_{\theta_0}$ be the ideal distribution. The set of all distributions of the form \begin{center} $ \displaystyle \mathcal{U}_c(\theta_0):=\{U_c(\theta_0,r) \ \vert \ r \in [0,\infty[\} $ \end{center} is referred to as convex contamination model, consisting of convex contamination balls \begin{center} $ \displaystyle U_c(\theta_0,r)=\{(1-r)_+P_{\theta_0}+\min(1,r)Q \ | \ Q \in \mathcal{M}_1(\mathcal{A}) \}$. \end{center} Here, $\mathcal{M}_1(\mathcal{A})$ is the set of probability distributions on $\mathcal{A}$. The probability $r$ is called the \enquote{contamination radius}. \end{Def}

Apart from the convex contamination model, other types of contamination models can be found in \cite{rieder}. On contaminated data, the least-squares estimator may become unreliable. Robust statistics quantifies the robustness of an estimator by its breakdown point (BDP). More precisely, the finite-sample BDP (\cite{huber83}) is defined as follows.

\begin{Def} Let $Z_n$ be a data sample consisting of $n$ instances $(X_i,Y_i), i=1,...,n$. For an estimator $\hat \theta \in \Theta$, the \textbf{finite-sample breakdown point} is  \begin{equation} \label{fsbdp} \varepsilon^*(\hat \theta,Z_n)=\min\left\{\frac{m}{n} \ \bigg\vert \ \sup_{Z_n^m}(||\hat \theta(Z_n^m)||)=\infty \right\}, \end{equation} where $Z_n^m$ denotes the set that has exactly $(n-m)$ instances in common with the original sample $Z_n$ and where $\hat \theta(Z_n^m)$ denotes the estimated coefficient on $Z_n^m$. \end{Def} 

The finite-sample BDP has to be interpreted in the sense that by contaminating $m$ out of $n$ instances, one has full control over the estimator, so, in a worst-case setting, can make the norm of the estimator larger than any bound. The least-squares estimator is an M-estimator with the loss function $\rho(r)=r^2$, whose derivative is $\psi(r)=2r$. It is well-known that M-estimators with monotone derivative of the loss function have a BDP of zero, e.g., \cite{maronna}, including the least-squares estimator. 

Several robust linear regression estimators have been proposed in the literature. Classically, one replaces the squared loss function with a loss function whose derivative is bounded. Huber regression optimizes the Huber loss (\cite{huber64}) \begin{center} $ \displaystyle L^{\text{Huber}}_{\delta}(r)=\begin{cases} r^2/2, \ \ \ \vert r\vert \le \delta \\ \delta \vert r\vert-\delta^2/2, \ \ \ \vert r\vert>\delta \end{cases}, $ \end{center} whose derivative is bounded but still monotone. Its BDP is at most 25\%, but can be arbitrarily close to zero, depending on the distribution of the predictors (see \cite[Sec. 6.4]{hampel}). A loss function with an even redescending derivative, i.e., $\lim_{|r| \rightarrow \infty}(\psi(r))=0$, is the Tukey biweight loss function, \begin{center} $ \displaystyle L^{\text{Tukey}}_k(r)=\begin{cases} 1-[1-(r/k)^2]^3, \ \ \ \vert r\vert \le k \\ 1, \ \ \ \vert r\vert>k \end{cases}. $ \end{center} 

Another strategy is followed by the least trimmed squares (LTS) estimator proposed in \cite{rous84}. Instead of optimizing the sum of the squared residuals of all $n$ instances, \cite{rous84} propose to replace the full sum by a truncated sum, i.e., \begin{center} $ \displaystyle \hat \beta^{\text{LTS}}=\argmin_{\beta \in \R^p}\left(\frac{1}{h}\sum_{i=1}^h r(\beta)_{i:n}^2\right), $ \end{center} where $r(\beta)=Y-X\beta$ and where $z_{i:n}$ denotes the $i$-th smallest element of the vector $z$. In other words, one optimizes the squared residuals over a \enquote{clean} subset of cardinality $h$. \cite{rous84} show that its BDP is $\lfloor n/2 \rfloor+\lfloor (p+1)/2\rfloor$, if one selects the size of the clean subset by $h=\lfloor n/2\rfloor+1$.

\subsection{The case $p>n$}

For the overparametrized setting, i.e., $p>n$, due to the singularity of $X^TX$, the inverse in the least-squares solution in \ref{lssol} is replaced by the Moore-Penrose inverse, leading to the closed-form solution $X^+Y$.  

As for the robust counterparts that optimize the Huber or the Tukey loss, due to the lack of a closed-form representation of the solution, one usually applies the iterative reweighted least-squares algorithm (IRWLS, \cite{huber}). A weighted least-squares estimator is given by  \begin{equation}\label{wlssol} \hat \beta^{{WLS}}=\argmin_{\beta \in \R^p}\left(\frac{1}{n}\sum_{i=1}^n w_i(Y_i-X_i\beta)^2\right)=(X^TWX)^{-1}X^TWY,  \end{equation} where $W$ is a diagonal matrix with the $i$-th diagonal element $w_i$. The weights $w_i$ satisfy $w_i \ge 0$ and $\sum_i w_i=1$. For $p>n$, the weighted least-squares solution is replaced by $(\tilde X)^+Y$, where $\tilde X=W^{1/2}X$. Therefore, initializing the IRWLS scheme with the least-squares estimator reduces any IRWLS procedure to minimum $l_2$-norm interpolation, since the least-squares estimator already interpolates the data. 

As a consequence, in our simulations, it was not possible to use the implementation of the IRWLS estimators that minimize the Huber loss and the Tukey loss, respectively, provided in the $\mathsf{R}$-package \texttt{robustreg}. Therefore, we use the $\mathsf{R}$-package \texttt{MTE} (\cite{qin17}, \cite{MTE}) where a given loss function is optimized by gradient descent. We increased the maximum number of iterations to 100. 

The computation of the LTS estimator is done by iteratively updating the clean subset and by computing the least squares estimator on this updated set. For $p>hn$, one faces again the problem that the fit is perfect on the clean subset. Optimizing another loss function such as the Huber or Tukey loss function on the clean subsets would not be meaningful as it would correspond to a double robustification, both in the sense of replacing the squared loss with a loss function with bounded derivative and trimming, which would at least considerably reduce the efficiency of the estimator and increase the computational costs. In addition, we observe in our experiments that at least the Huber M-estimator interpolates the training data once $p$ is sufficiently large, rendering any identification of a clean subset meaningless. 

We propose to apply a robust regression technique that allows for data with $p>n$ while still optimizing the least-squares loss and decide to consider two candidates: The sparse LTS (\cite{alfons13}), provided in the $\mathsf{R}$-package \texttt{robustHD} (\cite{alfons16}), and the robust Boosting algorithm RRBoost from \cite{ju21}, provided in the $\mathsf{R}$-package \texttt{RRBoost} ( \cite{rrboost}). In order to perform minimum $l_2$-interpolation on a clean subset, we apply either the sparse LTS or the robust Boosting algorithm on the whole training set and identify the clean subset based on the squared residuals. On this clean subset, we compute the minimum $l_2$-norm interpolator.

\section{Simulation} \label{sec:sim}

\subsection{Data generation}

In this paper, we solely consider linear regression. Motivated by the experiments in \cite{kobak} and \cite{muthu20}, we use a sparse true coefficient vector and generate the predictors according to a Gaussian distribution, either with independent features or with a spiked covariance matrix.

The predictors $X_i$ are i.i.d. realizations from a multivariate normal distribution $\mathcal{N}_p(\mu \cdot 1_p,\Sigma)$, where $1_p$ denotes the vector of length $p$ that only consists of ones, where $\mu \in \R$ and where $\Sigma$ is some covariance matrix. In the independent case, we set $\Sigma=I_{p \times p}$, denoting the identity matrix of dimension $p \times p$. In the spiked covariance case, we set $\Sigma=I_{p \times p}+\rho 1_p1_p^T$. We always use $\rho=0.25$. 

The true coefficient vector $\beta$ is generated randomly, either with Gaussian components, i.e., $\beta \sim \mathcal{N}_p(0_p,I_{p \times p})$, or with uniformly distributed components, i.e., $\beta_j \sim U([1,2])$ i.i.d.. In our simulations, in order to generate the test loss curve w.r.t. $p$, we use different $p$ but always let the true number of predictors be $s=20$. For $p>s$, $p-s$ components of $\beta$ are randomly selected and set to zero. For $p<s$, we generate a data set where $X$ has $s$ columns, generate the responses from it as well as the noise vector, and randomly select only $p$ predictor columns as input for the regression algorithm.

We first generate the responses $Y_i=X_i\beta$ and add a Gaussian noise vector $\epsilon \sim \mathcal{N}_n(0_n,\sigma^2 I_{n \times n})$ where we set $\sigma$ such that on the generated sample, a specified signal-to-noise ratio is maintained. 

In order to generate contaminated data, we distinguish between $Y$-contamination where only the response vector is contaminated, and $X$-contamination, where only the predictor matrix $X$ is contaminated. In both cases, the contamination is injected after having computed $Y=X\beta+\epsilon$, so the signal-to-noise ratio corresponds to the clean data. In the case of $Y$-contamination, we specify a contamination radius, $r$, and randomly select $\lfloor rn \rfloor$ components of $Y$. We consider additive contamination and add a fixed value of $c_{out}$ to each of the selected components. In the case of $X$-contamination, we randomly select $\lfloor rn \rfloor$ rows of the predictor matrix. We then add a fixed value of $c_{out}$ to $\lfloor 0.1p\rfloor$ randomly selected cells in the respective rows, where the selection of the column indices is done individually for each row. 

\begin{rem} Note that, although the rows are not fully contaminated, the contamination scheme is covered by case-wise $X$-contamination by limiting the number of contaminated rows. We do not need to consider cell-wise robust approaches that would be necessary in cell-wise contamination (\cite{alqallaf}) where the contamination radius refers to the fraction of contaminated cells, allowing for more than 50\% of the rows to be contaminated. \end{rem}

We always generate a data set of size $n=n_{train}+n_{test}$ and split it into a training set of size $n_{train}$ and a test set of size $n_{test}$. In this paper, we always consider clean test data, i.e., the $Y$- or $X$-contamination is injected only to the training set. For example, although contaminated test data have been considered in \cite{TW24} and are undeniably important when working with real data, the purpose of this simulation study is to assess whether overparametrized regression can generalize to unseen clean data, even if the training data were contaminated.

\subsection{Algorithms}

The minimum $l_2$-norm interpolator is computed via the formula $\hat \beta^{LS}=X^+Y$. The Moore-Penrose pseudo-inverse $X^+$ is computed using the function \texttt{ginv} from the $\mathsf{R}$-package \texttt{MASS} (\cite{venables}).

When optimizing the Huber loss or the Tukey loss, we use the gradient descent algorithm from the $\mathsf{R}$-package \texttt{MTE} (\cite{qin17,MTE}), more precisely, the function \texttt{huber.reg} in the implementation (in the version from April 9, 2023) provided in the Github repository \url{https://github.com/shaobo-li/MTE}. Here, we allow for 100 iterations at most. Otherwise, the algorithm terminates once the difference of the current and the previous coefficient vector, quantified in the $||\cdot ||_{\infty}$-norm, is smaller than $10^{-4}$. The Huber loss is already implemented in this package. We use the hyperparameter $\delta=1.345$. As for the Tukey loss, we implemented it ourselves and just apply the gradient iterations to this loss function. Here, we use $k=4.685$. These choices for $\delta$ and $k$ correspond to an asymptotic efficiency of 95\% in the Gaussian setting.

The sparse LTS (SLTS) is implemented in the $\mathsf{R}$-package \texttt{robustHD} (\cite{alfons16}). We use the function \texttt{sparseLTS} in the default settings except for \texttt{alpha} which we set to 0.5, corresponding to a clean subset of size $h=\lfloor n/2\rfloor$. We use the clean subset identified by SLTS in order to compute the minimum $l_2$-norm interpolator on this subset. In addition, we also report the performance of the SLTS model itself.

The robust Boosting algorithm from \cite{ju21} is implemented in the $\mathsf{R}$-package \texttt{RRBoost} (\cite{rrboost}). We use the function \texttt{Boost} with default settings. Similarly as for SLTS, we use RRBoost in order to identify a clean subset, but also report the performance of RRBoost itself.

\subsection{Evaluation}

We consider several scenarios with different values for $n_{train}$, $p$, $\mu$, the signal-to-noise ratio, $r$ and the additive perturbation value $c_{out}$. Moreover, we distinguish between clean data, $Y$-contaminated and $X$-contaminated training data. 

As for the number of predictors, we use each $p$ from the set \begin{equation} \label{pset} \begin{split}\{5,10,20,30,40,50,60,80,100,150,200,250,300,400,500,\\ 750,1000,1250,1500,1750,2000,3000,4000,5000\}. \end{split} \end{equation} When applying the robust Boosting algorithm, where we only consider values up to $1250$ for $p$ for numerical reasons. 

\begin{table}
\begin{center}
\begin{small}
\begin{tabular}{|p{1.5cm}|p{1cm}|p{1cm}|p{2cm}|p{0.5cm}|p{2.5cm}|p{1cm}|p{2cm}|p{1cm}|} \hline 
$p$ & $n_{train}$ & $n_{test}$ & $\SNR$ & $\mu$ & $\beta$ & Cont. & $r$ & $c_{out}$ \\ \hline
Eq. \ref{pset} & 50 & 50 & $\{0.1,0.5,2,5\}$  & 0 & Normal/uniform & None, $X$, $Y$ & $\{0.1,0.25,0.5\}$ & 100  \\ \hline
\end{tabular}
\end{small}
\caption[Data generation specification]{Basic scenario specifications} \label{scentable}
\end{center}
\end{table}

Apart from the basic scenarios, specified in Tab. \ref{scentable}, where we always use both the independent design and the spiked covariance design, we also consider some alternative scenarios. For the independent design, we set $\mu=5$, so the predictors are not centered, and apply the minimum $l_2$-norm interpolator, Huber-based interpolation and SLTS-based interpolation. Moreover, we increase $n_{train}$ and $n_{test}$ to 200 and consider $Y$-contamination with $r \in \{0.1,0.25,0.5,0.75,0.9\}$. Here, we apply minimum $l_2$-norm interpolation and Huber-loss interpolation. For computational reasons, the maximum $p$ is $4000$. Finally, we consider the same scenarios but with $c_{out}=10000$ and again apply these two algorithms.

We repeat all simulations $B=500$ times for each scenario and compute the mean test MSE. We then plot the mean training MSE, the mean test MSE and the $l_1$-norm differences of the estimated and true coefficients, i.e., $||\hat \beta-\beta||_1$, against the number $p$ of predictors. For the iterative gradient-based Huber- and Tukey-loss minimization algorithms, we also visualize the number of iterations in dependence of $p$.

\newpage

\section{Test errors} \label{sec:testerr}

\subsection{Independent features, $\mu=0$}

\subsubsection{Minimum $l_2$-norm interpolation}

\begin{figure}[H]
\begin{center}
\includegraphics[width=7.5cm]{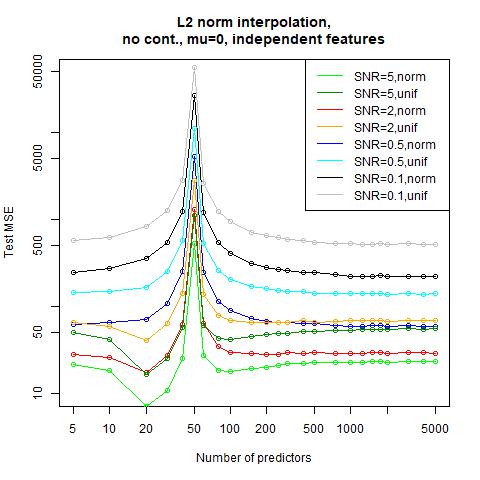} \\ 
\end{center}
\caption{Test MSE of minimum $l_2$-norm interpolation when trained on clean training data.} \label{fig:minl2mu0indep}
\end{figure}

\begin{figure}[H]
\begin{center}
\includegraphics[width=5.25cm]{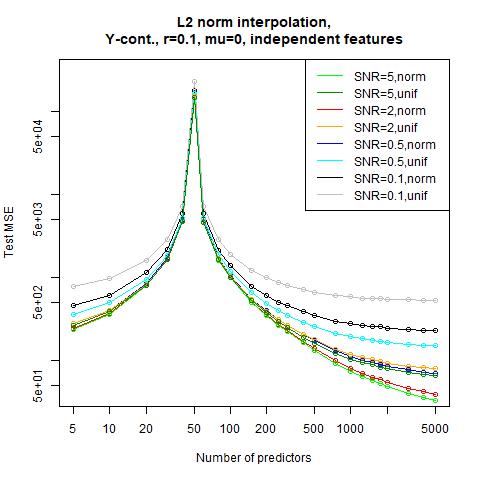} 
\includegraphics[width=5.25cm]{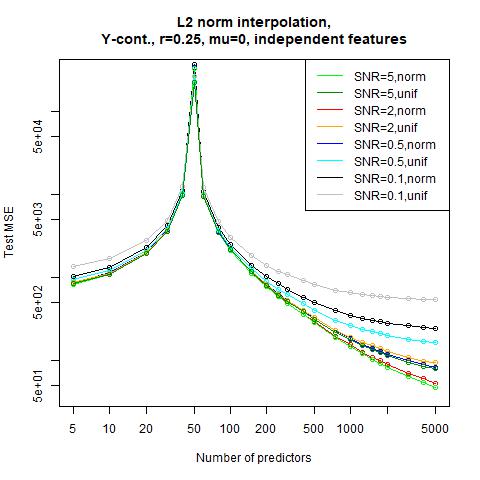} \\
 \includegraphics[width=5.25cm]{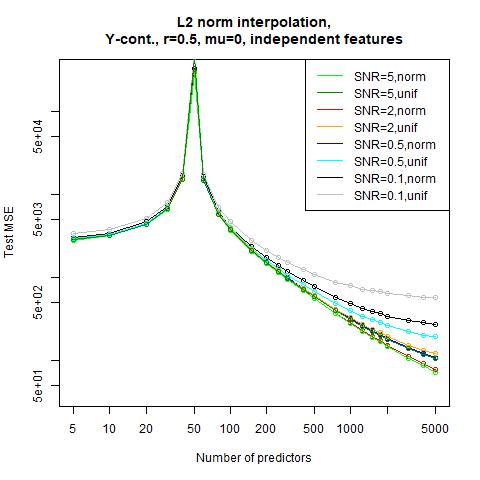} 
\includegraphics[width=5.25cm]{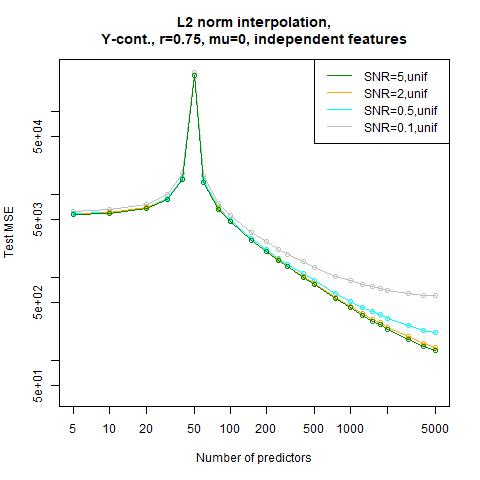}\\
 \includegraphics[width=5.25cm]{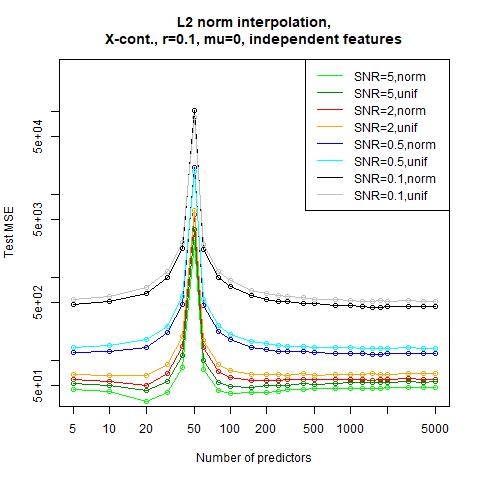} 
\includegraphics[width=5.25cm]{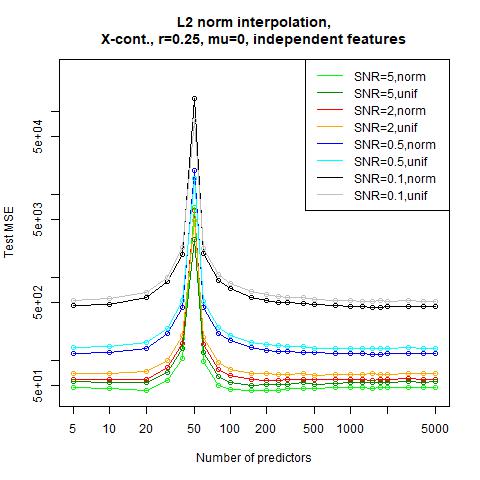} \\ 
\includegraphics[width=5.25cm]{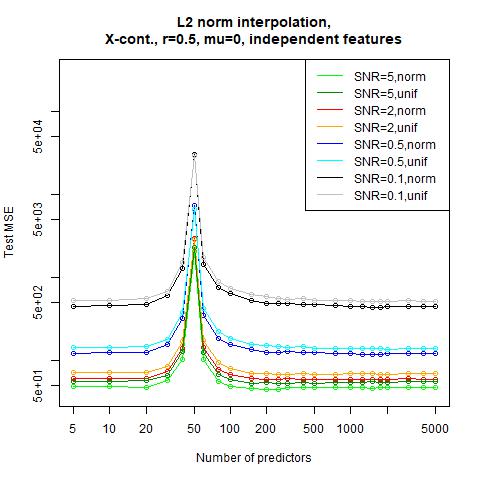}  
\includegraphics[width=5.25cm]{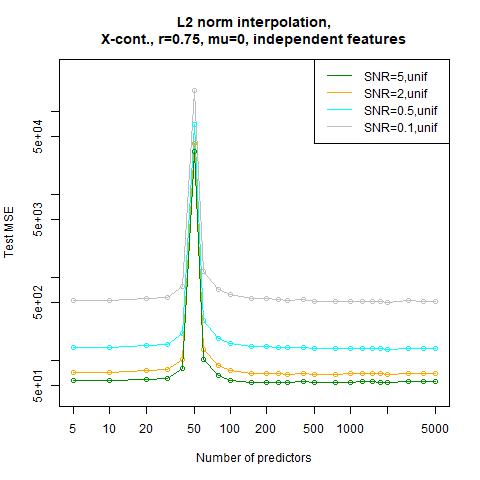} 
\end{center}
\caption{Test MSE of minimum $l_2$-norm interpolation when trained on contaminated training data.} \label{fig:minl2mu0indepcont}
\end{figure}

In Fig. \ref{fig:minl2mu0indep}, the test MSE curves attain their minimum at $p=s=20$ if the SNR is 2 or 5. After the peak at $p=n$, the test MSE decreases, and stays nearly constant afterwards, while always being higher than the minimum at $p=s$. For a SNR smaller than 1, the test MSE first increases until $p=n$, and decreases afterwards, even attaining smaller values than for $p<n$. In the case of $X$-contamination, as visualized in Fig. \ref{fig:minl2mu0indepcont}, the test MSE curves resemble those from the case of clean training data.

A completely different behavior can be observed in the case of $Y$-contamination. For $p<n$, the test MSE curve is strictly increasing, disregarding the SNR. After the peak, the curve is strictly decreasing. For larger $r$, the test MSE is higher for $p<n$ than for smaller $r$. For $p=5000$, the test MSE is comparable to the test MSE for $p=5000$ achieved when training the model on clean data. Therefore, the test MSE decrease is steeper for larger $r$.

\subsubsection{Huber-loss interpolation}

\begin{figure}[H]
\begin{center}
\includegraphics[width=7.5cm]{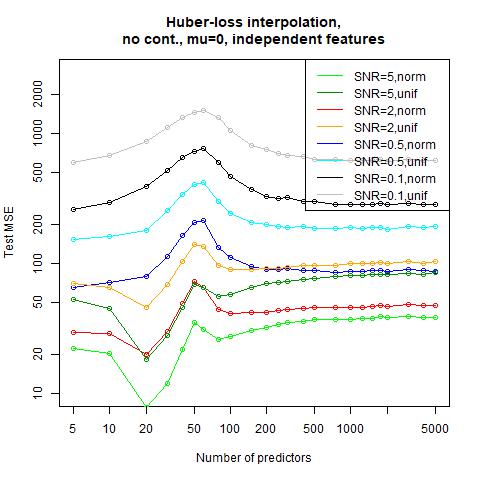}  
\end{center}
\caption{Test MSE of Huber-loss interpolation when trained on clean training data.} \label{fig:hubermu0indep}
\end{figure}

\begin{figure}[H]
\begin{center}
\includegraphics[width=5.25cm]{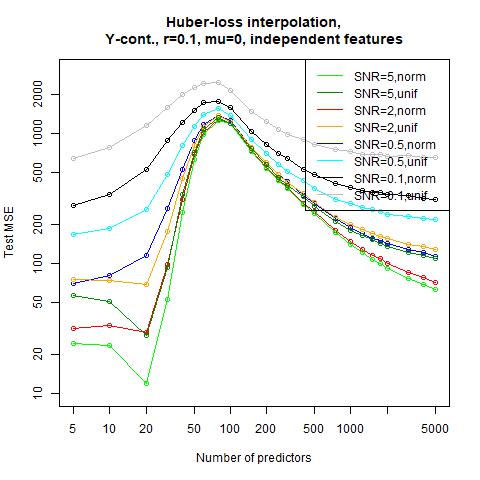} \includegraphics[width=5.25cm]{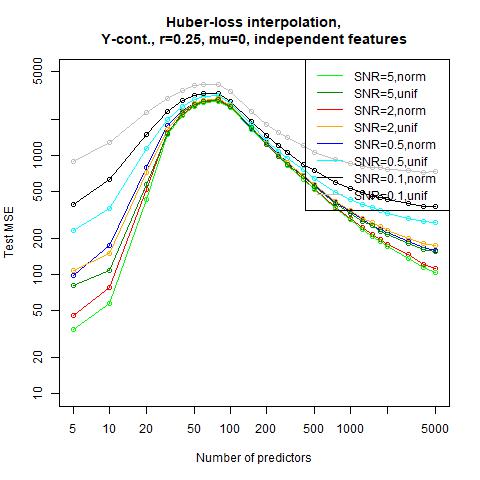} \\ \includegraphics[width=5.25cm]{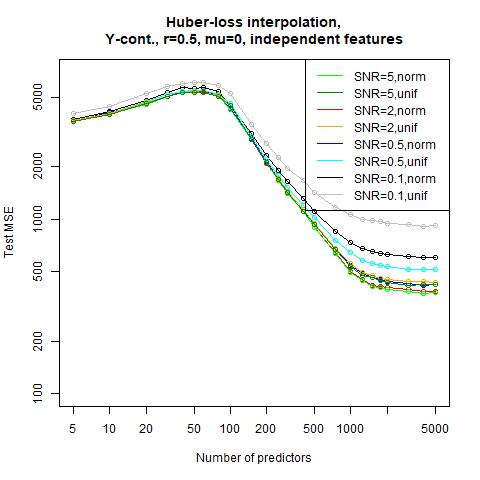} \includegraphics[width=5.25cm]{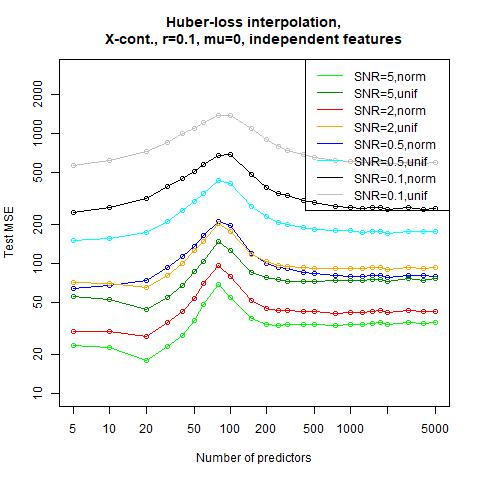} \\ \includegraphics[width=5.25cm]{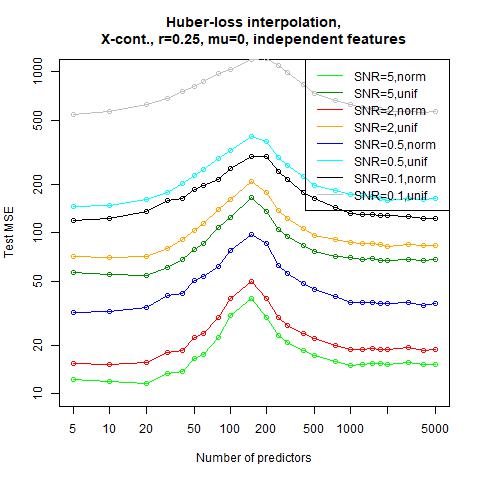} \includegraphics[width=5.25cm]{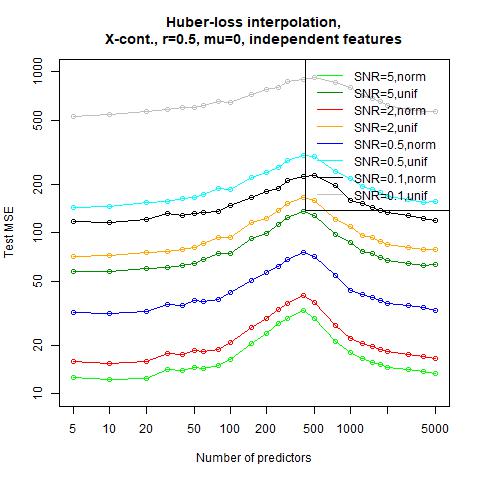} 
\end{center}
\caption{Test MSE of Huber-loss interpolation when trained on contaminated training data.} \label{fig:hubermu0indepcont}
\end{figure}

The main difference between the MSE curves in Fig. \ref{fig:minl2mu0indep} and Fig. \ref{fig:hubermu0indepcont} in comparison with Fig. \ref{fig:hubermu0indep} and Fig. \ref{fig:hubermu0indepcont} is that for the minimum $l_2$-norm interpolator, the peak at $p=n$ is much more elevated. This may be explained by the singularity of the predictor matrix at $p=n$, which does not affect the gradient descent algorithm for the Huber-loss minimization to the same extent. For $X$-contamination, when optimizing the Huber loss, the peak moves to the right once $r$ increases, while the MSE values for very low and very large $p$ remain nearly unaffected.

For $Y$-contamination, the minimum of the test MSE is still attained at $p=s$ if the SNR is larger than 1 and $r=0.1$. For $r=0.25$ and $r=0.5$, the test MSE is strictly increasing until attaining the peak (either at $p=n$ or at $p=100$), and decreasing afterwards. However, the MSE values at $p=5000$ are larger in the contaminated case than in the case with clean training data, and increase with $r$. For $X$-contamination, the peak is attained for $p=100$ for $r=0.1$ and is shifted to $p=500$ for $r=0.5$. The MSE values for low $p$ and large $p$ are comparable with the MSE values for the case of clean training data.

\subsubsection{Tukey-loss interpolation}

\begin{figure}[H]
\begin{center}
\includegraphics[width=7.5cm]{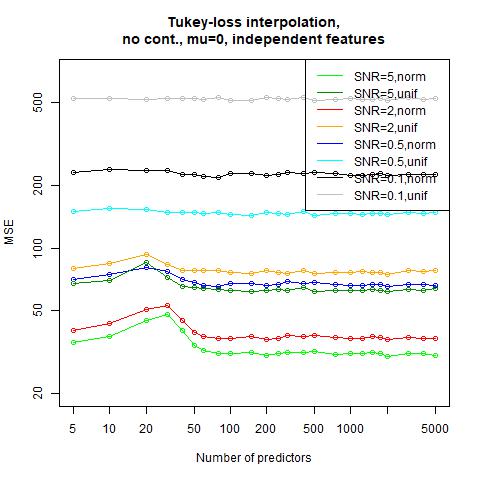} 
\end{center}
\caption{Test MSE of Tukey-loss interpolation when trained on clean training data.} \label{fig:tukeymu0indep}
\end{figure}

\begin{figure}[H]
\begin{center}
\includegraphics[width=5.25cm]{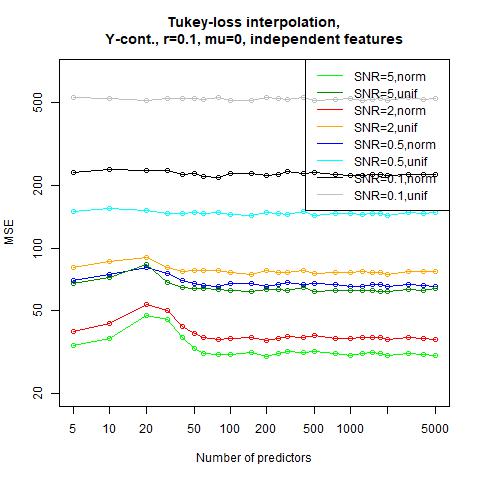} \includegraphics[width=5.25cm]{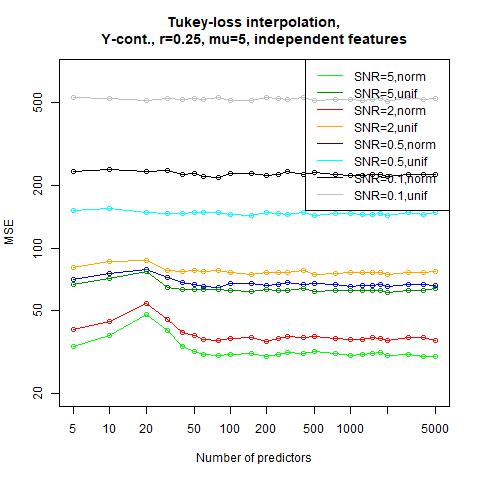} \\ \includegraphics[width=5.25cm]{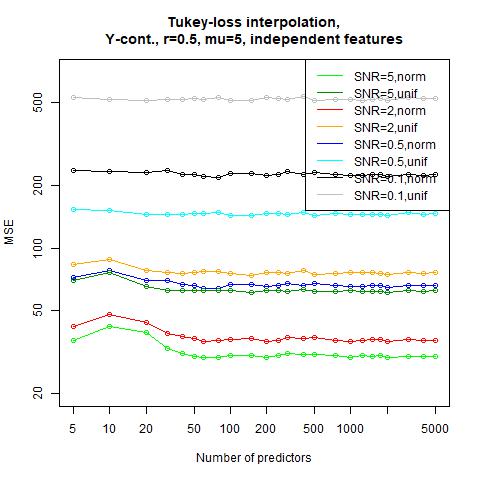} \includegraphics[width=5.25cm]{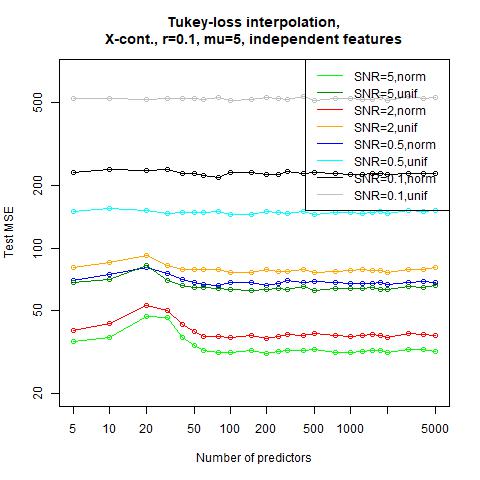} \\ \includegraphics[width=5.25cm]{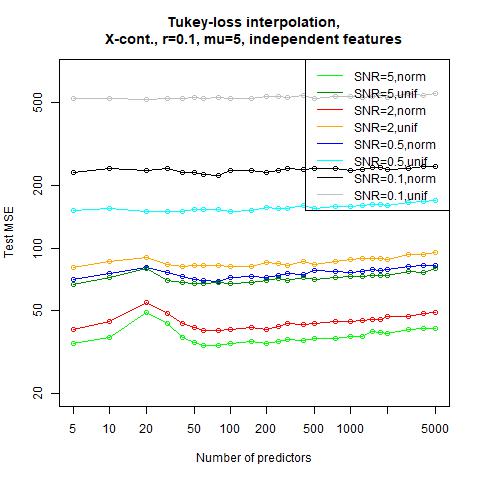} \includegraphics[width=5.25cm]{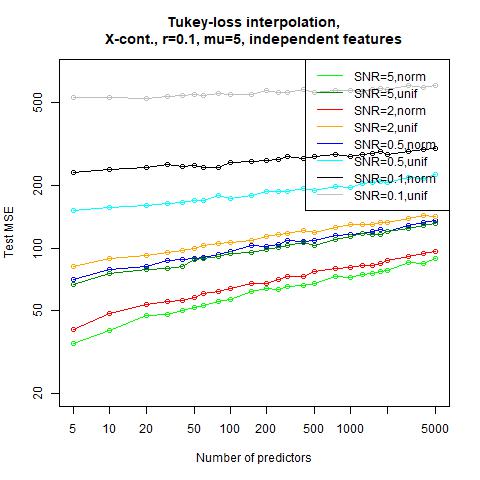} 
\end{center}
\caption{Test MSE of Tukey-loss interpolation when trained on contaminated training data.} \label{fig:tukeymu0indepcont}
\end{figure}

In Fig. \ref{fig:tukeymu0indep} and Fig. \ref{fig:tukeymu0indepcont}, one can observe at peak at $p=n$ or at $p=20$ if the SNR is larger than 1. Before the peak, the curves are strictly increasing, and remain constant after briefly decreasing after the peak. For a low SNR, the MSE curves are nearly constant. For $X$-contamination and large $r$, one can observe an increase of the MSE curves, in particular if the SNR is high. 

\subsubsection{SLTS-based interpolation}

\begin{figure}[H]
\begin{center}
\includegraphics[width=7.5cm]{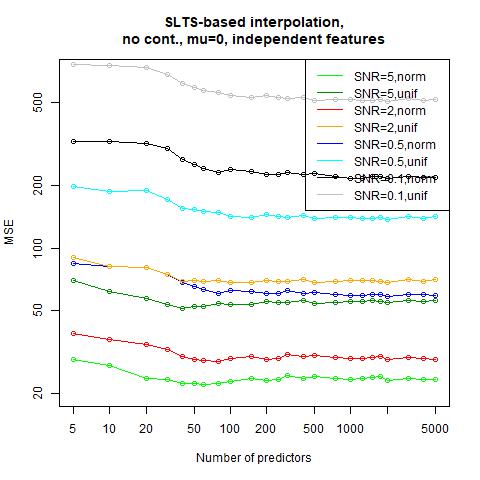} 
\end{center}
\caption{Test MSE of SLTS-based interpolation when trained on clean training data.} \label{fig:sltsmu0indep}
\end{figure}

\begin{figure}[H]
\begin{center}
\includegraphics[width=7.5cm]{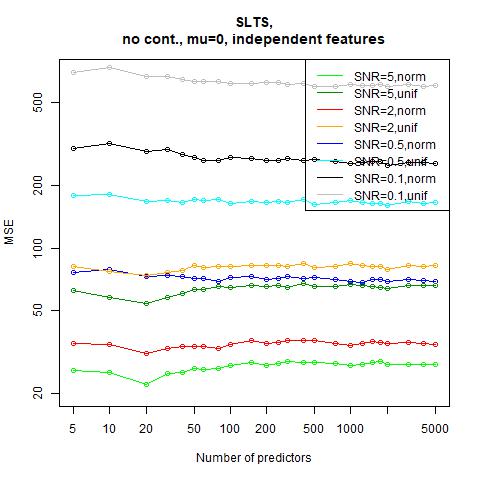} 
\end{center}
\caption{Test MSE of SLTS when trained on clean training data.} \label{fig:rawsltsmu0indep}
\end{figure}

\begin{figure}[H]
\begin{center}
\includegraphics[width=5.25cm]{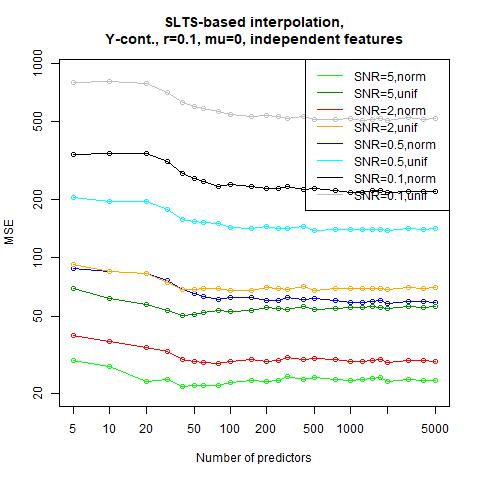} \includegraphics[width=5.25cm]{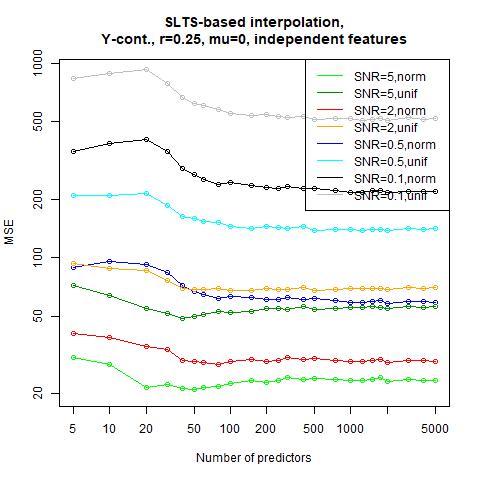} \\ \includegraphics[width=5.25cm]{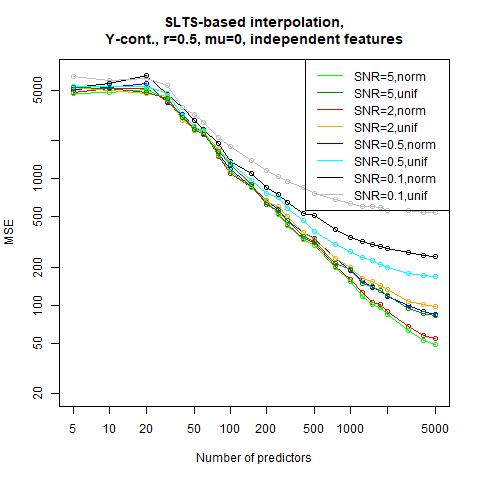} \includegraphics[width=5.25cm]{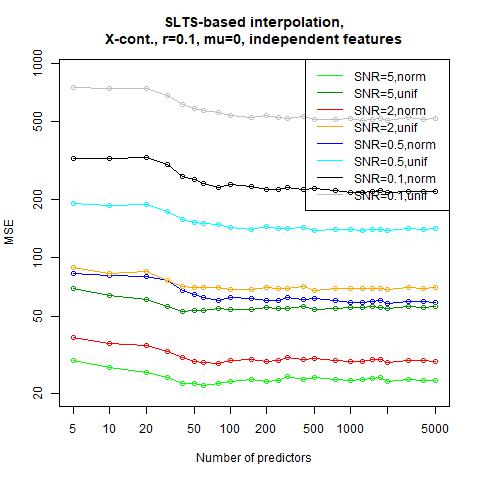} \\ \includegraphics[width=5.25cm]{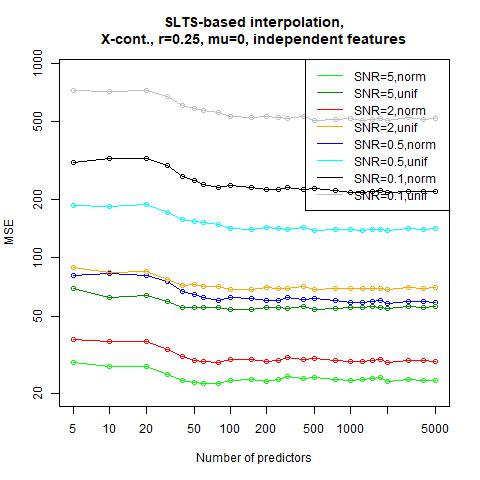} \includegraphics[width=5.25cm]{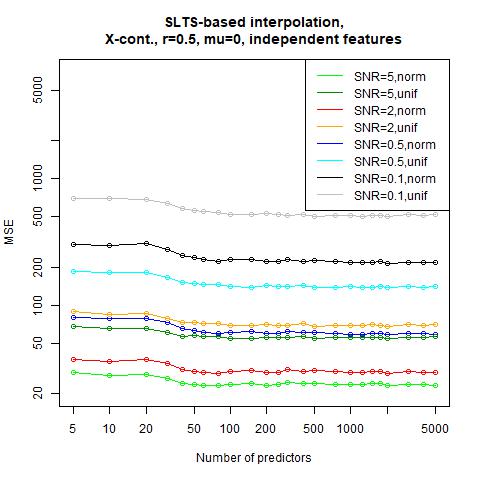} 
\end{center}
\caption{Test MSE of SLTS-based interpolation when trained on contaminated training data.} \label{fig:sltsmu0indepcont}
\end{figure}

In Fig. \ref{fig:sltsmu0indep}, one can observe a slight decrease in the MSE curves but without any visible peak or minimum. In Fig. \ref{fig:sltsmu0indepcont}, it is revealed that the behavior for $X$-contamination and $Y$-contamination with $r=0.1$ and $r=0.25$ leads to similar MSE curves. In contrast, in the case of $Y$-contamination and $r=0.5$, the MSE is much higher for low $p$ than in the other scenarios, attains a peak at $p=20$ and clearly decreases afterwards, attaining comparable values at $p=5000$ as in the other scenarios.

\begin{figure}[H]
\begin{center}
\includegraphics[width=5.25cm]{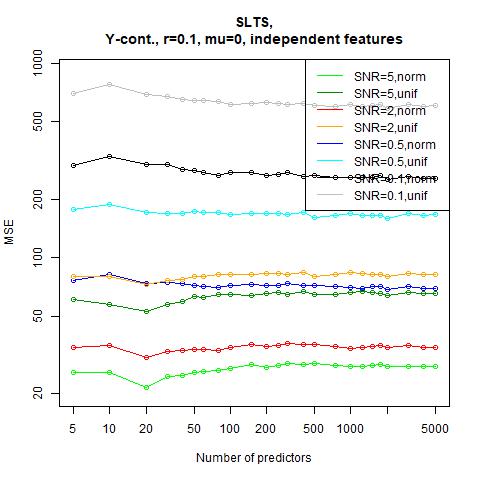} \includegraphics[width=5.25cm]{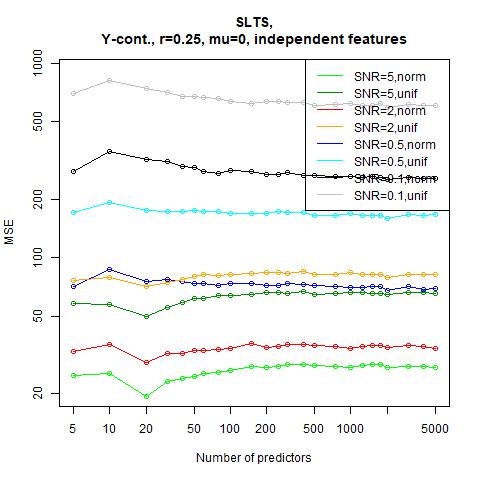} \\ \includegraphics[width=5.25cm]{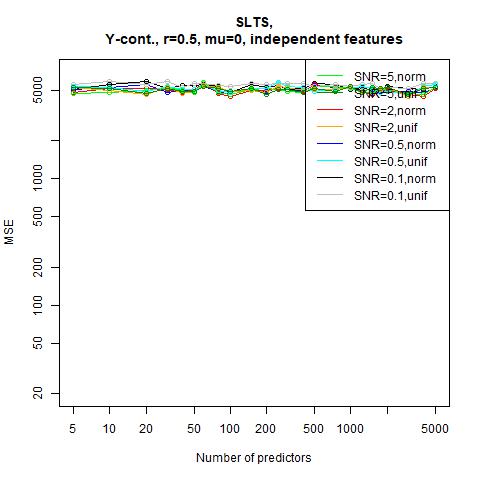} \includegraphics[width=5.25cm]{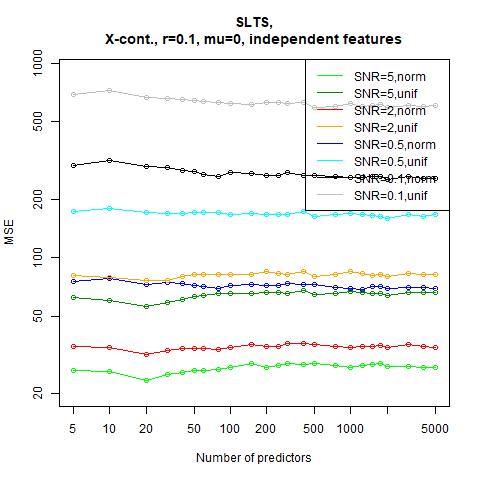} \\ \includegraphics[width=5.25cm]{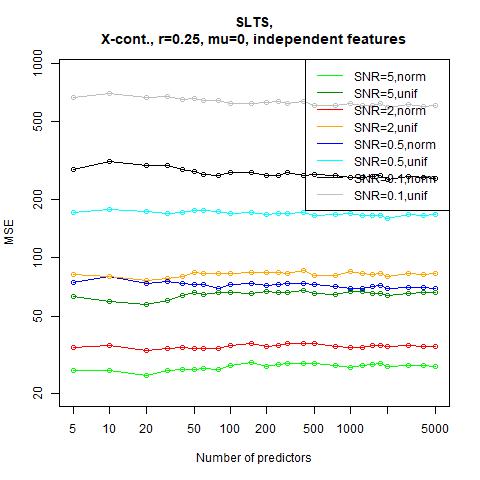} \includegraphics[width=5.25cm]{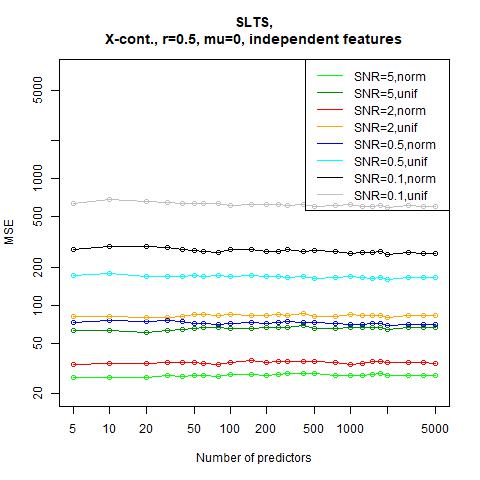} 
\end{center}
\caption{Test MSE of SLTS when trained on contaminated training data.}\label{fig:rawsltsmu0indepcont}
\end{figure}

Standard SLTS results in similar MSE curves when trained on clean data as SLTS-based interpolation, as it can be observed in Fig. \ref{fig:rawsltsmu0indep}. Standard SLTS leads to nearly constant MSE curves for $X$-contamination as seen in Fig. \ref{fig:rawsltsmu0indepcont}, while for SLTS-based interpolation, the MSE was higher for low $p$ and slightly decreased in order to remain nearly constant at larger $p$. A considerable difference can be observed for $r=0.5$ and $Y$-contamination. For SLTS, the MSE is always around 5000, while for SLTS-based interpolation, the curves started around 5000 as well in order to decrease significantly for growing $p$.

\subsubsection{Boosting-based interpolation}

\begin{figure}[H]
\begin{center}
\includegraphics[width=7.5cm]{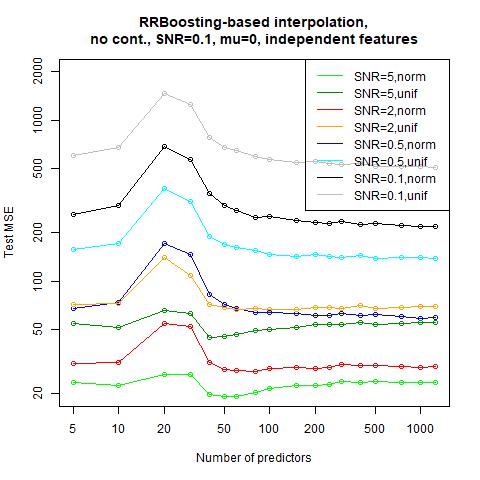} \\ 
\end{center}
\caption{Test MSE of RRBoost-based interpolation when trained on clean training data.} \label{fig:rrboostmu0indep}
\end{figure}

\begin{figure}[H]
\begin{center}
\includegraphics[width=7.5cm]{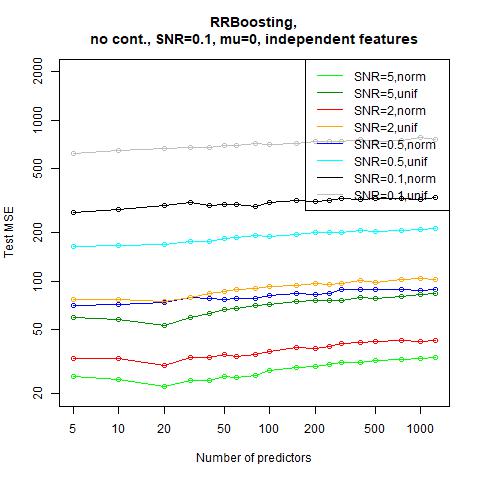} 
\end{center}
\caption{Test MSE of RRBoost when trained on clean training data.} \label{fig:rawrrboostmu0indep}
\end{figure}

\begin{figure}[H]
\begin{center}
\includegraphics[width=5.25cm]{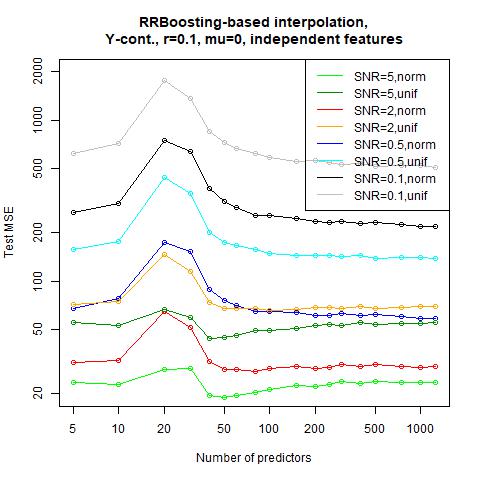} \includegraphics[width=5.25cm]{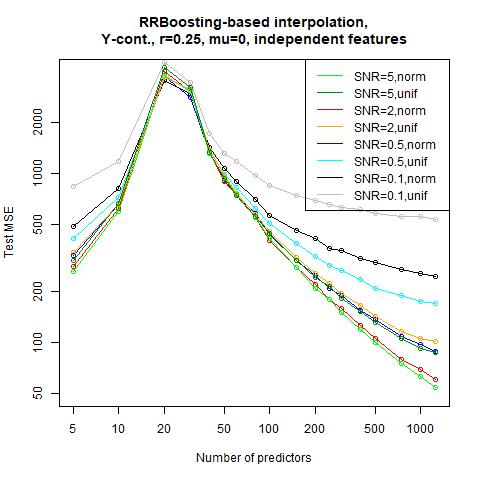} \\ \includegraphics[width=5.25cm]{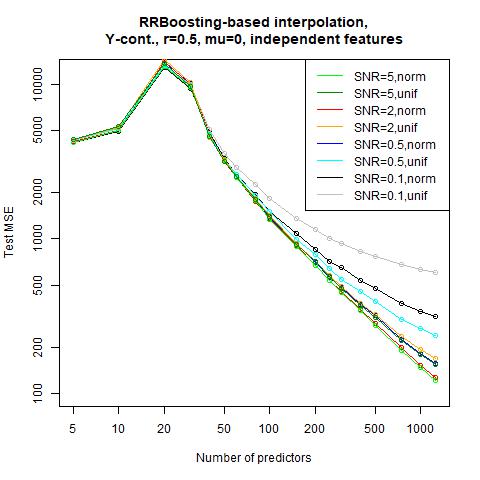} \includegraphics[width=5.25cm]{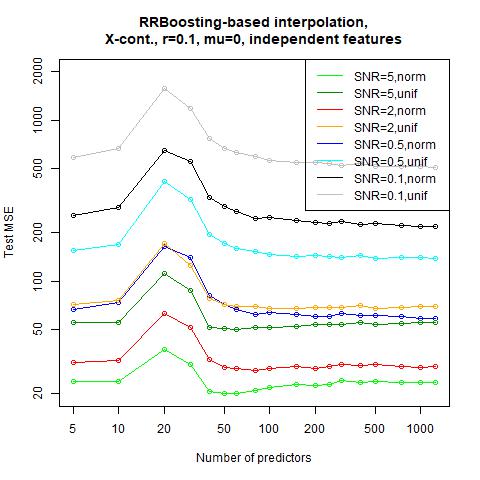} \\ \includegraphics[width=5.25cm]{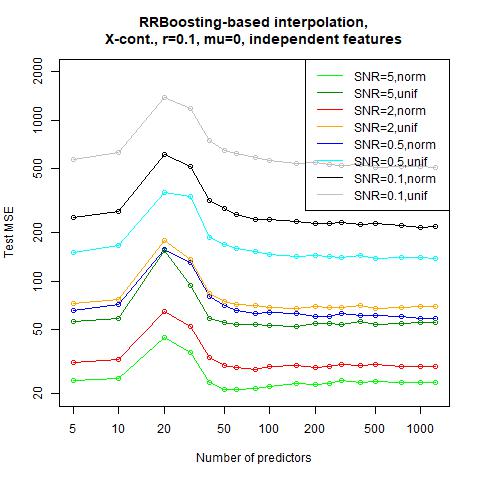} \includegraphics[width=5.25cm]{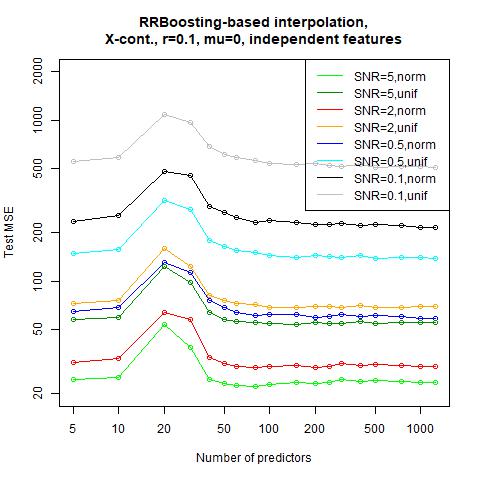} 
\end{center}
\caption{Test MSE of RRBoost-based interpolation when trained on contaminated training data.} \label{fig:rrboostmu0indepcont}
\end{figure}

The MSE curves in Fig. \ref{fig:rrboostmu0indep} and Fig. \ref{fig:rrboostmu0indepcont} first increase, attain at peak at $p=20$ and decrease afterwards. When training the model on clean data or on $X$-contaminated data, the MSE attains similar values for large $p$ as for low $p$. For $Y$-contamination and $r=0.25$ and $r=0.5$, the MSE is larger for low $p$ and significantly decreases after the peak, attaining however higher values at $p=5000$ than in the other scenarios.

\begin{figure}[H]
\begin{center}
\includegraphics[width=5.25cm]{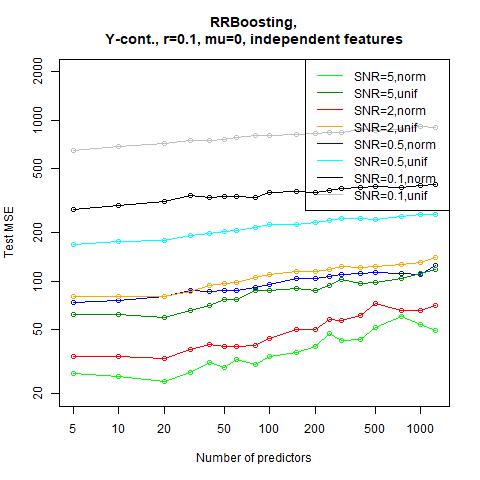} \includegraphics[width=5.25cm]{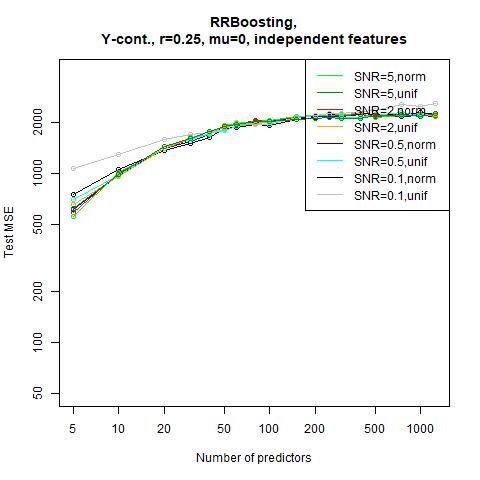} \\ \includegraphics[width=5.25cm]{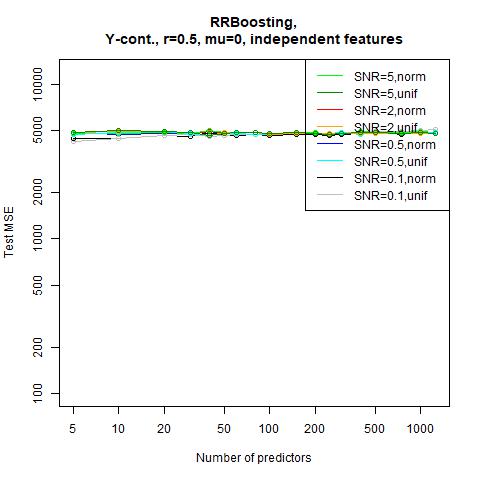} \includegraphics[width=5.25cm]{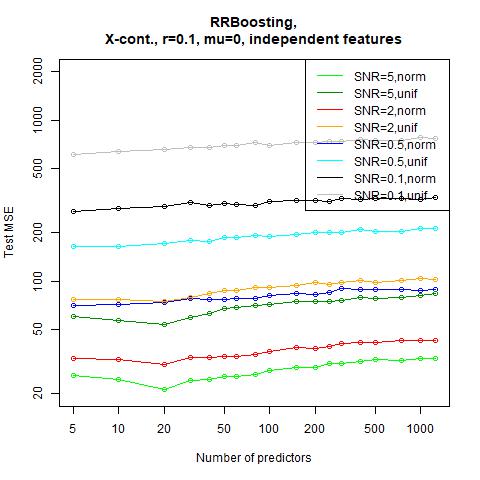} \\ \includegraphics[width=5.25cm]{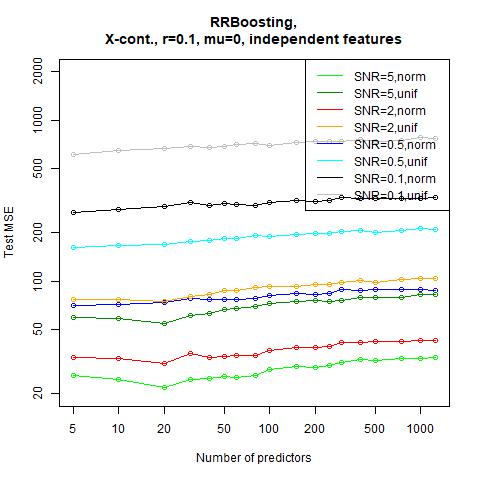} \includegraphics[width=5.25cm]{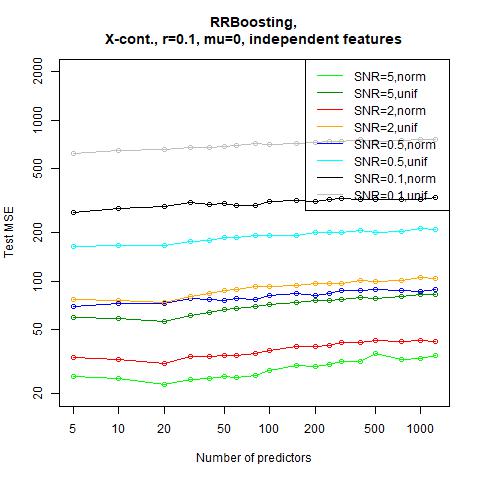} 
\end{center}
\caption{Test MSE of RRBoost when trained on contaminated training data.} \label{fig:rawrrboostmu0indepcont}
\end{figure}

In contrast to RRBoost-based interpolation, the MSE curves slightly increase for growing $p$ in Fig. \ref{fig:rawrrboostmu0indep} and Fig. \ref{fig:rawrrboostmu0indepcont}. For $Y$-contamination and $r=0.25$, the growth is steeper than for all other scenarios while for $r=0.5$, the MSE curves remain nearly constant.

\newpage

\subsection{Spiked covariance design, $\mu=0$}

\subsubsection{Minimum $l_2$-norm interpolation}

\begin{figure}[H]
\begin{center}
\includegraphics[width=7.5cm]{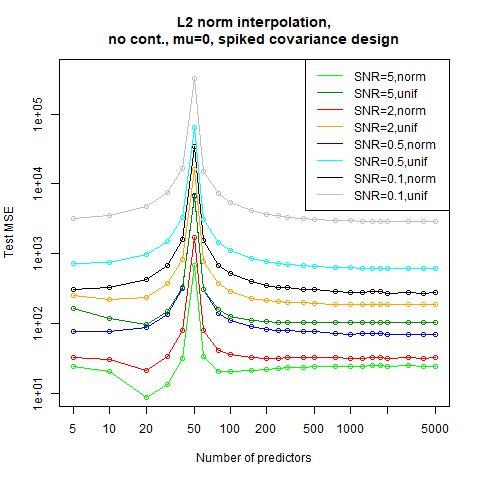} 
\end{center}
\caption{Test MSE of minimum $l_2$-norm interpolation when trained on clean training data.} \label{fig:minl2mu0spiked}
\end{figure}

\begin{figure}[H]
\begin{center}
\includegraphics[width=5.25cm]{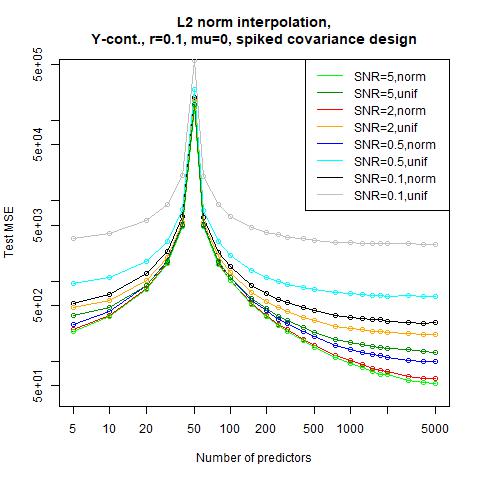} \includegraphics[width=5.25cm]{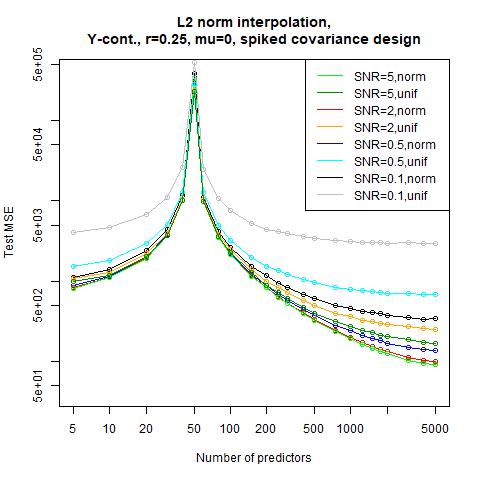} \\ \includegraphics[width=5.25cm]{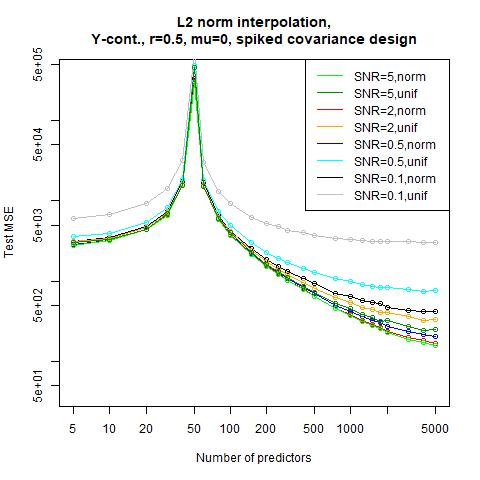} \includegraphics[width=5.25cm]{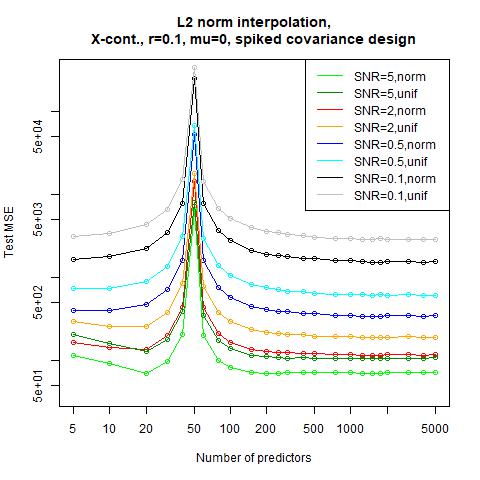} \\ \includegraphics[width=5.25cm]{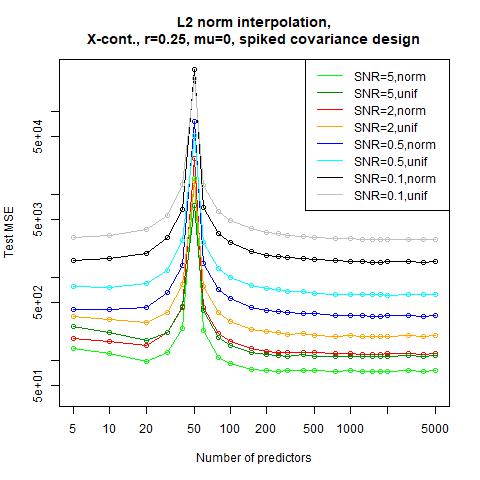} \includegraphics[width=5.25cm]{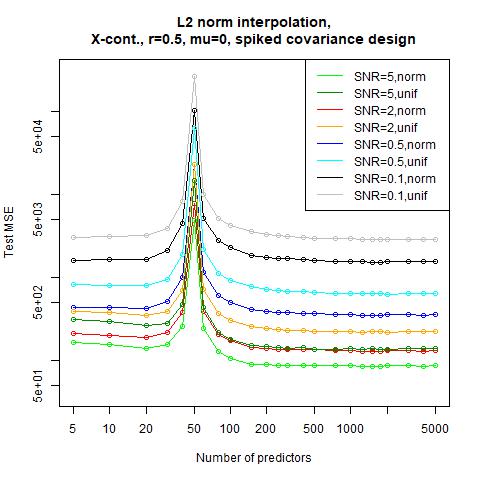} 
\end{center}
\caption{Test MSE of minimum $l_2$-norm interpolation when trained on contaminated training data.} \label{fig:minl2mu0spikedcont}
\end{figure}

\subsubsection{Huber-loss interpolation}

\begin{figure}[H]
\begin{center}
\includegraphics[width=7.5cm]{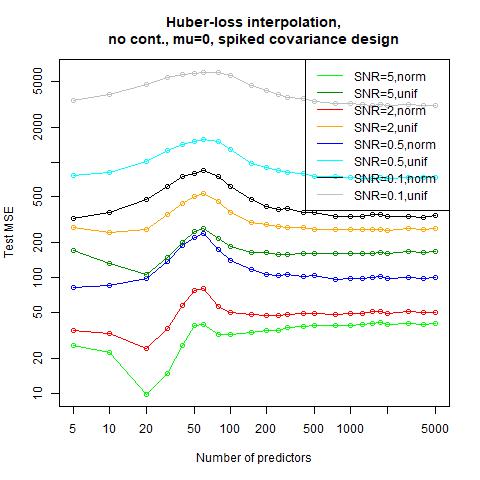} 
\end{center}
\caption{Test MSE of Huber-loss interpolation when trained on clean training data.} \label{fig:hubermu0indepspiked}
\end{figure}

\begin{figure}[H]
\begin{center}
\includegraphics[width=5.25cm]{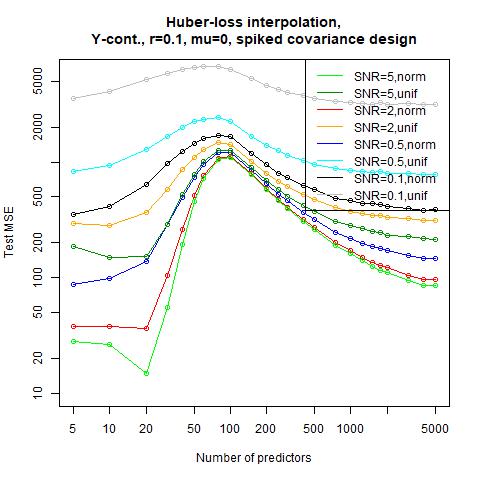} \includegraphics[width=5.25cm]{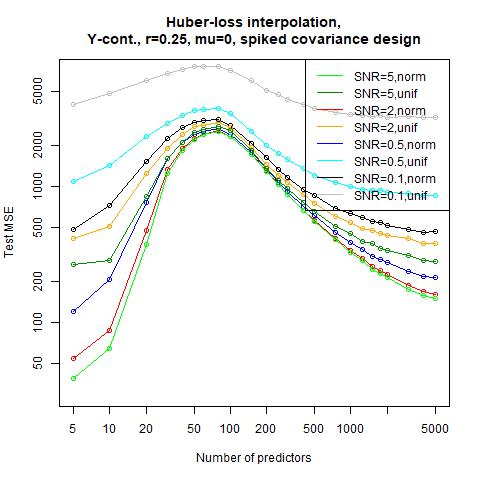} \\ \includegraphics[width=5.25cm]{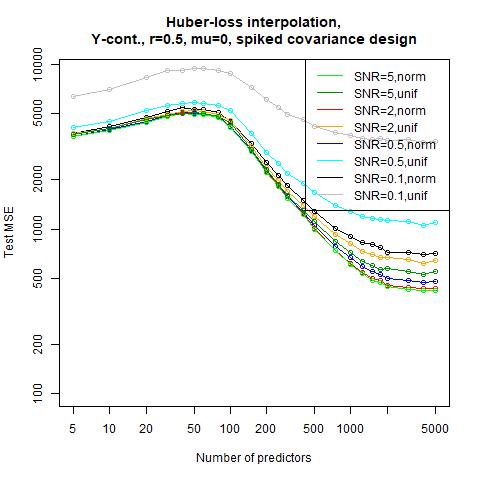} \includegraphics[width=5.25cm]{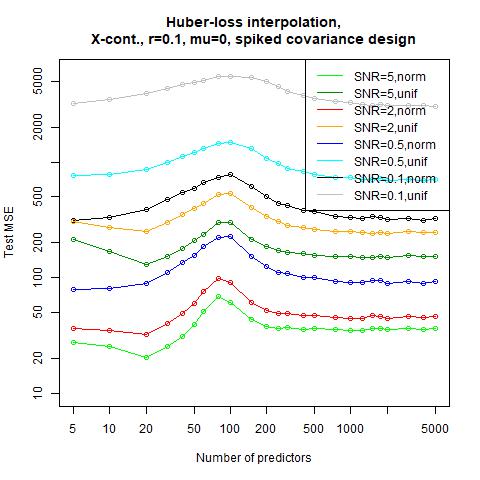} \\ \includegraphics[width=5.25cm]{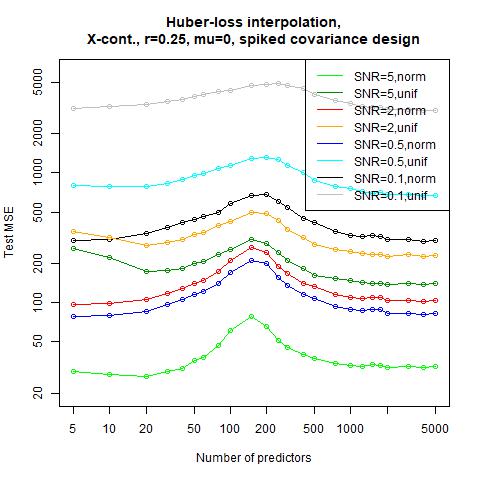} \includegraphics[width=5.25cm]{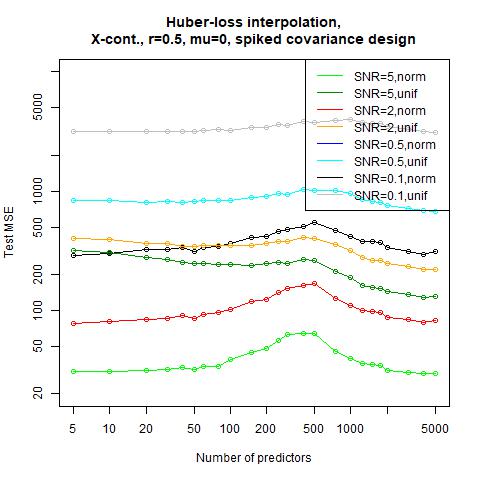} 
\end{center}
\caption{Test MSE of Huber-loss interpolation when trained on contaminated training data.} \label{fig:hubermu0indepspikedcont}
\end{figure}

\subsubsection{Tukey-loss interpolation}

\begin{figure}[H]
\begin{center}
\includegraphics[width=7.5cm]{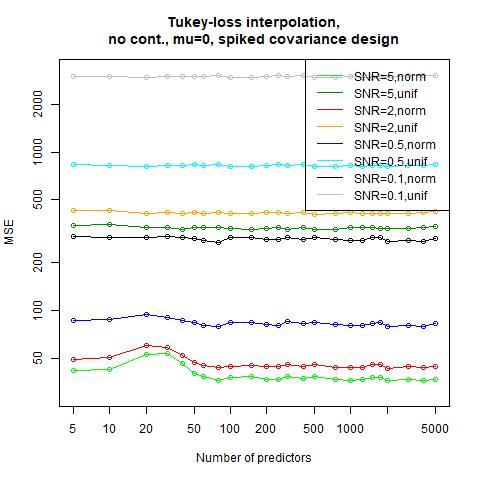} 
\end{center}
\caption{Test MSE of Tukey-loss interpolation when trained on clean training data.} \label{fig:tukeymu0spiked}
\end{figure}

\begin{figure}[H]
\begin{center}
\includegraphics[width=5.25cm]{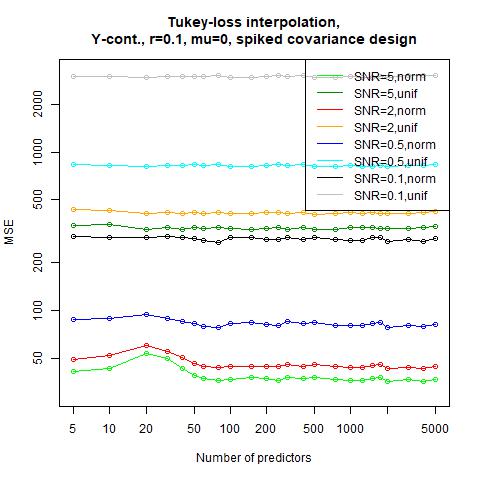} \includegraphics[width=5.25cm]{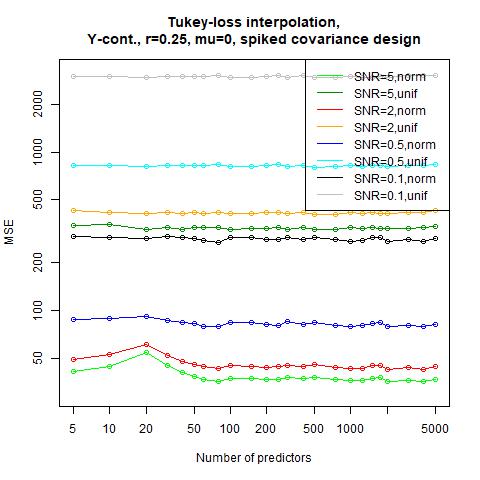} \\ \includegraphics[width=5.25cm]{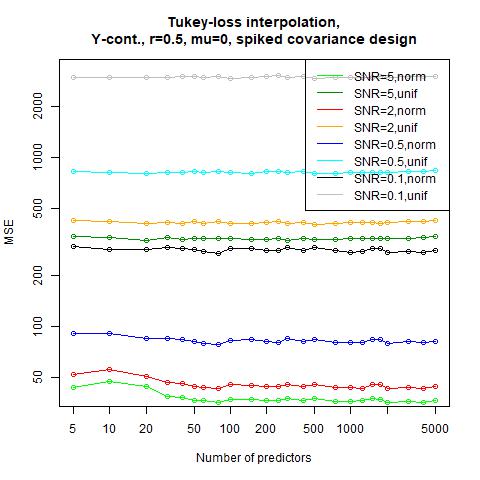} \includegraphics[width=5.25cm]{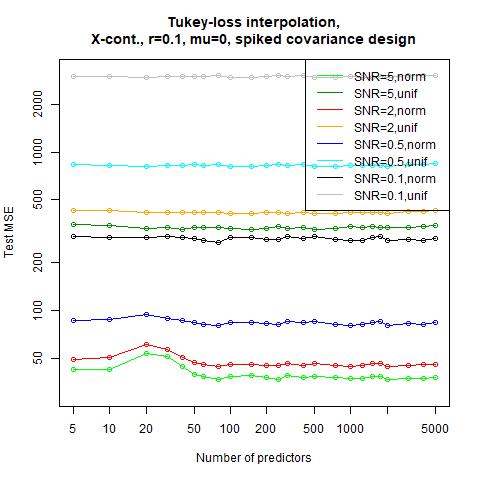} \\ \includegraphics[width=5.25cm]{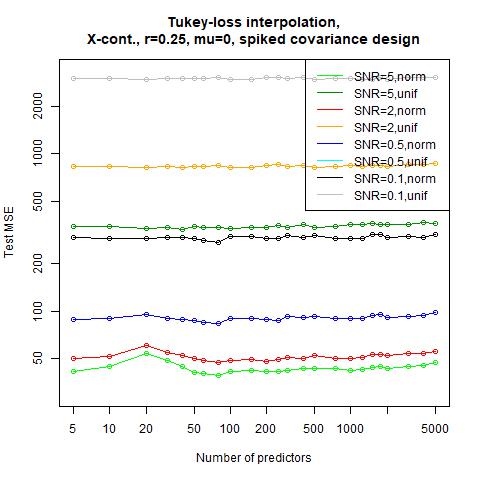} \includegraphics[width=5.25cm]{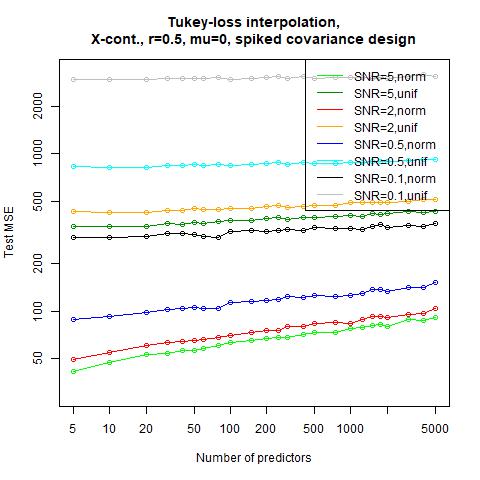} 
\end{center}
\caption{Test MSE of Tukey-loss interpolation when trained on contaminated training data.} \label{fig:tukeymu0spikedcont}
\end{figure}

\subsubsection{SLTS-based interpolation}

\begin{figure}[H]
\begin{center}
\includegraphics[width=7.5cm]{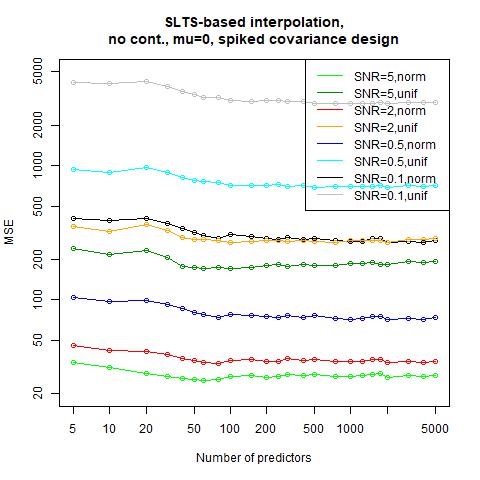} 
\end{center}
\caption{Test MSE of SLTS-based interpolation when trained on clean training data.} \label{fig:sltsmu0spiked}
\end{figure}

\begin{figure}[H]
\begin{center}
\includegraphics[width=7.5cm]{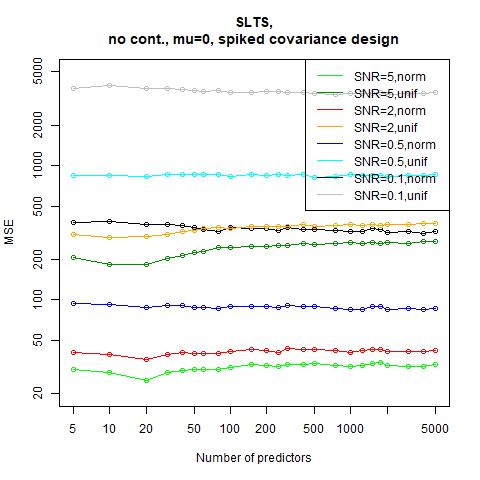} 
\end{center}
\caption{Test MSE of SLTS when trained on clean training data.}\label{fig:rawsltsmu0spiked}
\end{figure}

\begin{figure}[H]
\begin{center}
\includegraphics[width=5.25cm]{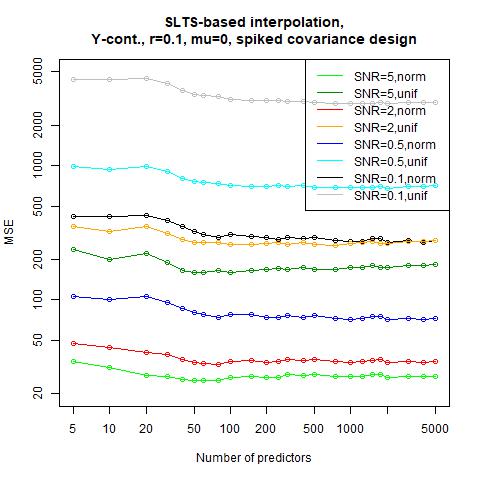} \includegraphics[width=5.25cm]{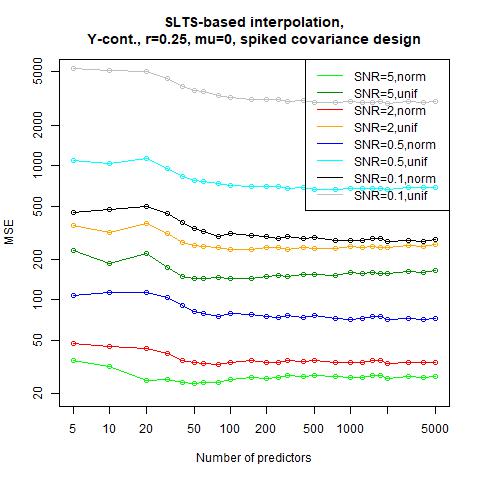} \\ \includegraphics[width=5.25cm]{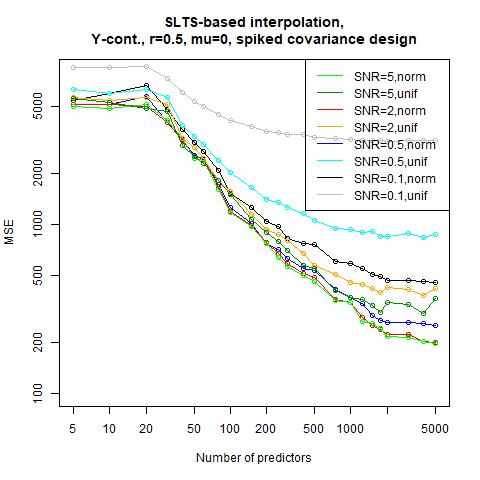} \includegraphics[width=5.25cm]{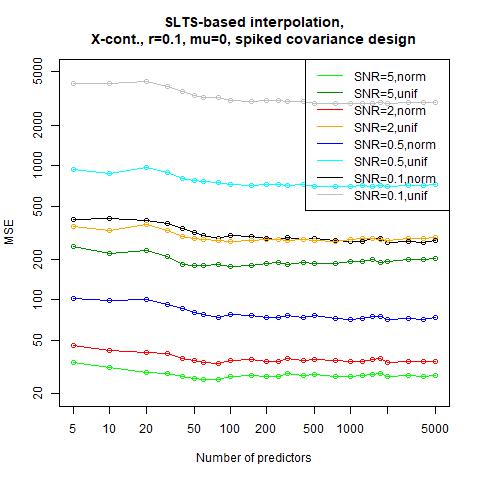} \\ \includegraphics[width=5.25cm]{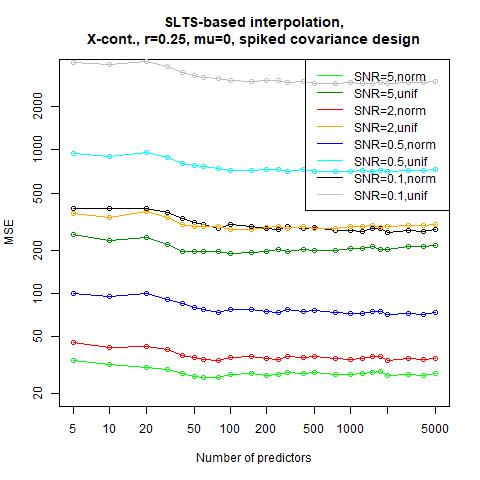} \includegraphics[width=5.25cm]{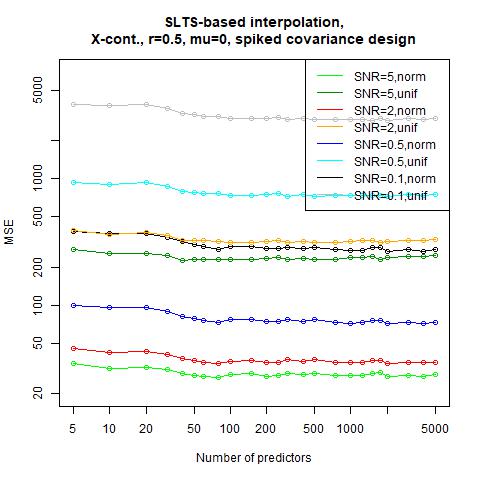} 
\end{center}
\caption{Test MSE of SLTS-based interpolation when trained on contaminated training data.} \label{fig:sltsmu0spikedcont}
\end{figure}

\begin{figure}[H]
\begin{center}
\includegraphics[width=5.25cm]{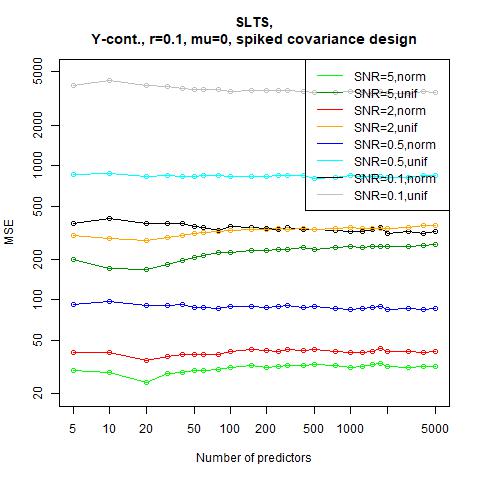} \includegraphics[width=5.25cm]{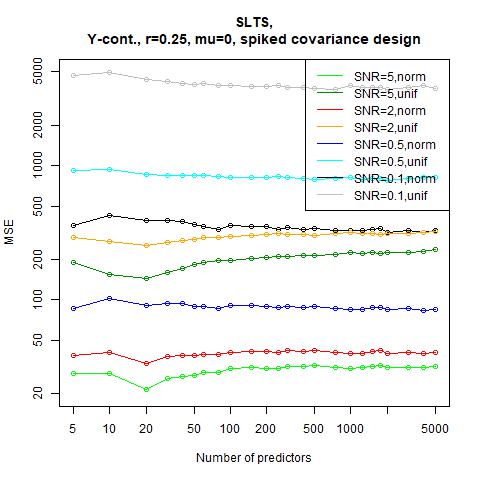} \\ \includegraphics[width=5.25cm]{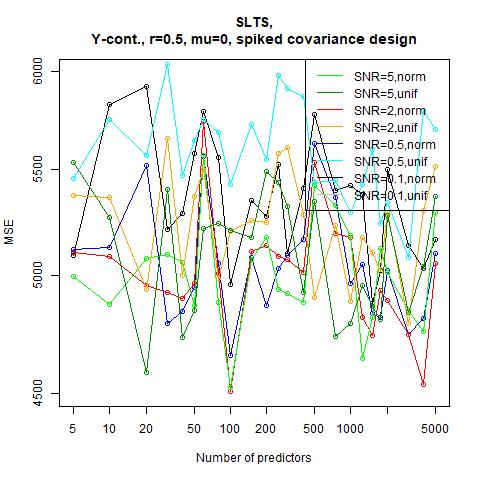} \includegraphics[width=5.25cm]{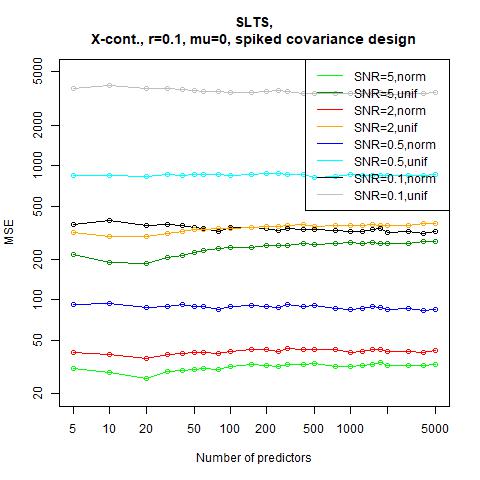} \\ \includegraphics[width=5.25cm]{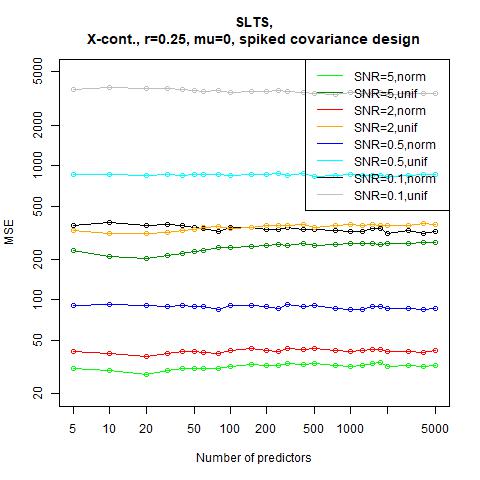} \includegraphics[width=5.25cm]{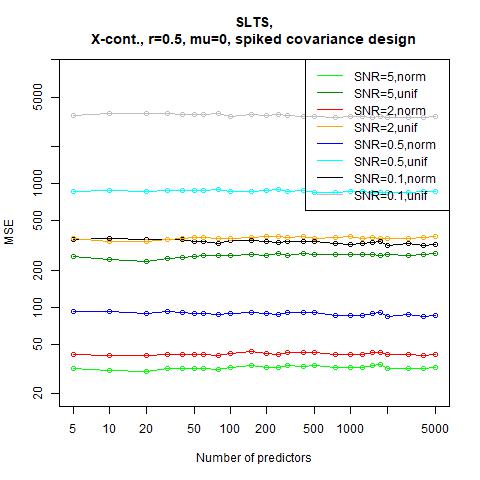} 
\end{center}
\caption{Test MSE of SLTS when trained on contaminated training data.} \label{fig:rawsltsmu0spikedcont}
\end{figure}

\subsubsection{Boosting-based interpolation}

\begin{figure}[H]
\begin{center}
\includegraphics[width=7.5cm]{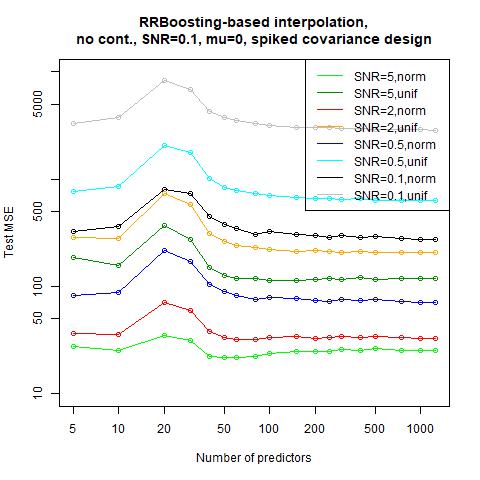} 
\end{center}
\caption{Test MSE of RRBoost-based interpolation when trained on clean training data.} \label{fig:rrboostmu0spiked}
\end{figure}

\begin{figure}[H]
\begin{center}
\includegraphics[width=7.5cm]{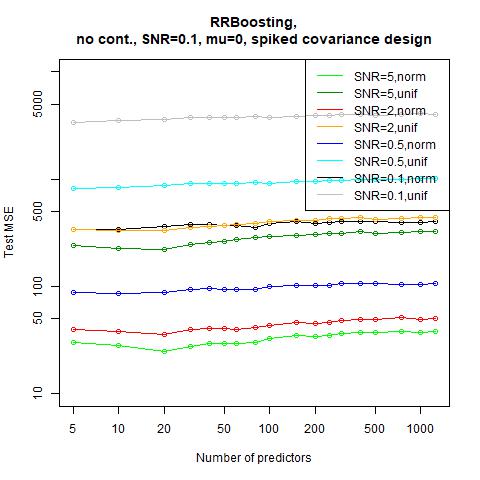} 
\end{center}
\caption{Test MSE of RRBoost when trained on clean training data.} \label{fig:rrboostmu0spikedraw}
\end{figure}

\begin{figure}[H]
\begin{center}
\includegraphics[width=5.25cm]{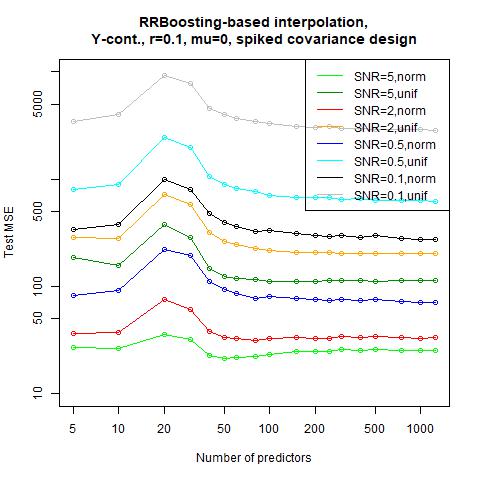} \includegraphics[width=5.25cm]{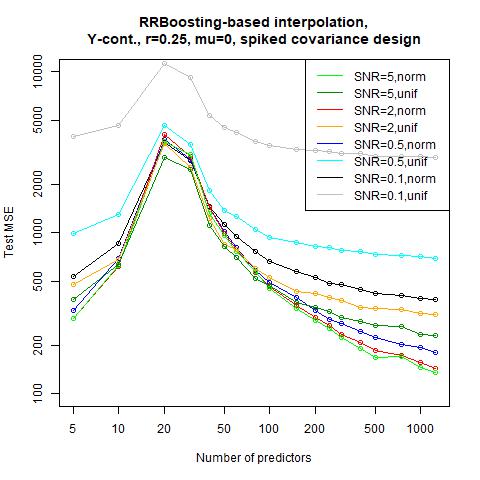} \\ \includegraphics[width=5.25cm]{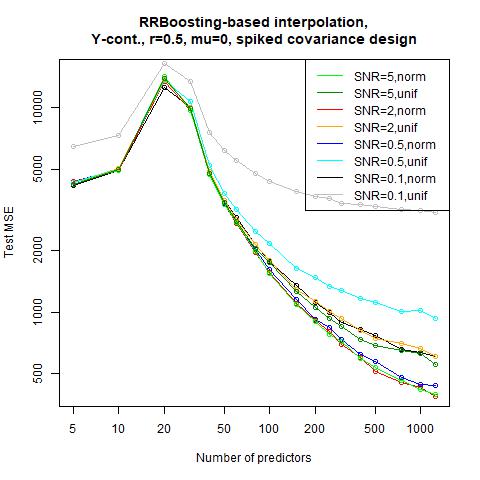} \includegraphics[width=5.25cm]{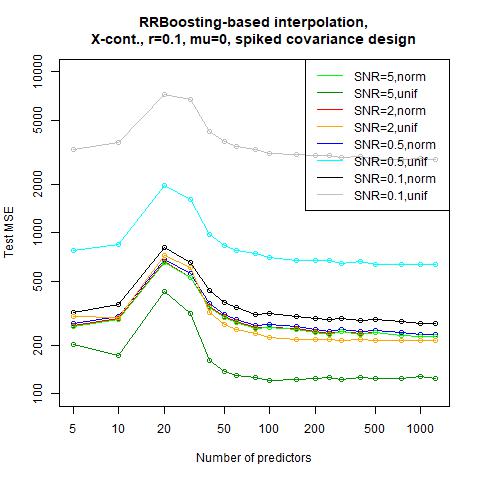} \\ \includegraphics[width=5.25cm]{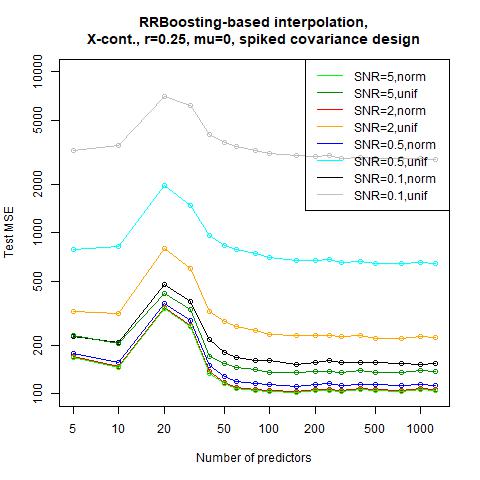} \includegraphics[width=5.25cm]{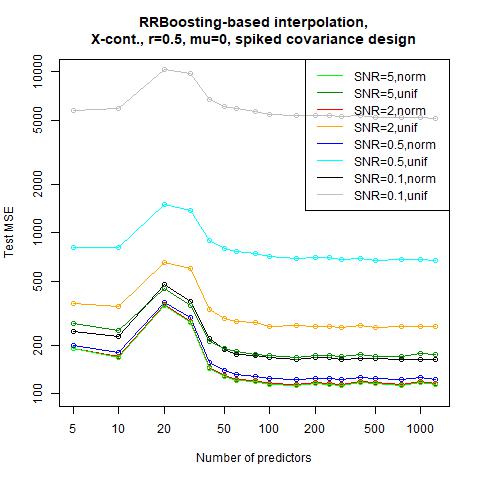} 
\end{center}
\caption{Test MSE of RRBoost-based interpolation when trained on contaminated training data.} \label{fig:rrboostmu0spikedcont}
\end{figure}

\begin{figure}[H]
\begin{center}
\includegraphics[width=5.25cm]{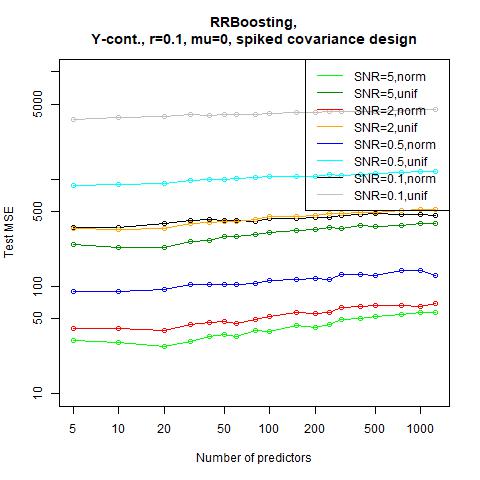} \includegraphics[width=5.25cm]{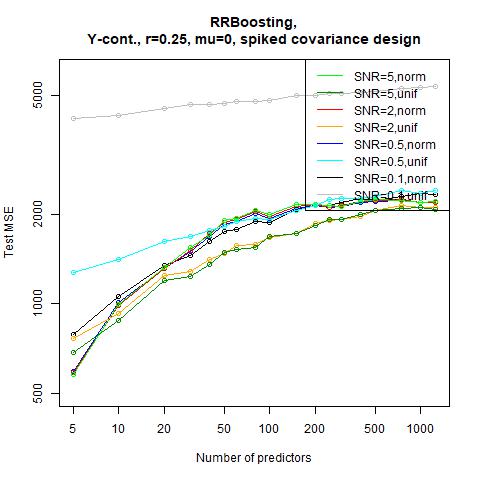} \\ \includegraphics[width=5.25cm]{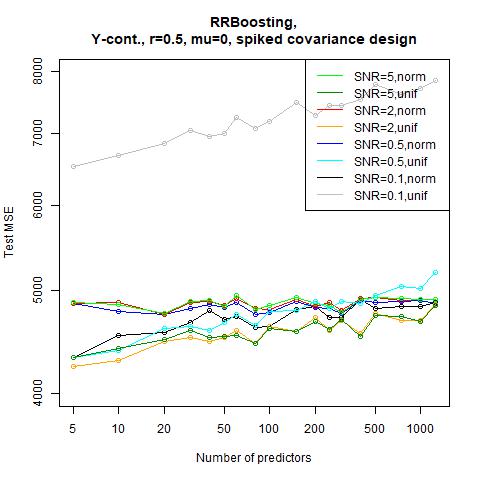} \includegraphics[width=5.25cm]{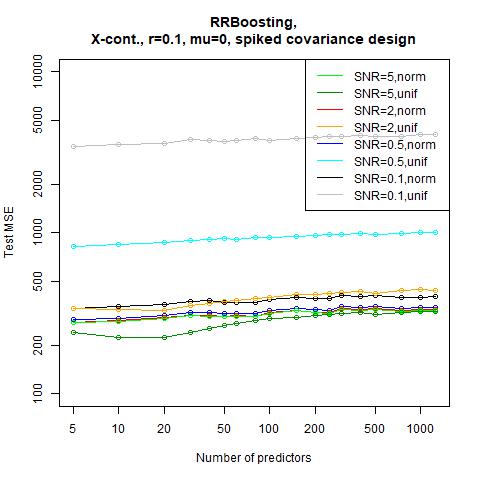} \\ \includegraphics[width=5.25cm]{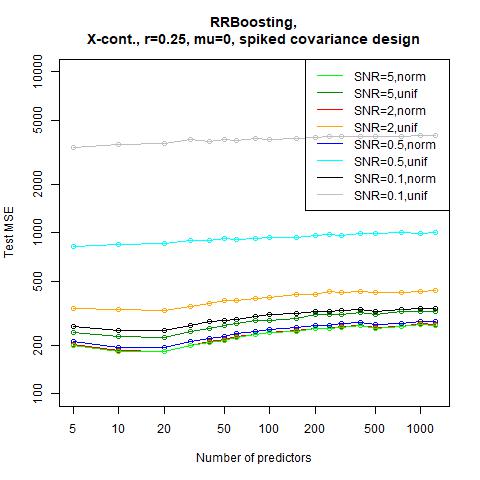} \includegraphics[width=5.25cm]{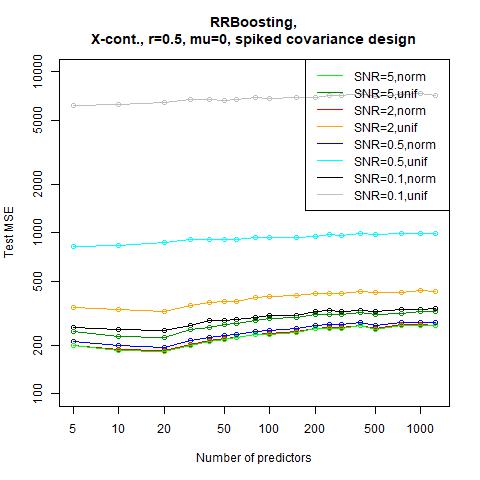} 
\end{center}
\caption{Test MSE of RRBoost when trained on contaminated training data.} \label{fig:rrboostmu0spikedcontraw}
\end{figure}

The curves in Fig. \ref{fig:minl2mu0spiked}, Fig. \ref{fig:minl2mu0spikedcont}, Fig. \ref{fig:hubermu0indepspiked}, Fig. \ref{fig:hubermu0indepspikedcont}, Fig. \ref{fig:tukeymu0spiked}, Fig. \ref{fig:tukeymu0spikedcont}, Fig. \ref{fig:sltsmu0spiked}, Fig. \ref{fig:sltsmu0spikedcont}, Fig. \ref{fig:rawsltsmu0spiked}, Fig. \ref{fig:rrboostmu0spiked}, Fig. \ref{fig:rrboostmu0spikedcont} and Fig. \ref{fig:rrboostmu0spikedraw} resemble those for the independent design. The only difference is that the MSE values are higher for the spiked covariance design and that, in the case of an MSE decrease for growing $p$, the attained MSE values for $p=5000$ are considerably higher in the contaminated settings than in the clean setting. In Fig. \ref{fig:rawsltsmu0spikedcont} and Fig. \ref{fig:rrboostmu0spikedcontraw}, the MSE curves fluctuate for $Y$-contamination and $r=0.5$, in contrast to the nearly constant curves when having independent design.

\subsection{Independent features, $\mu=5$}

\subsubsection{Minimum $l_2$-norm interpolation}

\begin{figure}[H]
\begin{center}
\includegraphics[width=7.5cm]{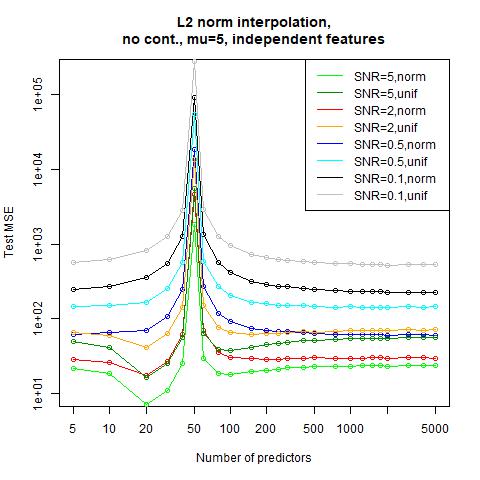} 
\end{center}
\caption{Test MSE of minimum $l_2$-norm interpolation when trained on clean training data with $\mu=5$.} \label{fig:minl2mu5indep}
\end{figure}

\begin{figure}[H]
\begin{center}
\includegraphics[width=5.25cm]{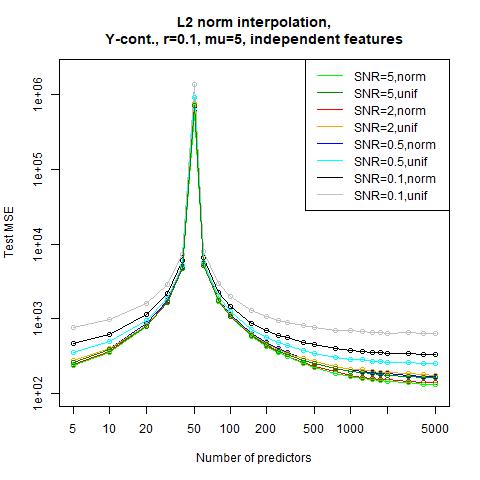} \includegraphics[width=5.25cm]{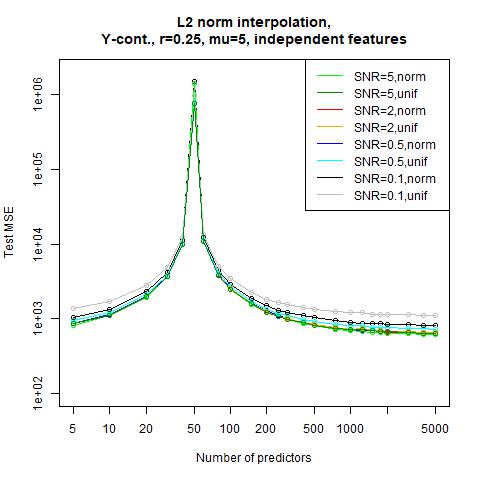} \\ \includegraphics[width=5.25cm]{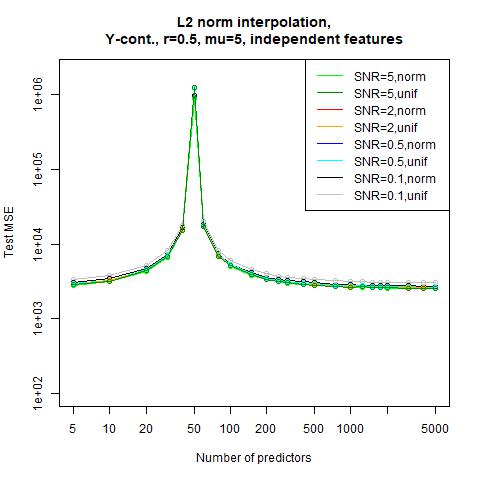} \includegraphics[width=5.25cm]{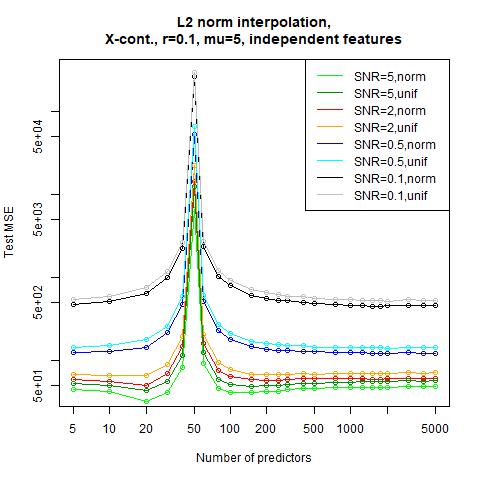} \\ \includegraphics[width=5.25cm]{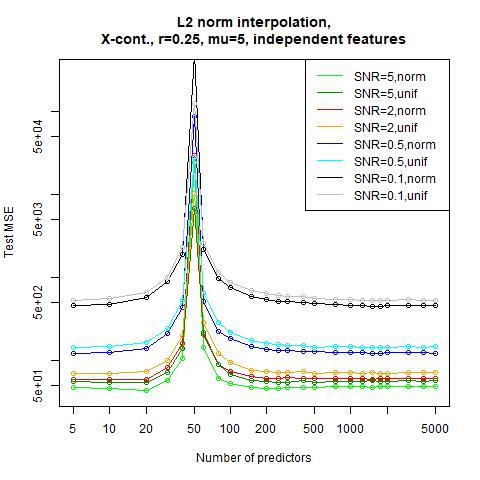} \includegraphics[width=5.25cm]{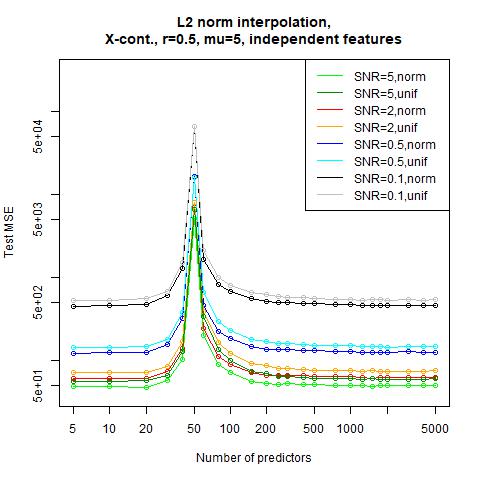} 
\end{center}
\caption{Test MSE of minimum $l_2$-norm interpolation when trained on contaminated training data with $\mu=5$.} \label{fig:minl2mu5indepcont}
\end{figure}

In contrast to the case $\mu=0$, the MSE curves for $Y$-contamination remain around the MSE values for small $p$ after the peak. In the case of $X$-contamination and clean training data, the curves resemble those from the case $\mu=0$, with the difference that the MSE values are higher, as depicted in Fig. \ref{fig:minl2mu5indep} and Fig. \ref{fig:minl2mu5indepcont}. 

\subsubsection{Huber-loss interpolation}

\begin{figure}[H]
\begin{center}
\includegraphics[width=7.5cm]{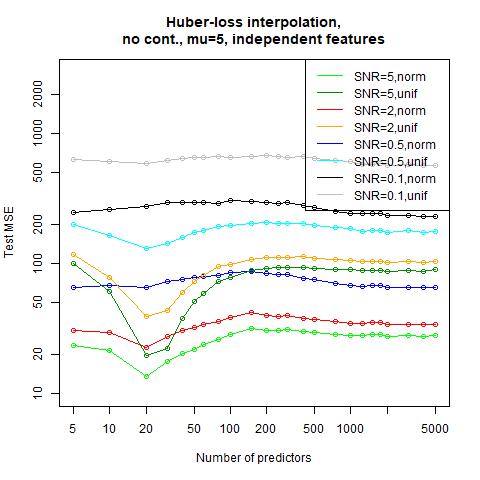} 
\end{center}
\caption{Test MSE of Huber-loss interpolation when trained on clean training data.} \label{fig:hubermu5indep}
\end{figure}

\begin{figure}[H]
\begin{center}
\includegraphics[width=5.25cm]{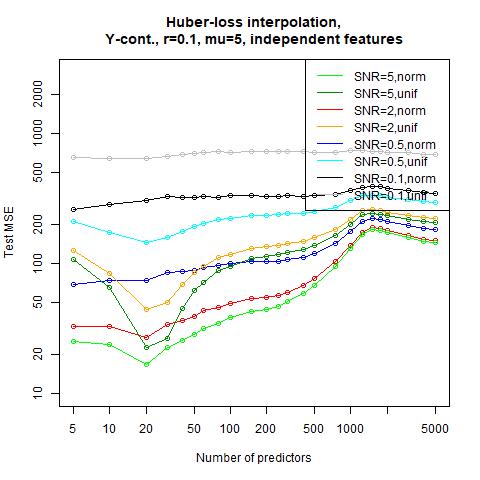} \includegraphics[width=5.25cm]{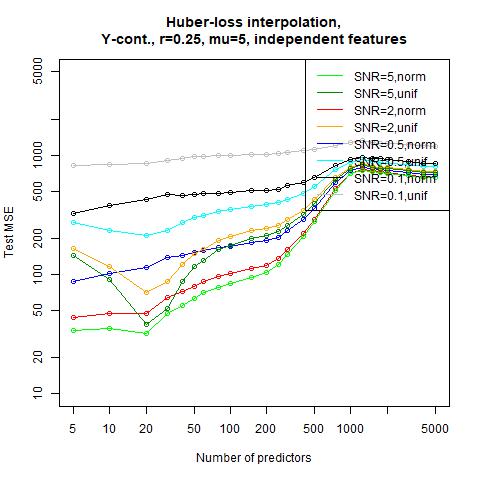} \\ \includegraphics[width=5.25cm]{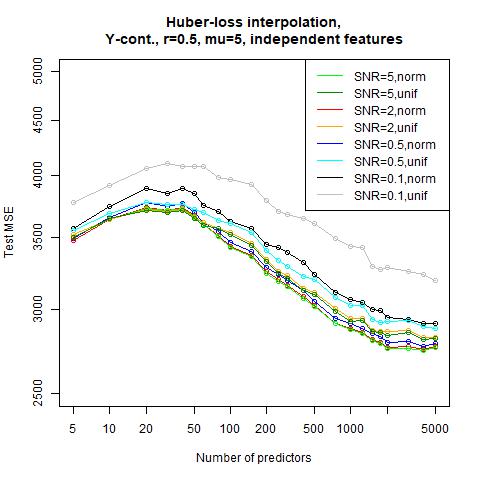} \includegraphics[width=5.25cm]{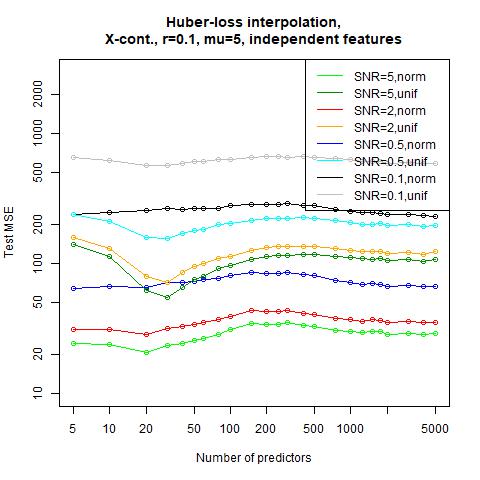} \\ \includegraphics[width=5.25cm]{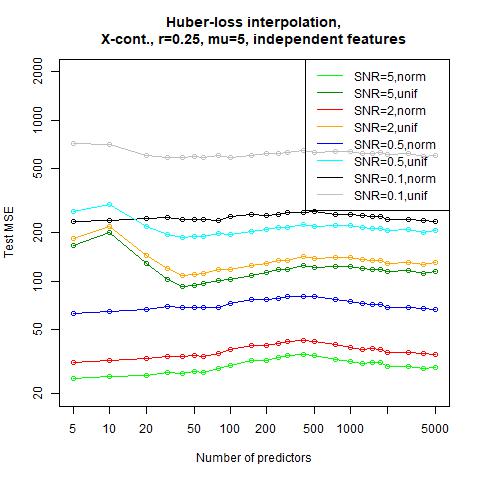} \includegraphics[width=5.25cm]{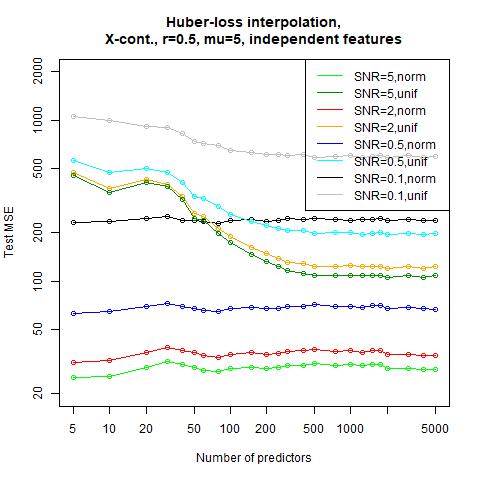} 
\end{center}
\caption{Test MSE of Huber-loss interpolation when trained on contaminated training data.} \label{fig:hubermu5indepcont}
\end{figure}

As one can observe in Fig. \ref{fig:hubermu5indep} and Fig. \ref{fig:hubermu5indepcont}, in contrast to the case $\mu=0$, there is no clear peak in the MSE curves for $X$-contamination and clean training data. In particular, for $Y$-contamination and $r=0.1$ and $r=0.25$, the MSE increases for growing $p$, while for $\mu=0$, it decreased after attaining a peak. 

\subsubsection{SLTS-based interpolation}

\begin{figure}[H]
\begin{center}
\includegraphics[width=7.5cm]{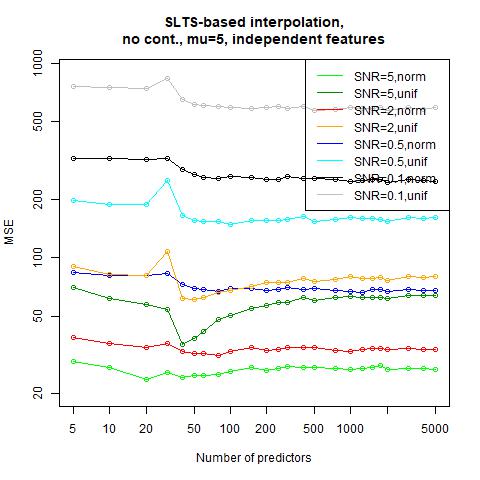} 
\end{center}
\caption{Test MSE of SLTS-based interpolation when trained on clean training data.} \label{fig:sltsmu5indep}
\end{figure}

\begin{figure}[H]
\begin{center}
\includegraphics[width=7.5cm]{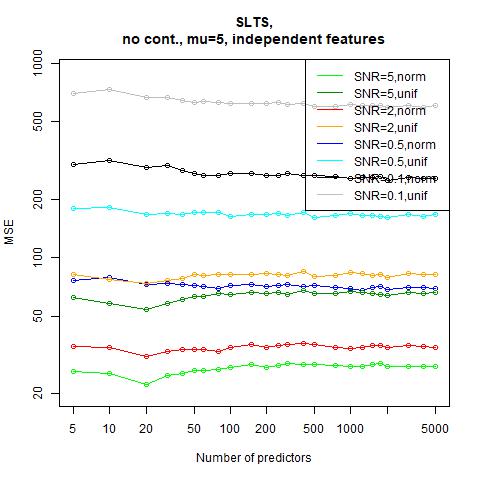} 
\end{center}
\caption{Test MSE of SLTS when trained on clean training data.}\label{fig:rawsltsmu5indep}
\end{figure}

\begin{figure}[H]
\begin{center}
\includegraphics[width=5.25cm]{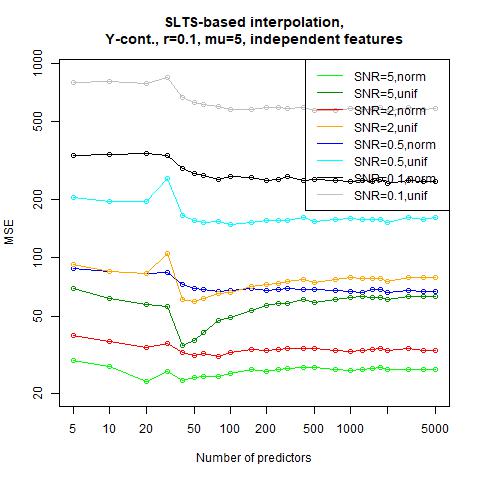} \includegraphics[width=5.25cm]{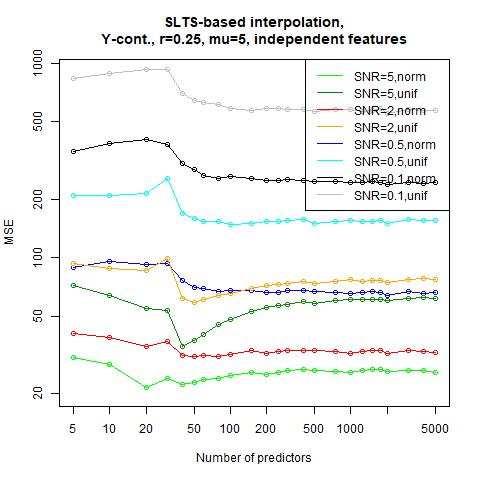} \\ \includegraphics[width=5.25cm]{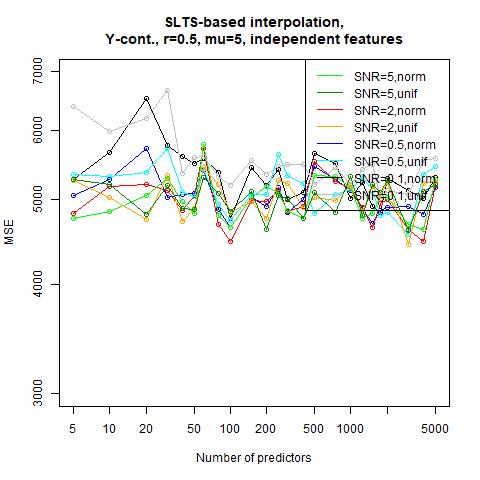} \includegraphics[width=5.25cm]{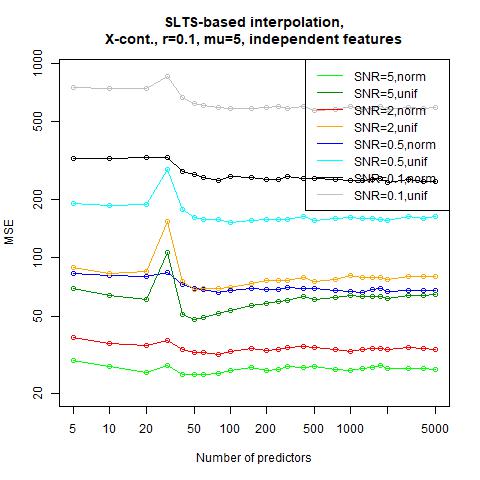} \\ \includegraphics[width=5.25cm]{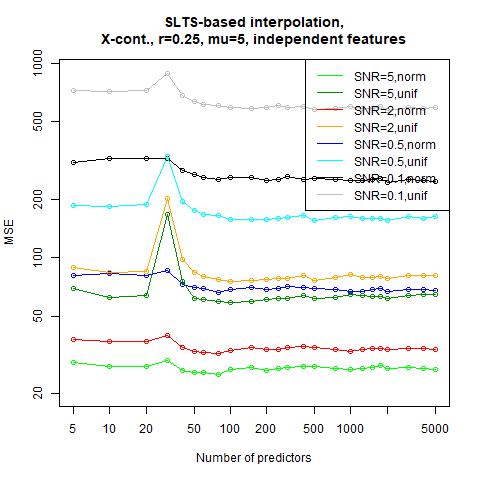} \includegraphics[width=5.25cm]{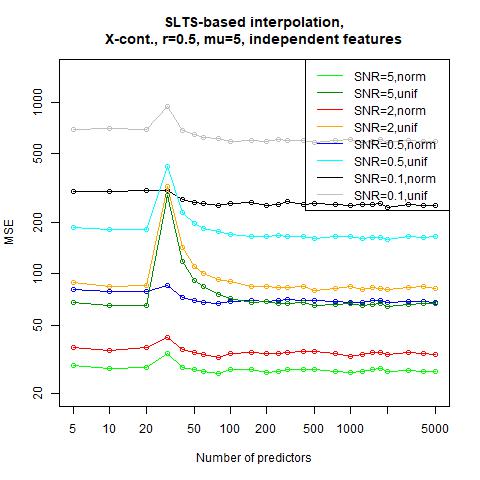} 
\end{center}
\caption{Test MSE of SLTS-based interpolation when trained on contaminated training data.}\label{fig:sltsmu5indepcont}
\end{figure}

\begin{figure}[H]
\begin{center}
\includegraphics[width=5.25cm]{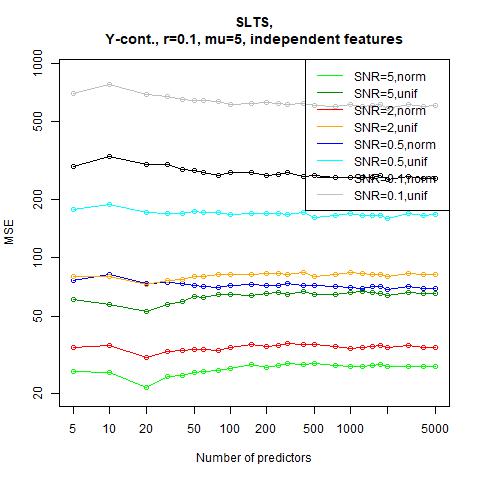} \includegraphics[width=5.25cm]{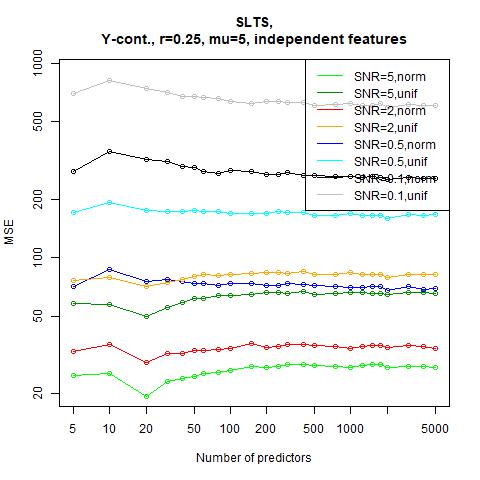} \\ \includegraphics[width=5.25cm]{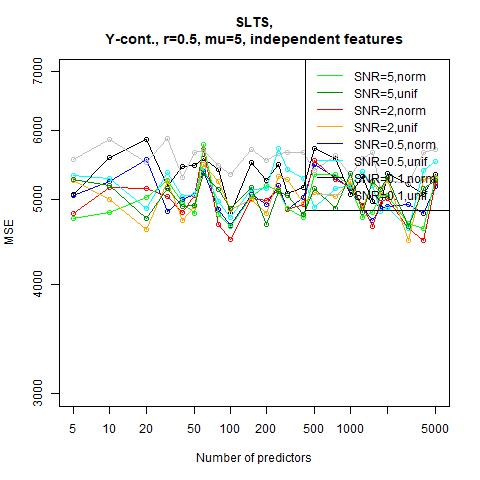} \includegraphics[width=5.25cm]{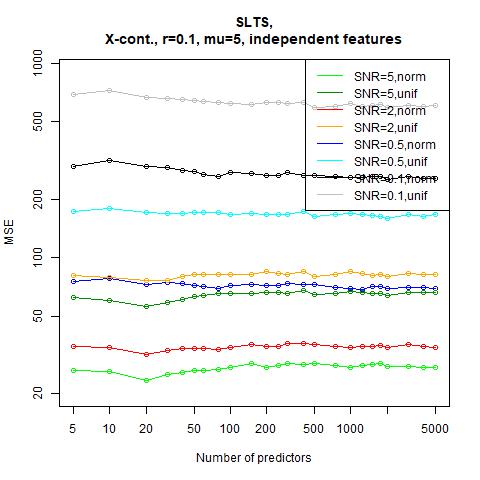} \\ \includegraphics[width=5.25cm]{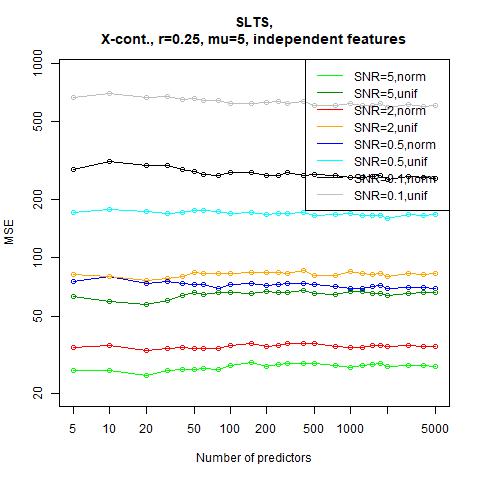} \includegraphics[width=5.25cm]{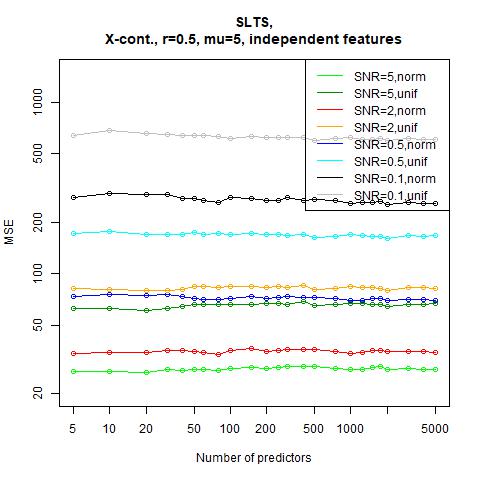} 
\end{center}
\caption{Test MSE of SLTS when trained on contaminated training data.} \label{fig:rawsltsmu5indepcont}
\end{figure}

The MSE curves depicted in Fig. \ref{fig:sltsmu5indep}, Fig. \ref{fig:sltsmu5indepcont}, Fig. \ref{fig:rawsltsmu5indep} and Fig. \ref{fig:rawsltsmu5indepcont} resemble those from the case $\mu=0$, except for the case of $X$-contamination and $r=0.5$, where the curves fluctuate instead of being nearly constant for raw SLTS and decreasing for larger $p$ for SLTS-based interpolation, respectively. 

\newpage

\subsection{$n=200$} \

\subsubsection{Minimum $l_2$-norm interpolation}

\begin{figure}[H]
\begin{center}
\includegraphics[width=5.25cm]{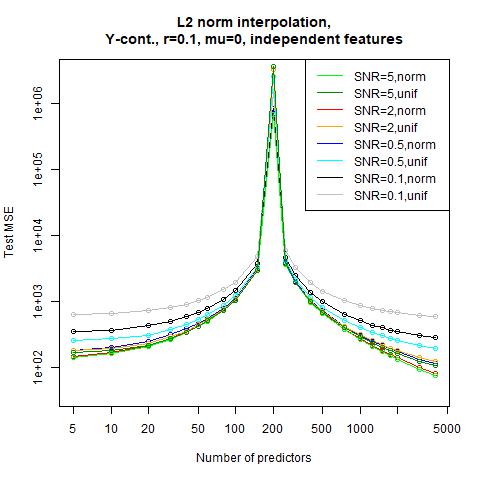} 
\includegraphics[width=5.25cm]{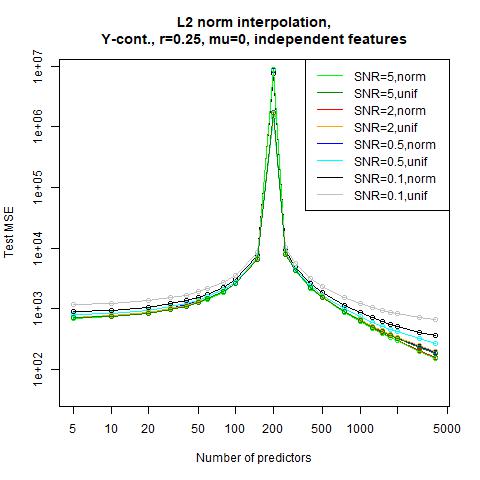} \\
 \includegraphics[width=5.25cm]{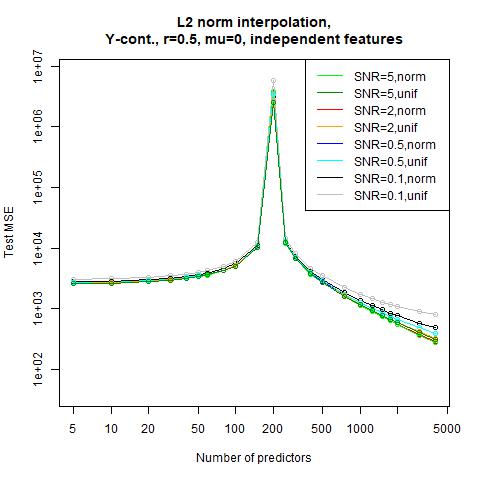} 
\includegraphics[width=5.25cm]{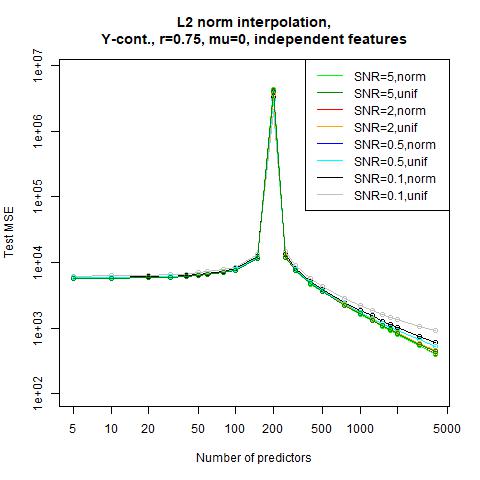}\\
 \includegraphics[width=5.25cm]{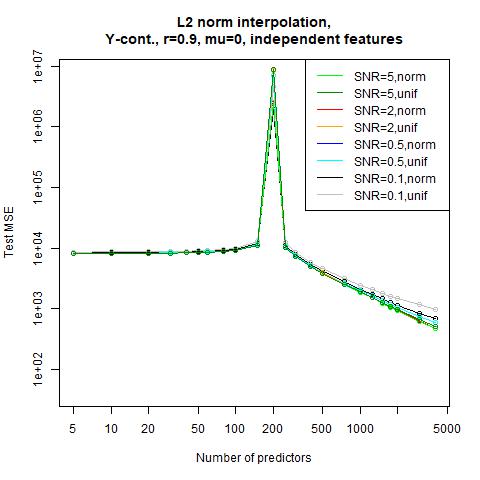} 
\end{center}
\caption{Test MSE of minimum $l_2$-norm interpolation when trained on $Y$-contaminated training data.} \label{fig:minl2mu0indepcontn200}
\end{figure}

In contrast to the case $n=50$ in Fig. \ref{fig:minl2mu0indepcont}, there is no longer a minimum at $p=s$ in the MSE curves depicted in Fig. \ref{fig:minl2mu0indepcontn200}, but a pronounced peak at $p=n$. After the peak, the MSE strictly decreases. The attained MSE values at $p=5000$ are larger than for the case $n=50$, but the curves seem to decrease further if $p$ grew larger.

\subsubsection{Huber-loss interpolation}

\begin{figure}[H]
\begin{center}
\includegraphics[width=5.25cm]{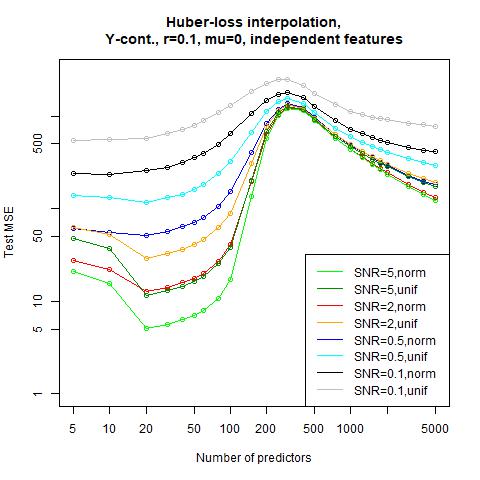} 
\includegraphics[width=5.25cm]{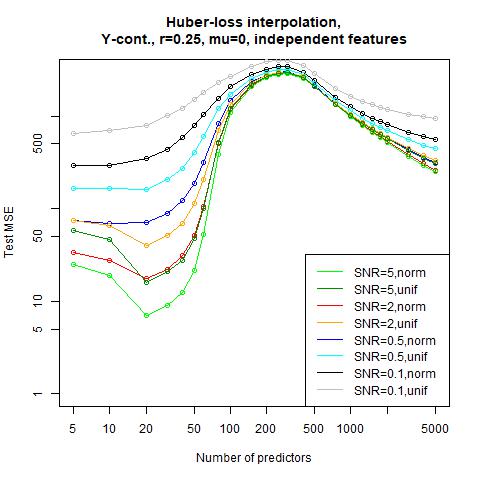} \\
 \includegraphics[width=5.25cm]{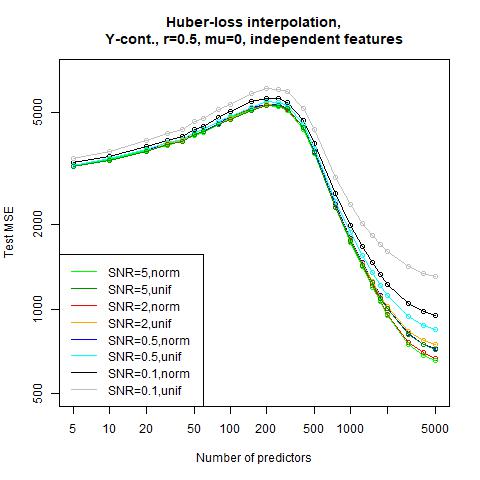} 
\includegraphics[width=5.25cm]{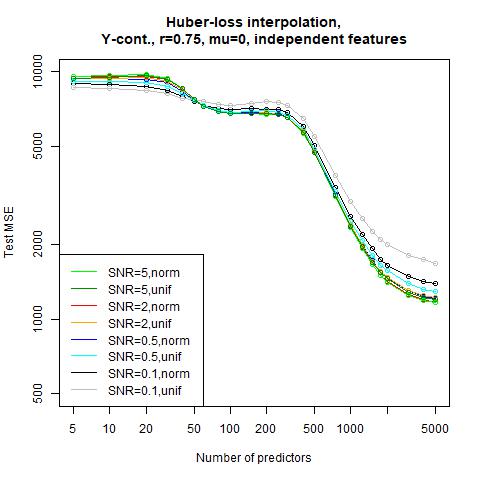}\\
 \includegraphics[width=5.25cm]{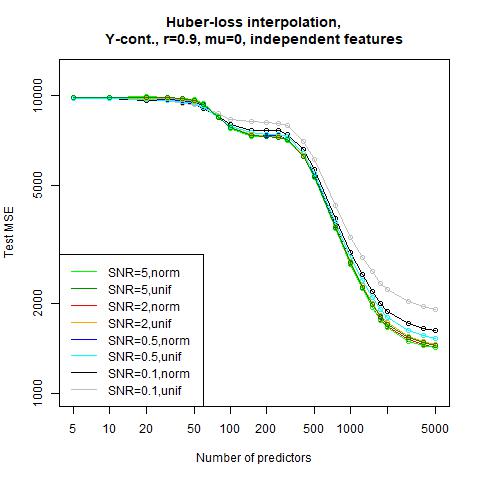} 
\end{center}
\caption{Test MSE of Huber-loss interpolation when trained on $Y$-contaminated training data.} \label{fig:hubermu0indepcontn200}
\end{figure}

The MSE curves in Fig. \ref{fig:hubermu0indepcontn200} are similary than those in Fig. \ref{fig:hubermu0indepcont} for $r \in \{0.1,0.25\}$, while the MSE values are slightly smaller. For $r=0.5$, the MSE values are larger, but the shape of the curves is similar as for the case $n=50$. The peak is attained shortly after $p=n$. For $r \in \{0.75,0.9\}$, the MSE is very large for small $p$, decreases slightly in order to remain at a plateau, and significantly decreases once $p>n$, although the MSE values at $p=5000$ are considerably larger than for smaller contamination radii.

\newpage

\subsection{$n=200$, $c_{out}=10000$} \

\subsubsection{Minimum $l_2$-norm interpolation}

\begin{figure}[H]
\begin{center}
\includegraphics[width=5.25cm]{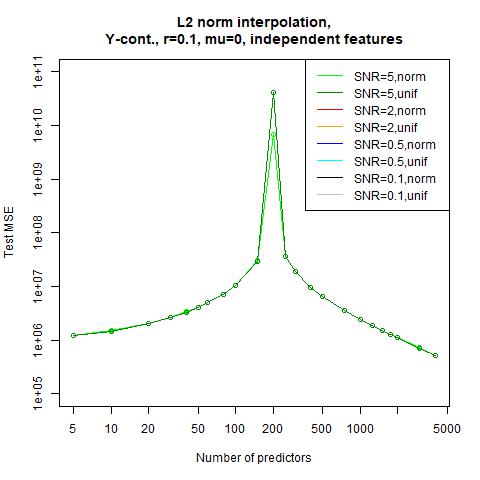} 
\includegraphics[width=5.25cm]{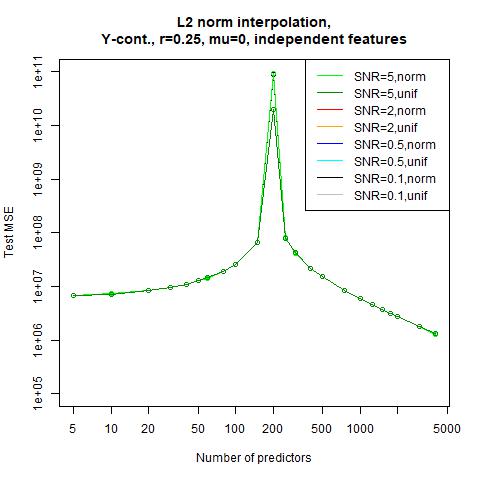} \\
 \includegraphics[width=5.25cm]{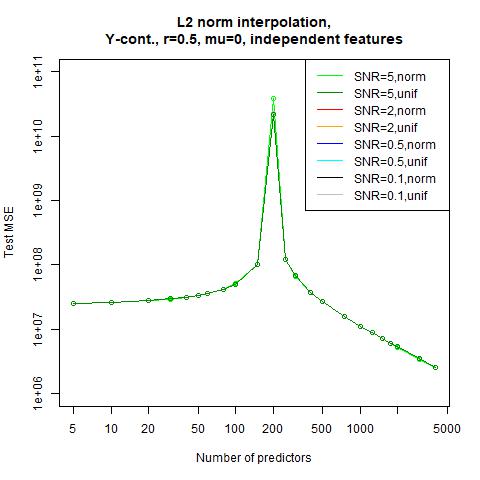} 
\includegraphics[width=5.25cm]{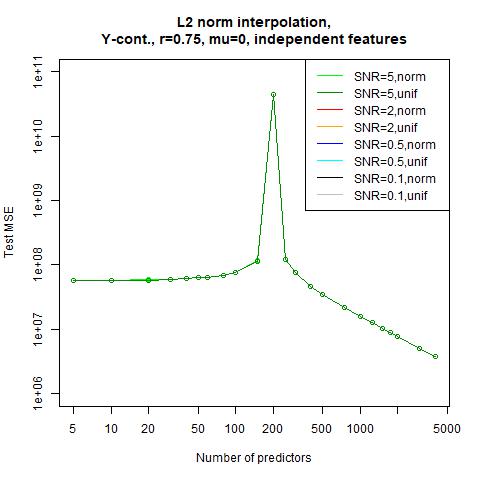}\\
 \includegraphics[width=5.25cm]{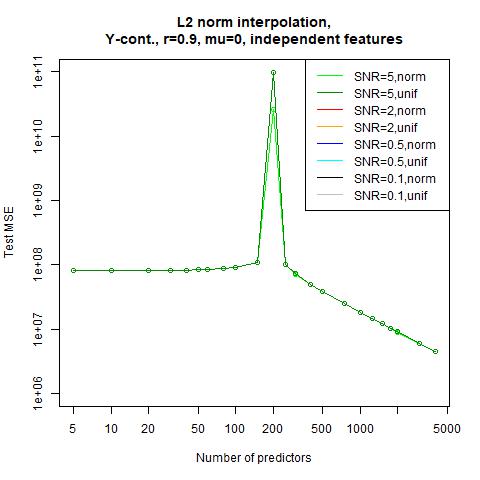} 
\end{center}
\caption{Test MSE of minimum $l_2$-norm interpolation when trained on $Y$-contaminated training data.} \label{fig:minl2mu0indepcontn200r10000}
\end{figure}

The MSE curves in Fig. \ref{fig:minl2mu0indepcontn200r10000} resemble those from Fig. \ref{fig:minl2mu0indepcontn200} for $c_{out}=100$, although the MSE values are clearly larger.

\subsubsection{Huber-loss interpolation}

\begin{figure}[H]
\begin{center}
\includegraphics[width=5.25cm]{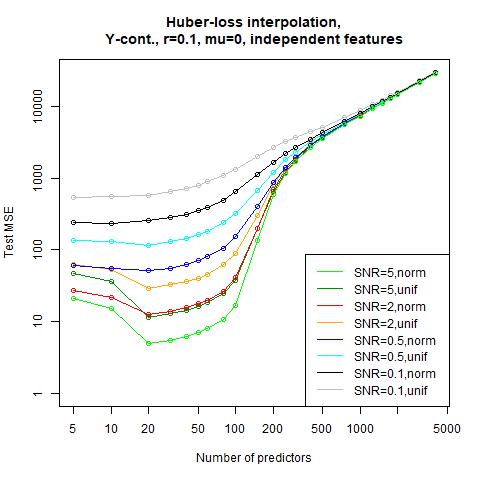} 
\includegraphics[width=5.25cm]{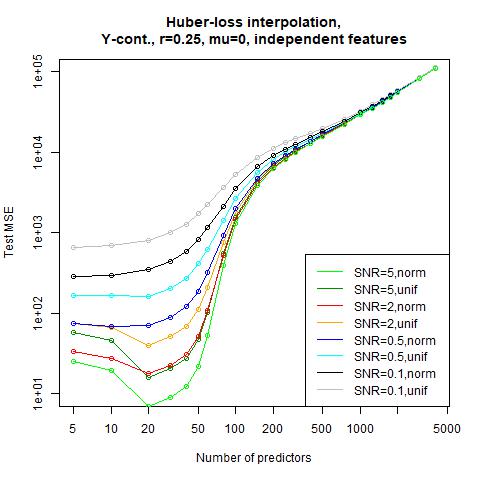} \\
 \includegraphics[width=5.25cm]{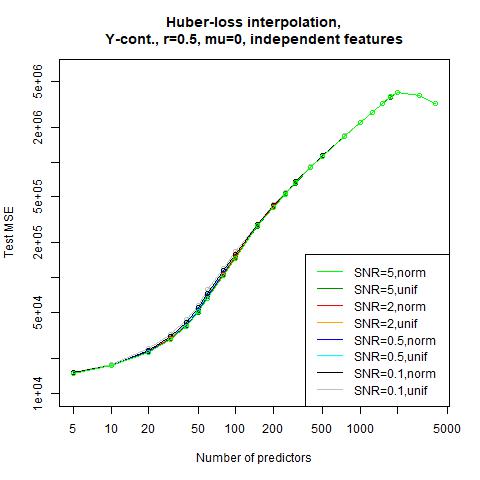} 
\includegraphics[width=5.25cm]{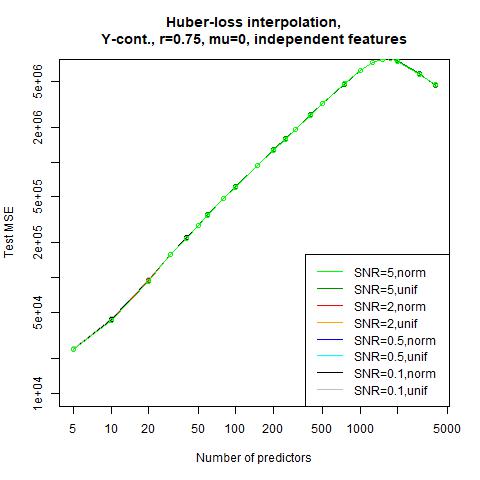}\\
 \includegraphics[width=5.25cm]{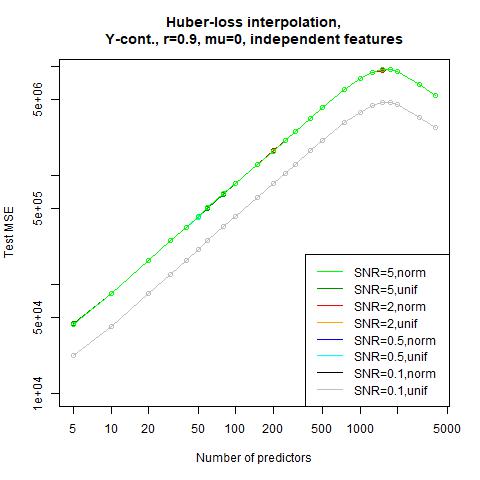} 
\end{center}
\caption{Test MSE of Huber-loss interpolation when trained on $Y$-contaminated training data.} \label{fig:hubermu0indepcontn200r10000}
\end{figure}

For $r \in \{0.1,0.25\}$, the MSE curves in Fig. \ref{fig:hubermu0indepcontn200r10000} attain a minimum at $p=s$ and monotonically increase thereafter. In contrast to the case $c_{out}=100$ visualized in Fig. \ref{fig:hubermu0indepcontn200}, the curves do not decrease even for the largest values of $p$ considered. For $r \ge 0.5$, the MSE monotonically increases and attains a peak at $p=2000$ in order to seemingly decrease as $p$ grows further.






\newpage

\section{Training errors} \label{sec:trainerr}

\subsection{Independent features, $\mu=0$}

\subsubsection{Minimum $l_2$-norm interpolation}

\begin{figure}[H]
\begin{center}
\includegraphics[width=7.5cm]{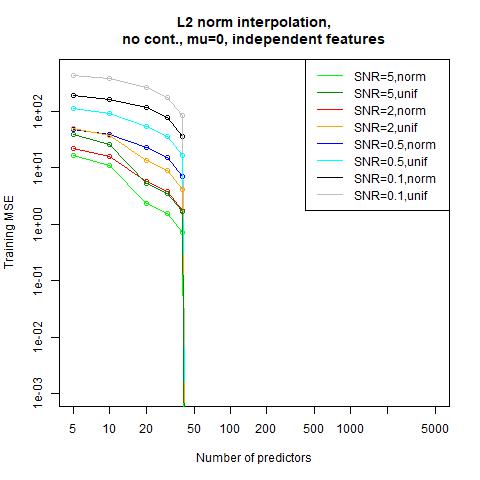} 
\end{center}
\caption{Training MSE of minimum $l_2$-norm interpolation when trained on clean training data.} \label{fig:minl2mu0indeptrain}
\end{figure}

\begin{figure}[H]
\begin{center}
\includegraphics[width=5.25cm]{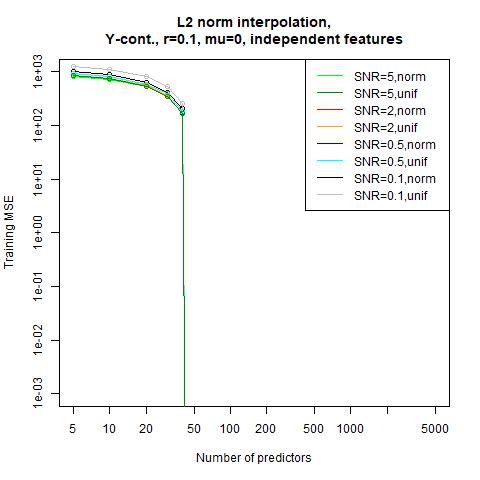} \includegraphics[width=5.25cm]{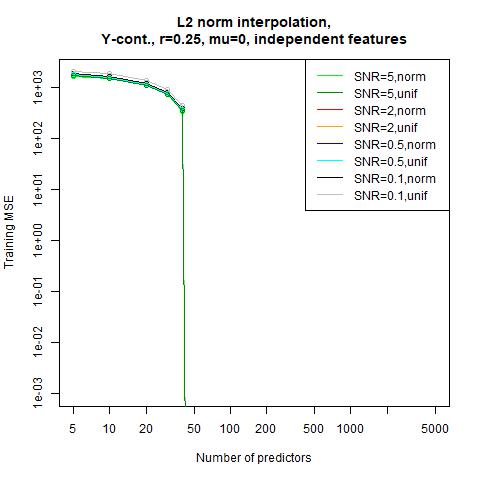} \\ \includegraphics[width=5.25cm]{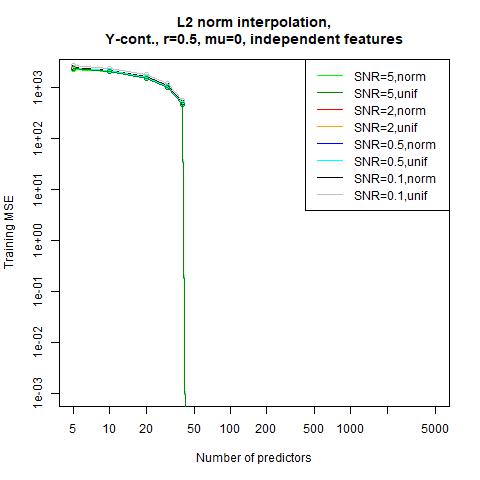} \includegraphics[width=5.25cm]{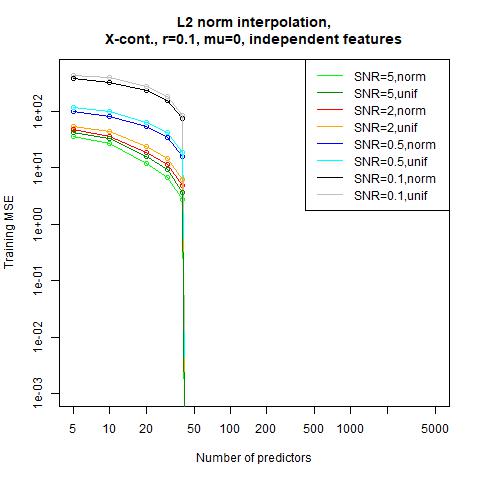} \\ \includegraphics[width=5.25cm]{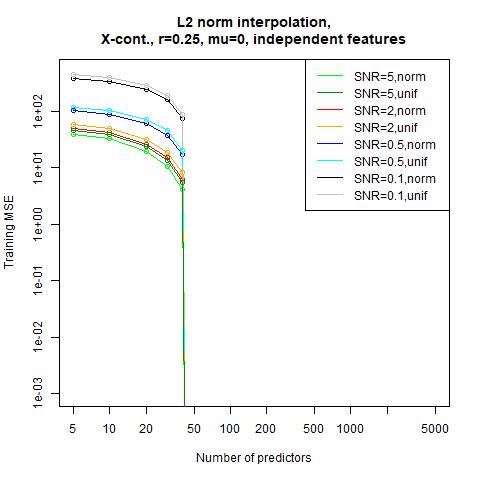} \includegraphics[width=5.25cm]{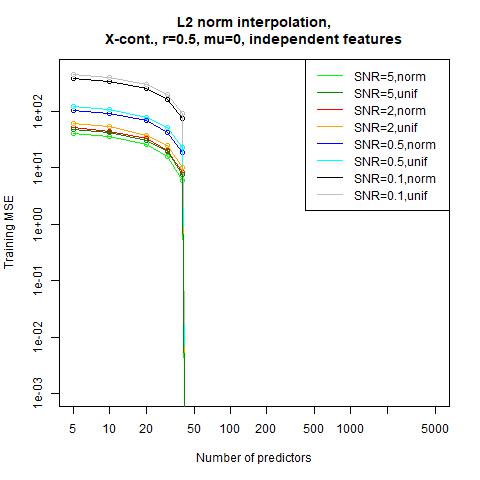} 
\end{center}
\caption{Training MSE of minimum $l_2$-norm interpolation when trained on contaminated training data.} \label{fig:minl2mu0indeptraincont}
\end{figure}

By interpolation, the training error vanishes once $p>n$, as depicted in Fig. \ref{fig:minl2mu0indeptrain} and Fig. \ref{fig:minl2mu0indeptraincont}. As expected, the training error is higher at low $p$ for $Y$-contaminated as for $X$-contaminated and clean data.

\subsubsection{Huber-loss interpolation}

\begin{figure}[H]
\begin{center}
\includegraphics[width=7.5cm]{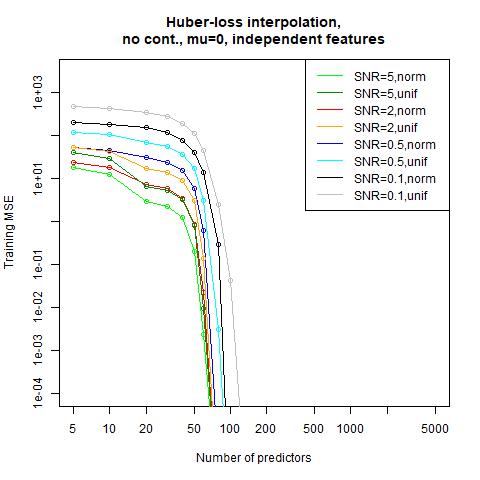} 
\end{center}
\caption{Training MSE of Huber-loss interpolation when trained on clean training data.} \label{fig:hubermu0indeptrain}
\end{figure}

\begin{figure}[H]
\begin{center}
\includegraphics[width=5.25cm]{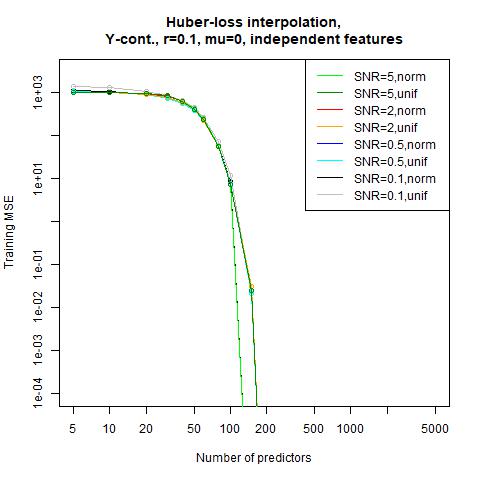} \includegraphics[width=5.25cm]{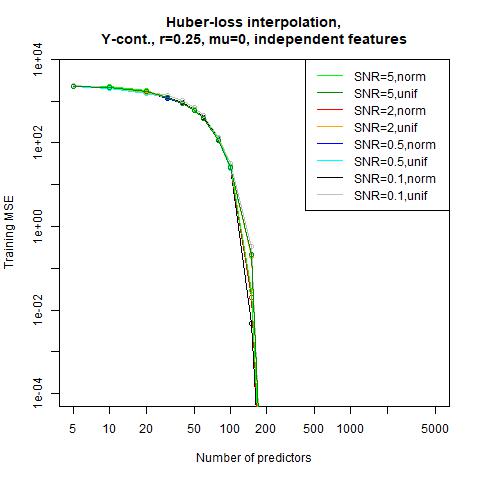} \\ \includegraphics[width=5.25cm]{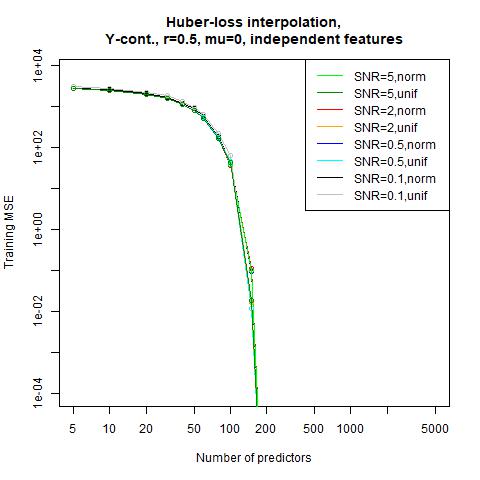} \includegraphics[width=5.25cm]{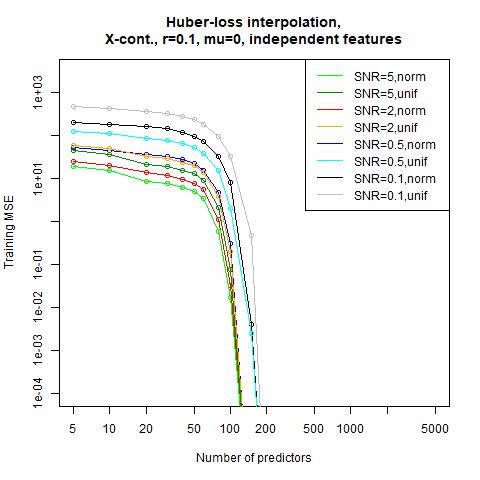} \\ \includegraphics[width=5.25cm]{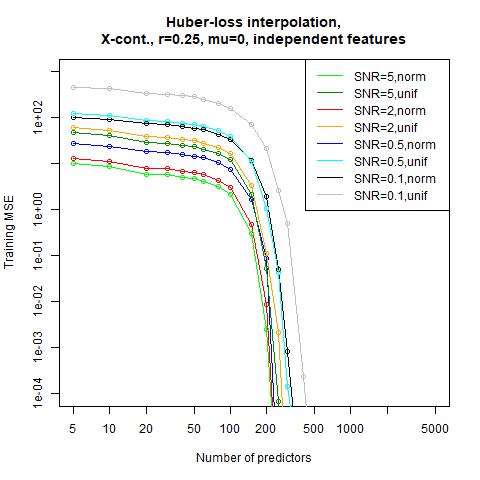} \includegraphics[width=5.25cm]{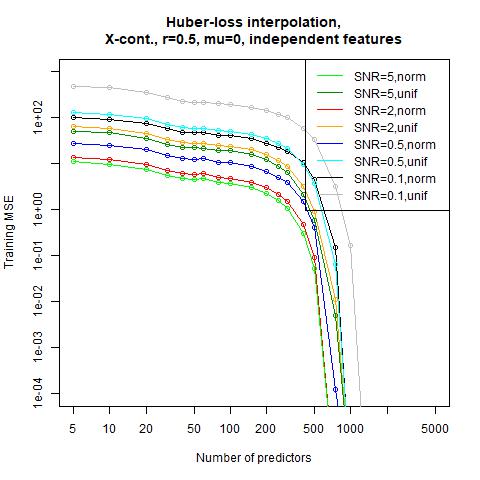} 
\end{center}
\caption{Training MSE of Huber-loss interpolation when trained on contaminated training data.}\label{fig:hubermu0indeptraincont}
\end{figure}

The training error curves in Fig. \ref{fig:hubermu0indeptrain} and Fig. \ref{fig:hubermu0indeptraincont} show that the training error does not vanish directly at $p=n$ but at some higher $p$, depending on the SNR and the contamination. In particular, $X$ contamination and a high $r$ lead to a late vanish of the training error, which requires up to $p=1250$. For small $p$ however, the training error is similar to that of minimum $l_2$-norm interpolation.

\subsubsection{Tukey-loss interpolation}

\begin{figure}[H]
\begin{center}
\includegraphics[width=7.5cm]{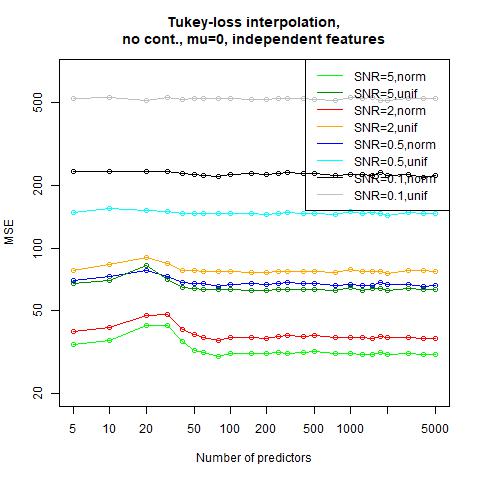} 
\end{center}
\caption{Training MSE of Tukey-loss interpolation when trained on clean training data.} \label{fig:tukeymu0indeptrain}
\end{figure}

\begin{figure}[H]
\begin{center}
\includegraphics[width=5.25cm]{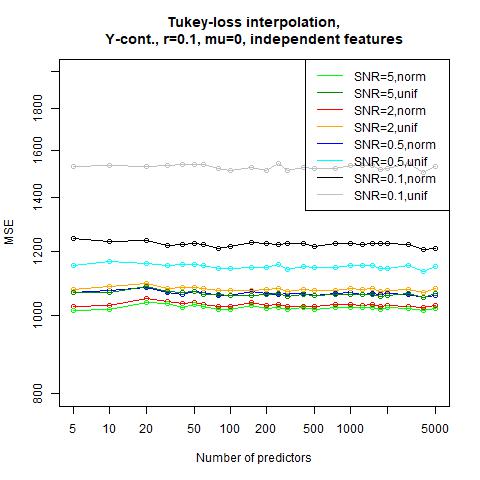} \includegraphics[width=5.25cm]{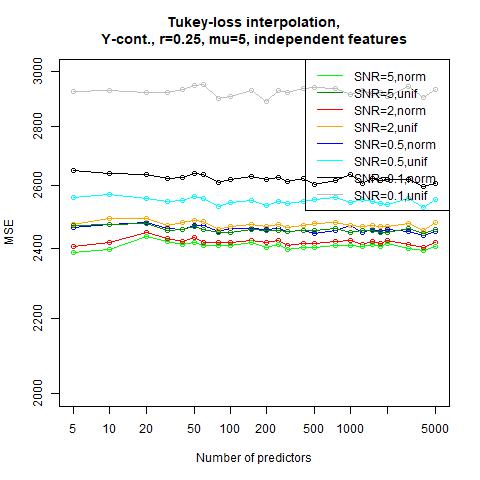} \\ \includegraphics[width=5.25cm]{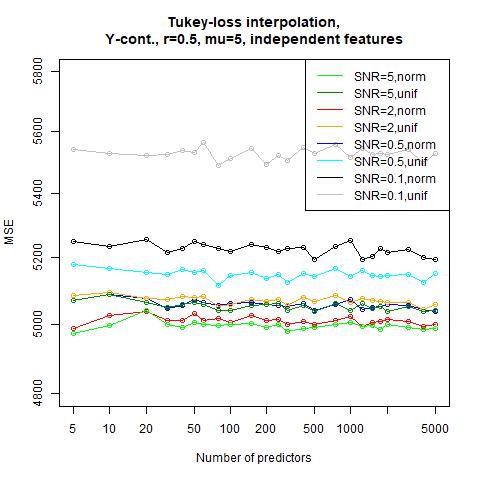} \includegraphics[width=5.25cm]{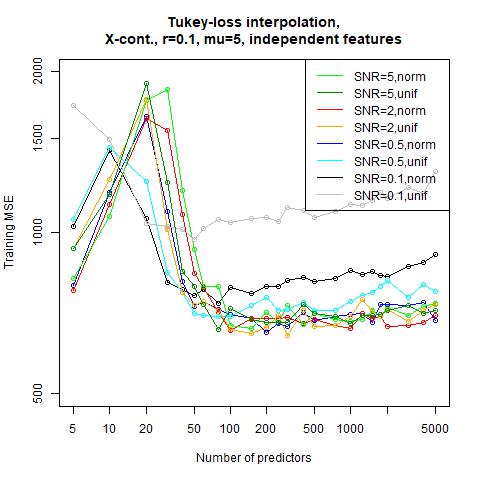} \\ \includegraphics[width=5.25cm]{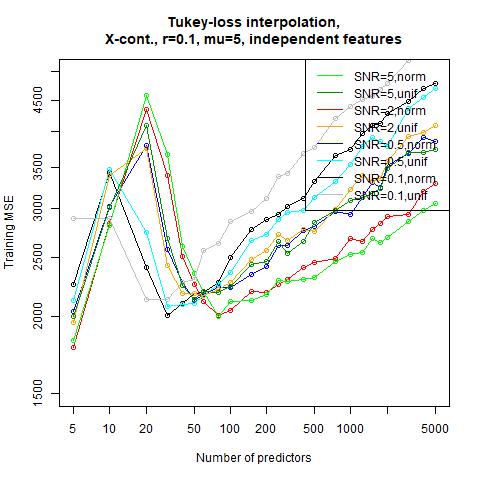} \includegraphics[width=5.25cm]{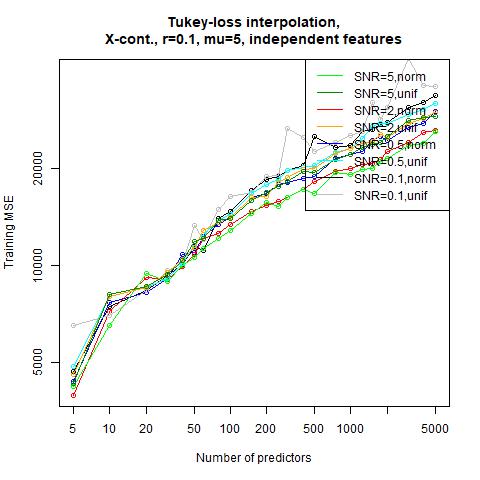} 
\end{center}
\caption{Training MSE of Tukey-loss interpolation when trained on contaminated training data.} \label{fig:tukeymu0indeptraincont}
\end{figure}

The training MSE for Tukey-loss interpolation, as depicted in Fig. \ref{fig:tukeymu0indeptrain} and Fig. \ref{fig:tukeymu0indeptraincont}, remains nearly constant for $Y$-contaminated and clean training data. It is not surprising that the training error does not vanish, as the Tukey loss is a redescending loss so that the data cannot be interpolated unless all residuals are smaller than the threshold in absolute value. As for $X$-contamination and $r=0.25$ and $r=0.5$, the training error even increases for growing $p$.

\subsubsection{SLTS-based interpolation}

\begin{figure}[H]
\begin{center}
\includegraphics[width=7.5cm]{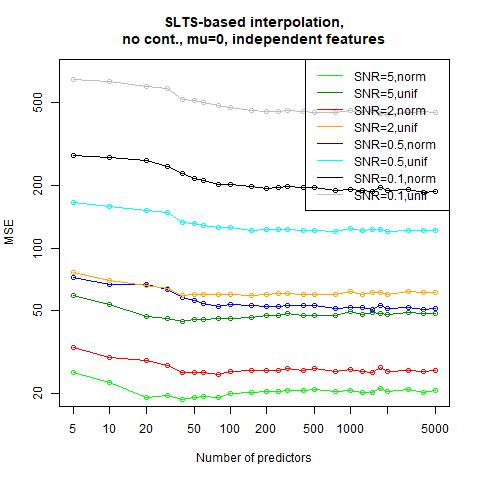} 
\end{center}
\caption{Training MSE of SLTS-based interpolation when trained on clean training data.} \label{fig:sltsmu0indeptrain}
\end{figure}

\begin{figure}[H]
\begin{center}
\includegraphics[width=7.5cm]{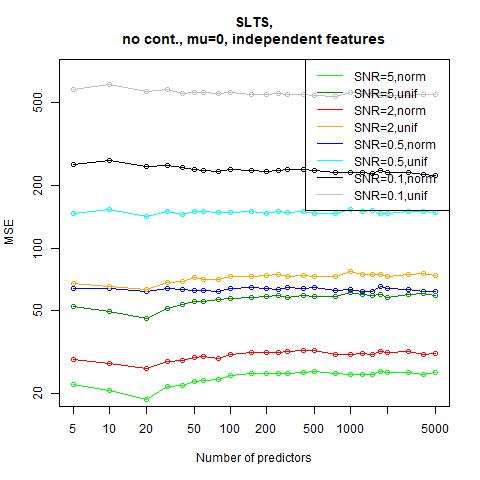} 
\end{center}
\caption{Training MSE of SLTS when trained on clean training data.}\label{fig:rawsltsmu0indeptrain}
\end{figure}

\begin{figure}[H]
\begin{center}
\includegraphics[width=5.25cm]{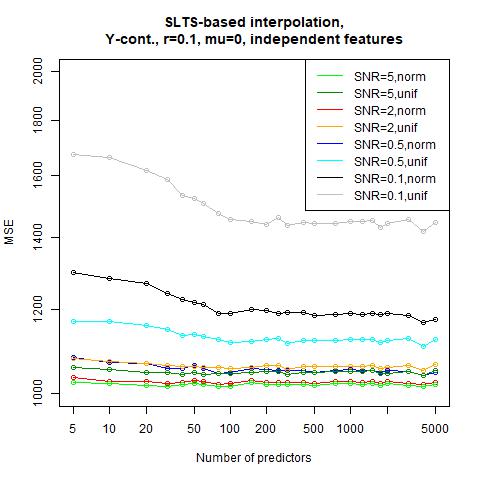} \includegraphics[width=5.25cm]{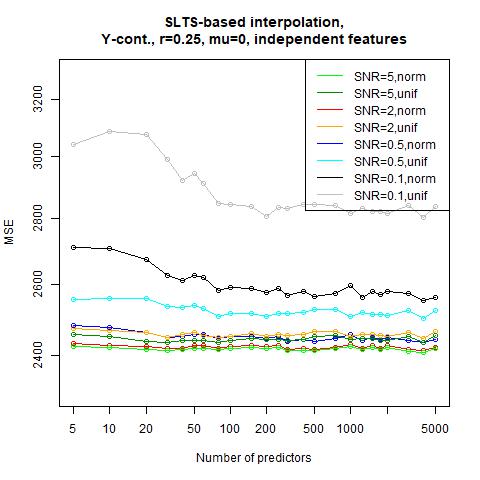} \\ \includegraphics[width=5.25cm]{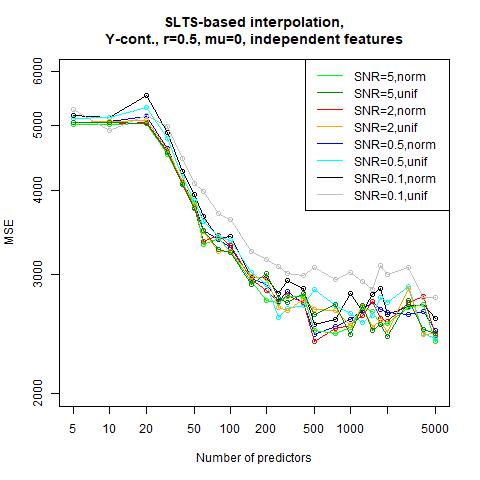} \includegraphics[width=5.25cm]{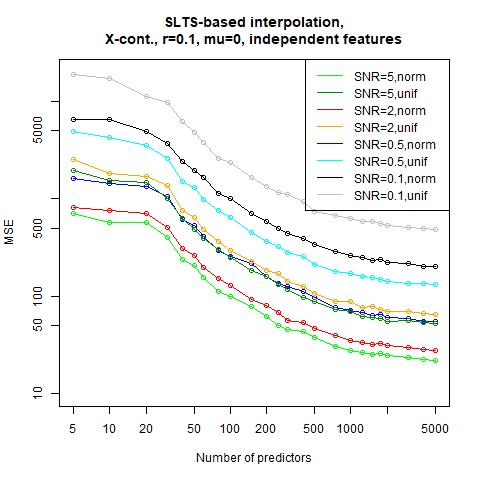} \\ \includegraphics[width=5.25cm]{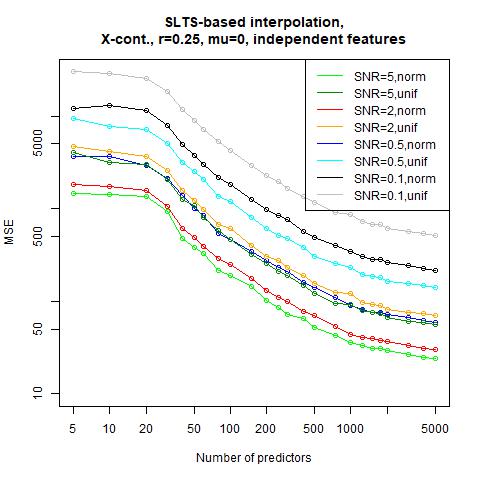} \includegraphics[width=5.25cm]{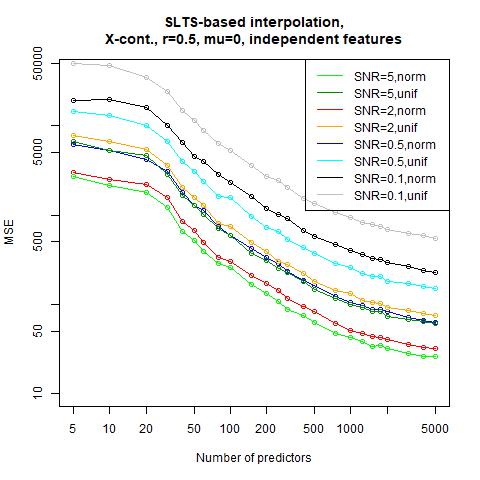} 
\end{center}
\caption{Training MSE of SLTS-based interpolation when trained on contaminated training data.}\label{fig:sltsmu0indeptraincont}
\end{figure}

\begin{figure}[H]
\begin{center}
\includegraphics[width=5.25cm]{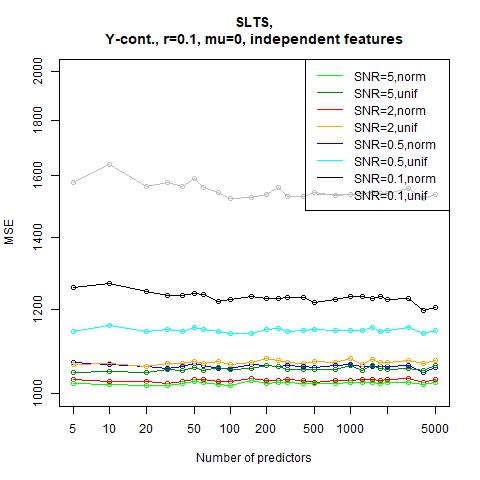} \includegraphics[width=5.25cm]{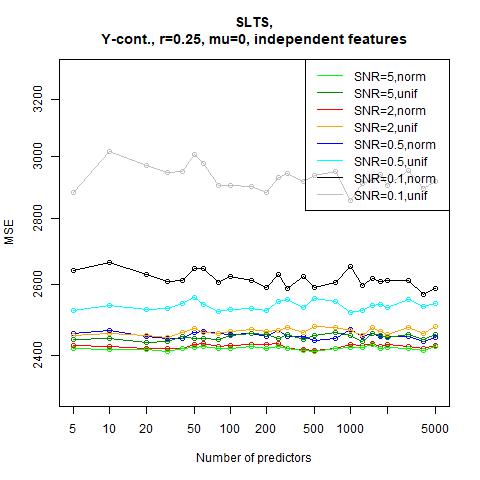} \\ \includegraphics[width=5.25cm]{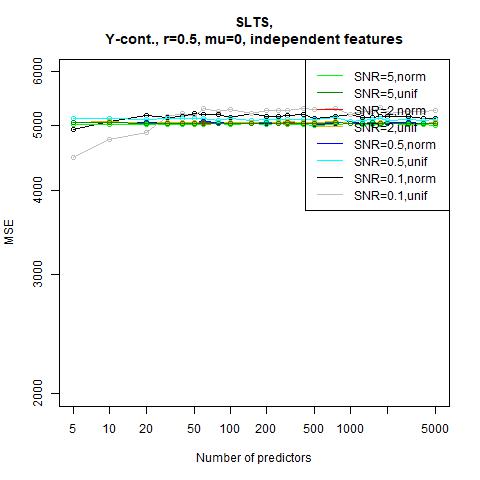} \includegraphics[width=5.25cm]{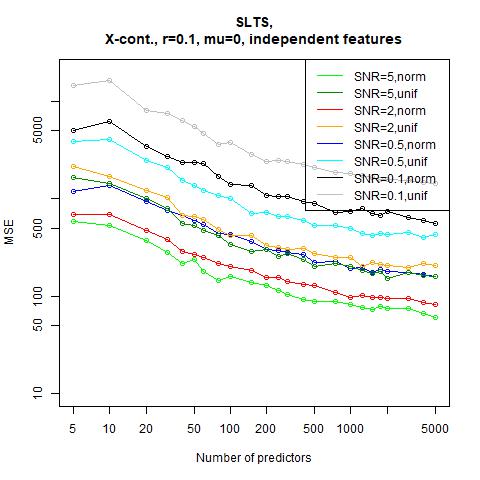} \\ \includegraphics[width=5.25cm]{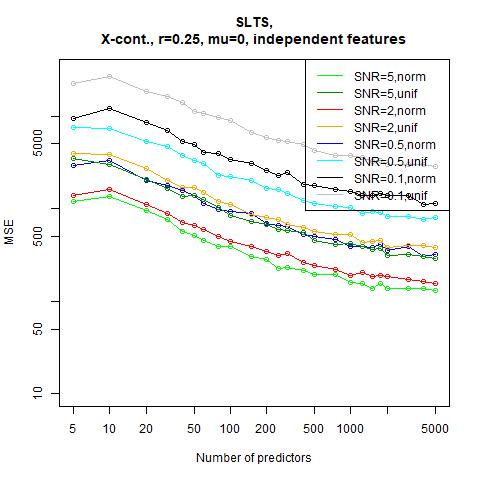} \includegraphics[width=5.25cm]{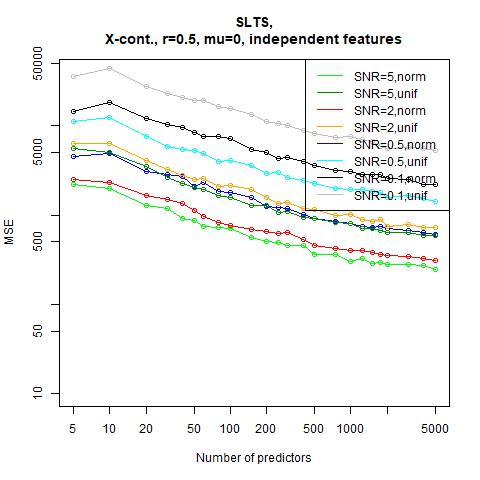} 
\end{center}
\caption{Training MSE of SLTS when trained on contaminated training data.}\label{fig:rawsltsmu0indeptraincont}
\end{figure}

Neither SLTS-based interpolation nor SLTS leads to a vanishing training loss, since the model is trained only on a clean subset. As for the case of clean training data, as depicted in Fig. \ref{fig:sltsmu0indeptrain} for SLTS-based interpolation, one can observe a slight decrease of the training error as $p$ increases, while the training MSE for SLTS increases after $p=s$, as visualized in Fig. \ref{fig:rawsltsmu0indeptrain}. For contaminated data, as seen in Fig. \ref{fig:sltsmu0indeptraincont} and Fig. \ref{fig:rawsltsmu0indeptraincont}, the training error decreases for growing $p$ for $X$-contamination, while remaining nearly constant for $Y$-contamination and $r=0.1$ and $r=0.25$, except for SLTS-based interpolation and an SNR of 0.1, where a slight decrease can be observed. For SLTS-based interpolation and $Y$-contamination with $r=0.5$, one can observe a light peak at $p=20$ and a considerable descent of the training error afterwards, while the training error remains constantly high for the raw SLTS.

\subsubsection{Boosting-based interpolation}

\begin{figure}[H]
\begin{center}
\includegraphics[width=7.5cm]{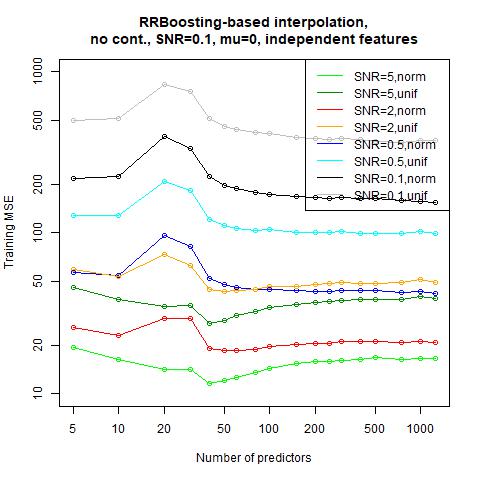} 
\end{center}
\caption{Training MSE of RRBoost-based interpolation when trained on clean training data.} \label{fig:rrboostmu0indeptrain}
\end{figure}

\begin{figure}[H]
\begin{center}
\includegraphics[width=7.5cm]{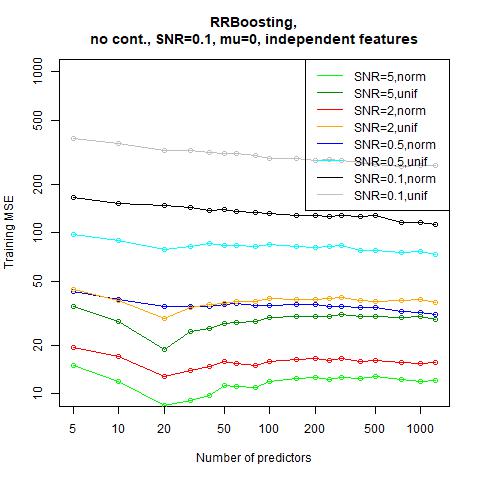} 
\end{center}
\caption{Training MSE of RRBoost when trained on clean training data.} \label{fig:rawrrboostmu0indeptrain}
\end{figure}

\begin{figure}[H]
\begin{center}
\includegraphics[width=5.25cm]{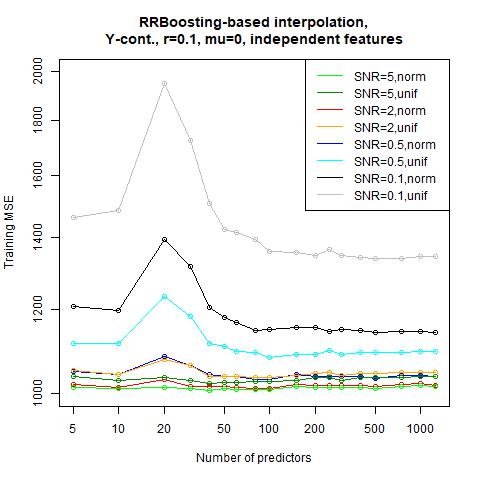} \includegraphics[width=5.25cm]{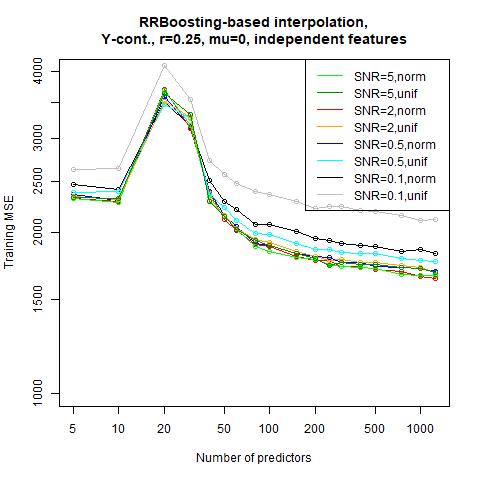} \\ \includegraphics[width=5.25cm]{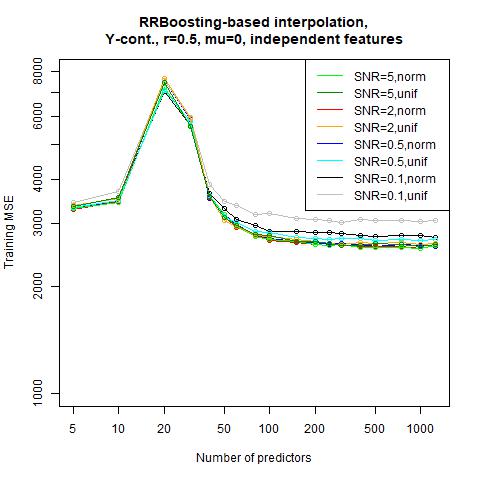} \includegraphics[width=5.25cm]{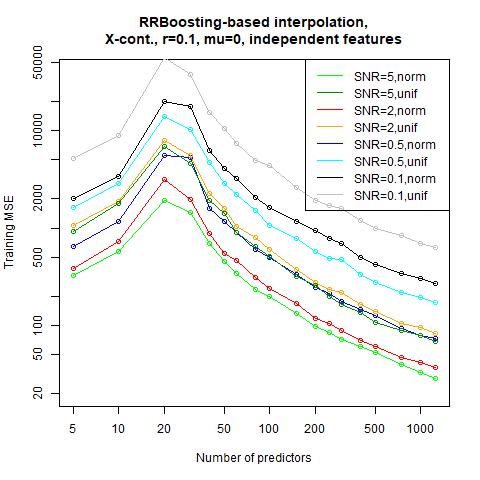} \\ \includegraphics[width=5.25cm]{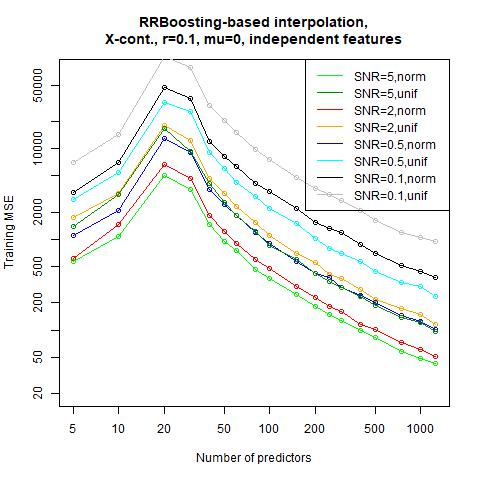} \includegraphics[width=5.25cm]{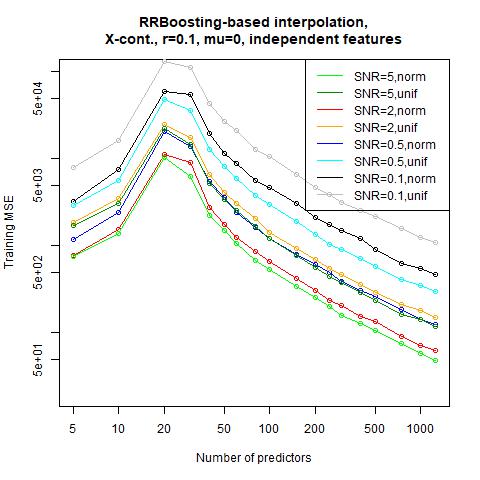} 
\end{center}
\caption{Training MSE of RRBoost-based interpolation when trained on contaminated training data.}\label{fig:rrboostmu0indeptraincont}
\end{figure}

\begin{figure}[H]
\begin{center}
\includegraphics[width=5.25cm]{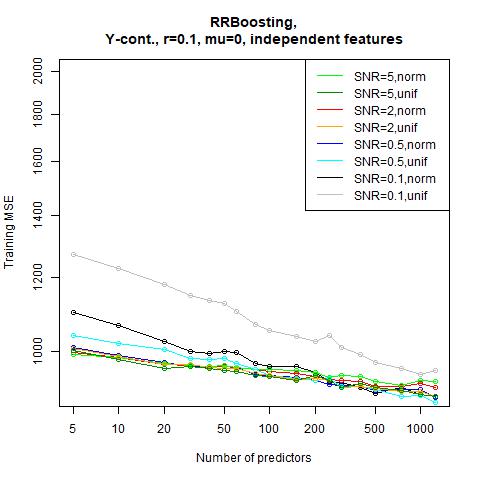} \includegraphics[width=5.25cm]{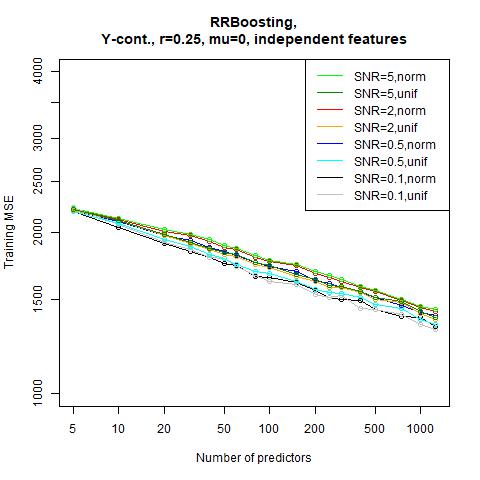} \\ \includegraphics[width=5.25cm]{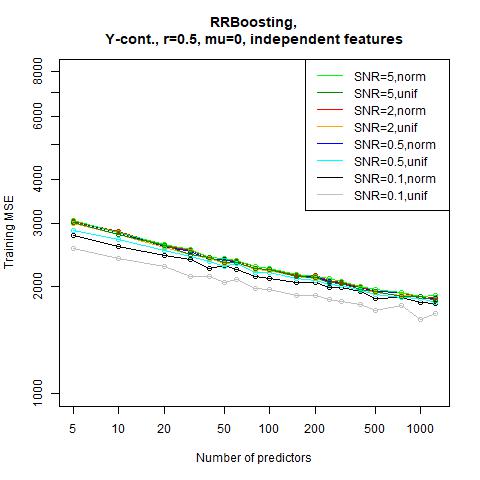} \includegraphics[width=5.25cm]{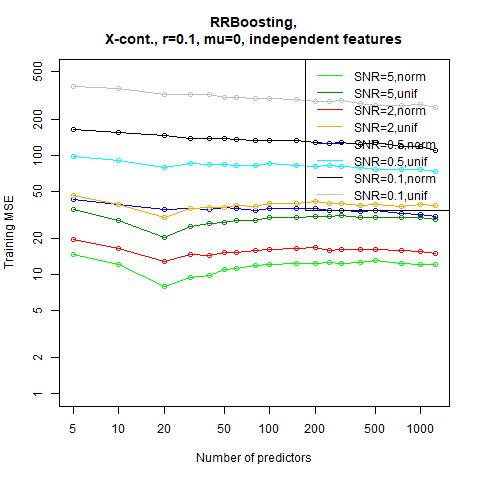} \\ \includegraphics[width=5.25cm]{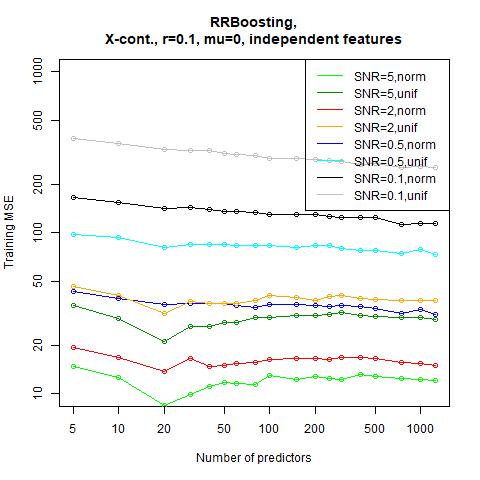} \includegraphics[width=5.25cm]{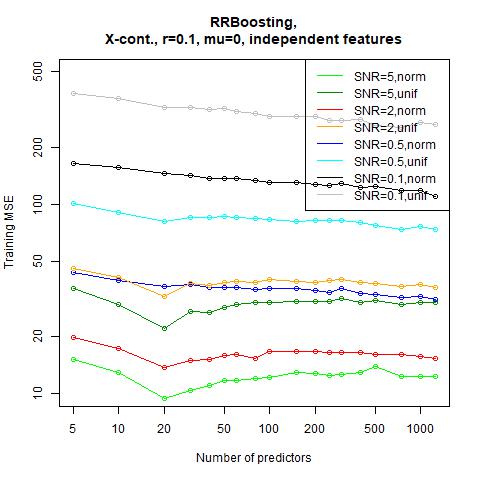} 
\end{center}
\caption{Training MSE of RRBoost when trained on contaminated training data.}\label{fig:rawrrboostmu0indeptraincont}
\end{figure}

For RRBoost-based interpolation, Fig. \ref{fig:rrboostmu0indeptrain} reveals that on clean data, one can observe a  peak at $p=20$ and a decrease of the training error afterwards for an SNR of 0.1 and 0.5, while for an SNR of 2 and 5, it slightly increases for growing $p$. The training MSE for RRBoost itself attains a minimum at $p=s$ and slightly increases afterwards, in order to remain constant for large $p$, as shown in Fig. \ref{fig:rawrrboostmu0indeptrain}.
One can see in Fig. \ref{fig:rrboostmu0indeptraincont} for RRBoost-based interpolation for $Y$-contamination, there is a pronounced peak at $p=20$ and a decrease of the training error afterwards. For $X$-contamination, the training error monotonically decreases after attaining a peak at $p=20$. For RRBoosting, as visualized in Fig. \ref{fig:rawrrboostmu0indeptraincont}, one can observe a nearly monotonically slight decrease of the training error for growing $p$, which is steeper for $Y$-contamination than for $X$-contamination and clean data.

\subsection{Spiked covariance design, $\mu=0$}

\subsubsection{Minimum $l_2$-norm interpolation}

\begin{figure}[H]
\begin{center}
\includegraphics[width=7.5cm]{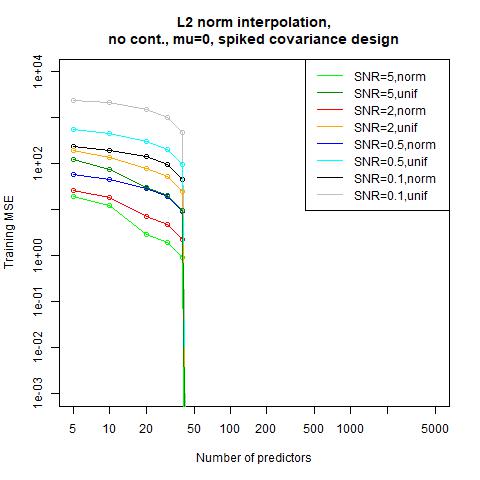} 
\end{center}
\caption{Training MSE of minimum $l_2$-norm interpolation when trained on clean training data.}\label{fig:minl2mu0spikedtrain}
\end{figure}

\begin{figure}[H]
\begin{center}
\includegraphics[width=5.25cm]{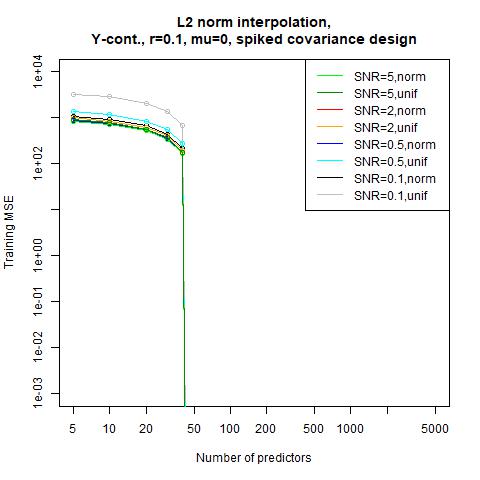} \includegraphics[width=5.25cm]{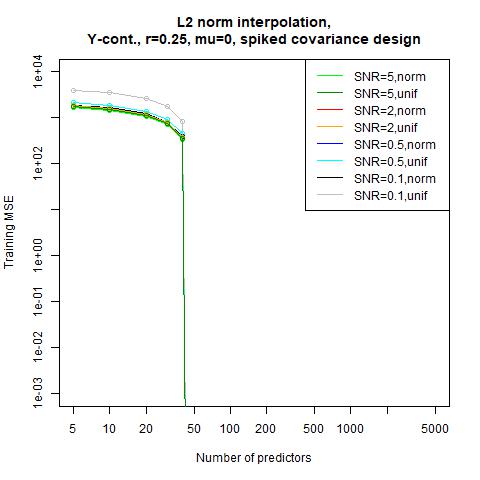} \\ \includegraphics[width=5.25cm]{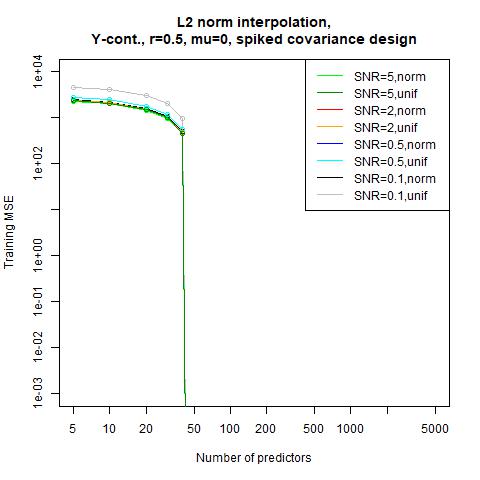} \includegraphics[width=5.25cm]{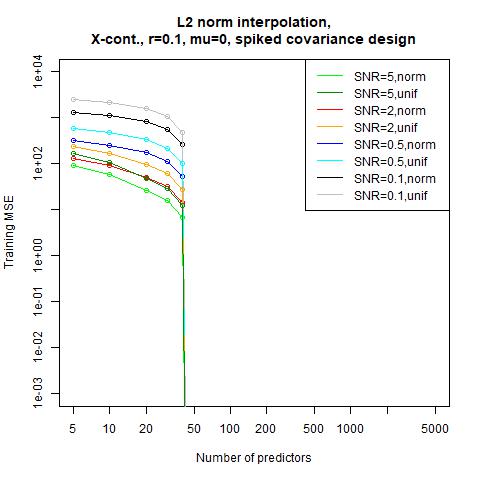} \\ \includegraphics[width=5.25cm]{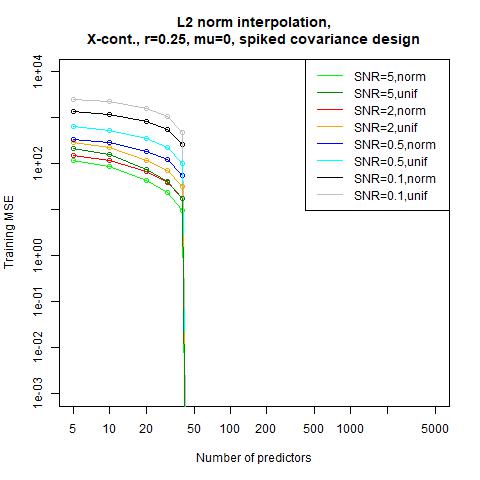} \includegraphics[width=5.25cm]{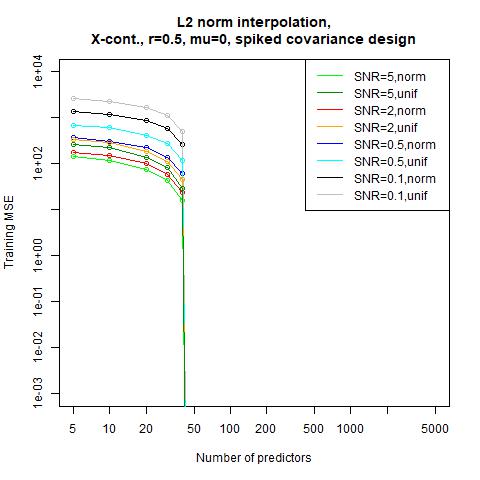} 
\end{center}
\caption{Training MSE of minimum $l_2$-norm interpolation when trained on contaminated training data.}\label{fig:minl2mu0spikedtraincont}
\end{figure}

\subsubsection{Huber-loss interpolation}

\begin{figure}[H]
\begin{center}
\includegraphics[width=7.5cm]{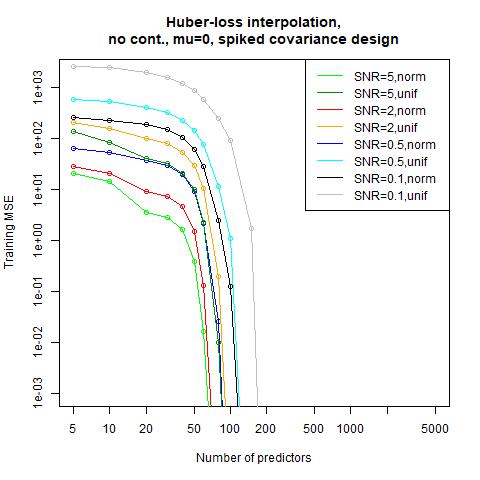} 
\end{center}
\caption{Training MSE of Huber-loss interpolation when trained on clean training data.}\label{fig:hubermu0indepspikedtrain}
\end{figure}

\begin{figure}[H]
\begin{center}
\includegraphics[width=5.25cm]{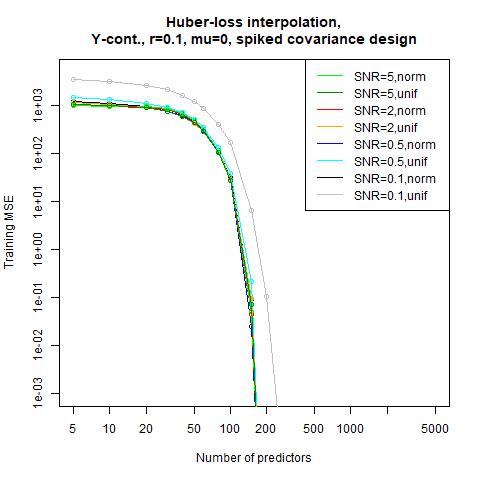} \includegraphics[width=5.25cm]{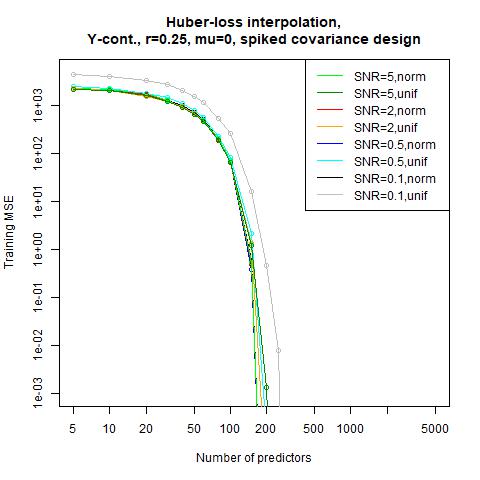} \\ \includegraphics[width=5.25cm]{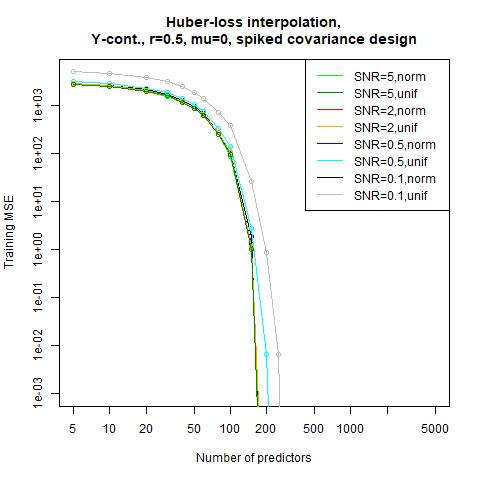} \includegraphics[width=5.25cm]{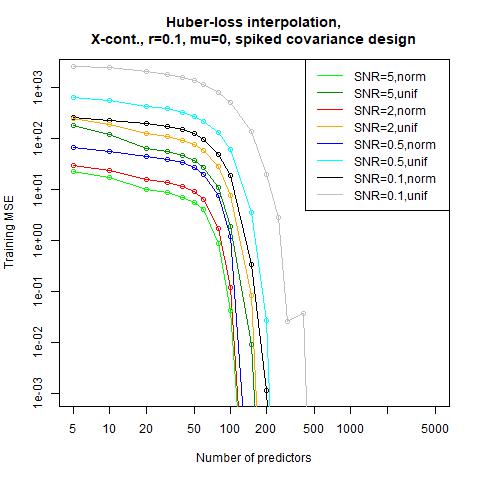} \\ \includegraphics[width=5.25cm]{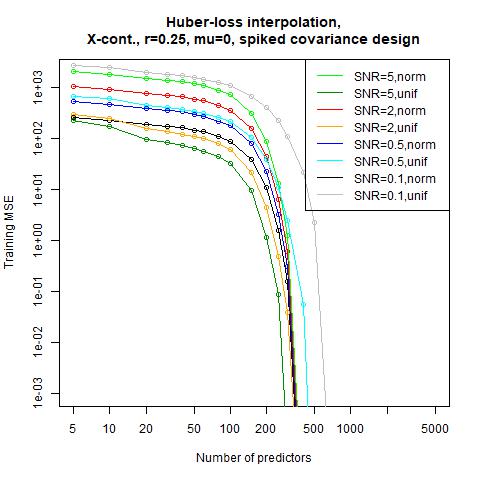} \includegraphics[width=5.25cm]{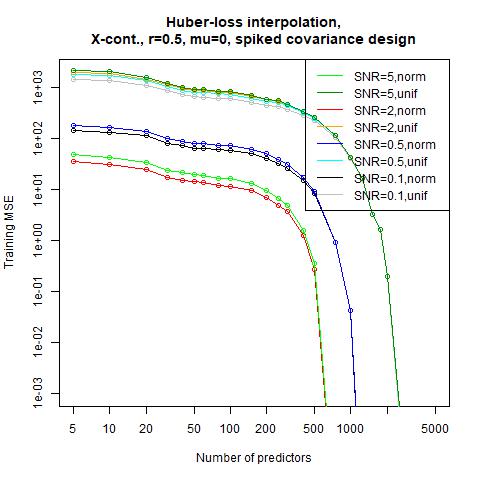} 
\end{center}
\caption{Training MSE of Huber-loss interpolation when trained on contaminated training data.} \label{fig:hubermu0indepspikedtraincont}
\end{figure}

\subsubsection{Tukey-loss interpolation}

\begin{figure}[H]
\begin{center}
\includegraphics[width=7.5cm]{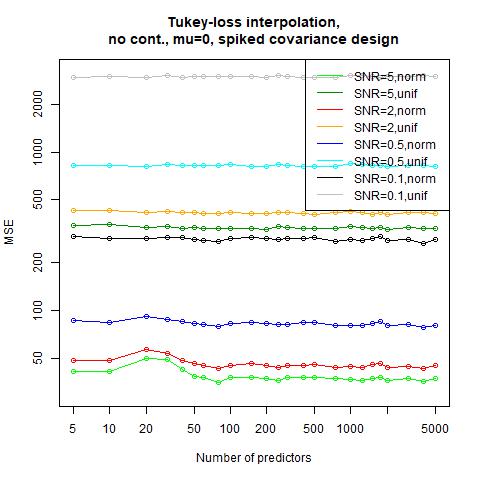} 
\end{center}
\caption{Training MSE of Tukey-loss interpolation when trained on clean training data.} \label{fig:tukeymu0spikedtrain}
\end{figure}

\begin{figure}[H]
\begin{center}
\includegraphics[width=5.25cm]{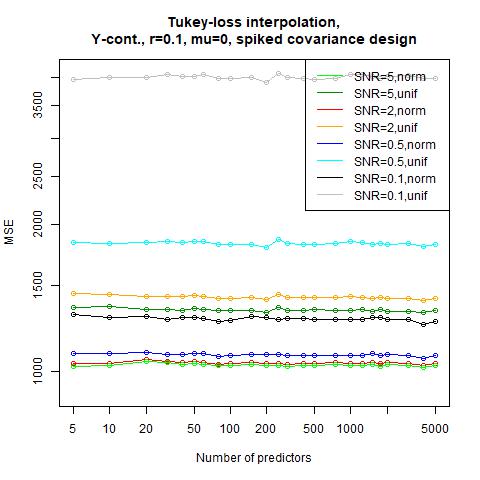} \includegraphics[width=5.25cm]{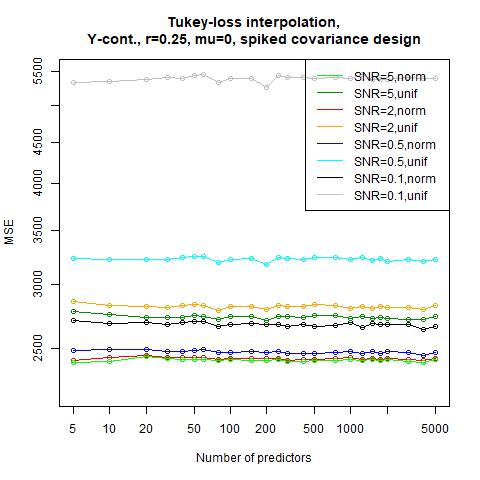} \\ \includegraphics[width=5.25cm]{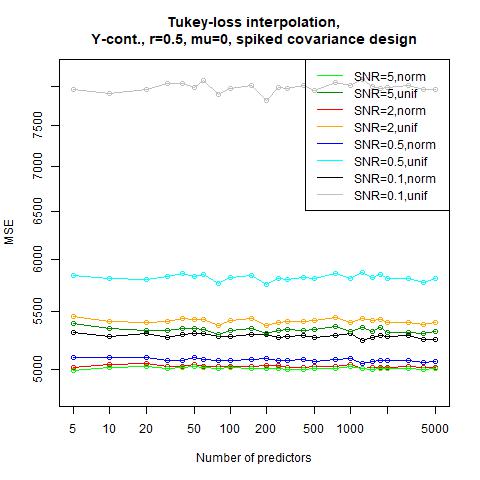} \includegraphics[width=5.25cm]{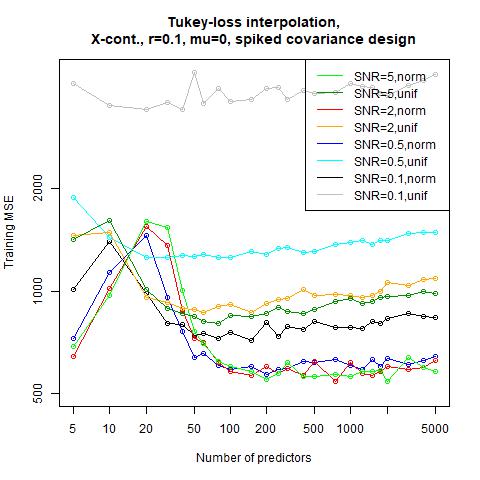} \\ \includegraphics[width=5.25cm]{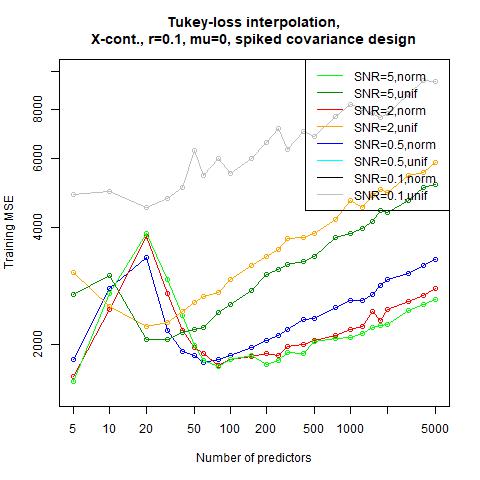} \includegraphics[width=5.25cm]{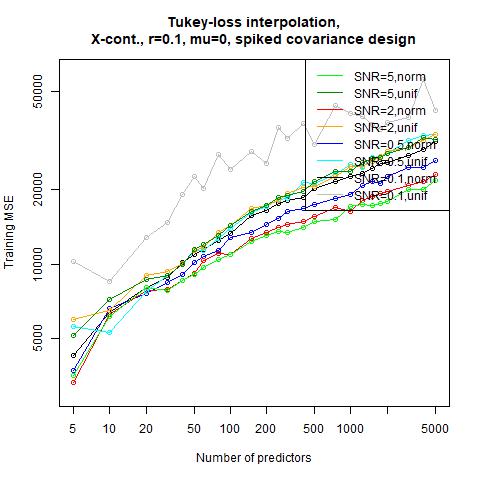} 
\end{center}
\caption{Training MSE of Tukey-loss interpolation when trained on contaminated training data.} \label{fig:tukeymu0spikedtraincont}
\end{figure}

\subsubsection{SLTS-based interpolation}

\begin{figure}[H]
\begin{center}
\includegraphics[width=7.5cm]{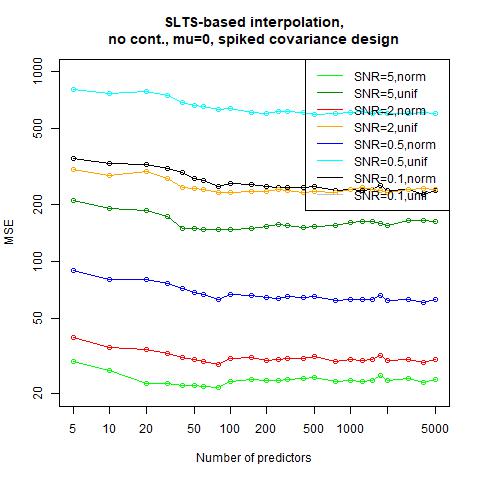} 
\end{center}
\caption{Training MSE of SLTS-based interpolation when trained on clean training data.}\label{fig:sltsmu0spikedtrain}
\end{figure}

\begin{figure}[H]
\begin{center}
\includegraphics[width=7.5cm]{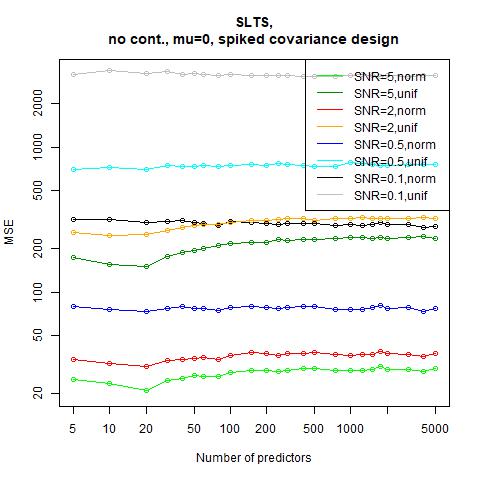} 
\end{center}
\caption{Training MSE of SLTS when trained on clean training data.}\label{fig:rawsltsmu0spikedtrain}
\end{figure}

\begin{figure}[H]
\begin{center}
\includegraphics[width=5.25cm]{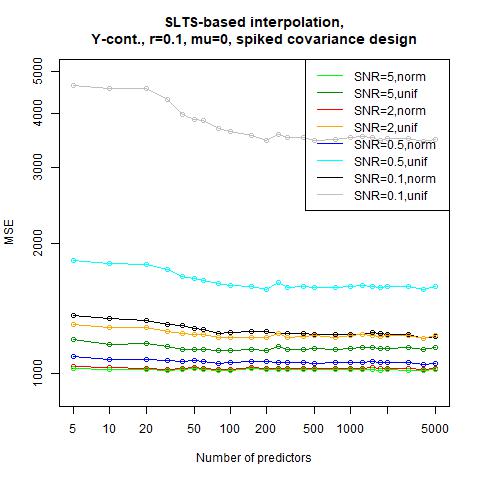} \includegraphics[width=5.25cm]{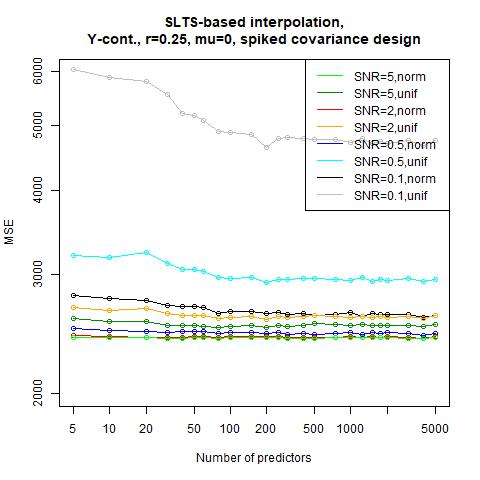} \\ \includegraphics[width=5.25cm]{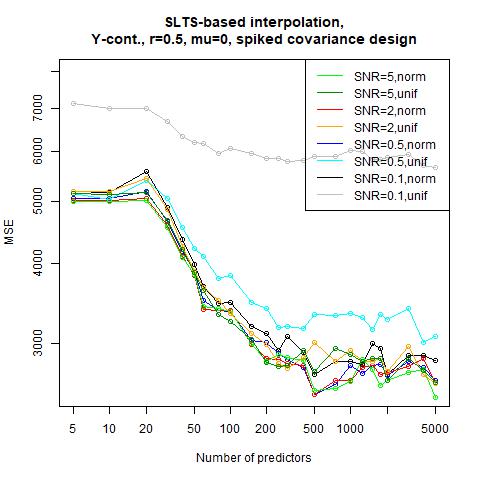} \includegraphics[width=5.25cm]{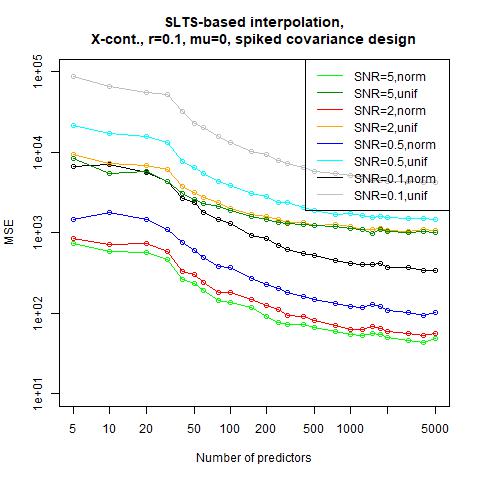} \\ \includegraphics[width=5.25cm]{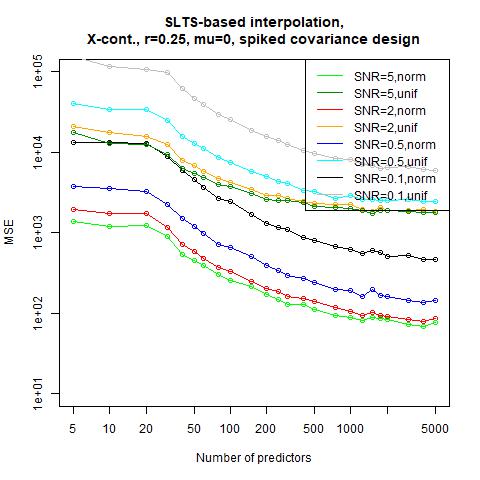} \includegraphics[width=5.25cm]{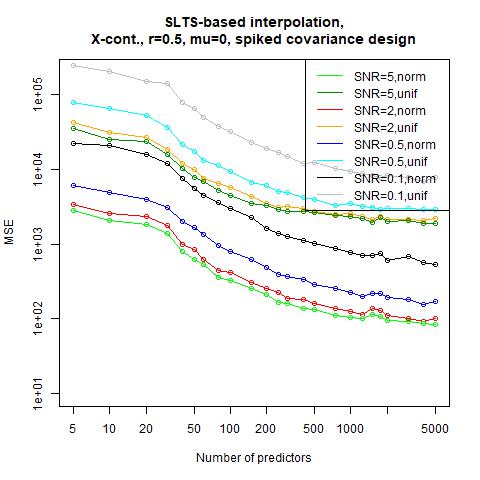} 
\end{center}
\caption{Training MSE of SLTS-based interpolation when trained on contaminated training data.}\label{fig:sltsmu0spikedtraincont}
\end{figure}

\begin{figure}[H]
\begin{center}
\includegraphics[width=5.25cm]{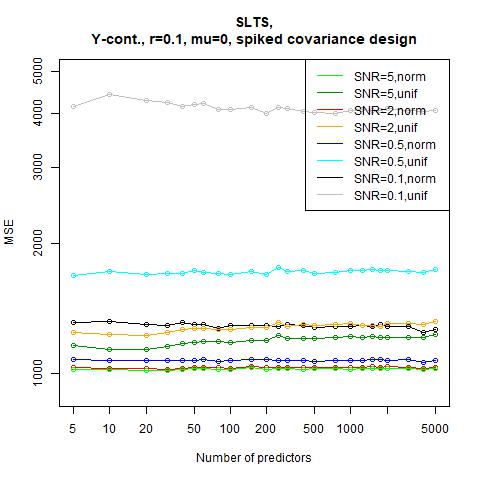} \includegraphics[width=5.25cm]{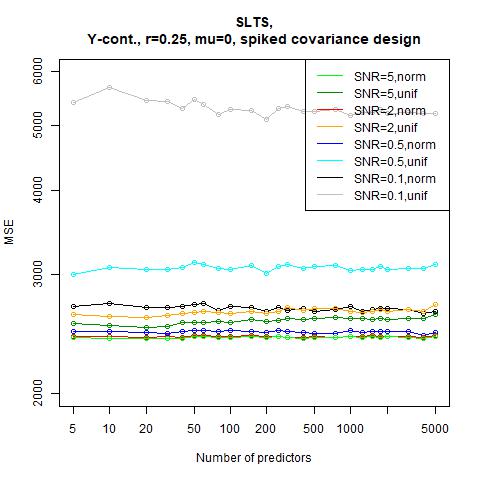} \\ \includegraphics[width=5.25cm]{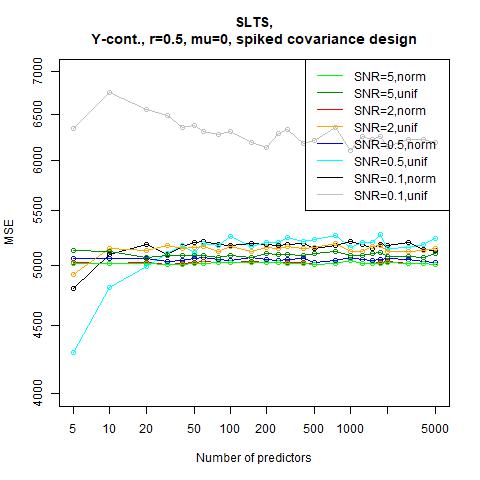} \includegraphics[width=5.25cm]{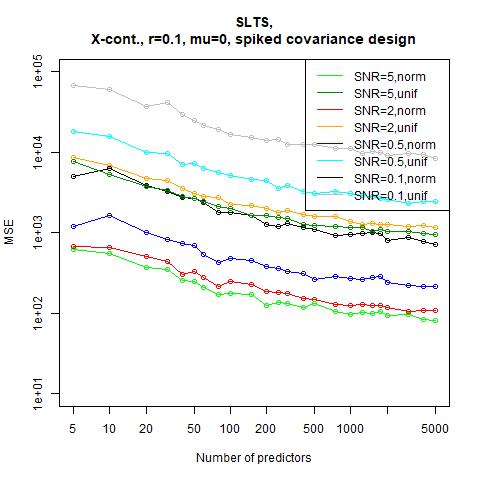} \\ \includegraphics[width=5.25cm]{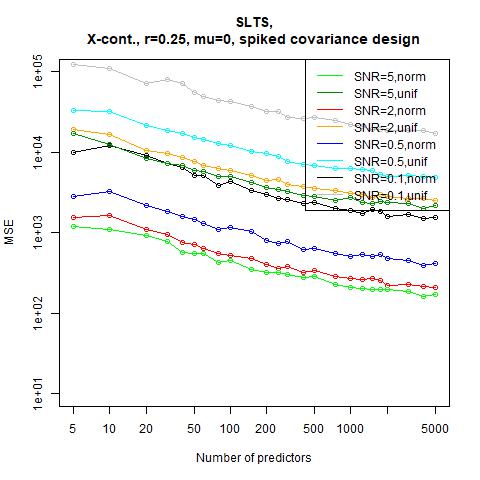} \includegraphics[width=5.25cm]{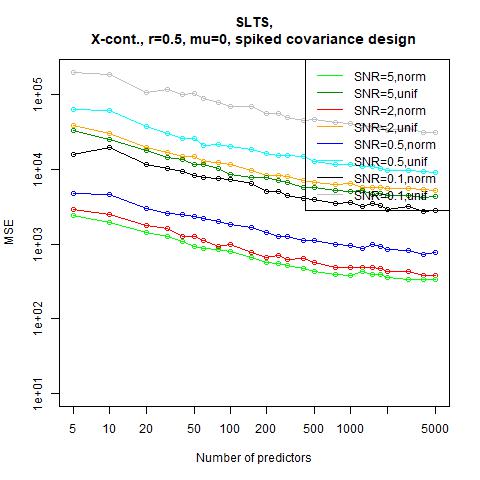} 
\end{center}
\caption{Training MSE of SLTS when trained on contaminated training data.} \label{fig:rawsltsmu0spikedtraincont}
\end{figure}

\subsubsection{Boosting-based interpolation}

\begin{figure}[H]
\begin{center}
\includegraphics[width=7.5cm]{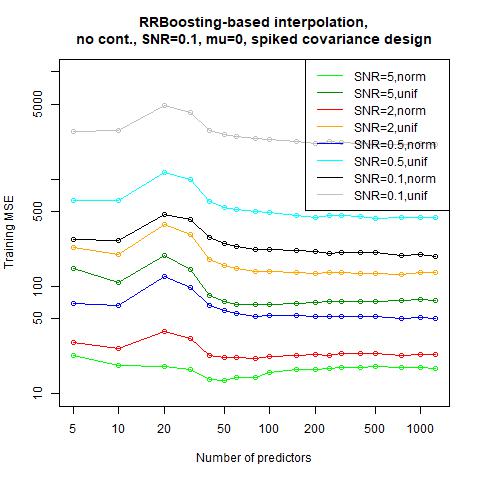} 
\end{center}
\caption{Training MSE of RRBoost-based interpolation when trained on clean training data.}\label{fig:rrboostmu0spikedtrain}
\end{figure}

\begin{figure}[H]
\begin{center}
\includegraphics[width=7.5cm]{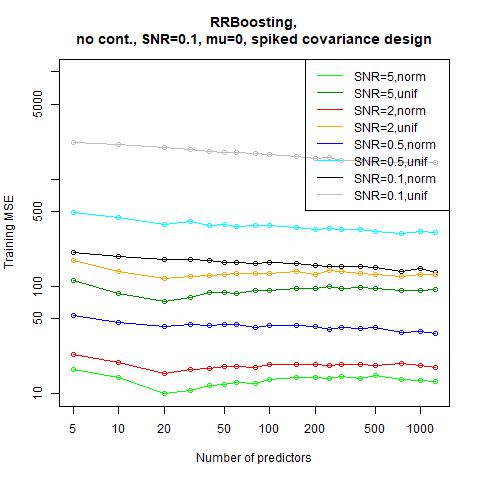} 
\end{center}
\caption{Training MSE of RRBoost-based interpolation when trained on clean training data.}\label{fig:rawrrboostmu0spikedtrain}
\end{figure}

\begin{figure}[H]
\begin{center}
\includegraphics[width=5.25cm]{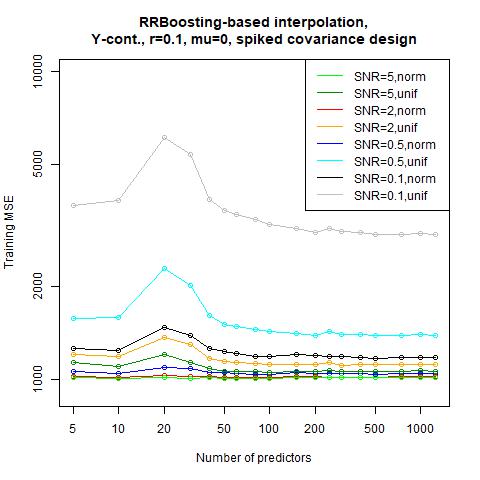} \includegraphics[width=5.25cm]{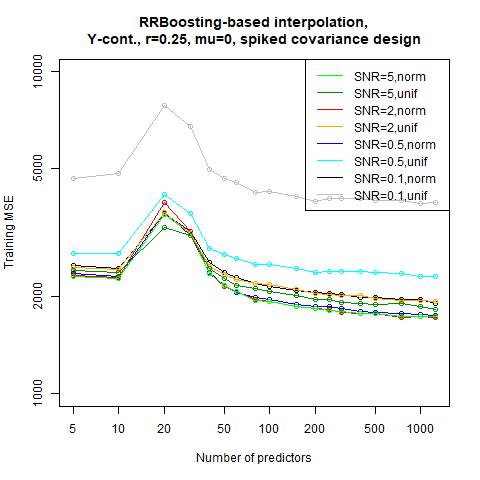} \\ \includegraphics[width=5.25cm]{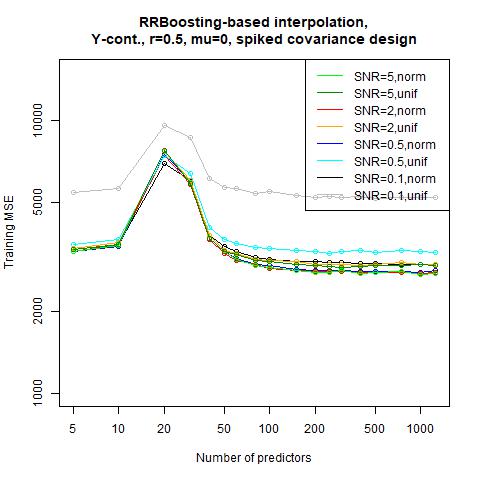} \includegraphics[width=5.25cm]{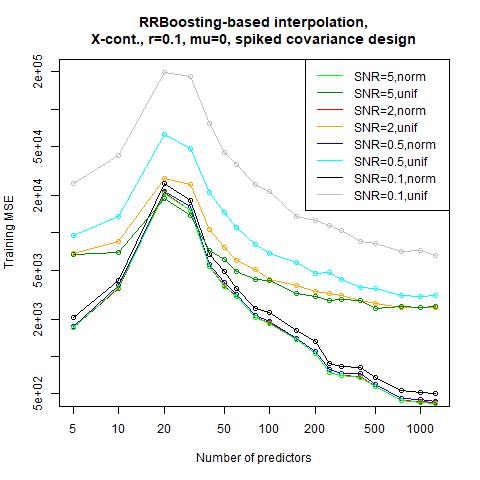} \\ \includegraphics[width=5.25cm]{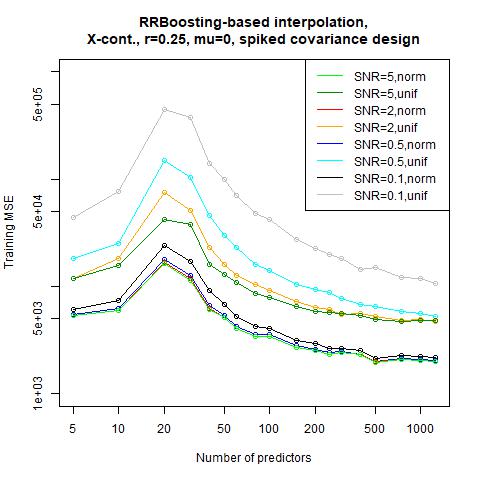} \includegraphics[width=5.25cm]{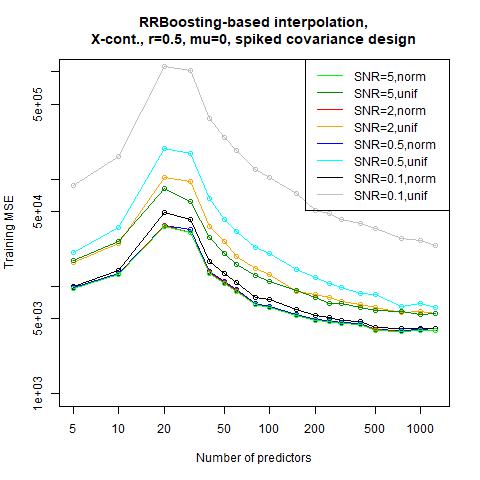} 
\end{center}
\caption{Training MSE of RRBoosting when trained on contaminated training data.}\label{fig:rrboostmu0spikedtraincont}
\end{figure}

\begin{figure}[H]
\begin{center}
\includegraphics[width=5.25cm]{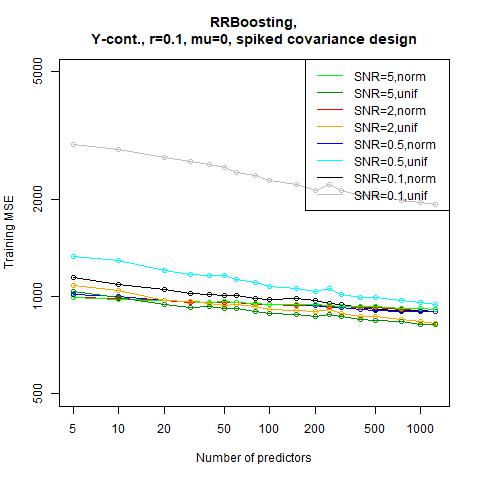} \includegraphics[width=5.25cm]{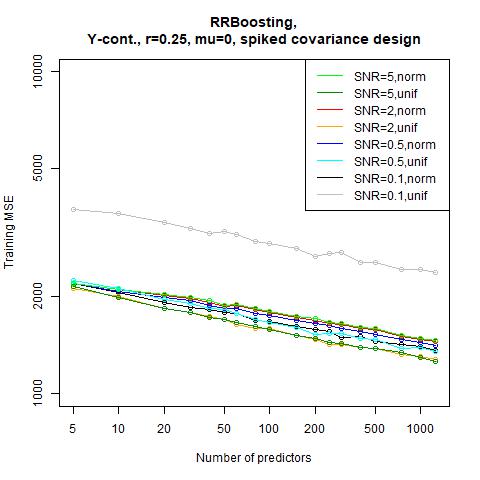} \\ \includegraphics[width=5.25cm]{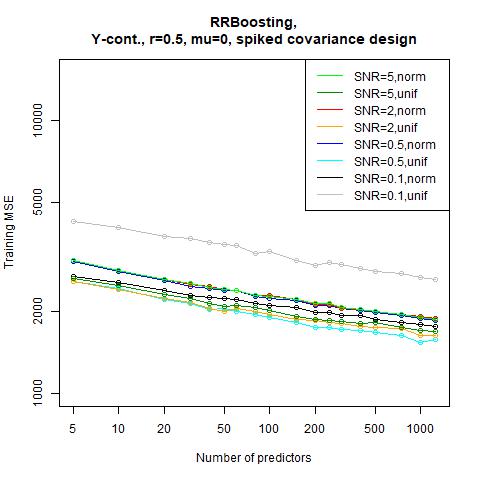} \includegraphics[width=5.25cm]{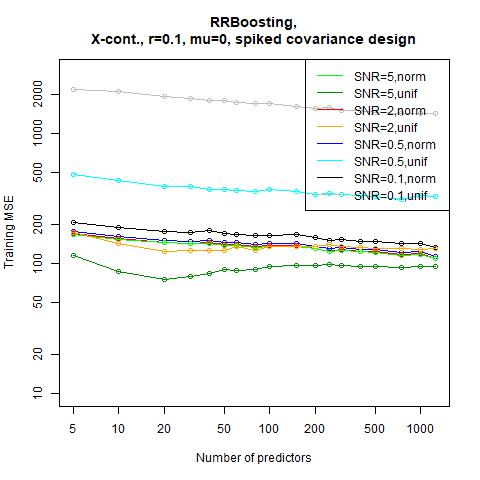} \\ \includegraphics[width=5.25cm]{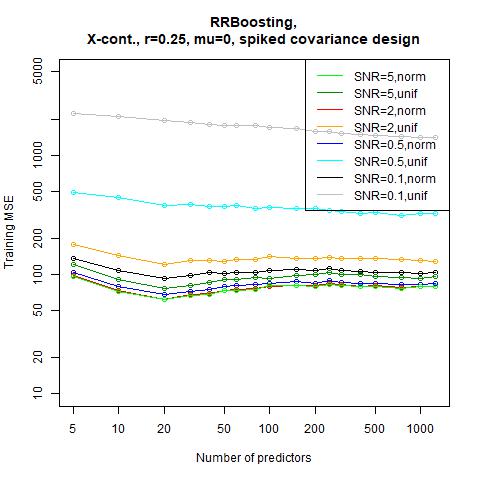} \includegraphics[width=5.25cm]{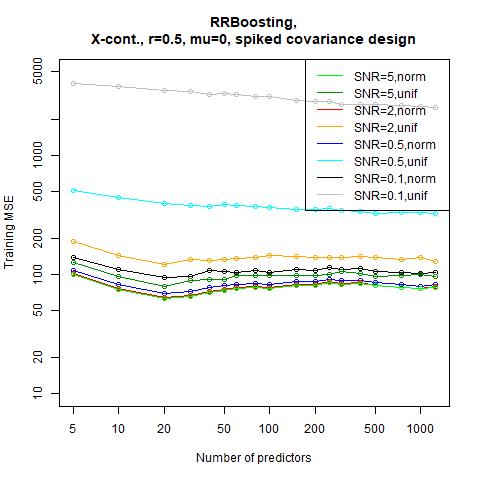} 
\end{center}
\caption{Training MSE of RRBoosting when trained on contaminated training data.}\label{fig:rawrrboostmu0spikedtraincont}
\end{figure}

It can be observed in Fig. \ref{fig:minl2mu0spikedtrain}, Fig. \ref{fig:minl2mu0spikedtraincont}, Fig. \ref{fig:hubermu0indepspikedtrain}, Fig. \ref{fig:hubermu0indepspikedtraincont}, Fig. \ref{fig:tukeymu0spikedtrain}, Fig. \ref{fig:tukeymu0spikedtraincont}, Fig. \ref{fig:sltsmu0spikedtrain}, Fig. \ref{fig:sltsmu0spikedtraincont}, Fig. \ref{fig:rawsltsmu0spikedtrain}, Fig. \ref{fig:rawsltsmu0spikedtraincont}, Fig. \ref{fig:rrboostmu0spikedtrain}, Fig. \ref{fig:rrboostmu0spikedtraincont}, Fig. \ref{fig:rawrrboostmu0spikedtrain} and Fig. \ref{fig:rawrrboostmu0spikedtraincont} that the training MSE curves resemble those from the case of independent design, although the MSE values themselves are higher.

\subsection{Independent features, $\mu=5$}

\subsubsection{Minimum $l_2$-norm interpolation}

\begin{figure}[H]
\begin{center}
\includegraphics[width=7.5cm]{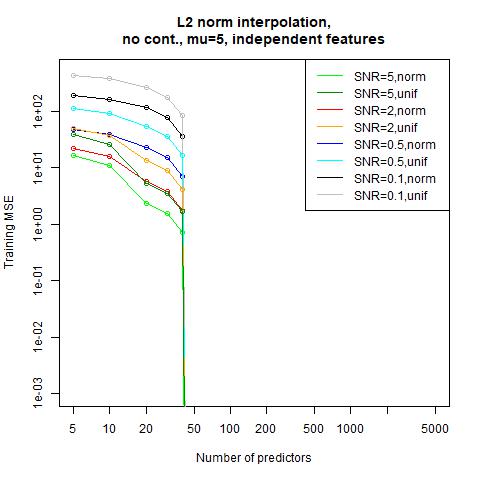} 
\end{center}
\caption{Training MSE of minimum $l_2$-norm interpolation when trained on clean training data.}\label{fig:minl2mu5indeptrain}
\end{figure}

\begin{figure}[H]
\begin{center}
\includegraphics[width=5.25cm]{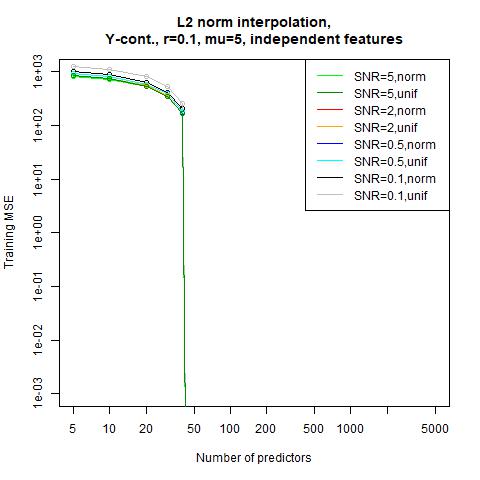} \includegraphics[width=5.25cm]{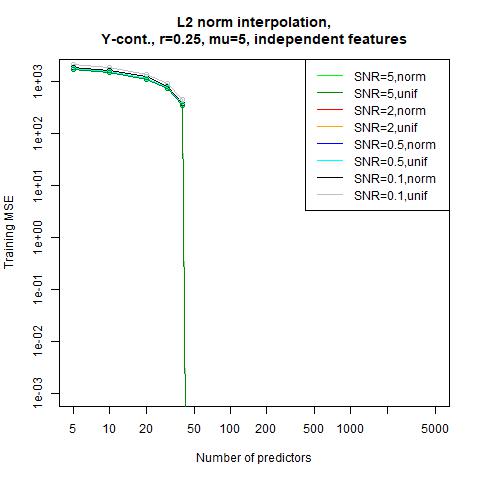} \\ \includegraphics[width=5.25cm]{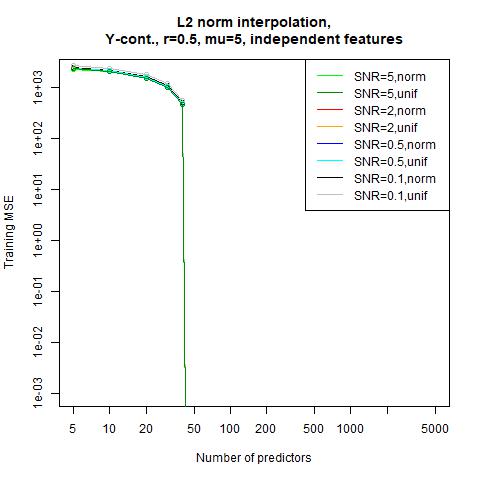} \includegraphics[width=5.25cm]{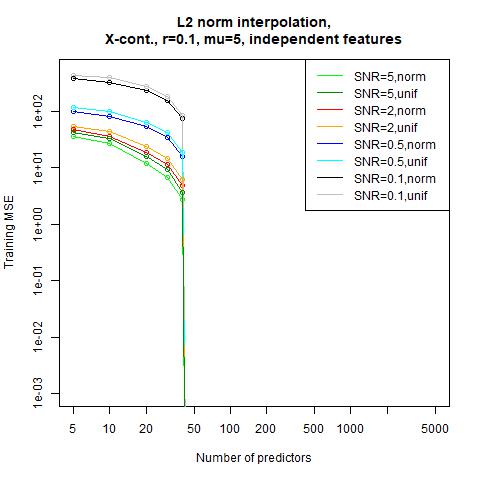} \\ \includegraphics[width=5.25cm]{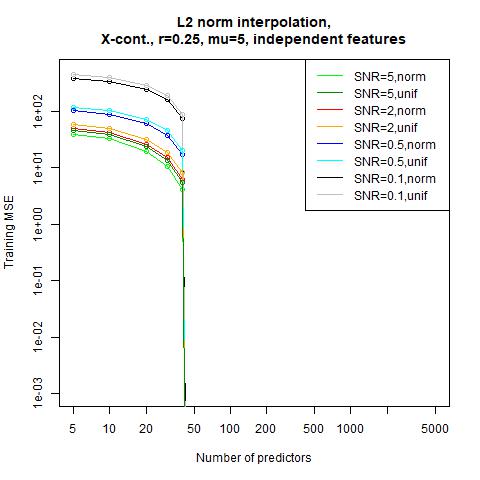} \includegraphics[width=5.25cm]{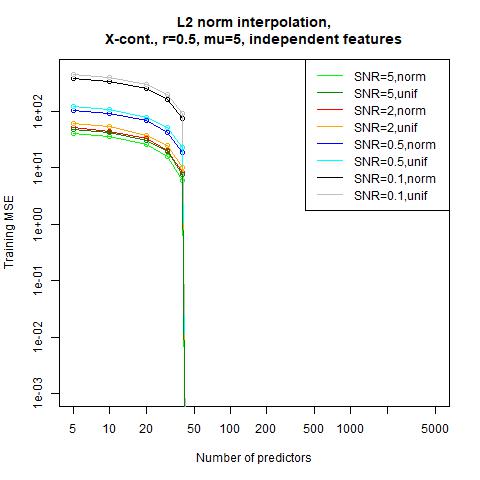} 
\end{center}
\caption{Training MSE of minimum $l_2$-norm interpolation when trained on contaminated training data.}\label{fig:minl2mu5indeptraincont}
\end{figure}

\subsubsection{Huber-loss interpolation}

\begin{figure}[H]
\begin{center}
\includegraphics[width=7.5cm]{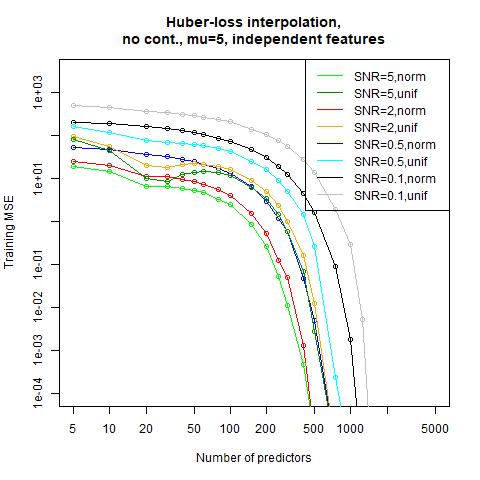} 
\end{center}
\caption{Training MSE of Huber-loss interpolation when trained on clean training data.}\label{fig:hubermu5indeptrain}
\end{figure}

\begin{figure}[H]
\begin{center}
\includegraphics[width=5.25cm]{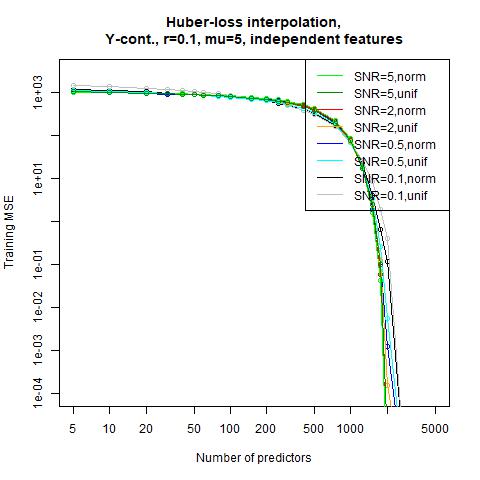} \includegraphics[width=5.25cm]{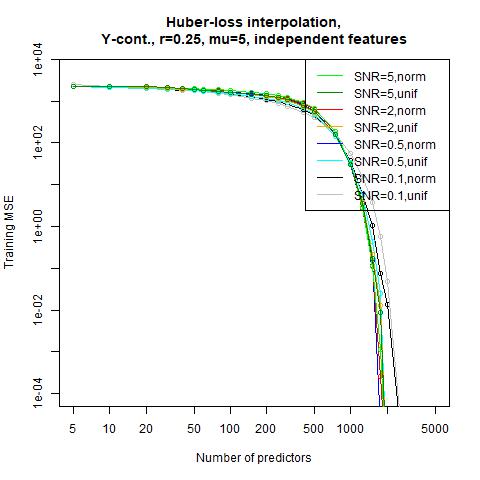} \\ \includegraphics[width=5.25cm]{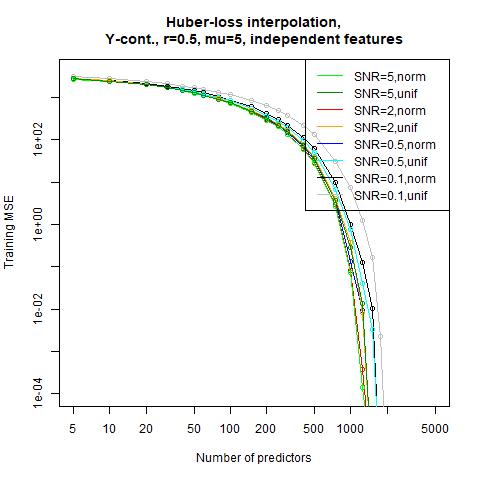} \includegraphics[width=5.25cm]{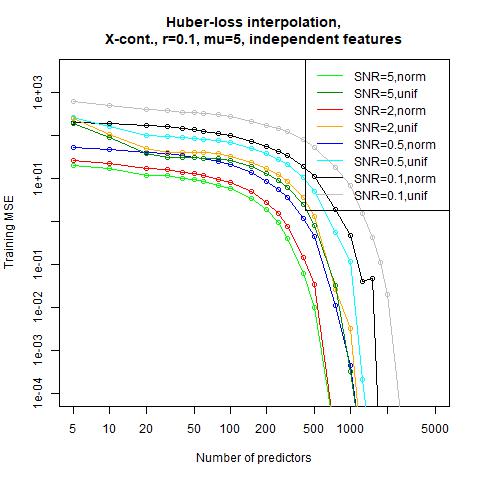} \\ \includegraphics[width=5.25cm]{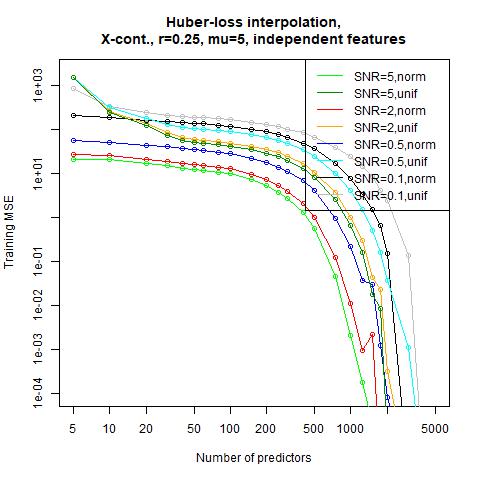} \includegraphics[width=5.25cm]{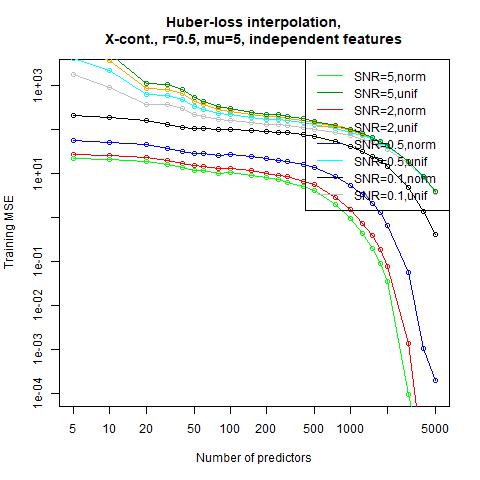} 
\end{center}
\caption{Training MSE of Huber-loss interpolation when trained on contaminated training data.}\label{fig:hubermu5indeptraincont}
\end{figure}

\subsubsection{SLTS-based interpolation}

\begin{figure}[H]
\begin{center}
\includegraphics[width=7.5cm]{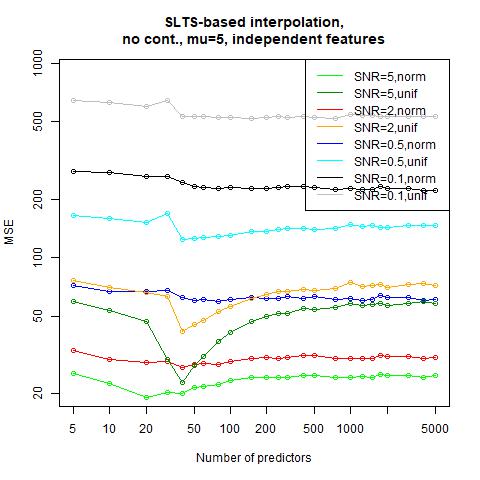} 
\end{center}
\caption{Training MSE of SLTS-based interpolation when trained on clean training data.}\label{fig:sltsmu5indeptrain}
\end{figure}

\begin{figure}[H]
\begin{center}
\includegraphics[width=7.5cm]{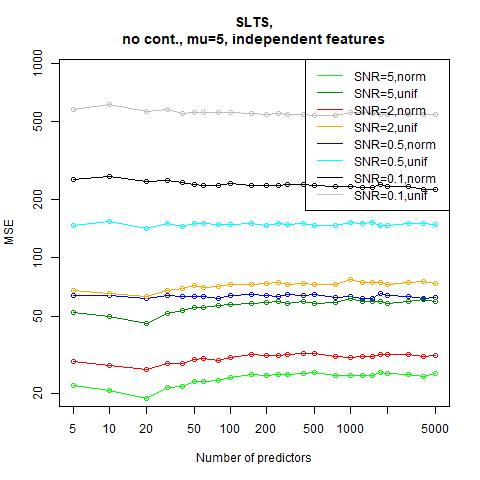} 
\end{center}
\caption{Training MSE of SLTS when trained on clean training data.}\label{fig:rawsltsmu5indeptrain}
\end{figure}

\begin{figure}[H]
\begin{center}
\includegraphics[width=5.25cm]{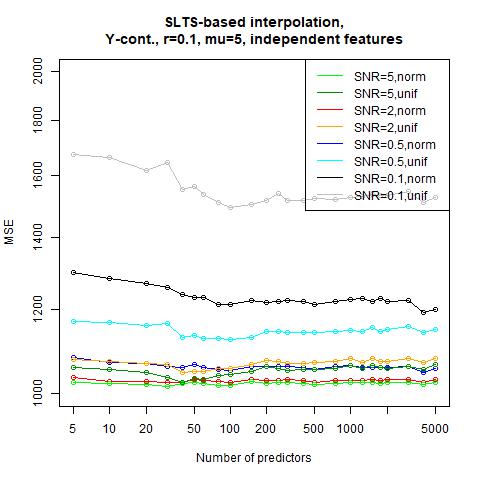} \includegraphics[width=5.25cm]{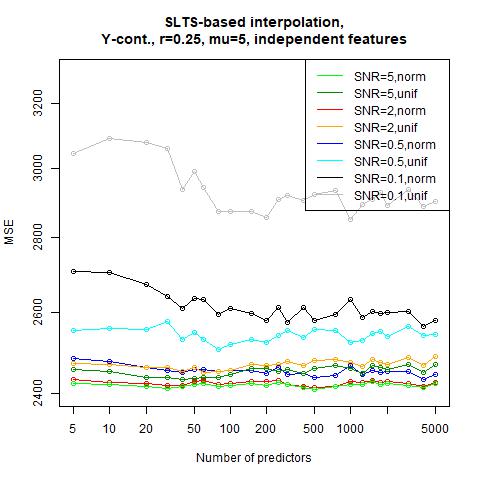} \\ \includegraphics[width=5.25cm]{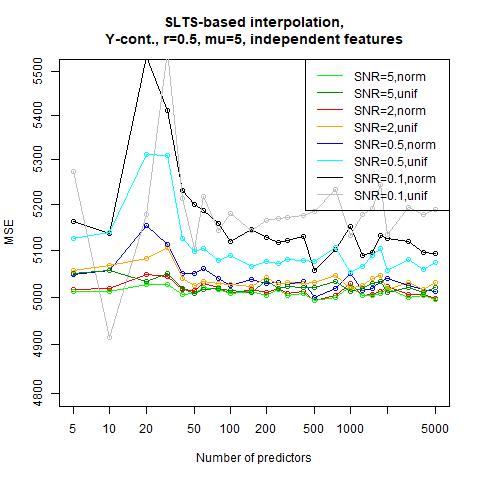} \includegraphics[width=5.25cm]{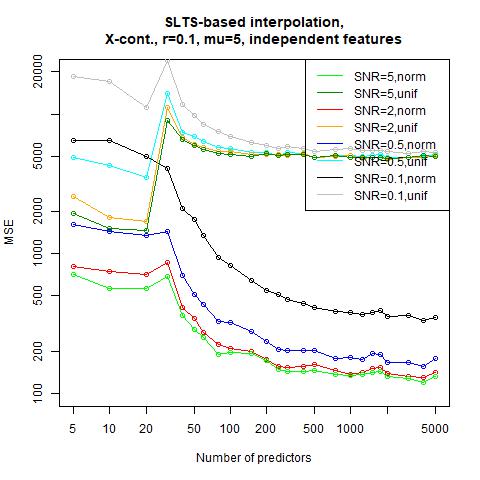} \\ \includegraphics[width=5.25cm]{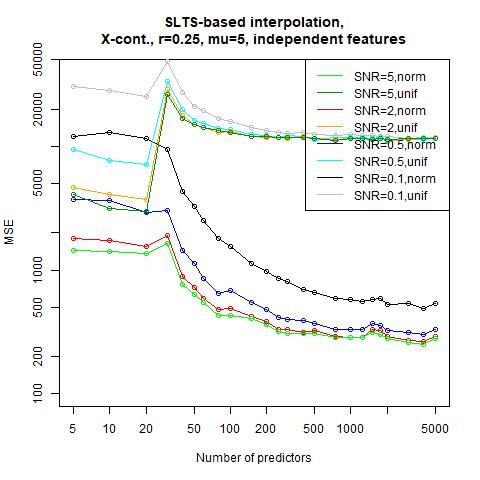} \includegraphics[width=5.25cm]{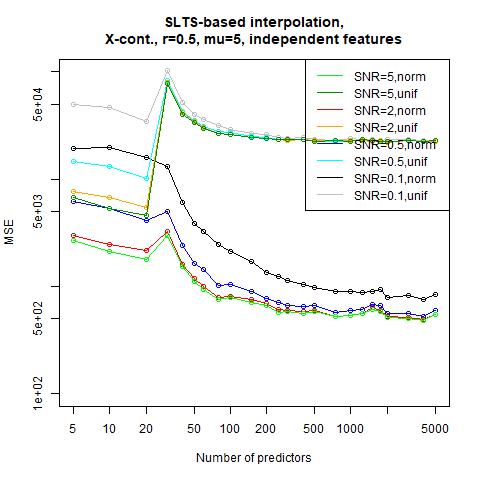} 
\end{center}
\caption{Training MSE of SLTS-based interpolation when trained on contaminated training data.}\label{fig:sltsmu5indeptraincont}
\end{figure}

\begin{figure}[H]
\begin{center}
\includegraphics[width=5.25cm]{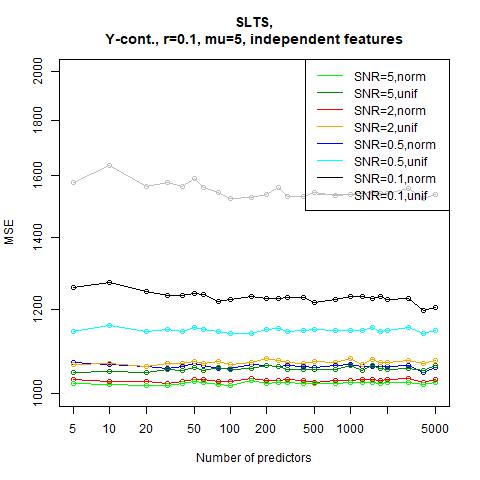} \includegraphics[width=5.25cm]{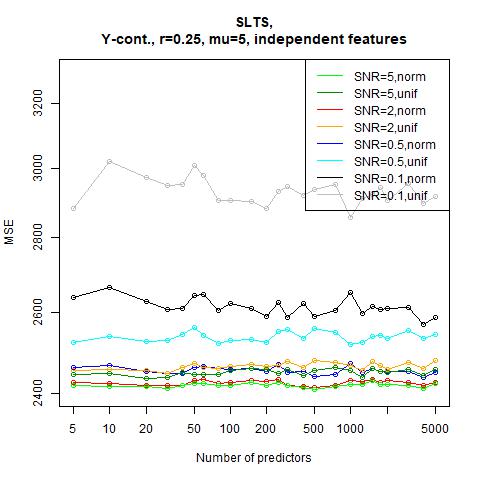} \\ \includegraphics[width=5.25cm]{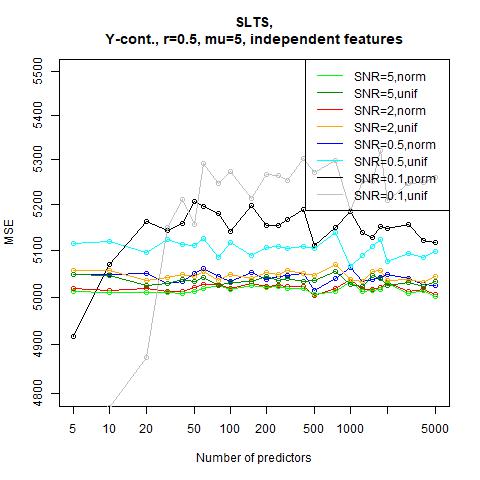} \includegraphics[width=5.25cm]{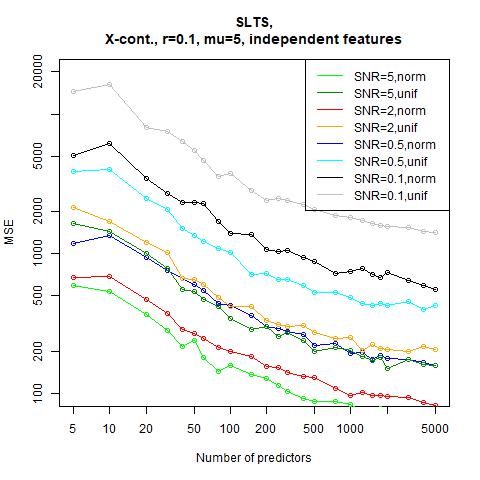} \\ \includegraphics[width=5.25cm]{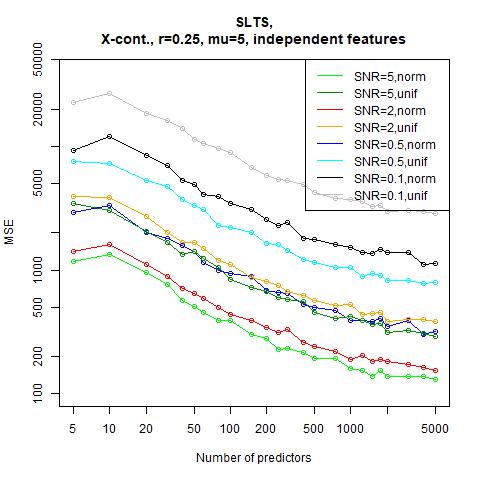} \includegraphics[width=5.25cm]{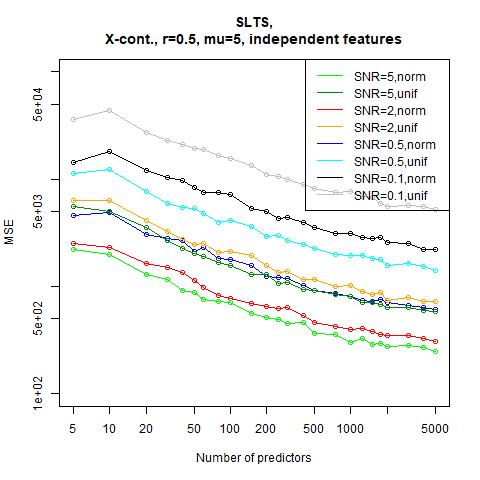} 
\end{center}
\caption{Training MSE of SLTS when trained on contaminated training data.}\label{fig:rawsltsmu5indeptraincont}
\end{figure}

It can be observed in Fig. \ref{fig:minl2mu5indeptrain}, Fig. \ref{fig:minl2mu5indeptraincont}, Fig. \ref{fig:hubermu5indeptrain}, Fig. \ref{fig:hubermu5indeptraincont}, Fig. \ref{fig:sltsmu5indeptrain}, Fig. \ref{fig:sltsmu5indeptraincont}, Fig. \ref{fig:rawsltsmu5indeptrain} and Fig. \ref{fig:rawsltsmu5indeptraincont} that the training MSE curves generally resemble those from the case $\mu=0$. For $X$-contamination and a low SNR and for $Y$-contamination and $r=0.5$, the MSE remains nearly constant after the peak, in contrast to the decreasing behavior in the setting $\mu=0$. In addition, the training error for Huber-loss interpolation vanishes much later than in the case $\mu=0$. For SLTS-based interpolation, the training MSE does not decrease with $p$ for $Y$-contamination and $r=0.5$ in Fig. \ref{fig:sltsmu5indeptraincont}, in contrast to the case $\mu=0$, see Fig. \ref{fig:sltsmu0spikedtraincont}. Moreover, the training MSE is larger for uniformly distributed coefficients than for Gaussian coefficients in the case of $X$-contamination for $\mu=5$, which is not the case for $\mu=0$.






\newpage 

\subsection{$n=200$} \

\subsubsection{Minimum $l_2$-norm interpolation}

\begin{figure}[H]
\begin{center}
\includegraphics[width=5.25cm]{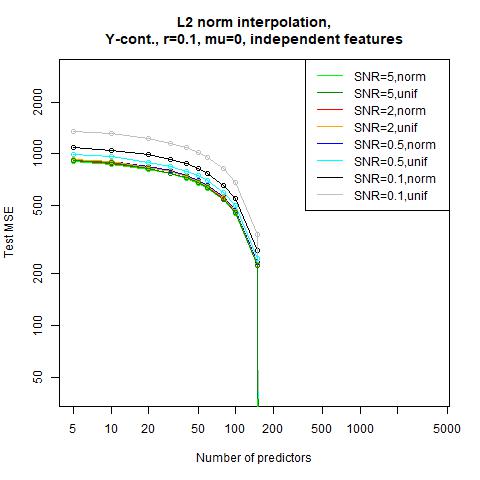} 
\includegraphics[width=5.25cm]{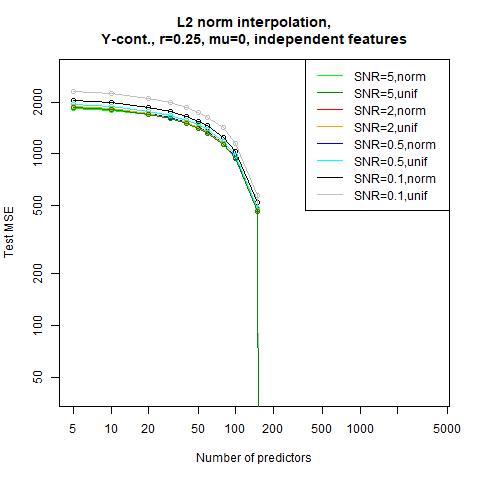} \\
 \includegraphics[width=5.25cm]{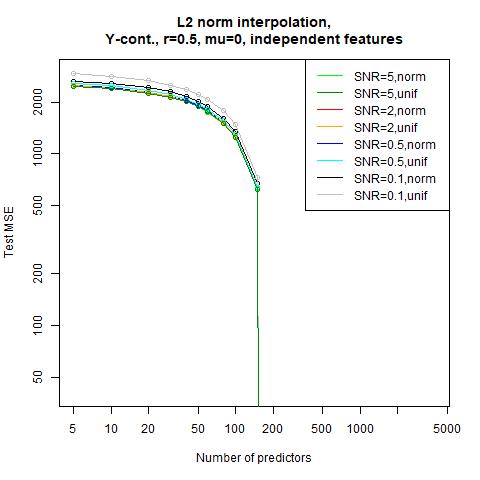} 
\includegraphics[width=5.25cm]{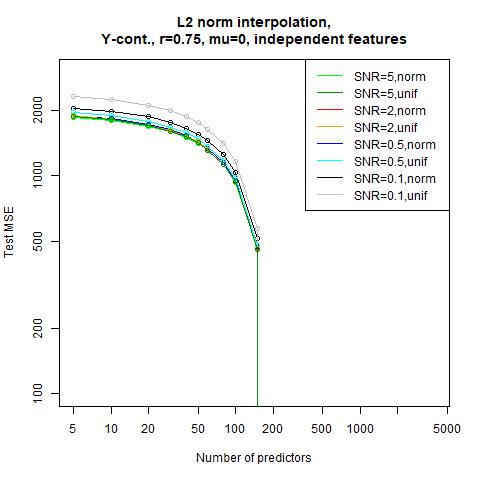}\\
 \includegraphics[width=5.25cm]{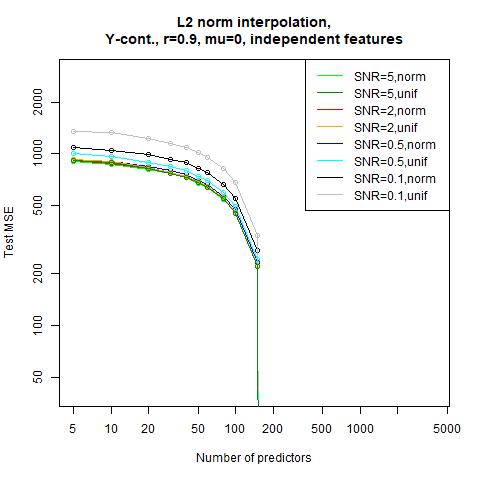} 
\end{center}
\caption{Training MSE of minimum $l_2$-norm interpolation when trained on $Y$-contaminated training data.}\label{fig:minl2mu0indepcontn200train}
\end{figure}

The training MSE vanishes at $p=n$, as expected (it is an issue of the \texttt{plot} function in $\mathsf{R}$ that it seems that the MSE vanishes earlier in Fig. \ref{fig:minl2mu0indepcontn200train}).

\subsubsection{Huber-loss interpolation}

\begin{figure}[H]
\begin{center}
\includegraphics[width=5.25cm]{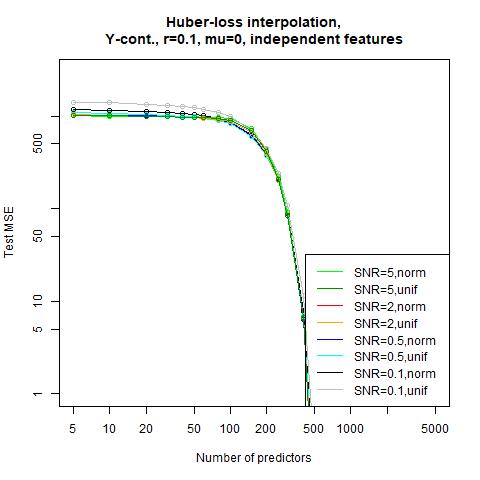} 
\includegraphics[width=5.25cm]{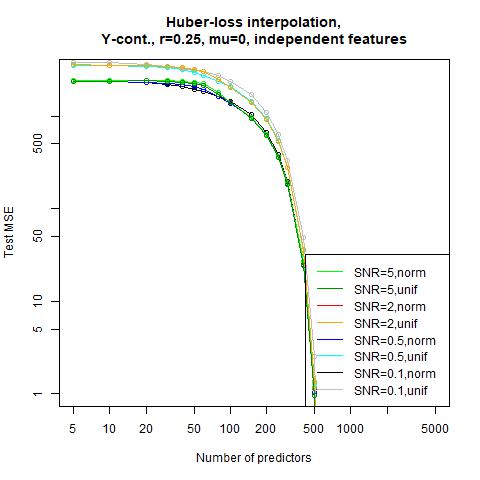} \\
 \includegraphics[width=5.25cm]{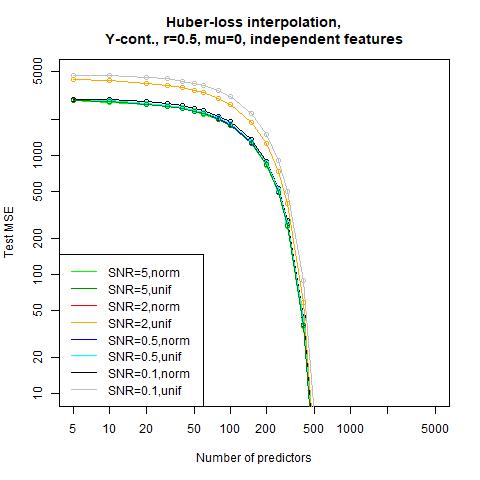} 
\includegraphics[width=5.25cm]{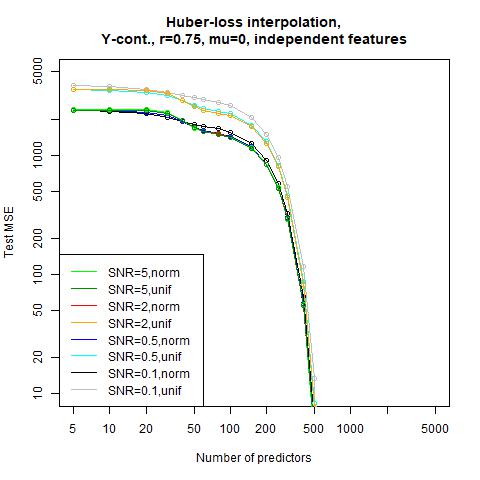}\\
 \includegraphics[width=5.25cm]{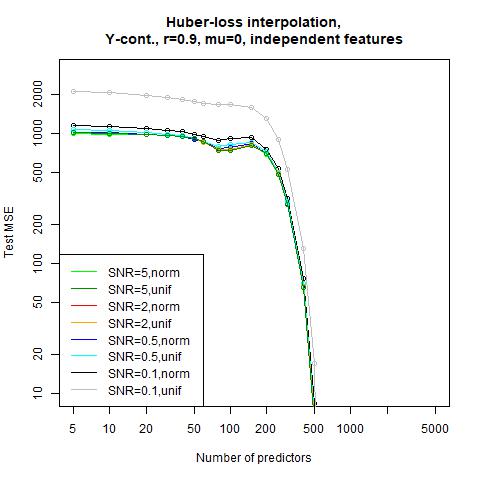} 
\end{center}
\caption{Training MSE of Huber-loss interpolation when trained on $Y$-contaminated training data.}\label{fig:hubermu0indepcontn200train}
\end{figure}

It already has been observed in Fig. \ref{fig:hubermu0indepspikedtrain} that for Huber-loss interpolation, the training MSE vanishes at some $p>n$. This is again the case in Fig. \ref{fig:hubermu0indepcontn200train}, where the training error vanishes at around $p=500$.

\newpage 

\subsection{$n=200$, $c_{out}=10000$} \

\subsubsection{Minimum $l_2$-norm interpolation}

\begin{figure}[H]
\begin{center}
\includegraphics[width=5.25cm]{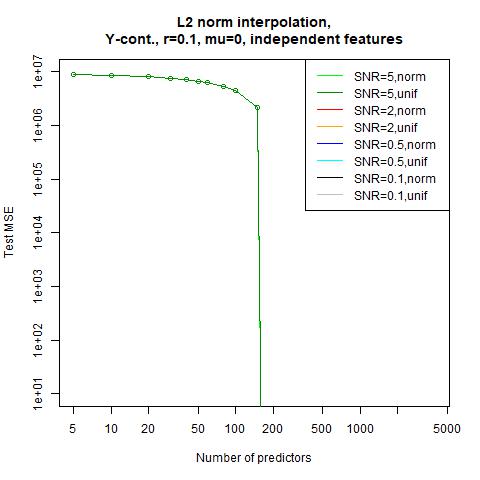} 
\includegraphics[width=5.25cm]{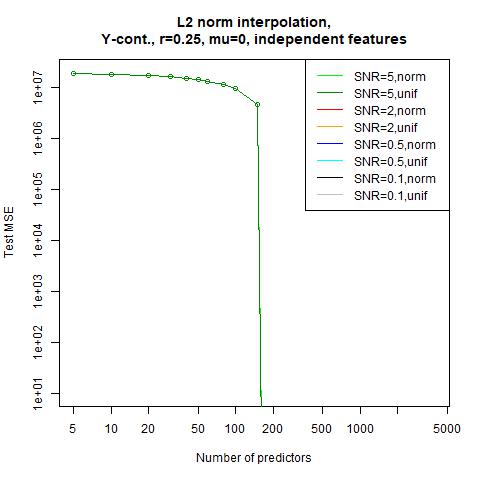} \\
 \includegraphics[width=5.25cm]{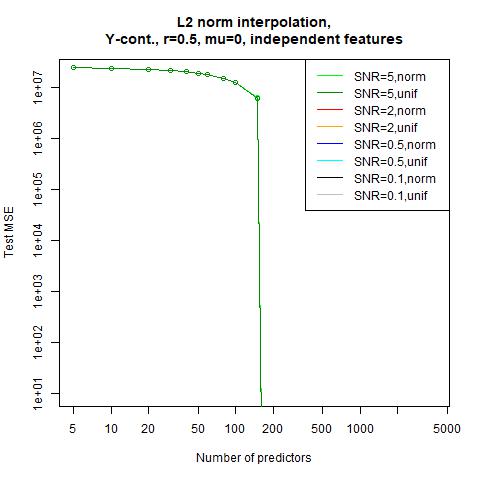} 
\includegraphics[width=5.25cm]{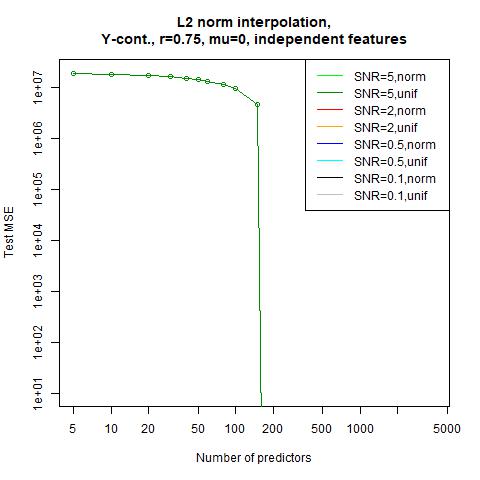}\\
 \includegraphics[width=5.25cm]{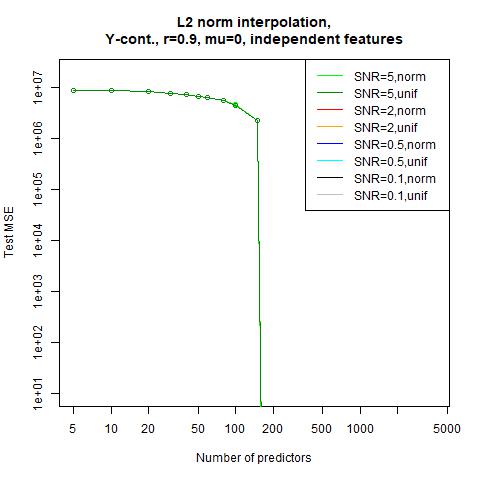} 
\end{center}
\caption{Training MSE of minimum $l_2$-norm interpolation when trained on $Y$-contaminated training data.}\label{fig:minl2mu0indepcontn200r10000train}
\end{figure}

The training MSE curves in Fig. \ref{fig:minl2mu0indepcontn200r10000train} resemble those from Fig. \ref{fig:minl2mu0indepcontn200train} for the case $c_{out}$, although the MSE values for $p<n$ are clearly larger.

\subsubsection{Huber-loss interpolation}

\begin{figure}[H]
\begin{center}
\includegraphics[width=5.25cm]{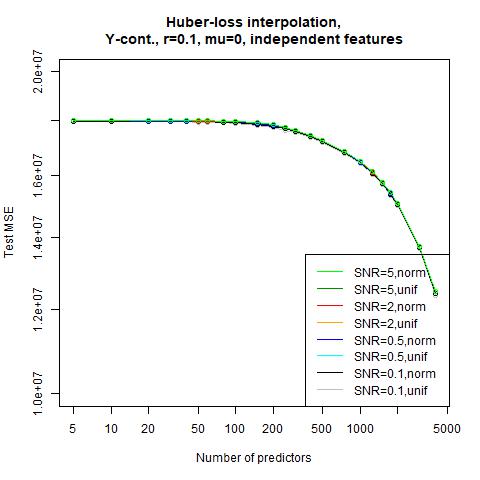} 
\includegraphics[width=5.25cm]{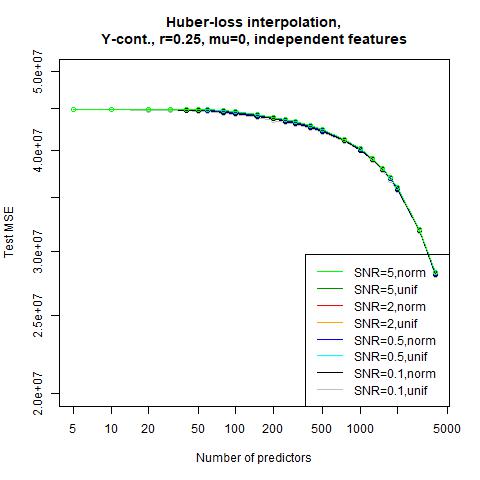} \\
 \includegraphics[width=5.25cm]{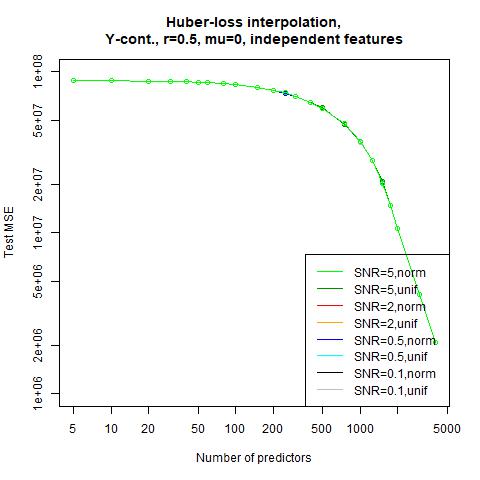} 
\includegraphics[width=5.25cm]{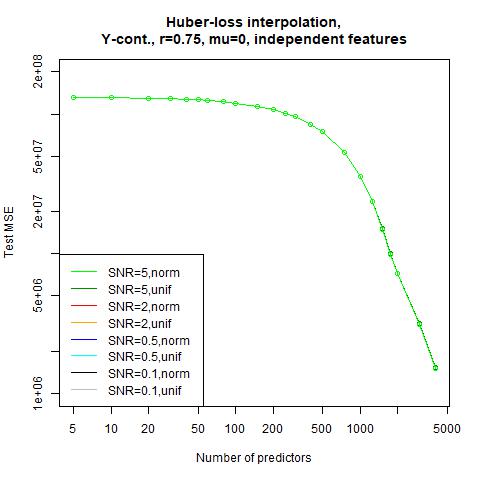}\\
 \includegraphics[width=5.25cm]{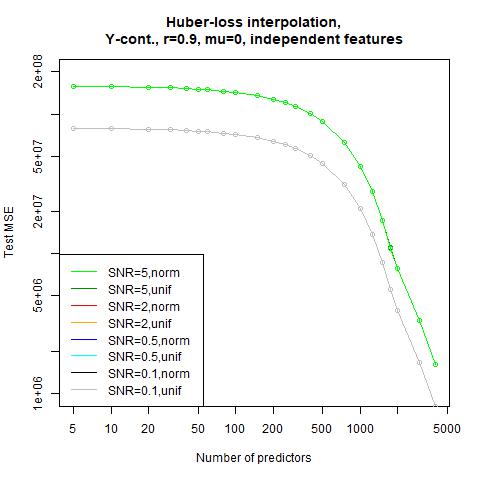} 
\end{center}
\caption{Training MSE of Huber-loss interpolation when trained on $Y$-contaminated training data.}\label{fig:hubermu0indepcontn200r10000train}
\end{figure}

The training MSE curves in Fig. \ref{fig:hubermu0indepcontn200r10000train} resemble those from Fig. \ref{fig:hubermu0indepcontn200train} for the case $c_{out}=100$, but they start much later to considerably decrease.

\section{$l_1$-norm coefficient differences} \label{sec:l1diff}

\subsection{Independent features, $\mu=0$}

\subsubsection{Minimum $l_2$-norm interpolation}

\begin{figure}[H]
\begin{center}
\includegraphics[width=7.5cm]{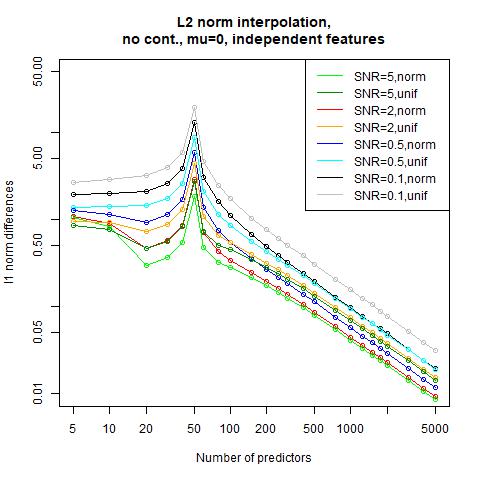} 
\end{center}
\caption{Differences $||\hat \beta-\beta||_1/n$ for the estimated coefficient vector $\hat \beta$ of minimum $l_2$-norm interpolation when trained on clean training data and the true coefficient vector $\beta$.}\label{fig:minl2mu0indepl1}
\end{figure}

\begin{figure}[H]
\begin{center}
\includegraphics[width=5.25cm]{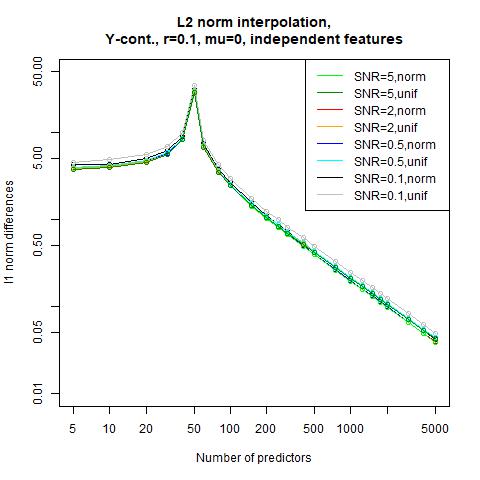} \includegraphics[width=5.25cm]{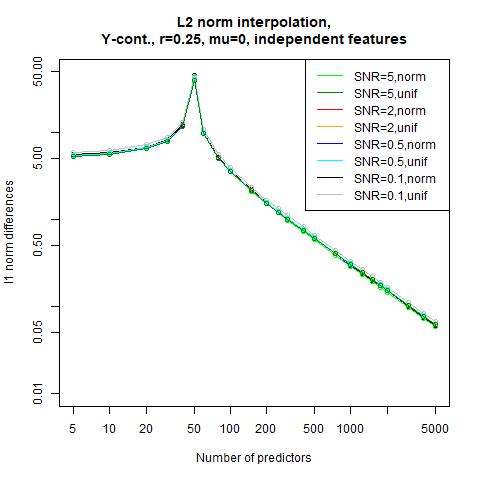} \\ \includegraphics[width=5.25cm]{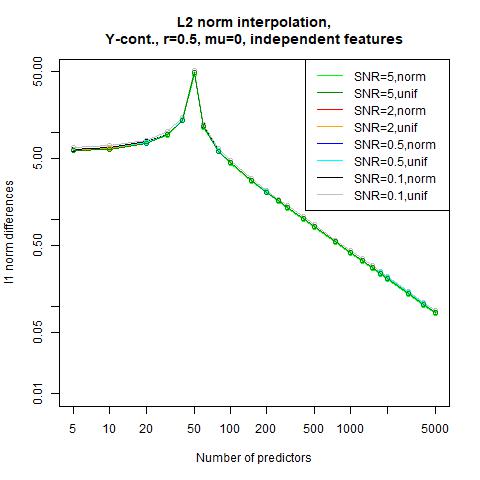} \includegraphics[width=5.25cm]{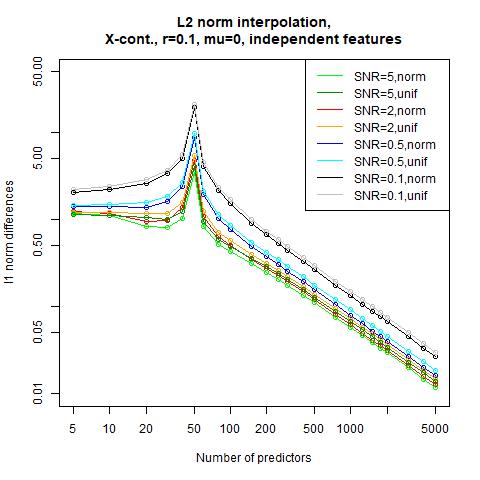} \\ \includegraphics[width=5.25cm]{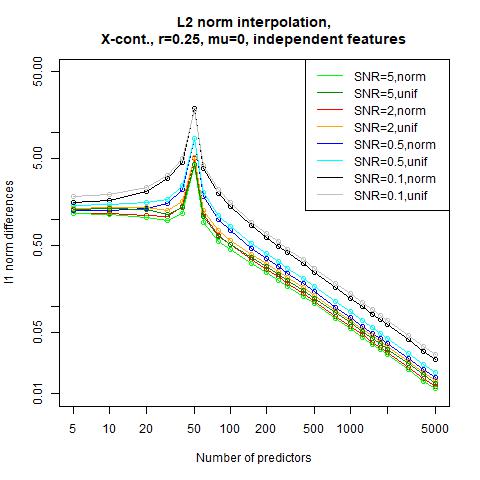} \includegraphics[width=5.25cm]{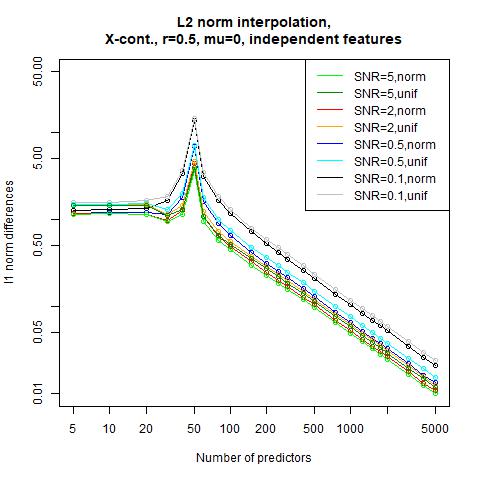} 
\end{center}
\caption{Differences $||\hat \beta-\beta||_1/n$ for the estimated coefficient vector $\hat \beta$ of minimum $l_2$-norm interpolation when trained on contaminated training data and the true coefficient vector $\beta$.}\label{fig:minl2mu0indepl1cont}
\end{figure}

One can observe in Fig. \ref{fig:minl2mu0indepl1} and Fig. \ref{fig:minl2mu0indepl1cont} that the $l_1$-norm coefficient differences attain a peak at $p=n$ and monotonically decrease for larger $p$. For high SNRs on clean data and $X$-contaminated data, one can also observe a local minimum at $p=20$ and $p \in \{30,40\}$, respectively.

\subsubsection{Huber-loss interpolation}

\begin{figure}[H]
\begin{center}
\includegraphics[width=7.5cm]{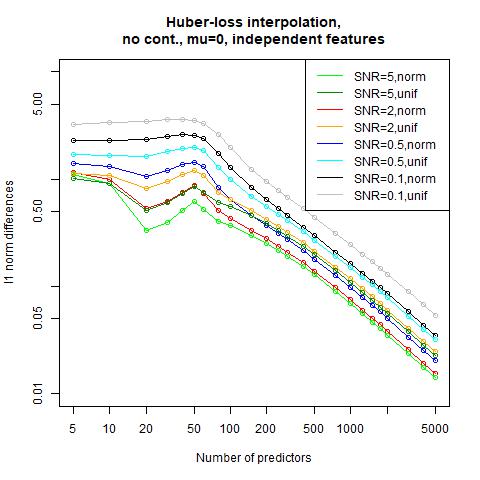} 
\end{center}
\caption{Differences $||\hat \beta-\beta||_1/n$ for the estimated coefficient vector $\hat \beta$ of Huber-loss interpolation when trained on clean training data and the true coefficient vector $\beta$.}\label{fig:hubermu0indepl1}
\end{figure}

\begin{figure}[H]
\begin{center}
\includegraphics[width=5.25cm]{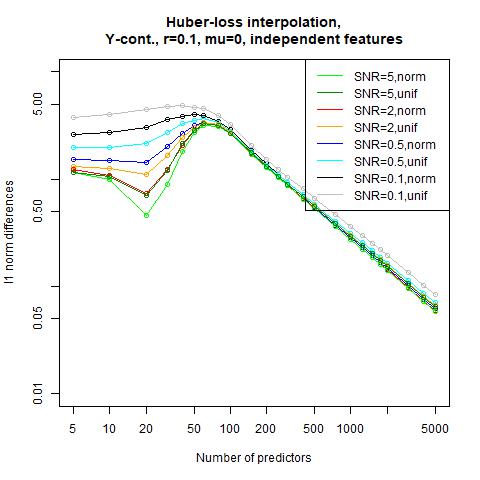} \includegraphics[width=5.25cm]{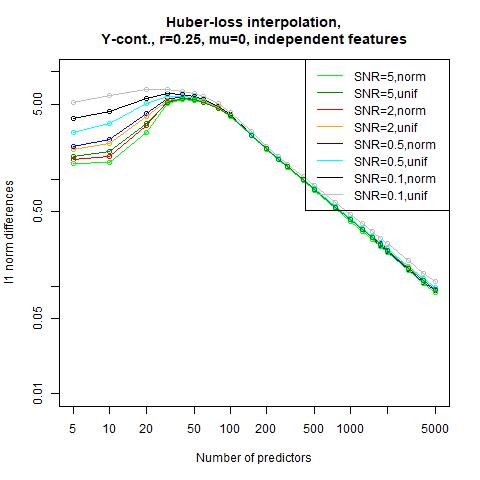} \\ \includegraphics[width=5.25cm]{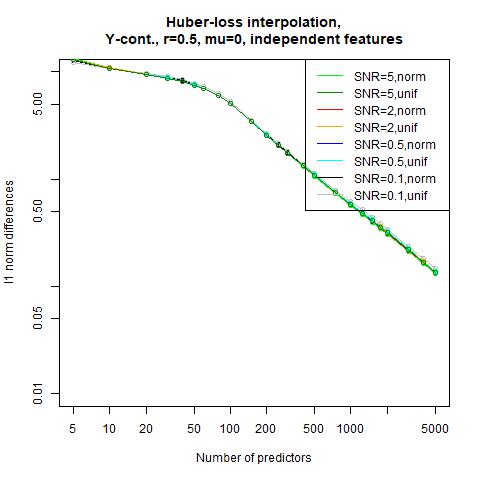} \includegraphics[width=5.25cm]{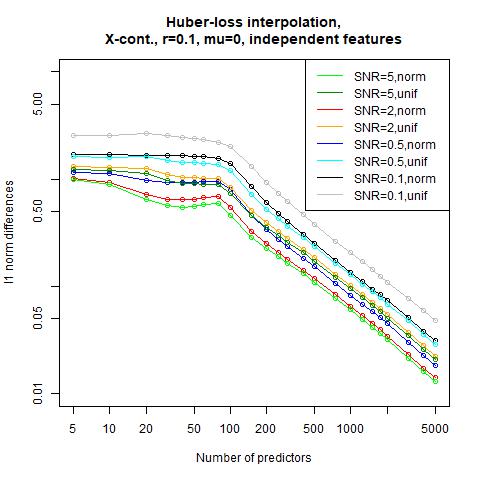} \\ \includegraphics[width=5.25cm]{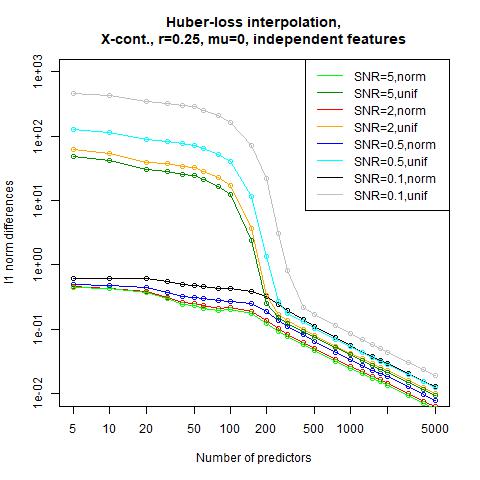} \includegraphics[width=5.25cm]{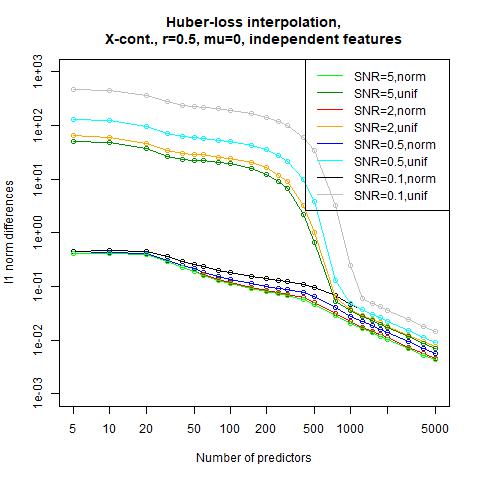} 
\end{center}
\caption{Differences $||\hat \beta-\beta||_1/n$ for the estimated coefficient vector $\hat \beta$ of Huber-loss interpolation when trained on contaminated training data and the true coefficient vector $\beta$.}\label{fig:hubermu0indepl1cont}
\end{figure}

The curves in Fig. \ref{fig:hubermu0indepl1} and Fig. \ref{fig:hubermu0indepl1cont} show that on clean data or contaminated data with low contamination radius, the differences first grow and monotonically decrease as $p$ further grows. For larger contamination radii, the curves are nearly constant until they decrease monotonically.

\subsubsection{Tukey-loss interpolation}

\begin{figure}[H]
\begin{center}
\includegraphics[width=7.5cm]{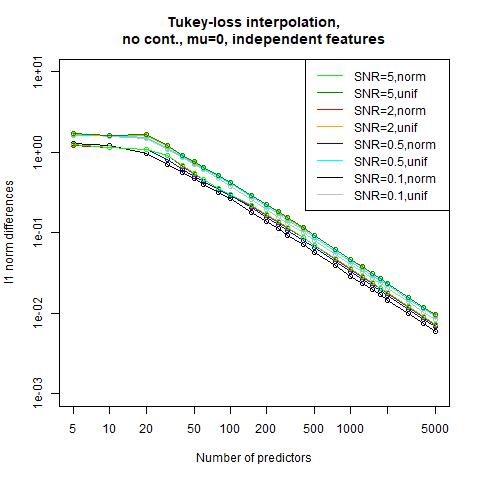} 
\end{center}
\caption{Differences $||\hat \beta-\beta||_1/n$ for the estimated coefficient vector $\hat \beta$ of Tukey-loss interpolation when trained on clean training data and the true coefficient vector $\beta$.}\label{fig:tukeymu0indepl1}
\end{figure}

\begin{figure}[H]
\begin{center}
\includegraphics[width=5.25cm]{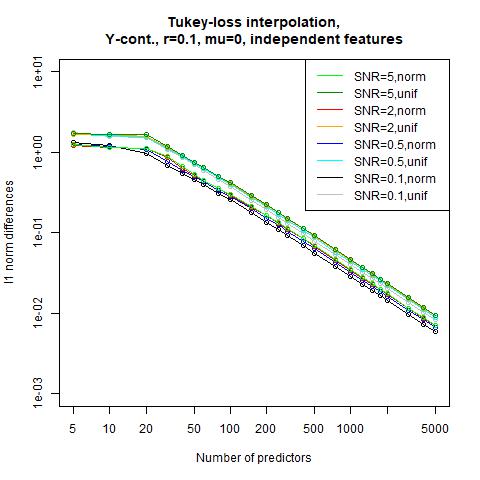} \includegraphics[width=5.25cm]{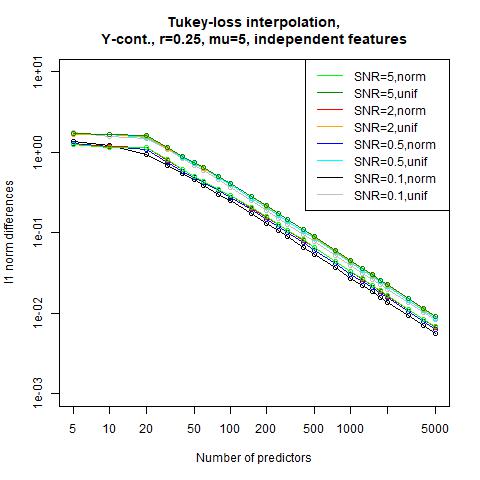} \\ \includegraphics[width=5.25cm]{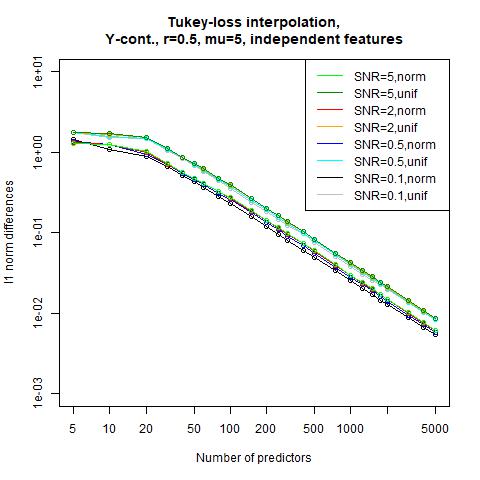} \includegraphics[width=5.25cm]{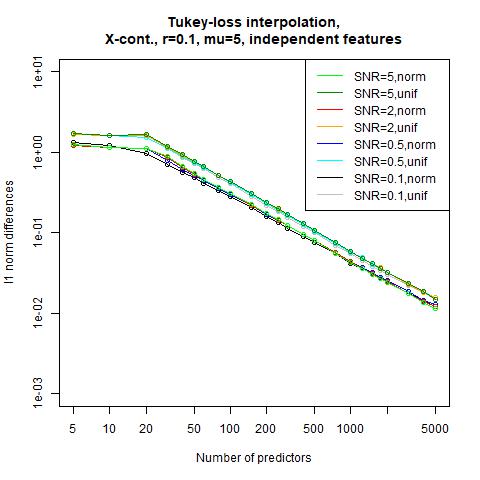} \\ \includegraphics[width=5.25cm]{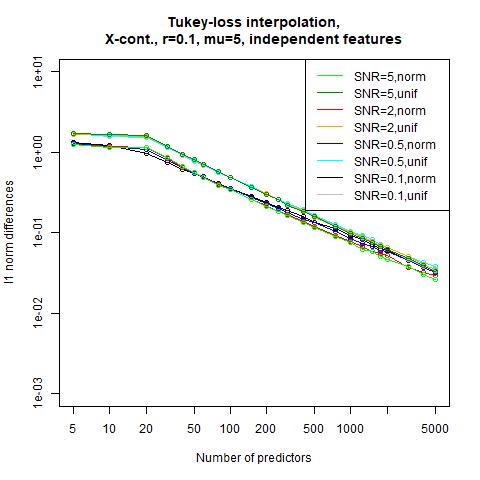} \includegraphics[width=5.25cm]{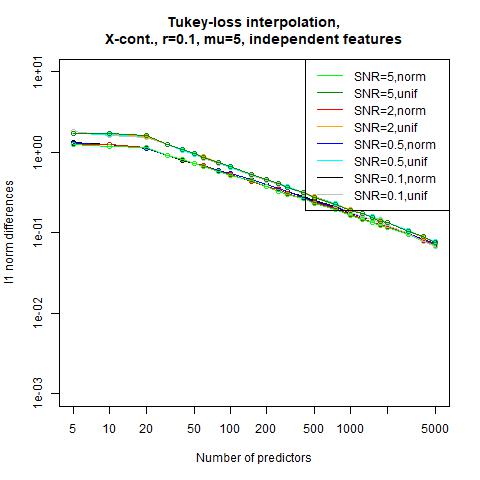} 
\end{center}
\caption{Differences $||\hat \beta-\beta||_1/n$ for the estimated coefficient vector $\hat \beta$ of Tukey-loss interpolation when trained on contaminated training data and the true coefficient vector $\beta$.}\label{fig:tukeymu0indepl1cont}
\end{figure}

Fig. \ref{fig:tukeymu0indepl1} and Fig. \ref{fig:tukeymu0indepl1cont} reveal that regardless of the contamination, the curves are nearly constant until $p=20$ and monotonically decrease as $p$ grows further.

\subsubsection{SLTS-based interpolation}

\begin{figure}[H]
\begin{center}
\includegraphics[width=7.5cm]{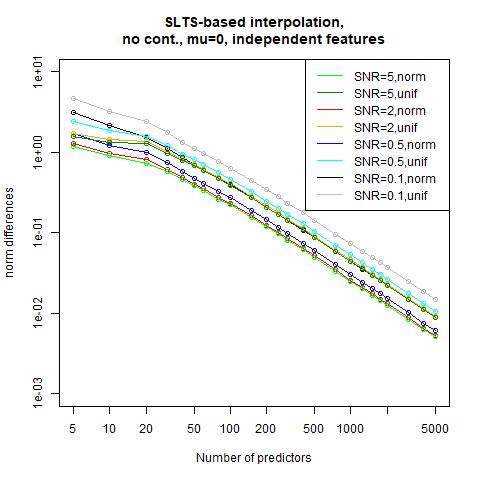} 
\end{center}
\caption{Differences $||\hat \beta-\beta||_1/n$ for the estimated coefficient vector $\hat \beta$ of SLTS-based interpolation when trained on clean training data and the true coefficient vector $\beta$.}\label{fig:sltsmu0indepl1}
\end{figure}

\begin{figure}[H]
\begin{center}
\includegraphics[width=7.5cm]{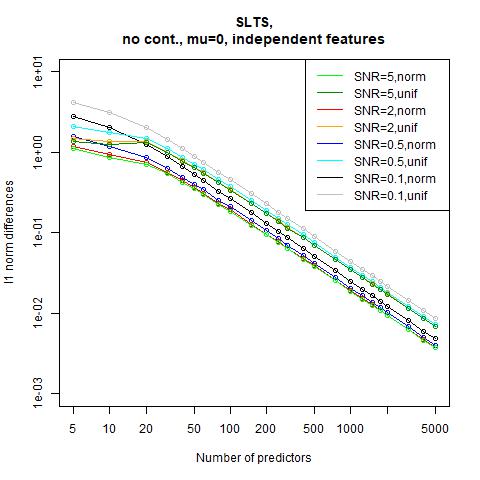} 
\end{center}
\caption{Differences $||\hat \beta-\beta||_1/n$ for the estimated coefficient vector $\hat \beta$ of SLTS when trained on clean training data and the true coefficient vector $\beta$.}\label{fig:rawsltsmu0indepl1}
\end{figure}

\begin{figure}[H]
\begin{center}
\includegraphics[width=5.25cm]{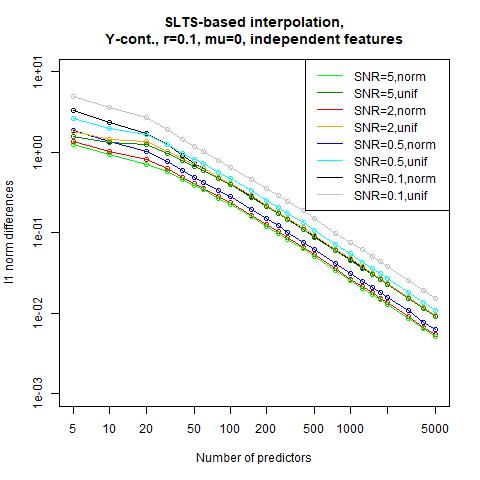} \includegraphics[width=5.25cm]{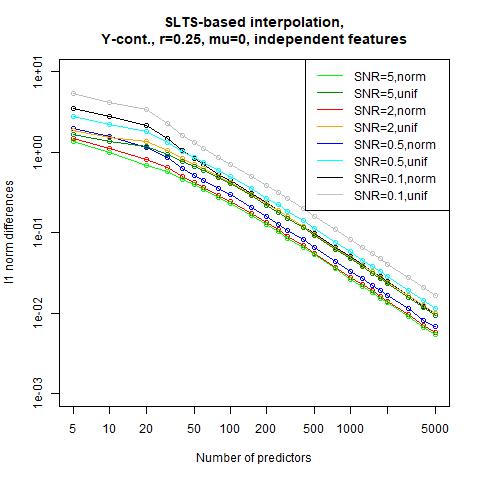} \\ \includegraphics[width=5.25cm]{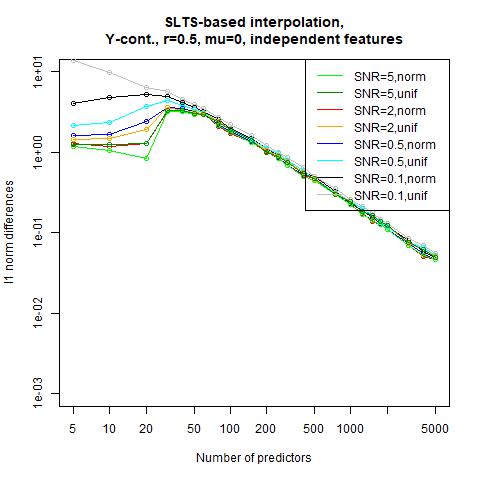} \includegraphics[width=5.25cm]{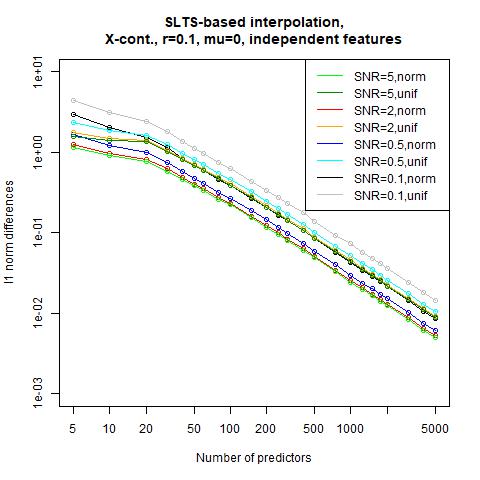} \\ \includegraphics[width=5.25cm]{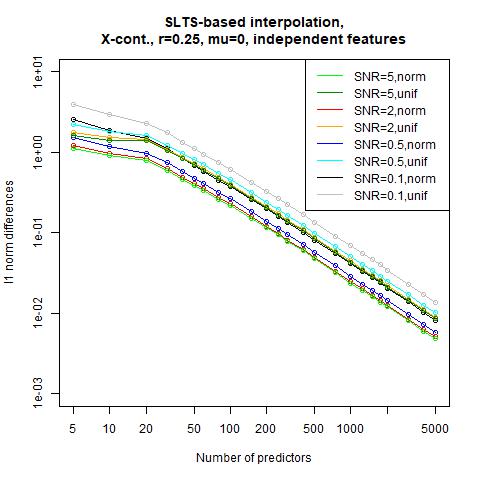} \includegraphics[width=5.25cm]{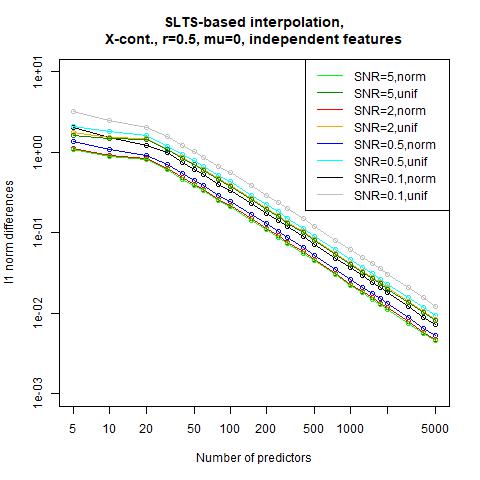} 
\end{center}
\caption{Differences $||\hat \beta-\beta||_1/n$ for the estimated coefficient vector $\hat \beta$ of SLTS-based interpolation when trained on contaminated training data and the true coefficient vector $\beta$.}\label{fig:sltsmu0indepl1cont}
\end{figure}

\begin{figure}[H]
\begin{center}
\includegraphics[width=5.25cm]{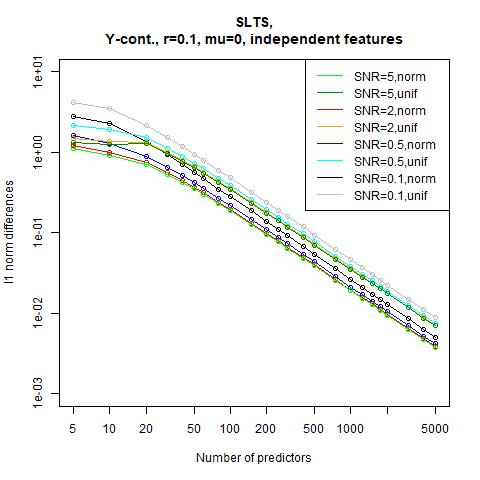} \includegraphics[width=5.25cm]{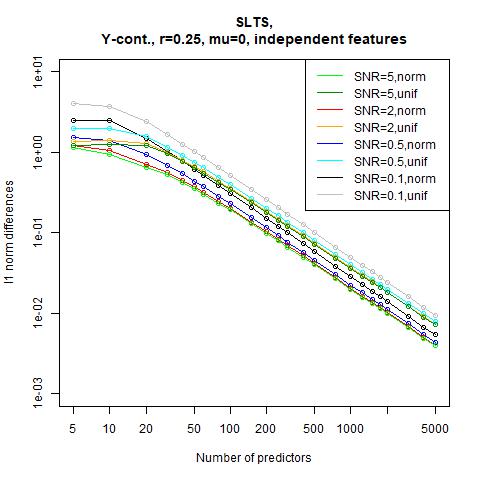} \\ \includegraphics[width=5.25cm]{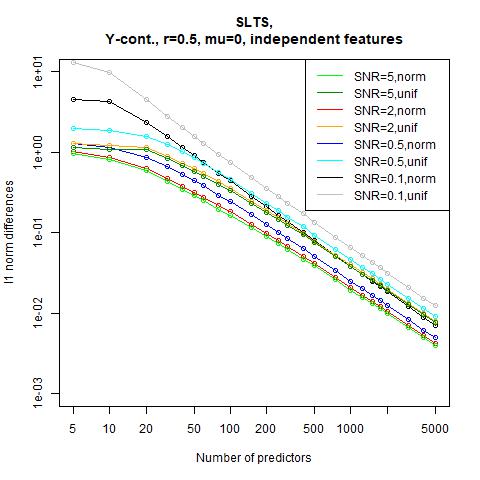} \includegraphics[width=5.25cm]{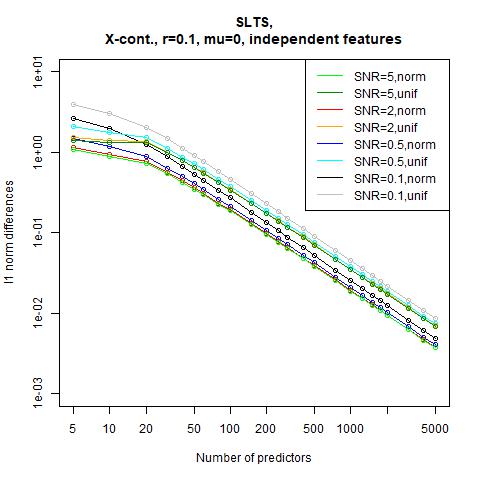} \\ \includegraphics[width=5.25cm]{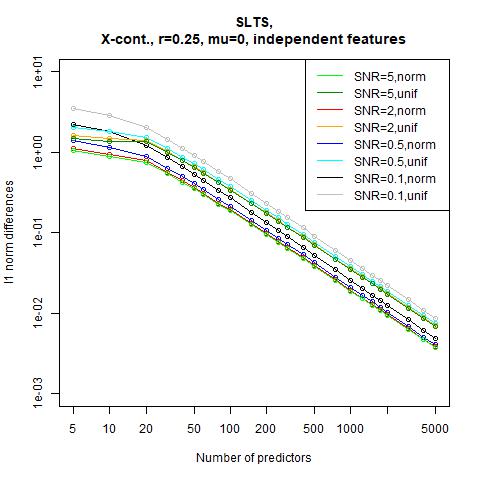} \includegraphics[width=5.25cm]{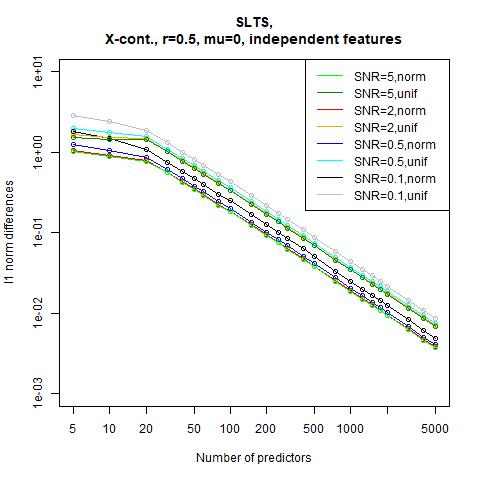} 
\end{center}
\caption{Differences $||\hat \beta-\beta||_1/n$ for the estimated coefficient vector $\hat \beta$ of SLTS when trained on contaminated training data and the true coefficient vector $\beta$.}\label{fig:rawsltsmu0indepl1cont}
\end{figure}

Fig. \ref{fig:sltsmu0indepl1} and Fig. \ref{fig:rawsltsmu0indepl1} show a nearly monotonically decreasing behavior of the coefficient differences on clean data. As for contaminated data, Fig. \ref{fig:sltsmu0indepl1cont} and Fig. \ref{fig:rawsltsmu0indepl1cont} reveal a similar structure as on clean data, both for SLTS and SLTS-based interpolation, except for $Y$-contamination and $r=0.5$, where the differences for SLTS-based interpolation first grow for high SNRs and monotonically decrease afterwards.

\subsubsection{Boosting-based interpolation}

\begin{figure}[H]
\begin{center}
\includegraphics[width=7.5cm]{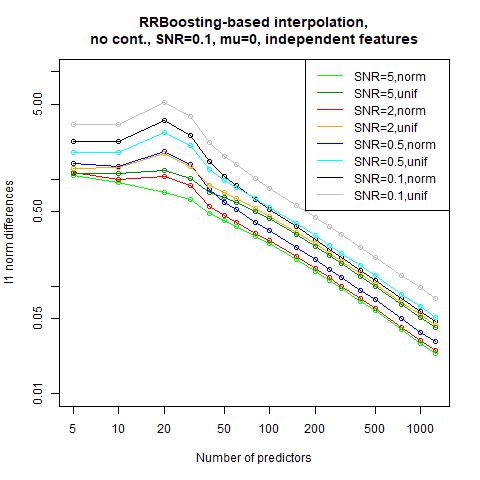} 
\end{center}
\caption{Differences $||\hat \beta-\beta||_1/n$ for the estimated coefficient vector $\hat \beta$ of RRBoosting-based interpolation when trained on clean training data and the true coefficient vector $\beta$.}\label{fig:rrboostmu0indepl1}
\end{figure}

\begin{figure}[H]
\begin{center}
\includegraphics[width=5.25cm]{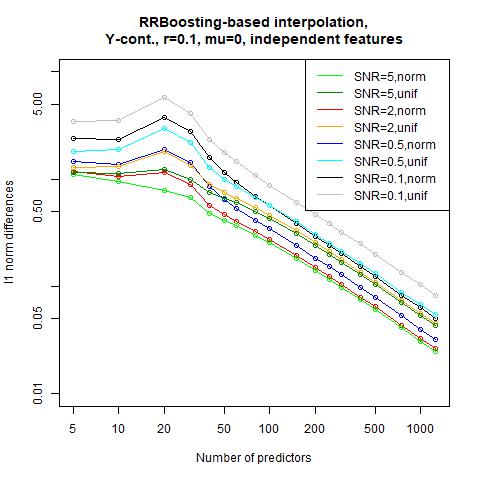} \includegraphics[width=5.25cm]{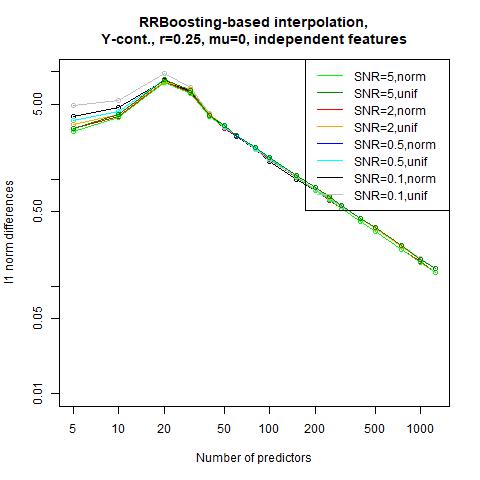} \\ \includegraphics[width=5.25cm]{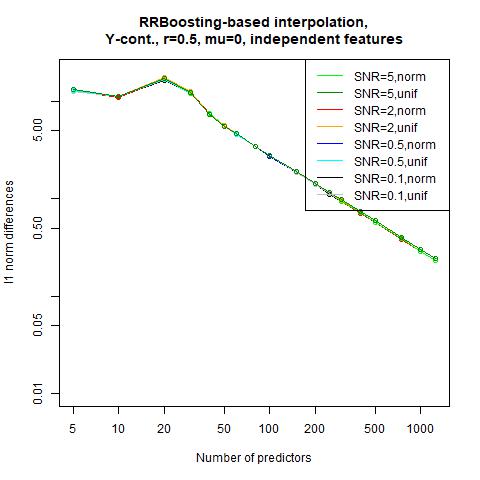} \includegraphics[width=5.25cm]{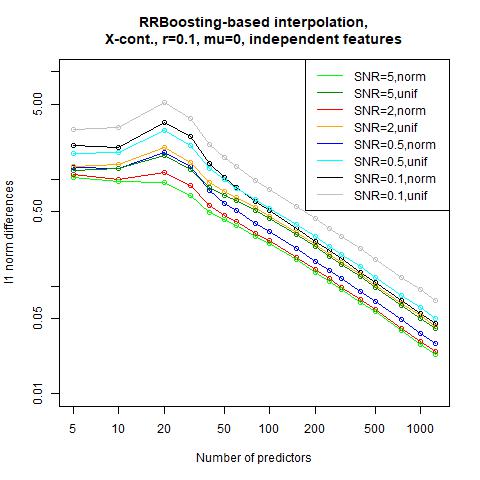} \\ \includegraphics[width=5.25cm]{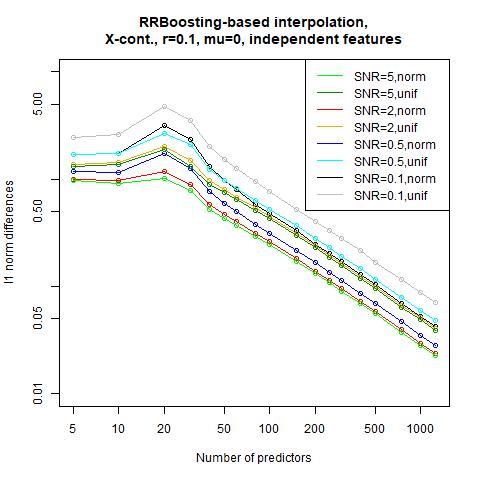} \includegraphics[width=5.25cm]{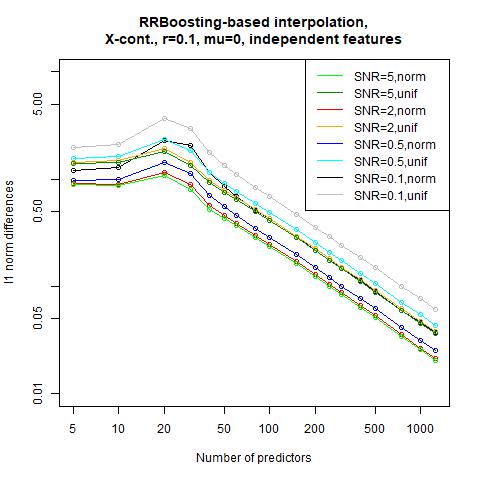} 
\end{center}
\caption{Differences $||\hat \beta-\beta||_1/n$ for the estimated coefficient vector $\hat \beta$ of RRBoosting-based interpolation when trained on contaminated training data and the true coefficient vector $\beta$.}\label{fig:rrboostmu0indepl1cont}
\end{figure}

Fig. \ref{fig:rrboostmu0indepl1} and Fig. \ref{fig:rrboostmu0indepl1cont} show that the $l_1$-differences attain a peak at $p=20$ and monotonically decrease a $p$ grows further.

\subsection{Spiked covariance design, $\mu=0$}

\subsubsection{Minimum $l_2$-norm interpolation}

\begin{figure}[H]
\begin{center}
\includegraphics[width=7.5cm]{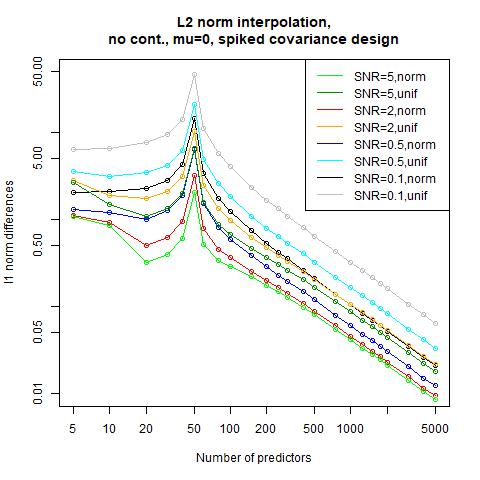} 
\end{center}
\caption{Differences $||\hat \beta-\beta||_1/n$ for the estimated coefficient vector $\hat \beta$ of minimum $l_2$-norm interpolation when trained on clean training data and the true coefficient vector $\beta$.}\label{fig:minl2mu0spikedl1}
\end{figure}

\begin{figure}[H]
\begin{center}
\includegraphics[width=5.25cm]{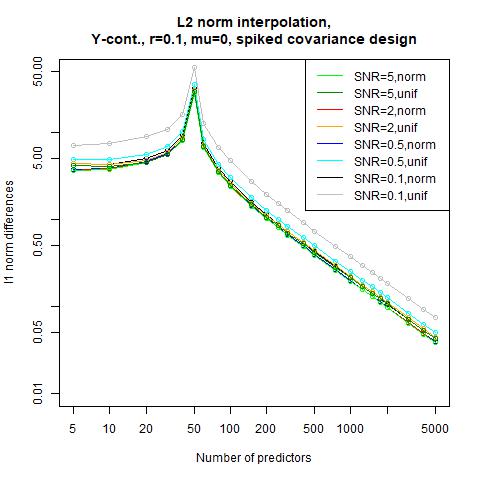} \includegraphics[width=5.25cm]{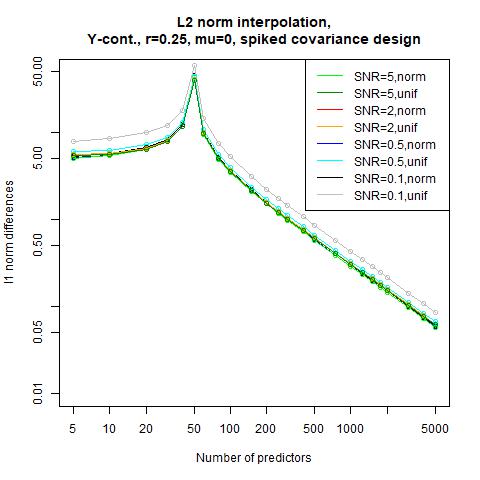} \\ \includegraphics[width=5.25cm]{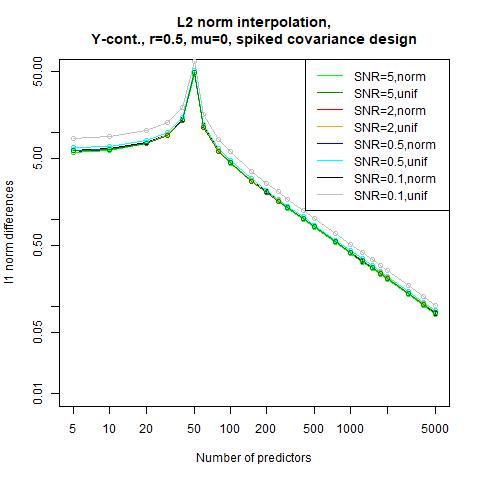} \includegraphics[width=5.25cm]{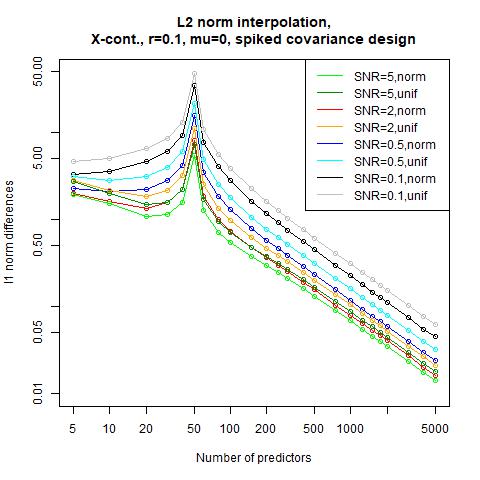} \\ \includegraphics[width=5.25cm]{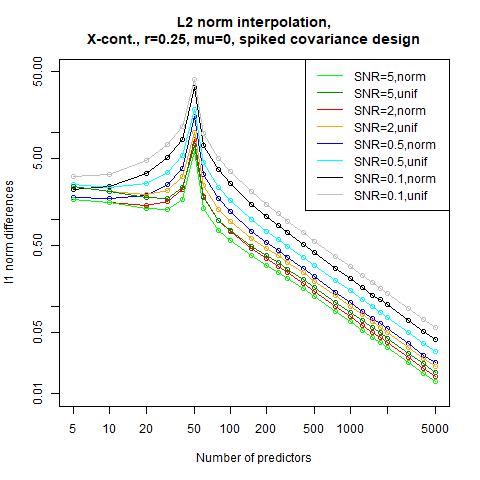} \includegraphics[width=5.25cm]{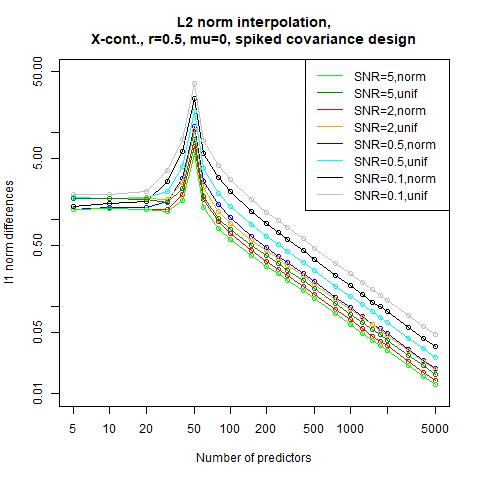} 
\end{center}
\caption{Differences $||\hat \beta-\beta||_1/n$ for the estimated coefficient vector $\hat \beta$ of minimum $l_2$-norm interpolation when trained on contaminated training data and the true coefficient vector $\beta$.}\label{fig:minl2mu0spikedl1cont}
\end{figure}

\subsubsection{Huber-loss interpolation}

\begin{figure}[H]
\begin{center}
\includegraphics[width=7.5cm]{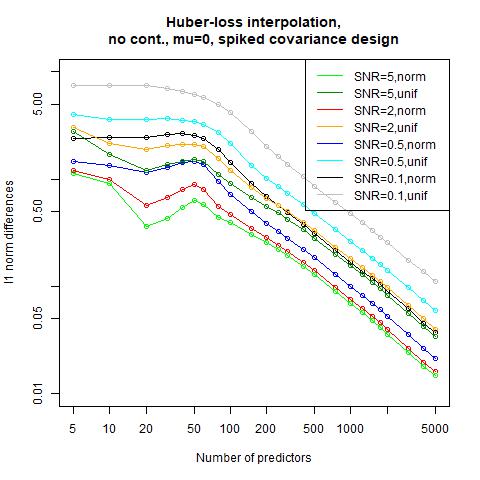} 
\end{center}
\caption{Differences $||\hat \beta-\beta||_1/n$ for the estimated coefficient vector $\hat \beta$ of Huber-norm interpolation when trained on clean training data and the true coefficient vector $\beta$.}\label{fig:hubermu0indepspikedl1}
\end{figure}

\begin{figure}[H]
\begin{center}
\includegraphics[width=5.25cm]{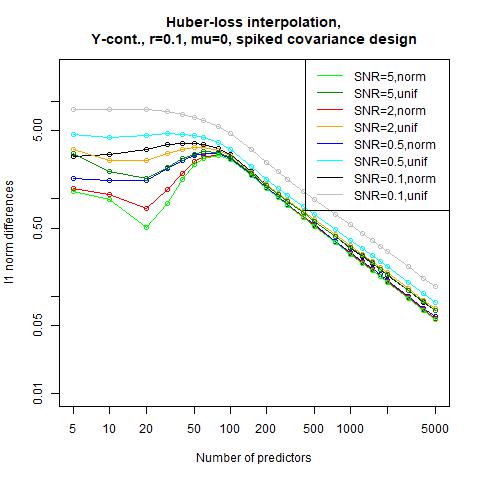} \includegraphics[width=5.25cm]{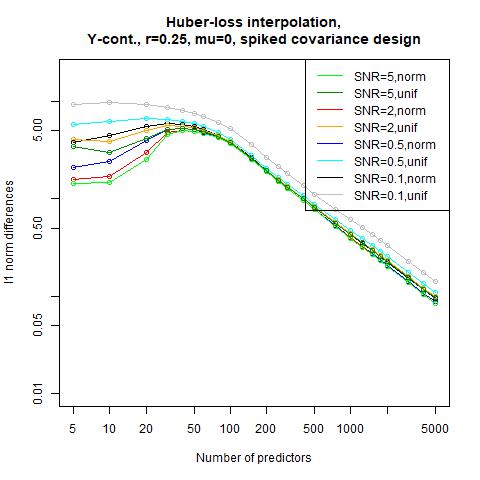} \\ \includegraphics[width=5.25cm]{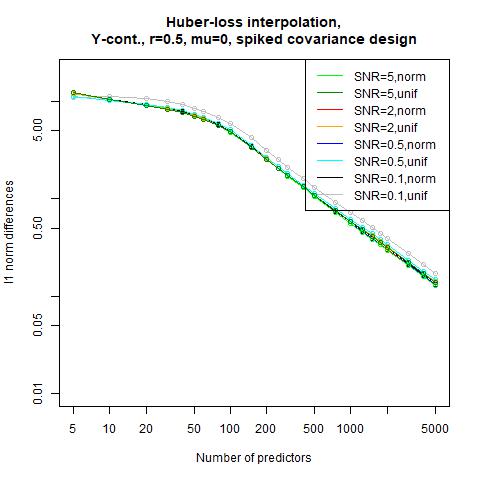} \includegraphics[width=5.25cm]{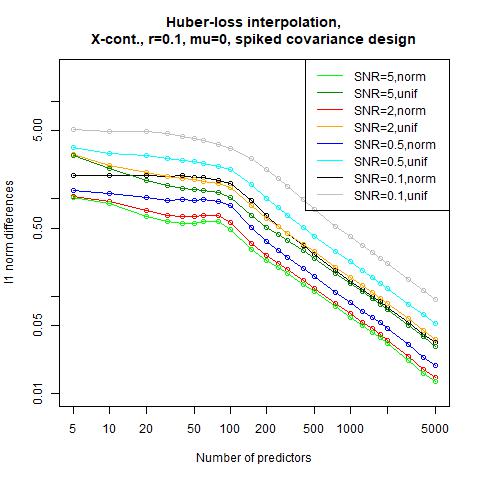} \\ \includegraphics[width=5.25cm]{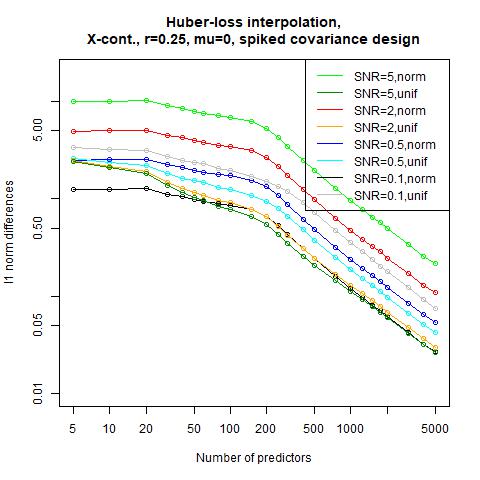} \includegraphics[width=5.25cm]{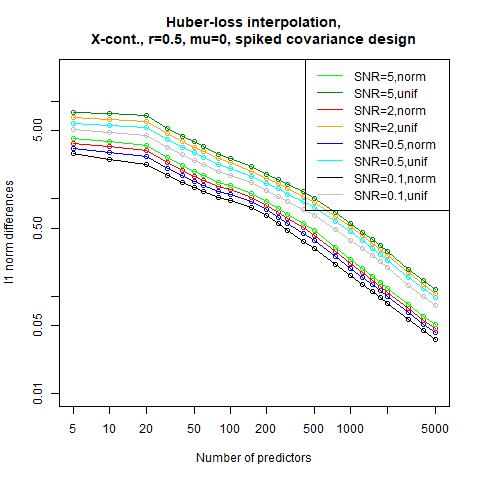} 
\end{center}
\caption{Differences $||\hat \beta-\beta||_1/n$ for the estimated coefficient vector $\hat \beta$ of Huber-norm interpolation when trained on contaminated training data and the true coefficient vector $\beta$.}\label{fig:hubermu0indepspikedl1cont}
\end{figure}

\subsubsection{Tukey-loss interpolation}

\begin{figure}[H]
\begin{center}
\includegraphics[width=7.5cm]{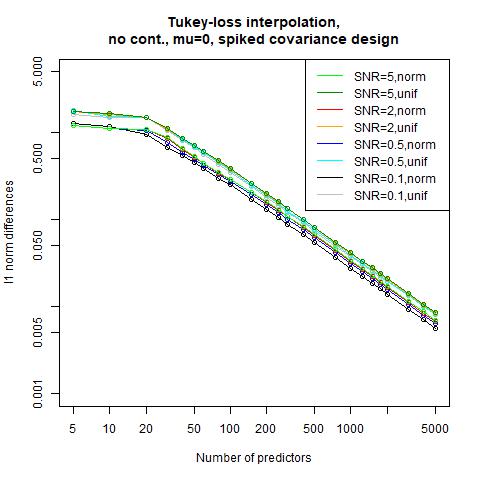} 
\end{center}
\caption{Differences $||\hat \beta-\beta||_1/n$ for the estimated coefficient vector $\hat \beta$ of Tukey-norm interpolation when trained on clean training data and the true coefficient vector $\beta$.}\label{fig:tukeymu0spikedl1}
\end{figure}

\begin{figure}[H]
\begin{center}
\includegraphics[width=5.25cm]{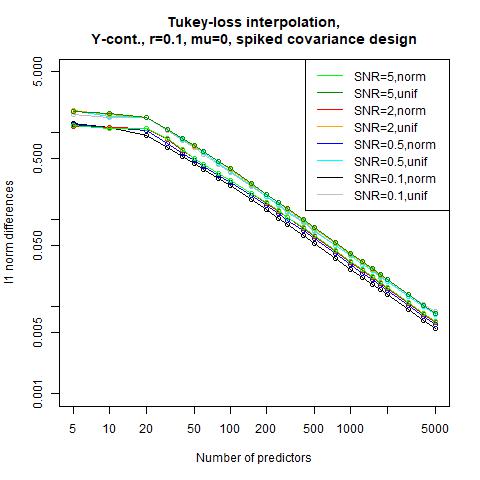} \includegraphics[width=5.25cm]{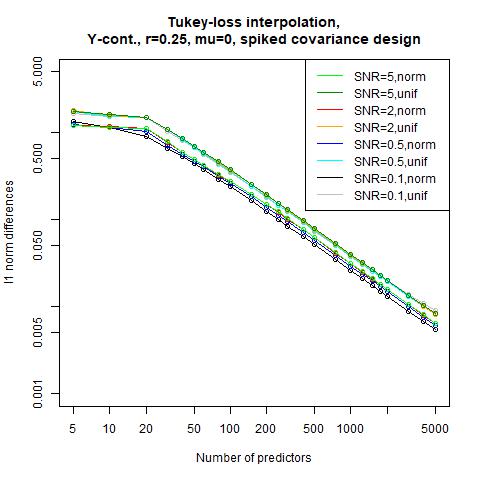} \\ \includegraphics[width=5.25cm]{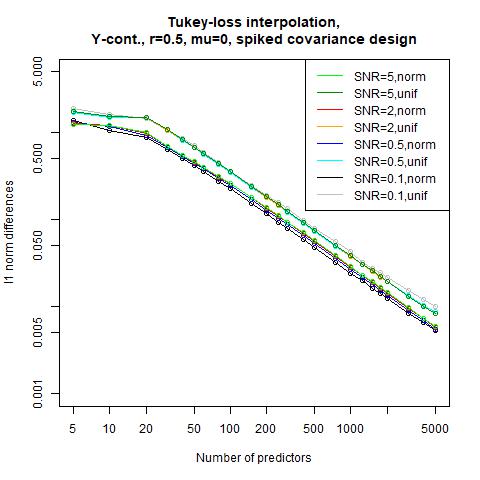} \includegraphics[width=5.25cm]{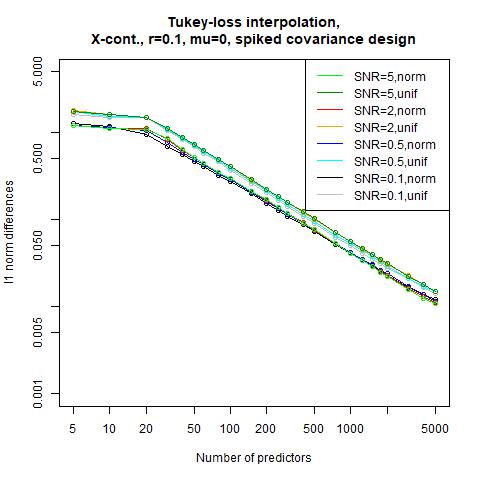} \\ \includegraphics[width=5.25cm]{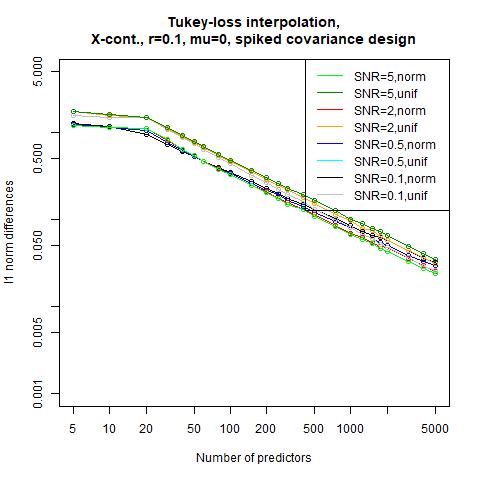} \includegraphics[width=5.25cm]{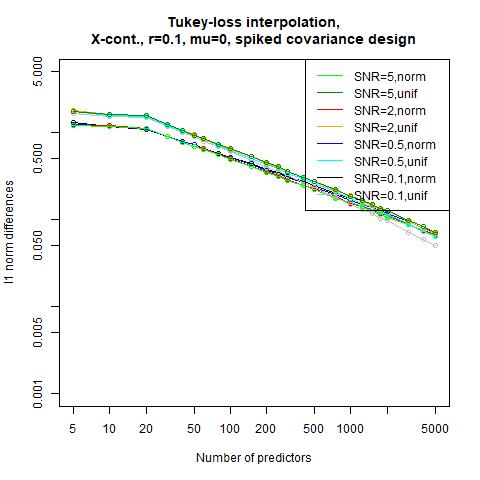} 
\end{center}
\caption{Differences $||\hat \beta-\beta||_1/n$ for the estimated coefficient vector $\hat \beta$ of Tukey-norm interpolation when trained on contaminated training data and the true coefficient vector $\beta$.}\label{fig:tukeymu0spikedl1cont}
\end{figure}

\subsubsection{SLTS-based interpolation}

\begin{figure}[H]
\begin{center}
\includegraphics[width=7.5cm]{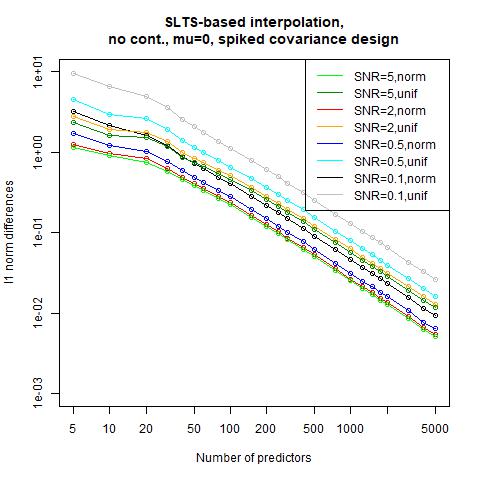} 
\end{center}
\caption{Differences $||\hat \beta-\beta||_1/n$ for the estimated coefficient vector $\hat \beta$ of SLTS-based interpolation when trained on clean training data and the true coefficient vector $\beta$.}\label{fig:sltsmu0spikedl1}
\end{figure}

\begin{figure}[H]
\begin{center}
\includegraphics[width=7.5cm]{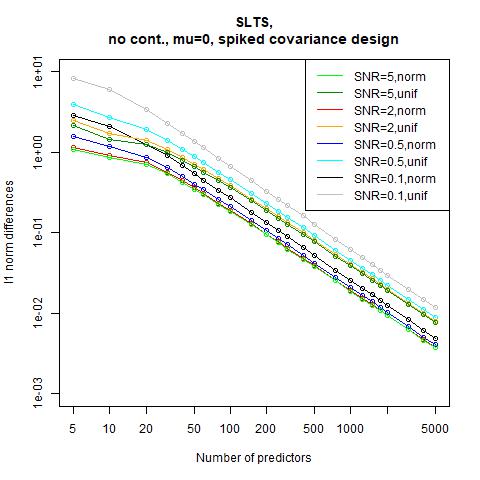} 
\end{center}
\caption{Differences $||\hat \beta-\beta||_1/n$ for the estimated coefficient vector $\hat \beta$ of SLTS when trained on clean training data and the true coefficient vector $\beta$.}\label{fig:rawsltsmu0spikedl1}
\end{figure}

\begin{figure}[H]
\begin{center}
\includegraphics[width=5.25cm]{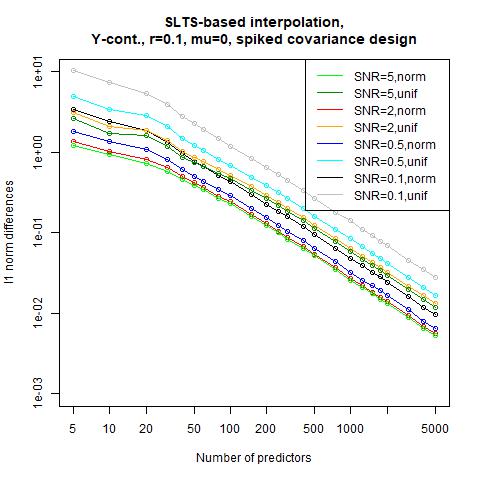} \includegraphics[width=5.25cm]{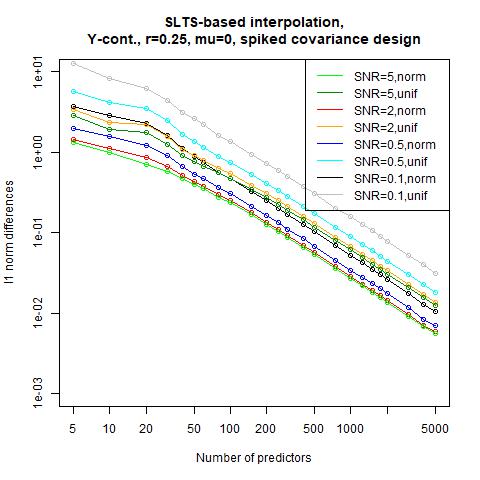} \\ \includegraphics[width=5.25cm]{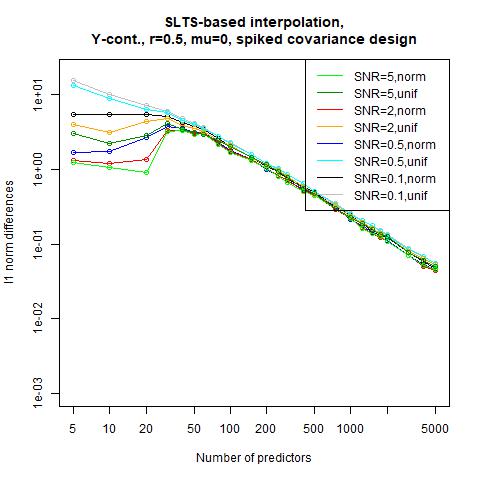} \includegraphics[width=5.25cm]{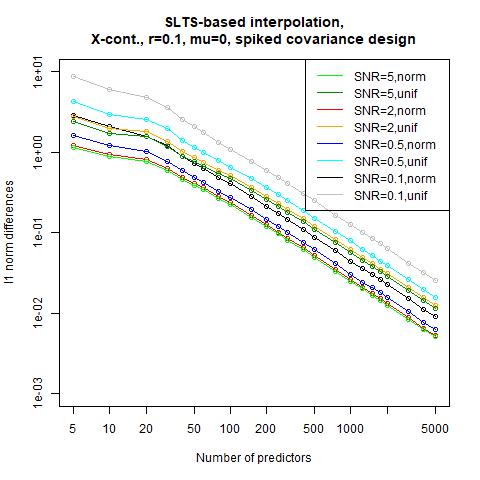} \\ \includegraphics[width=5.25cm]{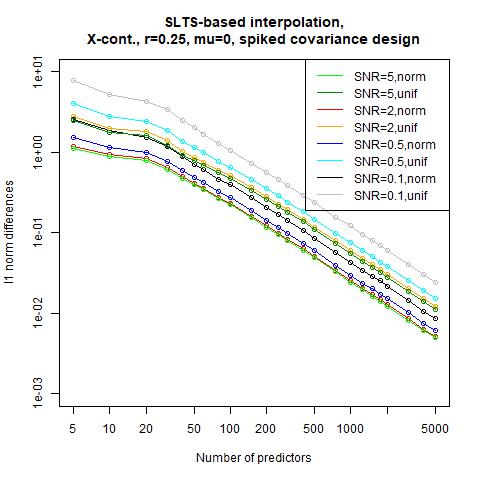} \includegraphics[width=5.25cm]{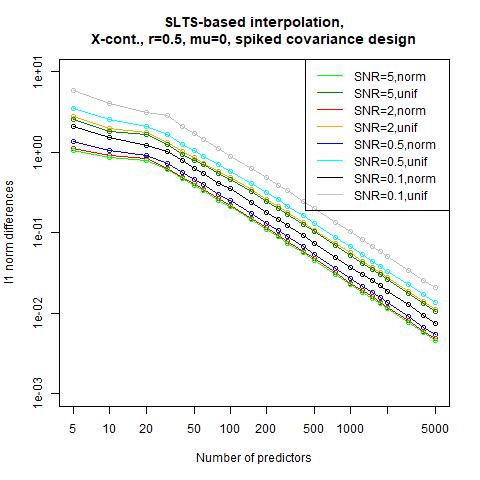} 
\end{center}
\caption{Differences $||\hat \beta-\beta||_1/n$ for the estimated coefficient vector $\hat \beta$ of SLTS-based interpolation when trained on contaminated training data and the true coefficient vector $\beta$.}\label{fig:sltsmu0spikedl1cont}
\end{figure}

\begin{figure}[H]
\begin{center}
\includegraphics[width=5.25cm]{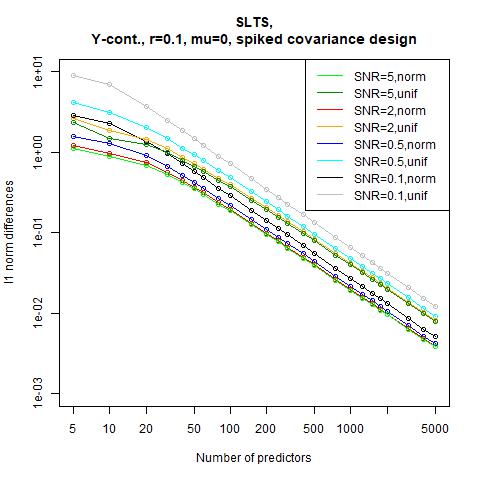} \includegraphics[width=5.25cm]{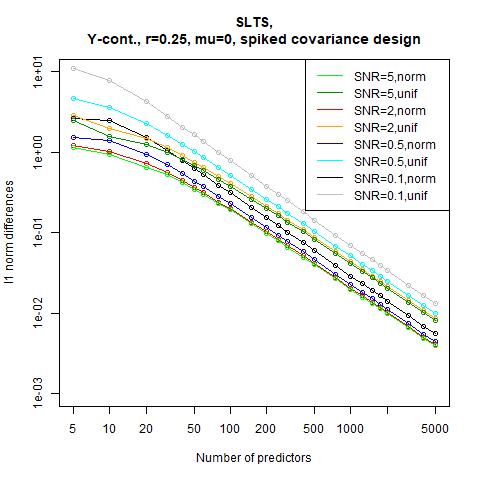} \\ \includegraphics[width=5.25cm]{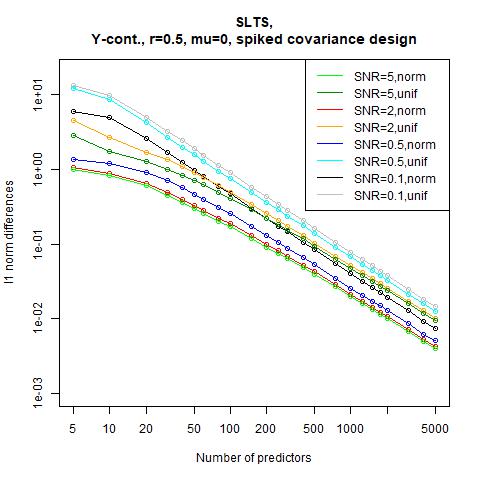} \includegraphics[width=5.25cm]{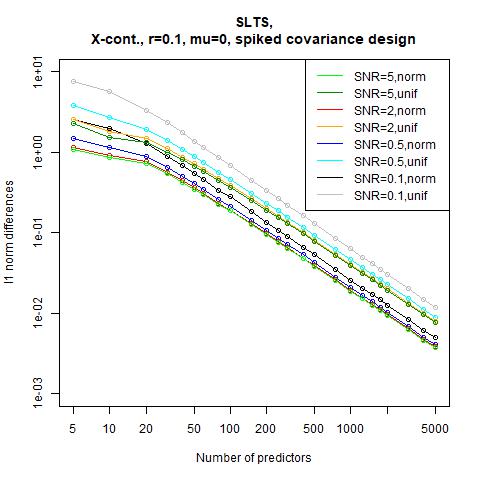} \\ \includegraphics[width=5.25cm]{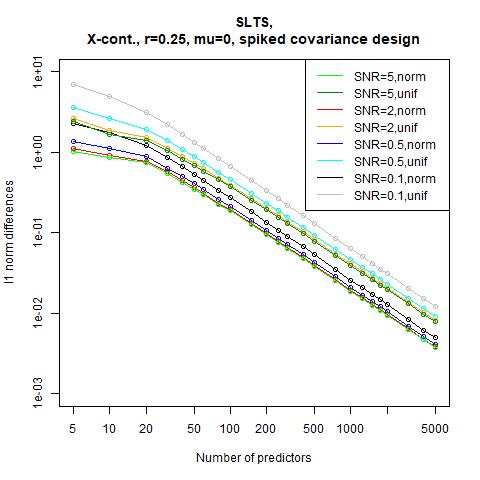} \includegraphics[width=5.25cm]{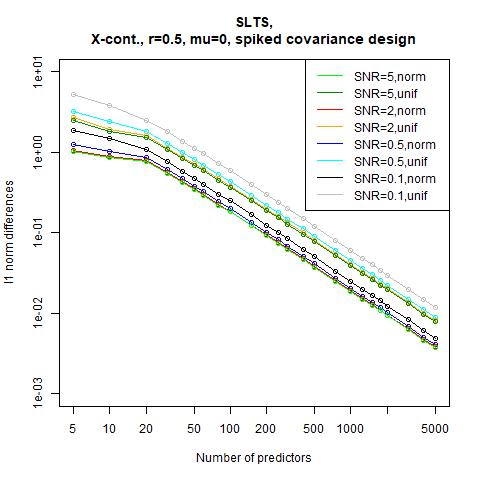} 
\end{center}
\caption{Differences $||\hat \beta-\beta||_1/n$ for the estimated coefficient vector $\hat \beta$ of SLTS when trained on contaminated training data and the true coefficient vector $\beta$.}\label{fig:rawsltsmu0spikedl1cont}
\end{figure}

\subsubsection{Boosting-based interpolation}

\begin{figure}[H]
\begin{center}
\includegraphics[width=7.5cm]{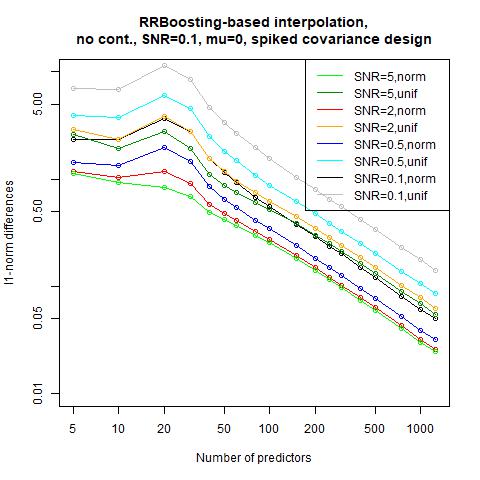} 
\end{center}
\caption{Differences $||\hat \beta-\beta||_1/n$ for the estimated coefficient vector $\hat \beta$ of RRBoosting-based interpolation when trained on clean training data and the true coefficient vector $\beta$.}\label{fig:rrboostmu0spikedl1}
\end{figure}

\begin{figure}[H]
\begin{center}
\includegraphics[width=5.25cm]{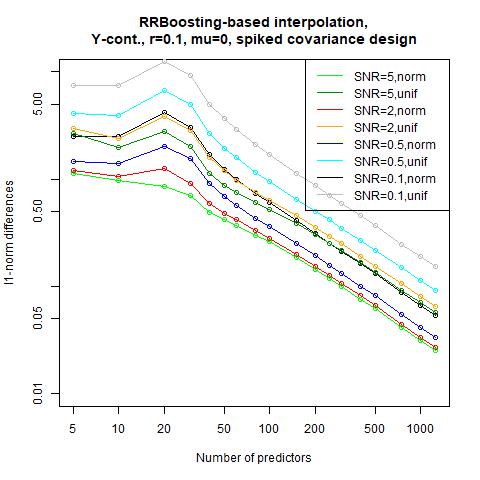} \includegraphics[width=5.25cm]{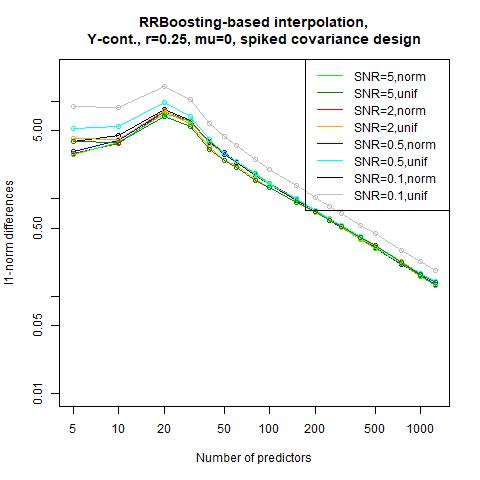} \\ \includegraphics[width=5.25cm]{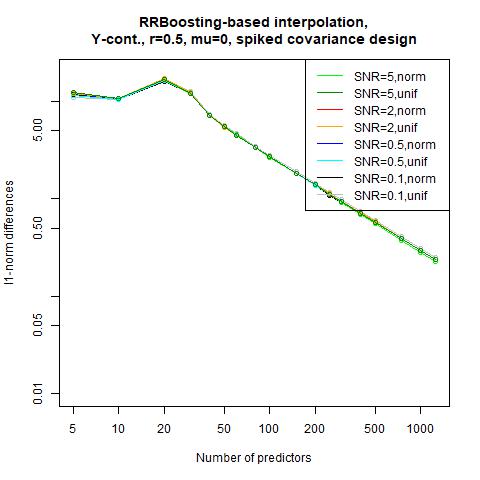} \includegraphics[width=5.25cm]{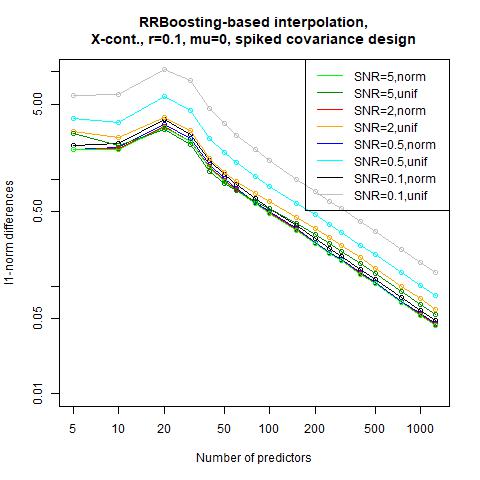} \\ \includegraphics[width=5.25cm]{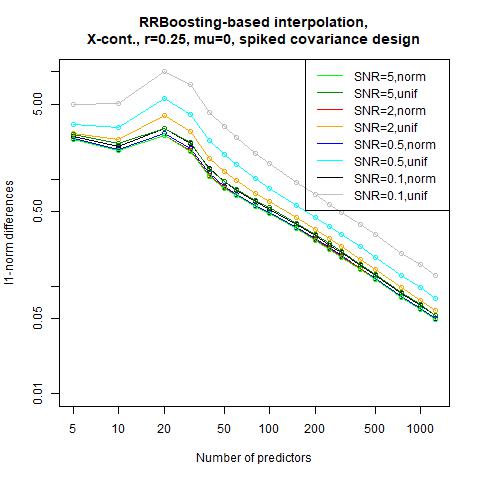} \includegraphics[width=5.25cm]{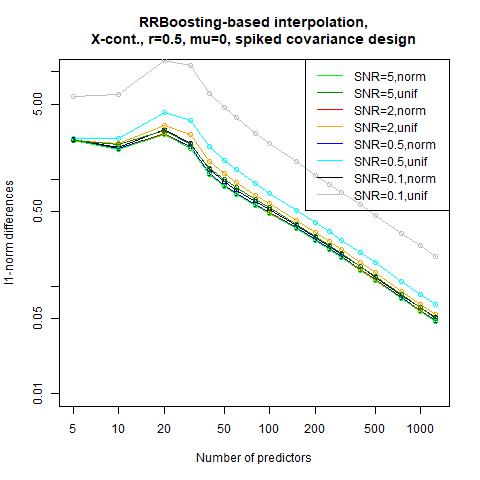} 
\end{center}
\caption{Differences $||\hat \beta-\beta||_1/n$ for the estimated coefficient vector $\hat \beta$ of RRBoosting-based interpolation when trained on contaminated training data and the true coefficient vector $\beta$.}\label{fig:rrboostmu0spikedl1cont}
\end{figure}

As one can observe in Fig. \ref{fig:minl2mu0spikedl1}, Fig. \ref{fig:minl2mu0spikedl1cont}, Fig. \ref{fig:hubermu0indepspikedl1}, Fig. \ref{fig:hubermu0indepspikedl1cont}, Fig. \ref{fig:tukeymu0spikedl1}, Fig. \ref{fig:tukeymu0spikedl1cont}, Fig. \ref{fig:sltsmu0spikedl1}, Fig. \ref{fig:sltsmu0spikedl1cont}, Fig. \ref{fig:rawsltsmu0spikedl1}, Fig. \ref{fig:rawsltsmu0spikedl1cont}, Fig. \ref{fig:rrboostmu0spikedl1} and Fig. \ref{fig:rrboostmu0spikedl1cont}, the norm difference curves resemble those for the independent design, with slightly higher values.

\subsection{Independent features, $\mu=5$}

\subsubsection{Minimum $l_2$-norm interpolation}

\begin{figure}[H]
\begin{center}
\includegraphics[width=7.5cm]{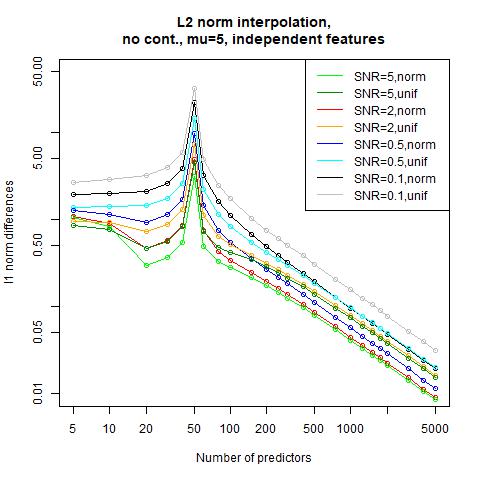} 
\end{center}
\caption{Differences $||\hat \beta-\beta||_1/n$ for the estimated coefficient vector $\hat \beta$ of minimum $l_2$-norm interpolation when trained on clean training data and the true coefficient vector $\beta$.}\label{fig:minl2mu5indepl1}
\end{figure}

\begin{figure}[H]
\begin{center}
\includegraphics[width=5.25cm]{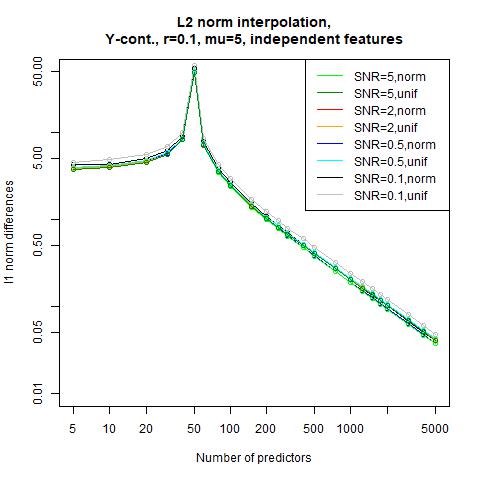} \includegraphics[width=5.25cm]{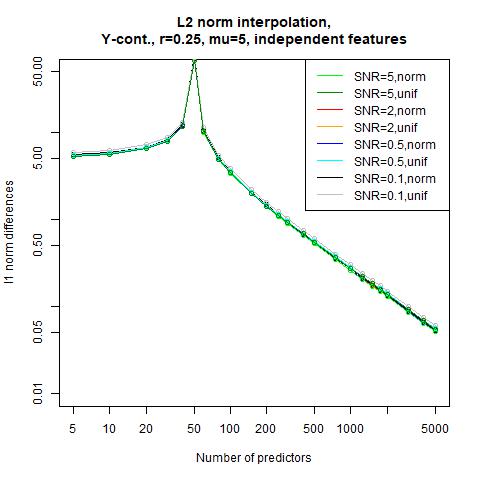} \\ \includegraphics[width=5.25cm]{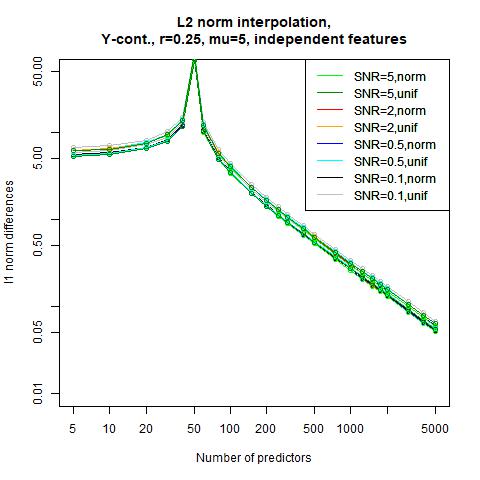} \includegraphics[width=5.25cm]{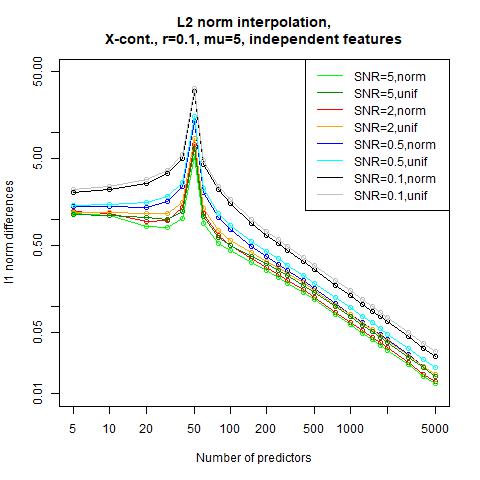} \\ \includegraphics[width=5.25cm]{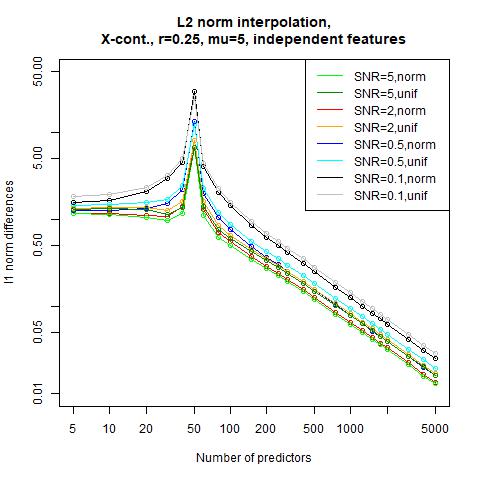} \includegraphics[width=5.25cm]{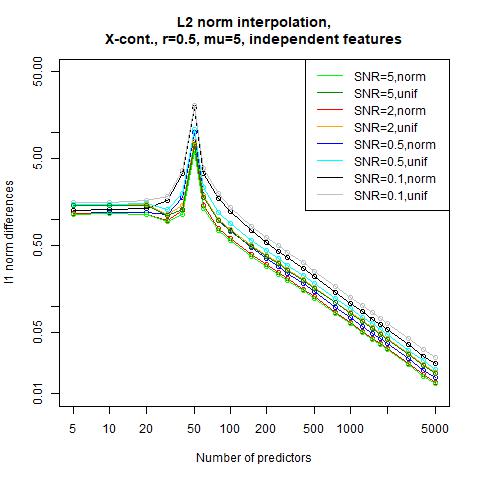} 
\end{center}
\caption{Differences $||\hat \beta-\beta||_1/n$ for the estimated coefficient vector $\hat \beta$ of minimum $l_2$-norm interpolation when trained on contaminated training data and the true coefficient vector $\beta$.}\label{fig:minl2mu5indepl1cont}
\end{figure}

\subsubsection{Huber-loss interpolation}

\begin{figure}[H]
\begin{center}
\includegraphics[width=7.5cm]{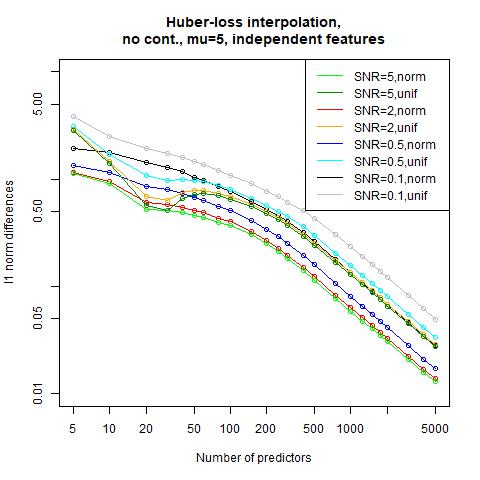} 
\end{center}
\caption{Differences $||\hat \beta-\beta||_1/n$ for the estimated coefficient vector $\hat \beta$ of Huber-loss interpolation when trained on clean training data and the true coefficient vector $\beta$.}\label{fig:hubermu5indepl1}
\end{figure}

\begin{figure}[H]
\begin{center}
\includegraphics[width=5.25cm]{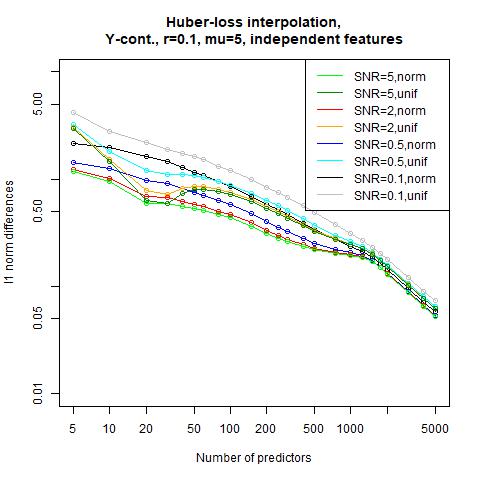} \includegraphics[width=5.25cm]{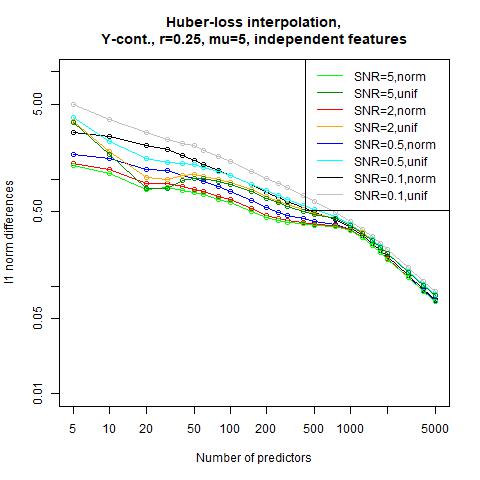} \\ \includegraphics[width=5.25cm]{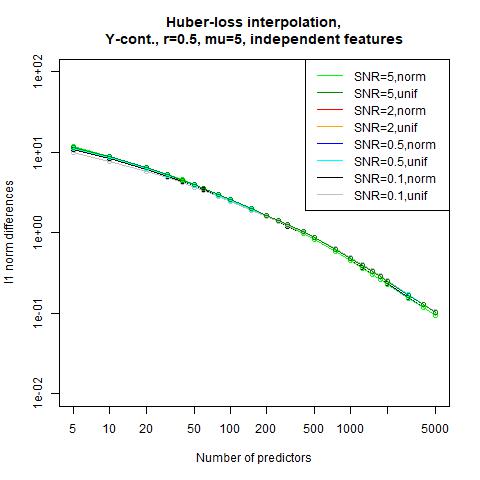} \includegraphics[width=5.25cm]{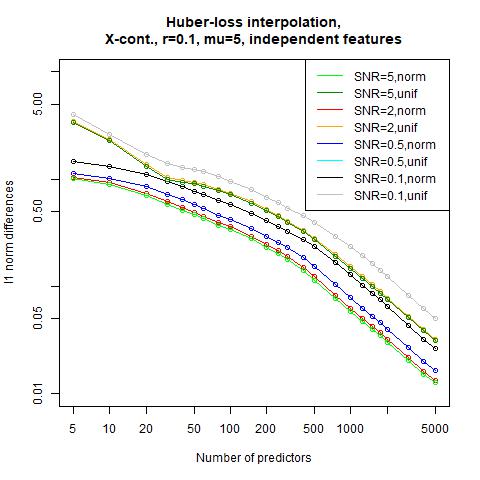} \\ \includegraphics[width=5.25cm]{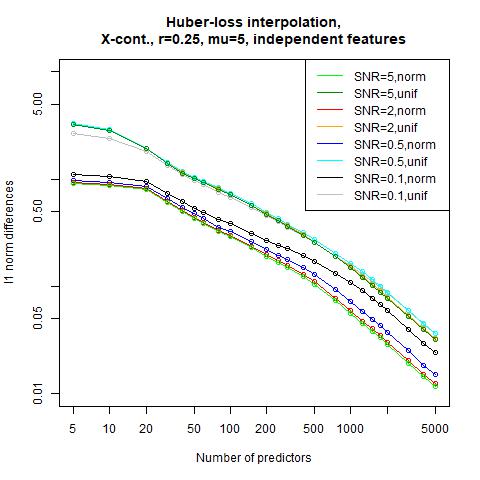} \includegraphics[width=5.25cm]{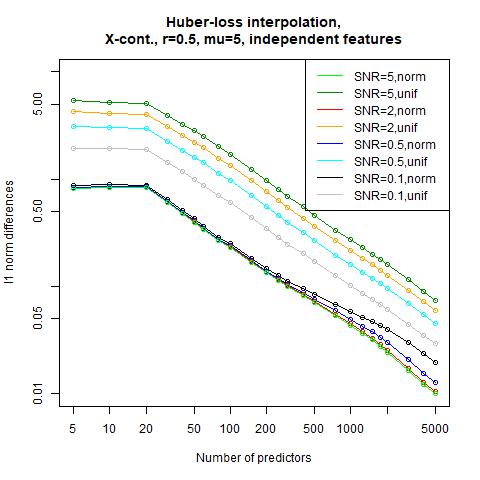} 
\end{center}
\caption{Differences $||\hat \beta-\beta||_1/n$ for the estimated coefficient vector $\hat \beta$ of Huber-loss interpolation when trained on contaminated training data and the true coefficient vector $\beta$.}\label{fig:hubermu5indepl1cont}
\end{figure}

\subsubsection{SLTS-based interpolation}

\begin{figure}[H]
\begin{center}
\includegraphics[width=7.5cm]{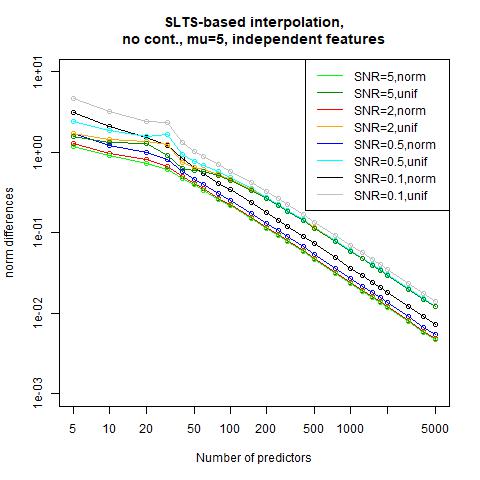} 
\end{center}
\caption{Differences $||\hat \beta-\beta||_1/n$ for the estimated coefficient vector $\hat \beta$ of SLTS-based interpolation when trained on clean training data and the true coefficient vector $\beta$.}\label{fig:sltsmu5indepl1}
\end{figure}

\begin{figure}[H]
\begin{center}
\includegraphics[width=7.5cm]{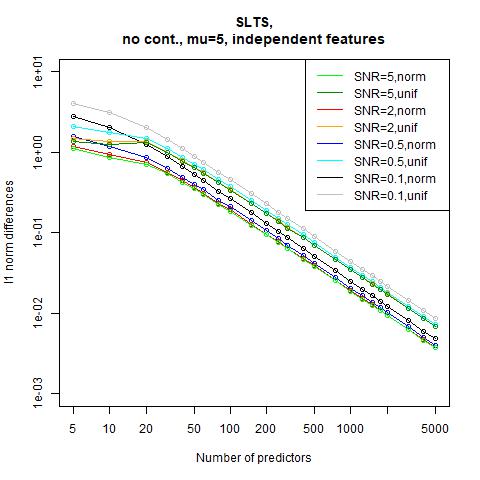} 
\end{center}
\caption{Differences $||\hat \beta-\beta||_1/n$ for the estimated coefficient vector $\hat \beta$ of SLTS when trained on clean training data and the true coefficient vector $\beta$.}\label{fig:rawsltsmu5indepl1}
\end{figure}

\begin{figure}[H]
\begin{center}
\includegraphics[width=5.25cm]{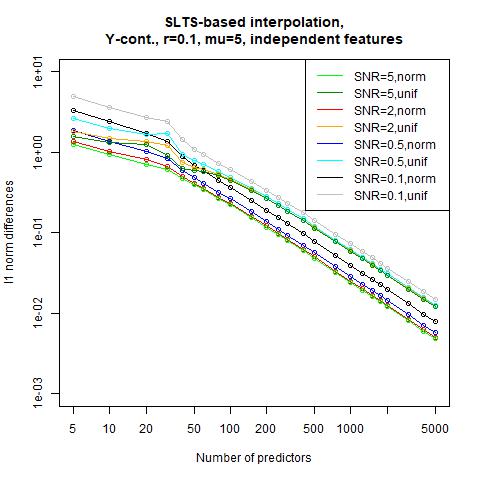} \includegraphics[width=5.25cm]{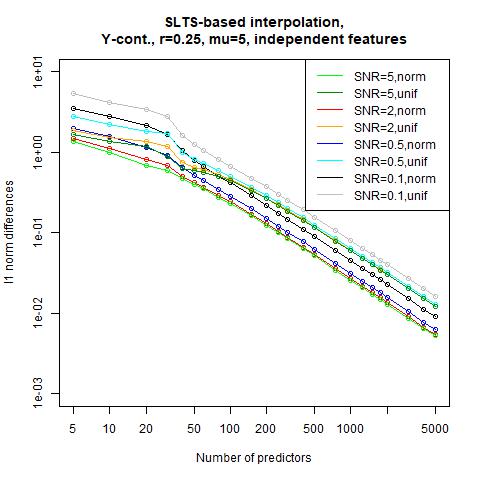} \\ \includegraphics[width=5.25cm]{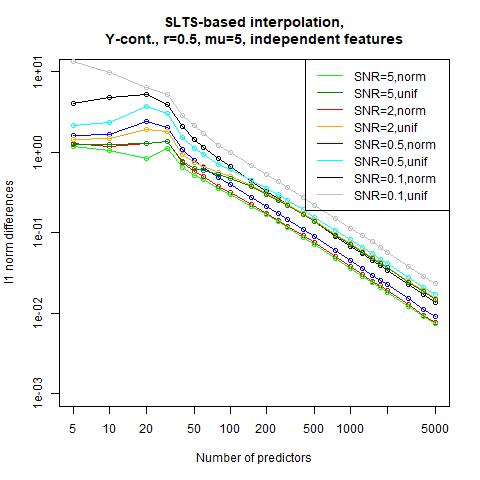} \includegraphics[width=5.25cm]{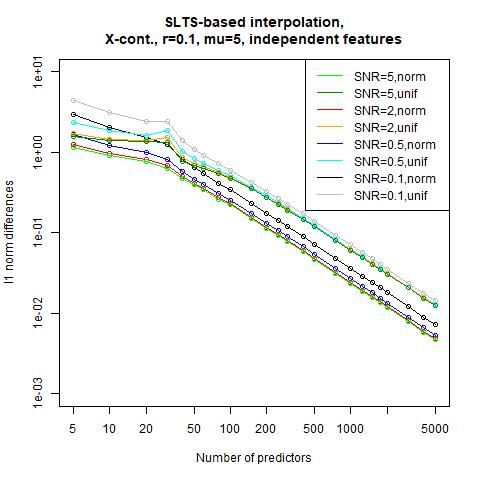} \\ \includegraphics[width=5.25cm]{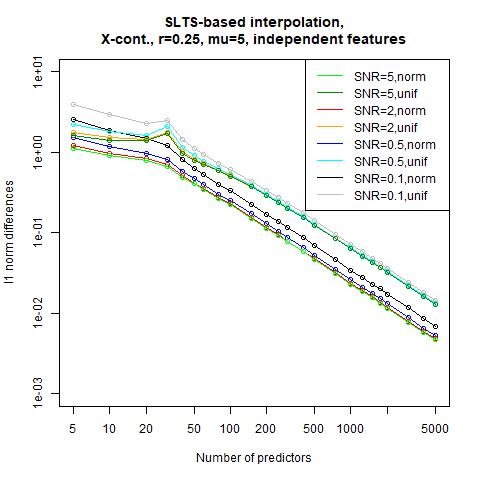} \includegraphics[width=5.25cm]{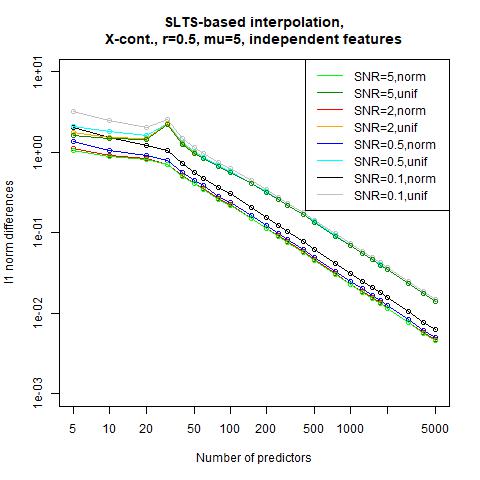} 
\end{center}
\caption{Differences $||\hat \beta-\beta||_1/n$ for the estimated coefficient vector $\hat \beta$ of SLTS-based interpolation when trained on contaminated training data and the true coefficient vector $\beta$.}\label{fig:sltsmu5indepl1cont}
\end{figure}

\begin{figure}[H]
\begin{center}
\includegraphics[width=5.25cm]{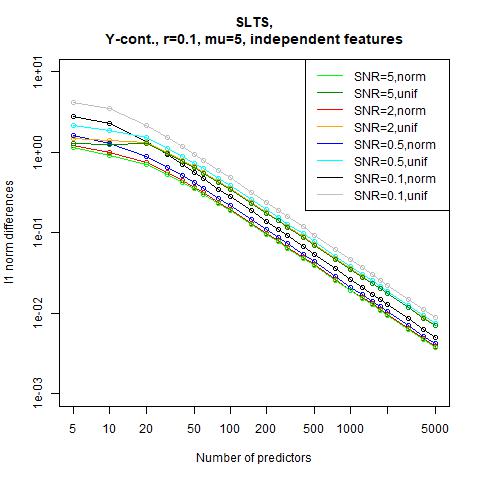} \includegraphics[width=5.25cm]{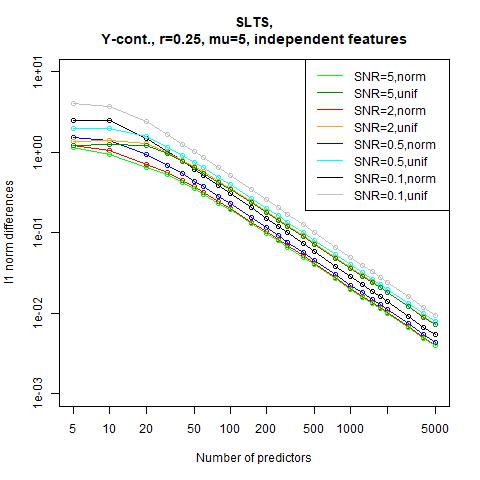} \\ \includegraphics[width=5.25cm]{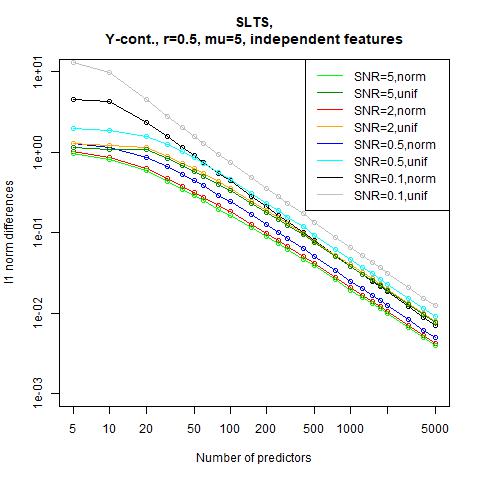} \includegraphics[width=5.25cm]{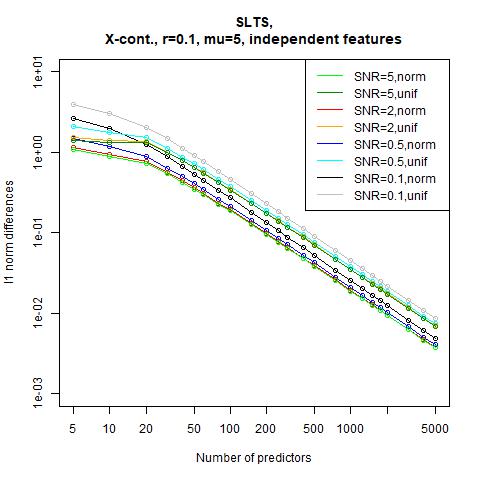} \\ \includegraphics[width=5.25cm]{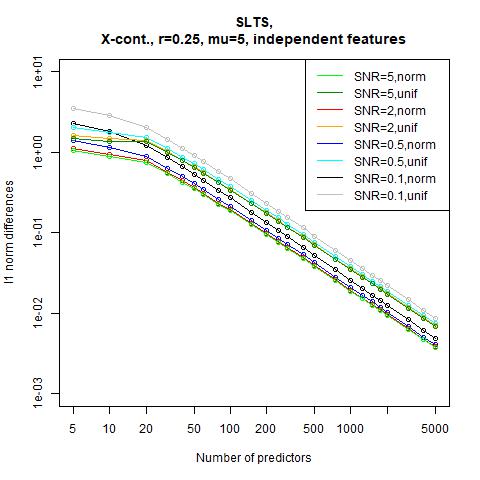} \includegraphics[width=5.25cm]{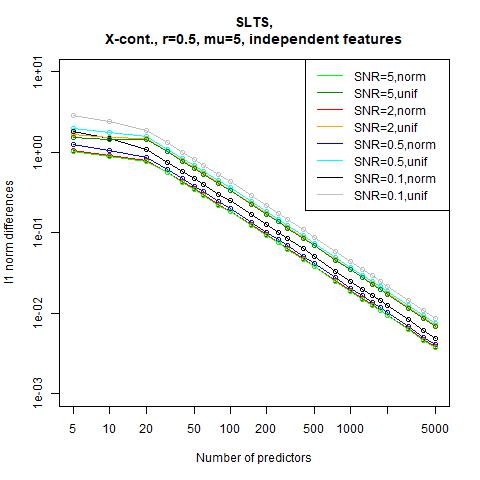} 
\end{center}
\caption{Differences $||\hat \beta-\beta||_1/n$ for the estimated coefficient vector $\hat \beta$ of SLTS when trained on contaminated training data and the true coefficient vector $\beta$.}\label{fig:rawsltsmu5indepl1cont}
\end{figure}

\subsubsection{Boosting-based interpolation}

\begin{figure}[H]
\begin{center}
\includegraphics[width=7.5cm]{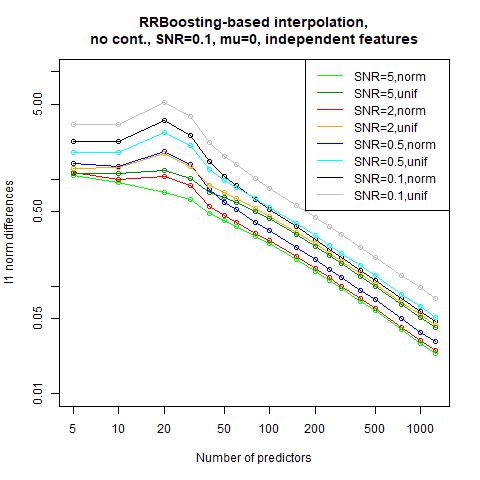} 
\end{center}
\caption{Differences $||\hat \beta-\beta||_1/n$ for the estimated coefficient vector $\hat \beta$ of RRBoosting-based interpolation when trained on clean training data and the true coefficient vector $\beta$.}\label{fig:rrboostmu0indepl1}
\end{figure}

\begin{figure}[H]
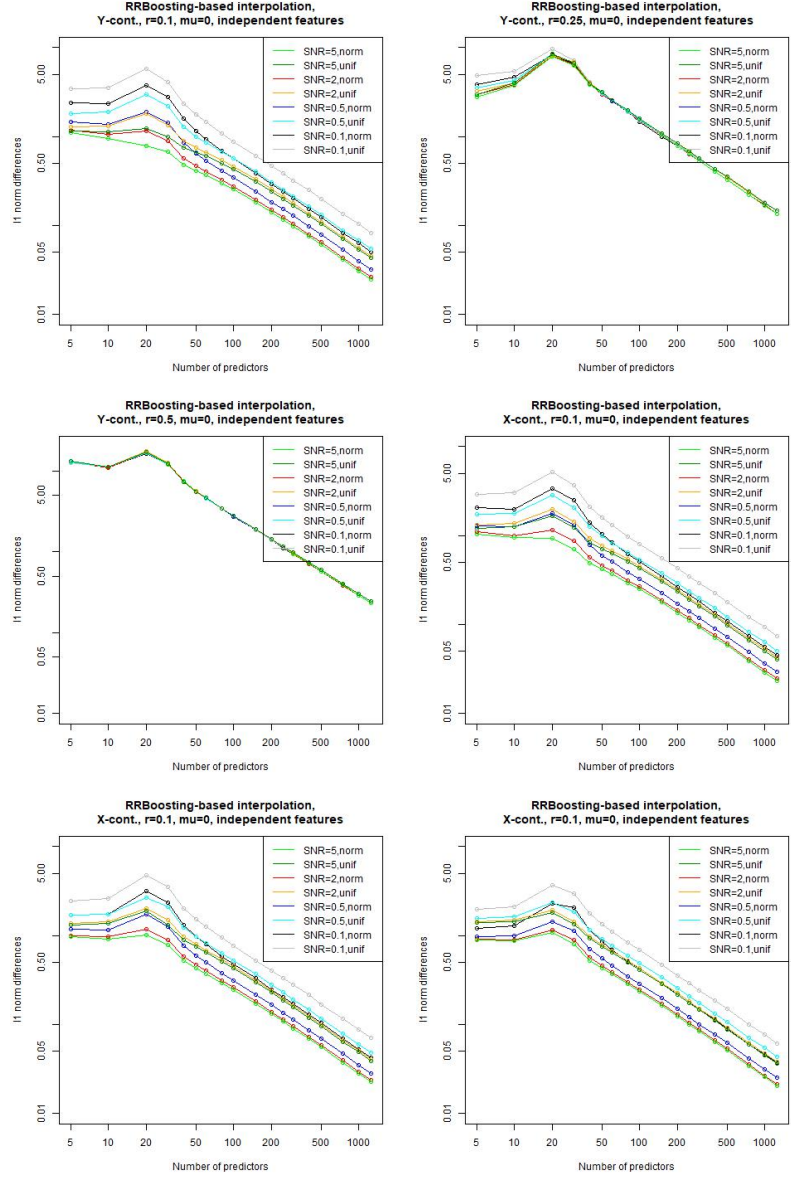

\begin{center}
\includegraphics[width=5.25cm]{DDCurveRRBoostn50mu0r01rad100l1diff.jpeg} \includegraphics[width=5.25cm]{DDCurveRRBoostn50mu0r025rad100l1diff.jpeg} \\ \includegraphics[width=5.25cm]{DDCurveRRBoostn50mu0r05rad100l1diff.jpeg} \includegraphics[width=5.25cm]{DDCurveRRBoostn50mu0Xr01rad100l1diff.jpeg} \\ \includegraphics[width=5.25cm]{DDCurveRRBoostn50mu0Xr025rad100l1diff.jpeg} \includegraphics[width=5.25cm]{DDCurveRRBoostn50mu0Xr05rad100l1diff.jpeg} 
\end{center}
\caption{Differences $||\hat \beta-\beta||_1/n$ for the estimated coefficient vector $\hat \beta$ of RRBoosting-based interpolation when trained on contaminated training data and the true coefficient vector $\beta$.} \label{fig:rrboostmu0indepl1cont}
\end{figure}

Fig. \ref{fig:minl2mu5indepl1}, Fig. \ref{fig:minl2mu5indepl1cont}, Fig. \ref{fig:hubermu5indepl1}, Fig. \ref{fig:hubermu5indepl1cont}, fig. \ref{fig:sltsmu5indepl1}, Fig. \ref{fig:sltsmu5indepl1cont}, Fig. \ref{fig:rawsltsmu5indepl1}, Fig. \ref{fig:rawsltsmu5indepl1cont}, Fig. \ref{fig:rrboostmu0indepl1} and Fig. \ref{fig:rrboostmu0indepl1cont} reveal that the coefficient difference curves resemble those from the case $\mu=0$.







\newpage

\subsection{$n=200$} \

\subsubsection{Minimum $l_2$-norm interpolation}

\begin{figure}[H]
\begin{center}
\includegraphics[width=5.25cm]{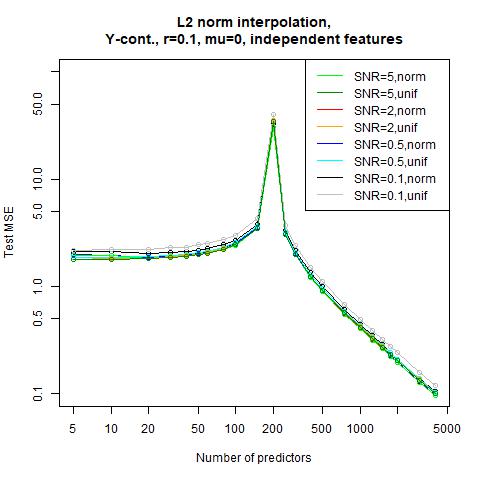} 
\includegraphics[width=5.25cm]{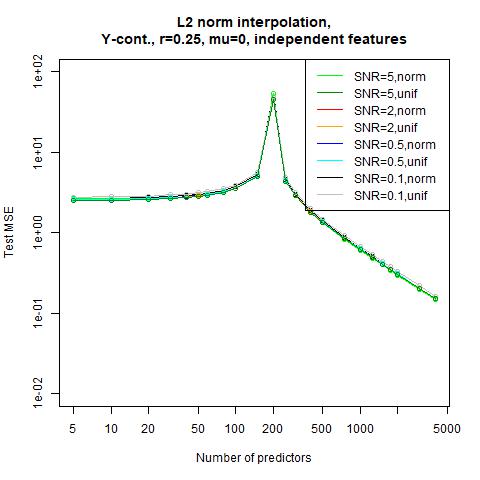} \\
 \includegraphics[width=5.25cm]{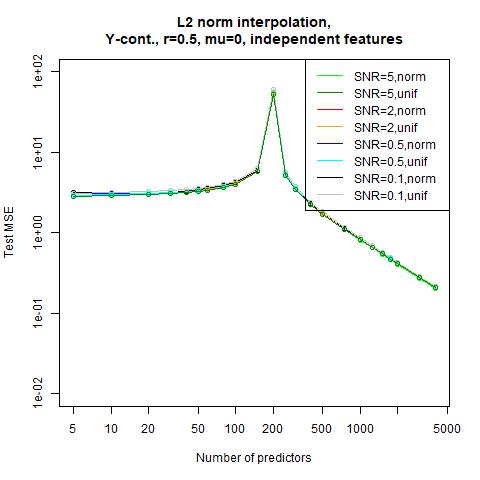} 
\includegraphics[width=5.25cm]{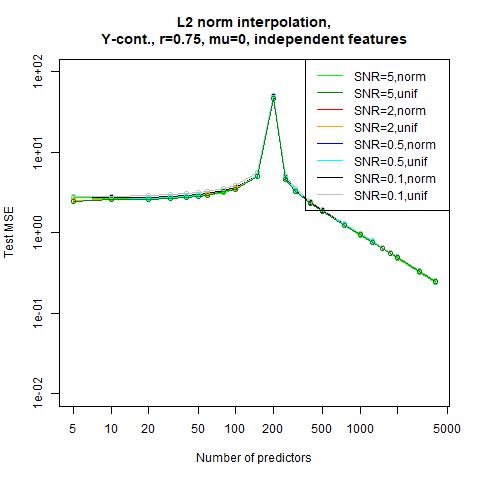}\\
 \includegraphics[width=5.25cm]{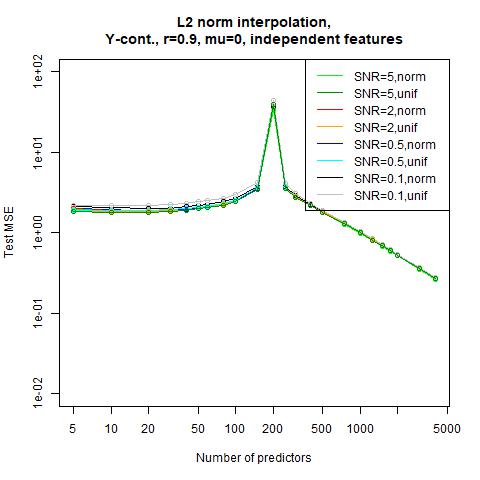} 
\end{center}
\caption{Differences $||\hat \beta-\beta||_1/n$ for the estimated coefficient vector $\hat \beta$ of minimum $l_2$-norm interpolation when trained on $Y$-contaminated training data.}\label{fig:minl2mu0indepcontn200l1}
\end{figure}

\subsubsection{Huber-loss interpolation}

\begin{figure}[H]
\begin{center}
\includegraphics[width=5.25cm]{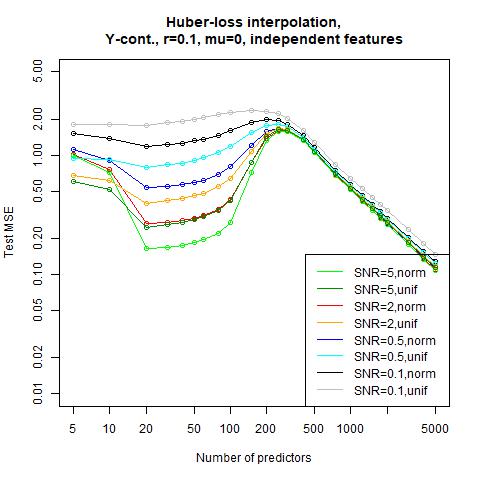} 
\includegraphics[width=5.25cm]{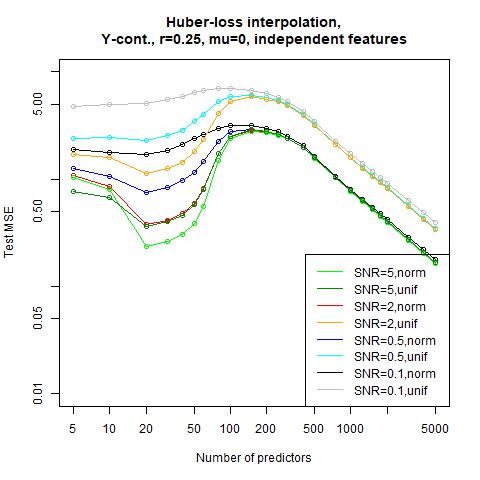} \\
 \includegraphics[width=5.25cm]{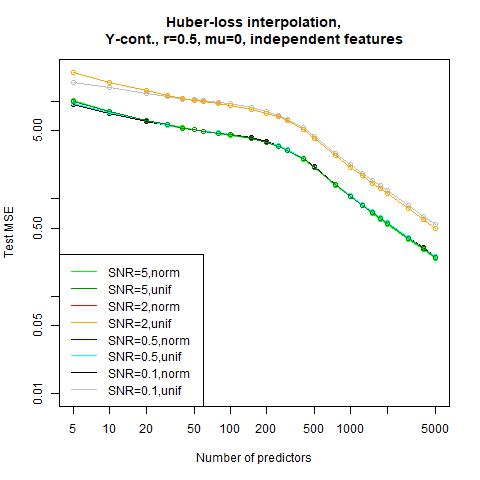} 
\includegraphics[width=5.25cm]{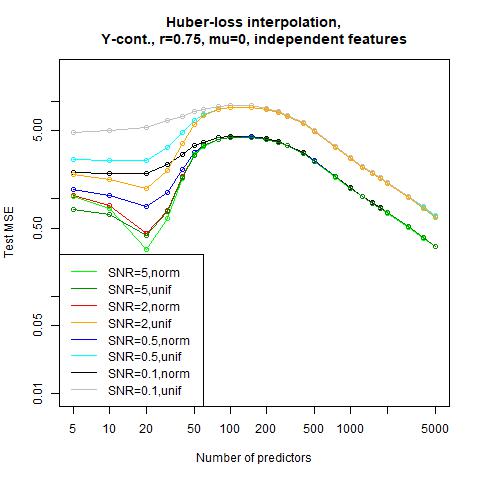}\\
 \includegraphics[width=5.25cm]{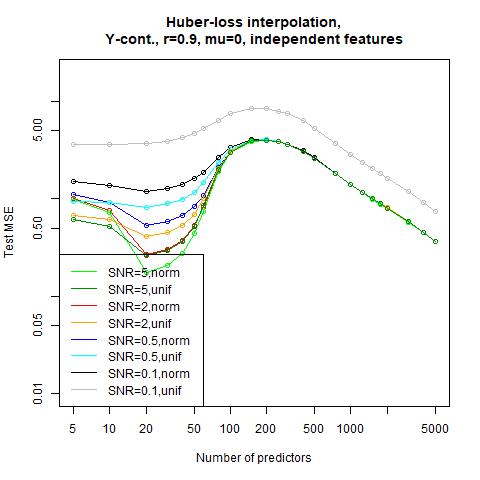} 
\end{center}
\caption{Differences $||\hat \beta-\beta||_1/n$ for the estimated coefficient vector $\hat \beta$ of Huber-loss interpolation when trained on $Y$-contaminated training data.}\label{fig:hubermu0indepcontn200l1}
\end{figure}

The curves in Fig. \ref{fig:minl2mu0indepcontn200l1} and Fig.  \ref{fig:hubermu0indepcontn200l1} resemble those from the case $n=50$, of course, the peak is attained later.

\newpage 

\subsection{$n=200$, $c_{out}=10000$} \

\subsubsection{Minimum $l_2$-norm interpolation}

\begin{figure}[H]
\begin{center}
\includegraphics[width=5.25cm]{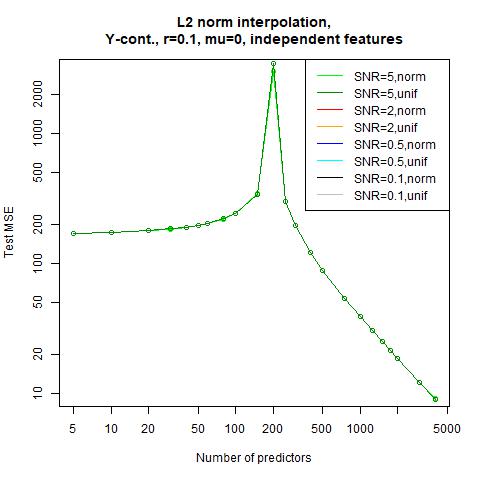} 
\includegraphics[width=5.25cm]{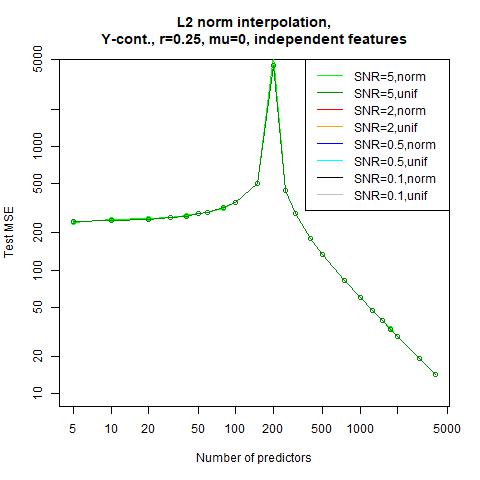} \\
 \includegraphics[width=5.25cm]{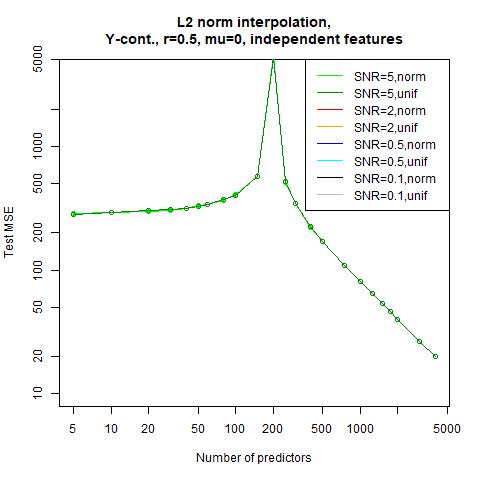} 
\includegraphics[width=5.25cm]{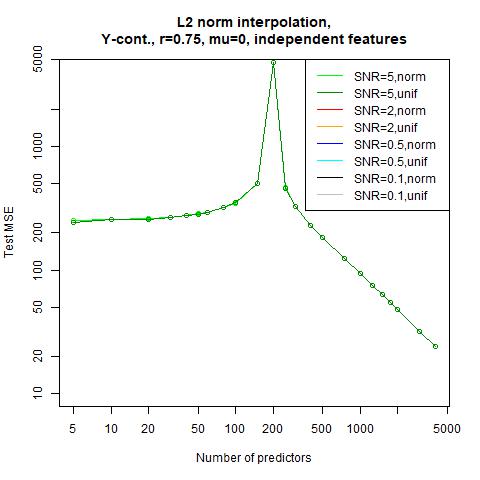}\\
 \includegraphics[width=5.25cm]{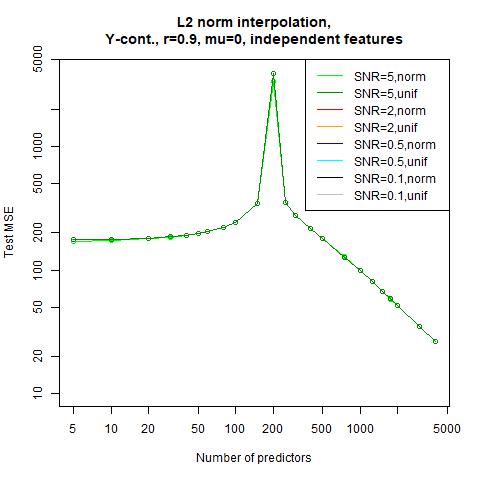} 
\end{center}
\caption{Differences $||\hat \beta-\beta||_1/n$ for the estimated coefficient vector $\hat \beta$ of minimum $l_2$-norm interpolation when trained on $Y$-contaminated training data.}\label{fig:minl2mu0indepcontn200r10000l1}
\end{figure}

The curves in Fig. \ref{fig:minl2mu0indepcontn200r10000l1} resemble those from the case $c_{out}=100$ in Fig. \ref{fig:minl2mu0indepcontn200l1}, but with much larger values.

\subsubsection{Huber-loss interpolation}

\begin{figure}[H]
\begin{center}
\includegraphics[width=5.25cm]{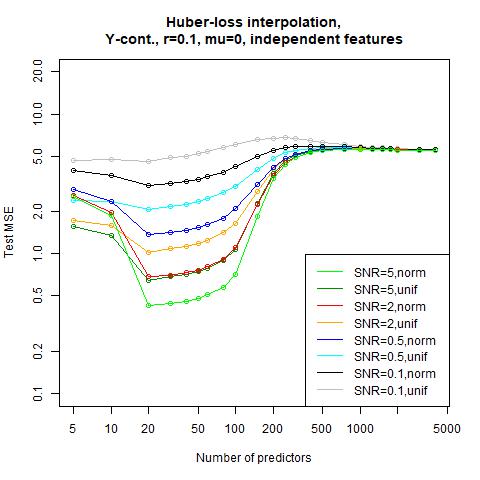} 
\includegraphics[width=5.25cm]{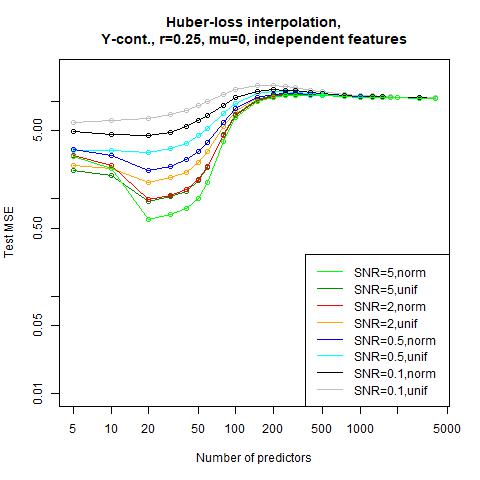} \\
 \includegraphics[width=5.25cm]{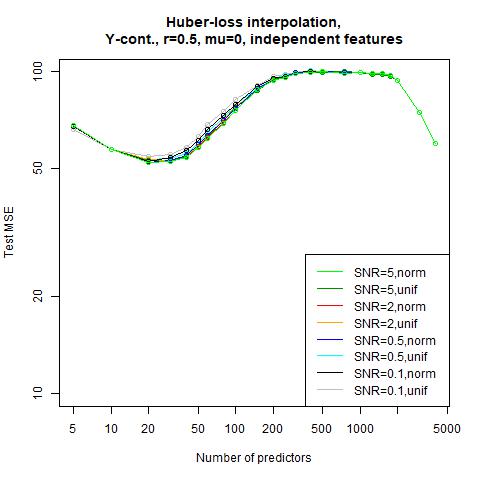} 
\includegraphics[width=5.25cm]{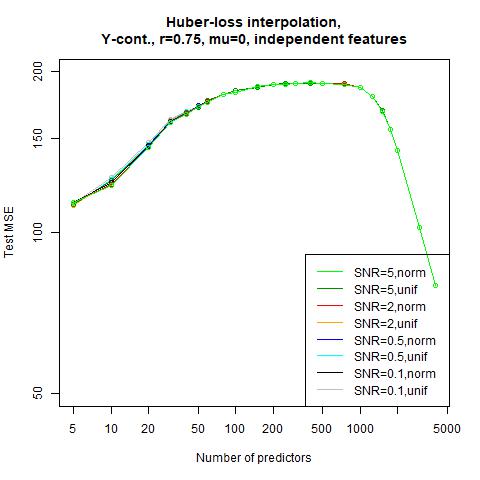}\\
 \includegraphics[width=5.25cm]{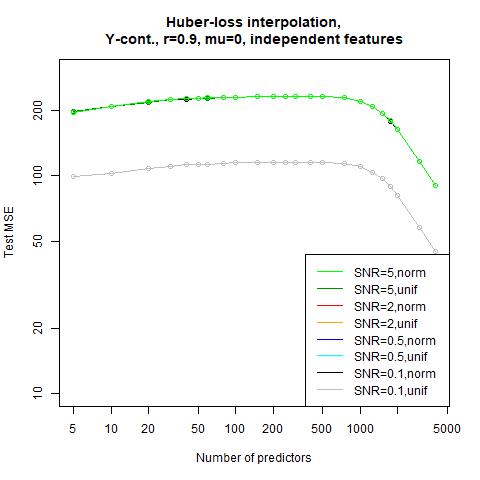} 
\end{center}
\caption{Differences $||\hat \beta-\beta||_1/n$ for the estimated coefficient vector $\hat \beta$ of Huber-loss interpolation when trained on $Y$-contaminated training data.}\label{fig:hubermu0indepcontn200r10000l1}
\end{figure}

For $r \in \{0.1,0.25\}$, the curves in Fig. \ref{fig:hubermu0indepcontn200r10000l1} first decrease, as in Fig. \ref{fig:hubermu0indepcontn200l1} for the case $c_{out}=100$, but stay constant for large $p$ instead of decreasing. For larger $r$, the curves decrease at larger $p$.

\section{Number of iterations} \label{sec:iter}

\subsection{Independent features, $\mu=0$}

\begin{figure}[H]
\begin{center}
\includegraphics[width=7.5cm]{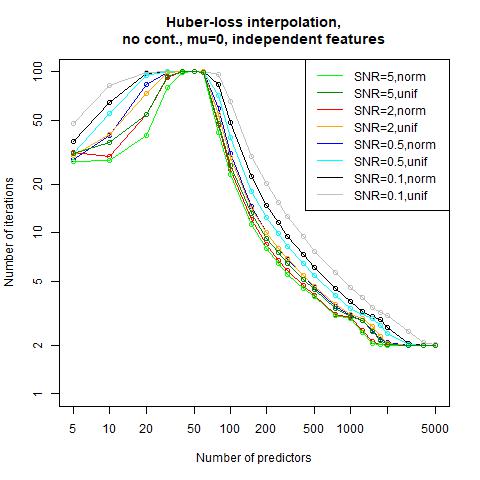} 
\end{center}
\caption{Mean number of iterations of Huber-loss interpolation when trained on clean training data.}\label{fig:hubermu0indepIter}
\end{figure}

\begin{figure}[H]
\begin{center}
\includegraphics[width=7.5cm]{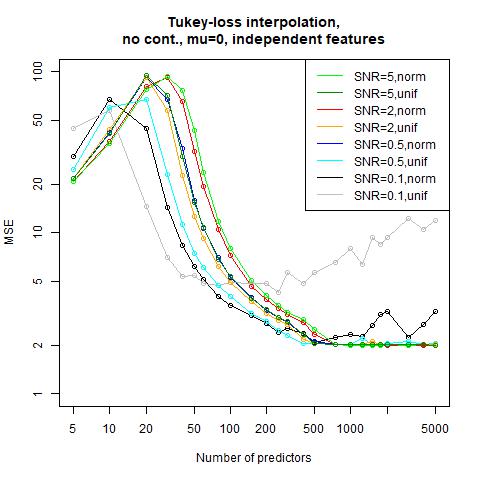} 
\end{center}
\caption{Mean number of iterations of Tukey-loss interpolation when trained on clean training data.}\label{fig:tukeymu0indepIter}
\end{figure}

\begin{figure}[H]
\begin{center}
\includegraphics[width=5.25cm]{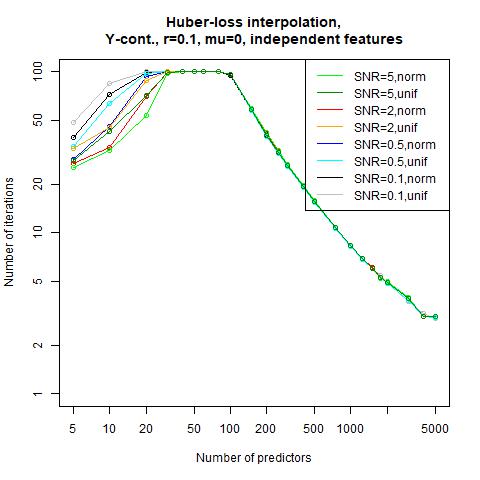} \includegraphics[width=5.25cm]{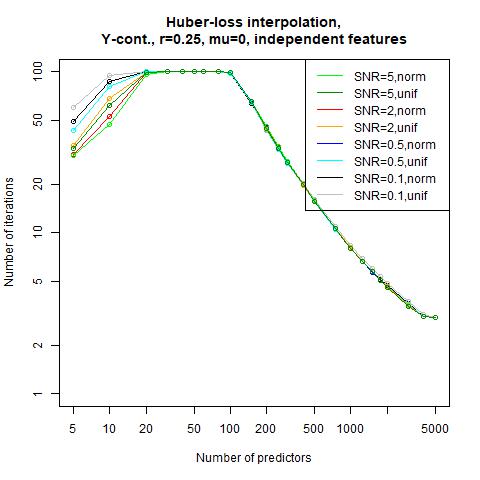} \\ \includegraphics[width=5.25cm]{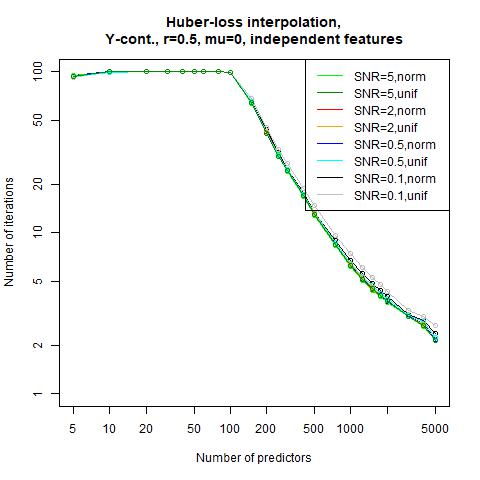} \includegraphics[width=5.25cm]{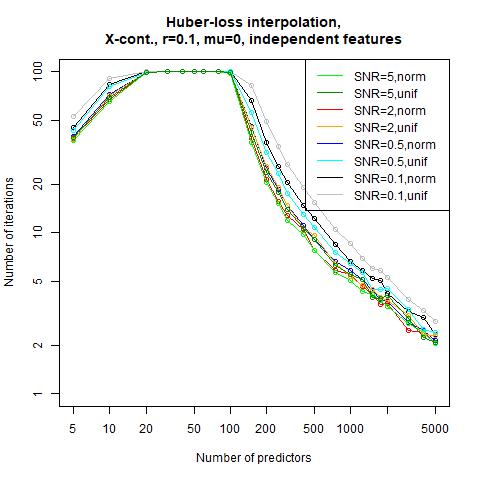} \\ \includegraphics[width=5.25cm]{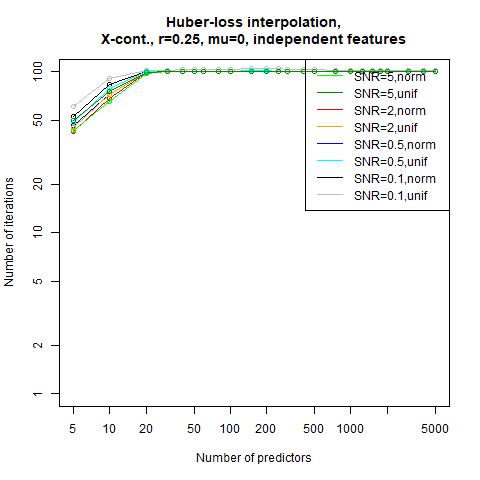} \includegraphics[width=5.25cm]{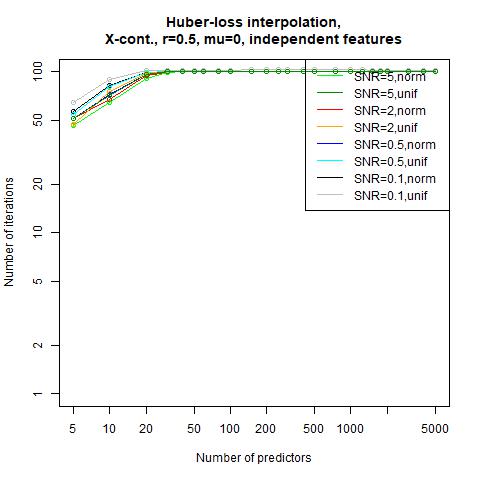} 
\end{center}
\caption{Mean number of iterations of Huber-loss interpolation when trained on contaminated training data.}\label{fig:hubermu0indepItercont}
\end{figure}

For Huber-loss based interpolation, as one can observe in Fig. \ref{fig:hubermu0indepIter} and Fig. \ref{fig:hubermu0indepItercont}, the number of iterations reaches the allowed maximum of 100 if $p$ is in the vicinity of $n$. This plateau is larger for higher $r$. As for $Y$-contamination, the number of iterations considerably decreases for growing $p$, reaching numbers below 10 eventually for $p=5000$. In the case of $X$-contamination, the number of iterations only decreases for $r=0.1$, while for larger contamination radii, the number remains at the maximum of 100.

\begin{figure}[H]
\begin{center}
\includegraphics[width=5.25cm]{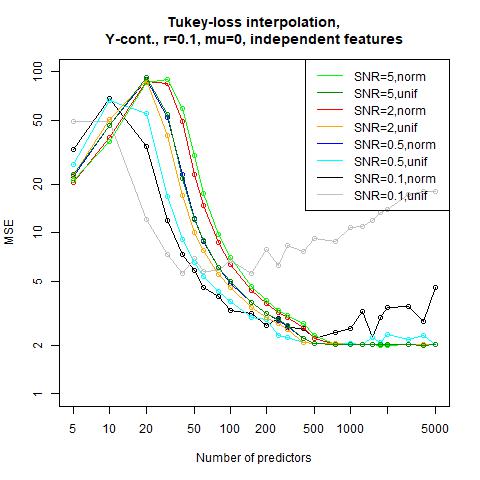} \includegraphics[width=5.25cm]{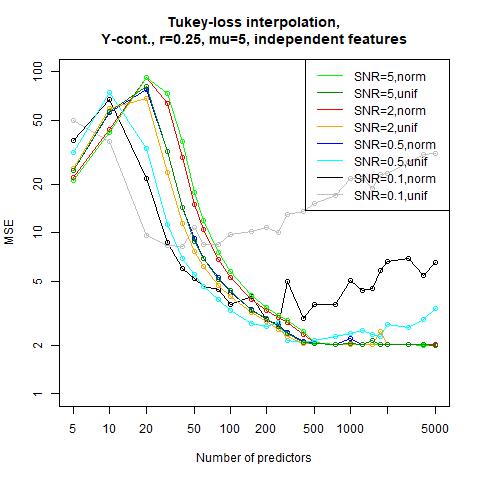} \\ \includegraphics[width=5.25cm]{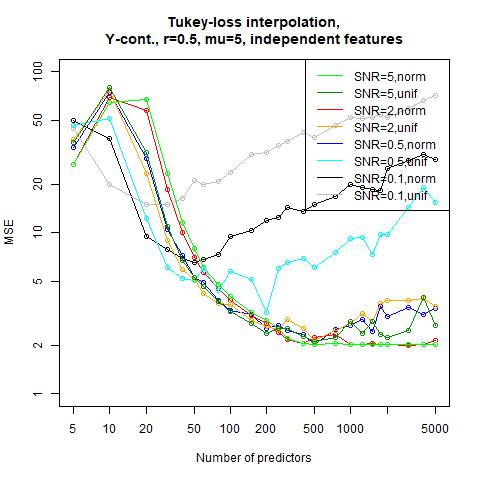} \includegraphics[width=5.25cm]{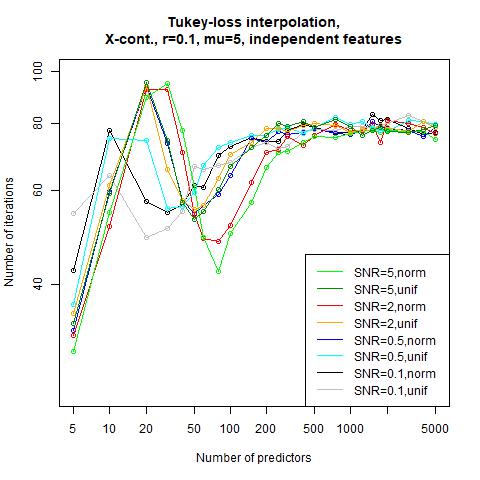} \\ \includegraphics[width=5.25cm]{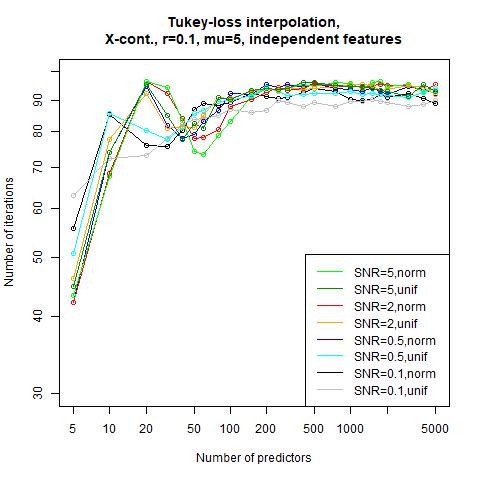} \includegraphics[width=5.25cm]{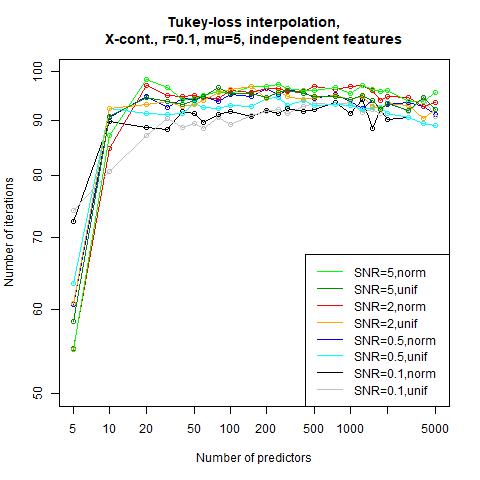} 
\end{center}
\caption{Mean number of iterations of Tukey-loss interpolation when trained on contaminated training data.}\label{fig:tukeymu0indepItercont}
\end{figure}

For Tukey-based interpolation, Fig. \ref{fig:tukeymu0indepIter} reveals that the number of iterations decreases after a peak, but for an SNR of $0.1$, it increases again for large $p$. This behaviour can also be observed for $Y$-contamination, as visualized in Fig. \ref{fig:tukeymu0indepItercont}. For $X$-contamination, the number of iterations increases and stays nearly the maximum number of iterations of 100.

\subsection{Independent features, $\mu=5$}

\begin{figure}[H]
\begin{center}
\includegraphics[width=7.5cm]{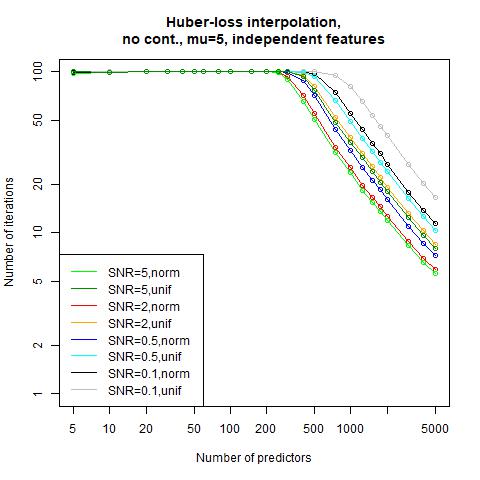} 
\end{center}
\caption{Mean number of iterations of Huber-loss interpolation when trained on clean training data.}\label{fig:hubermu0indepIter}
\end{figure}

\begin{figure}[H]
\begin{center}
\includegraphics[width=5.25cm]{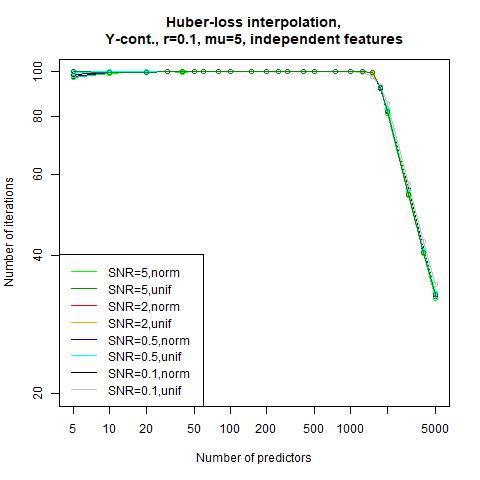} \includegraphics[width=5.25cm]{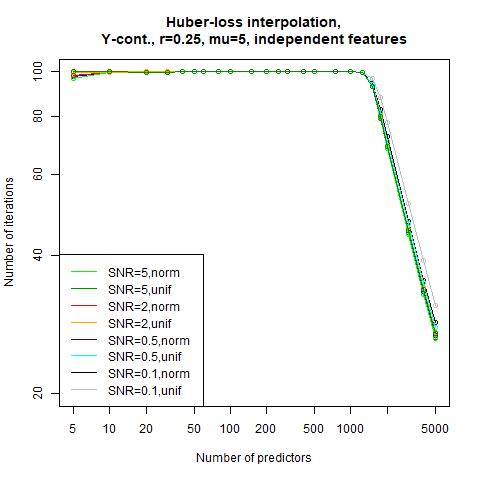} \\ \includegraphics[width=5.25cm]{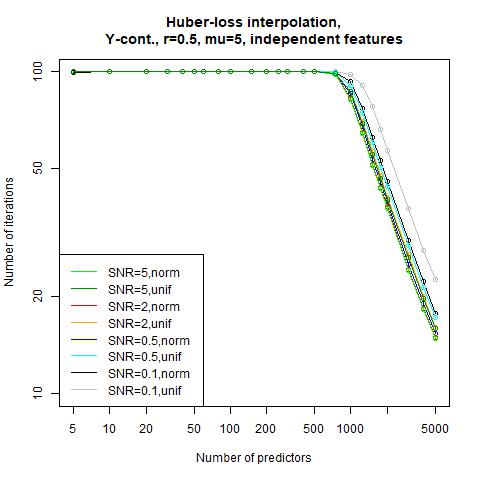} \includegraphics[width=5.25cm]{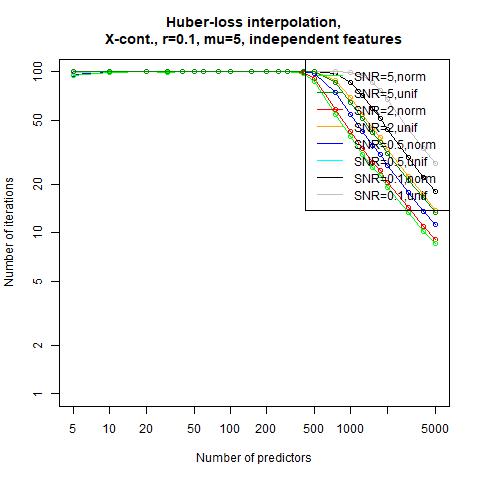} \\ \includegraphics[width=5.25cm]{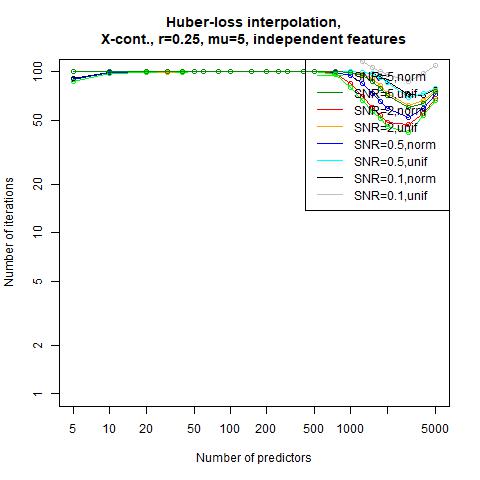} \includegraphics[width=5.25cm]{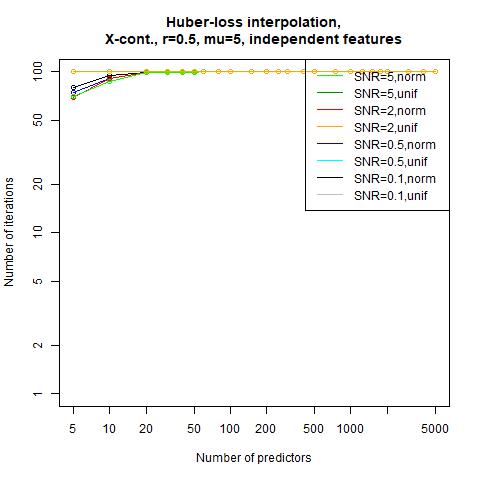} 
\end{center}
\caption{Mean number of iterations of Huber-loss interpolation when trained on contaminated training data.}\label{fig:hubermu0indepItercont}
\end{figure}

In contrast to the case $\mu=0$, the number of iterations stays much longer in the plateau and decreases for large $p$, as shown in Fig. \ref{fig:hubermu0indepIter} and Fig. \ref{fig:hubermu0indepItercont}. For $r=0.75$, it however increases again as $p$ grows larger. For $r=0.9$, the number of iterations stays at its maximum even for very large $p$.

\subsection{Spiked covariance design, $\mu=0$}

\begin{figure}[H]
\begin{center}
\includegraphics[width=7.5cm]{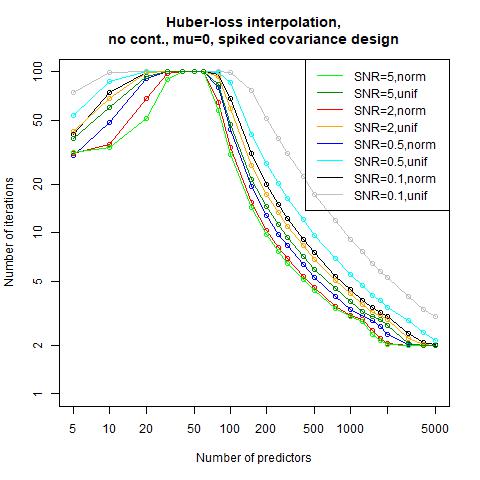} 
\caption{Mean number of iterations of Huber-loss interpolation when trained on clean training data.}\label{fig:hubermu0spikedIter}
\end{center}
\end{figure}

\begin{figure}[H]
\begin{center}
\includegraphics[width=7.5cm]{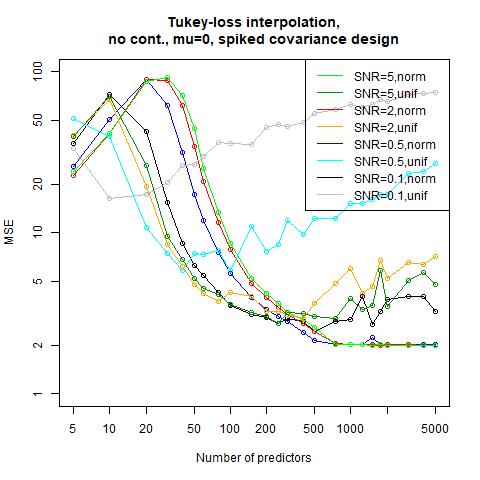} 
\end{center}
\caption{Mean number of iterations of Tukey-loss interpolation when trained on clean training data.}\label{fig:tukeymu0spikedIter}
\end{figure}

\begin{figure}[H]
\begin{center}
\includegraphics[width=5.25cm]{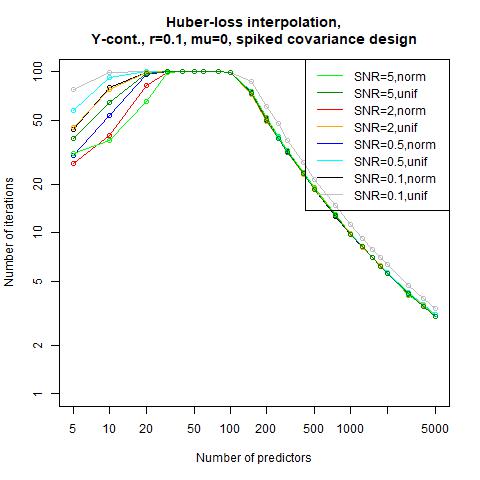} \includegraphics[width=5.25cm]{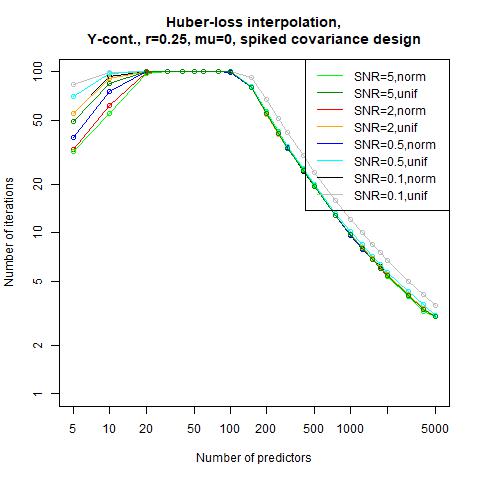} \\ \includegraphics[width=5.25cm]{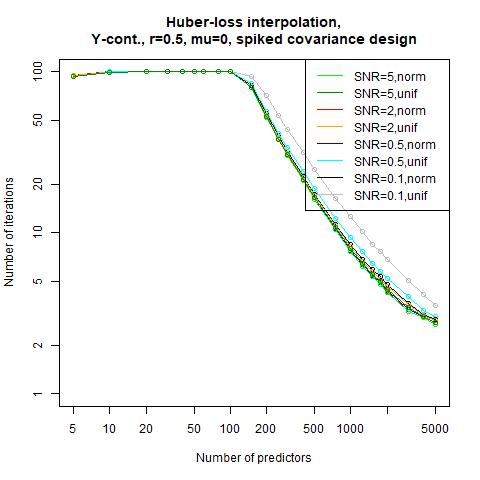} \includegraphics[width=5.25cm]{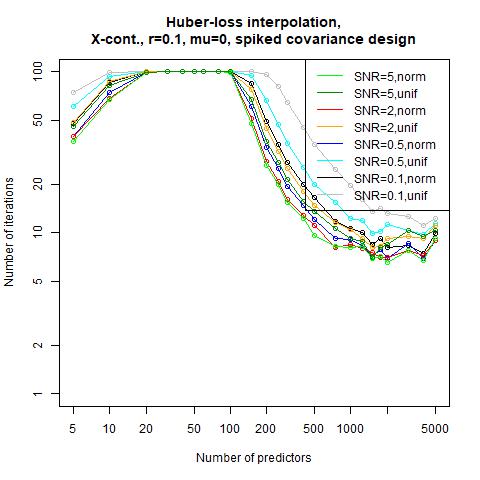} \\ \includegraphics[width=5.25cm]{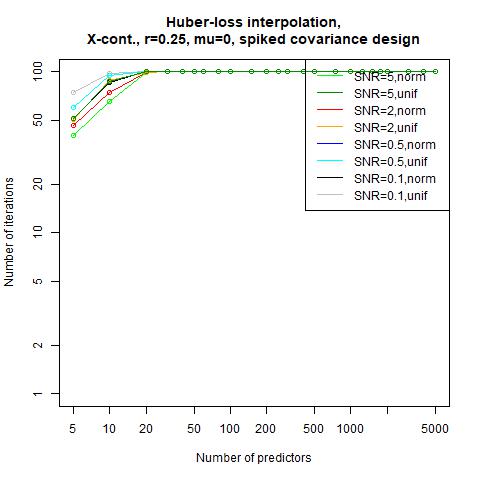} \includegraphics[width=5.25cm]{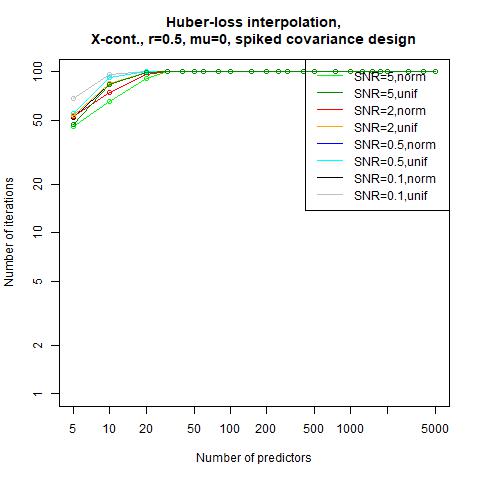} 
\end{center}
\caption{Mean number of iterations of Huber-loss interpolation when trained on contaminated training data.}\label{fig:hubermu0spikedItercont}
\end{figure}

\begin{figure}[H]
\begin{center}
\includegraphics[width=5.25cm]{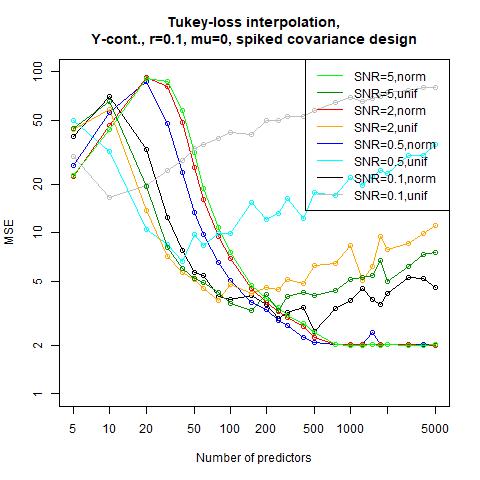} \includegraphics[width=5.25cm]{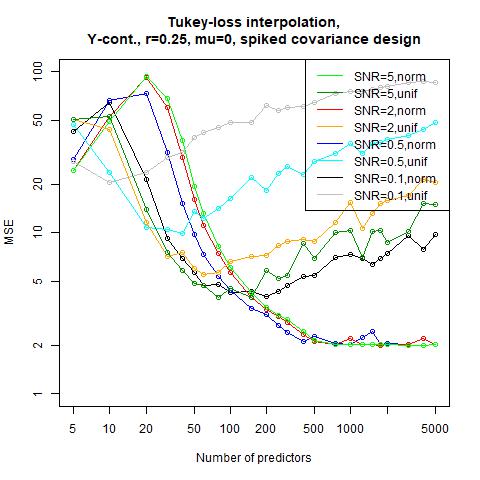} \\ \includegraphics[width=5.25cm]{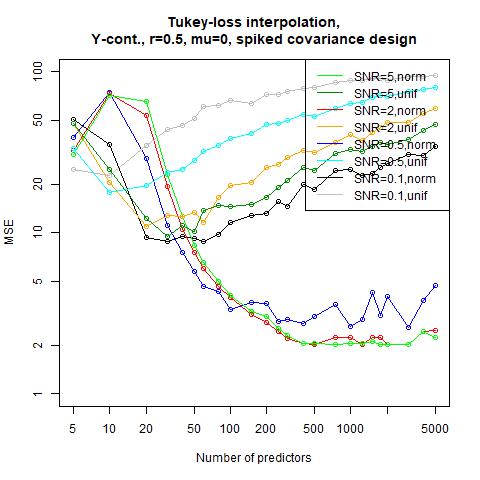} \includegraphics[width=5.25cm]{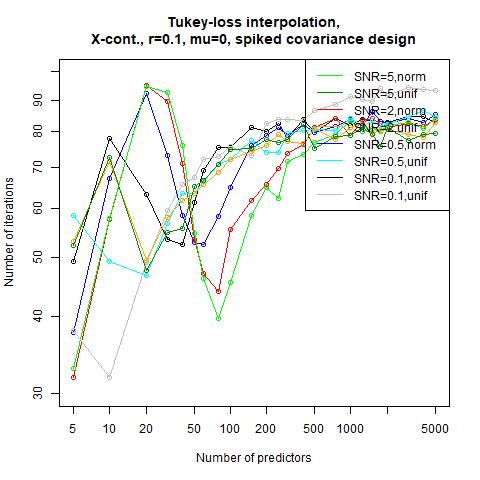} \\ \includegraphics[width=5.25cm]{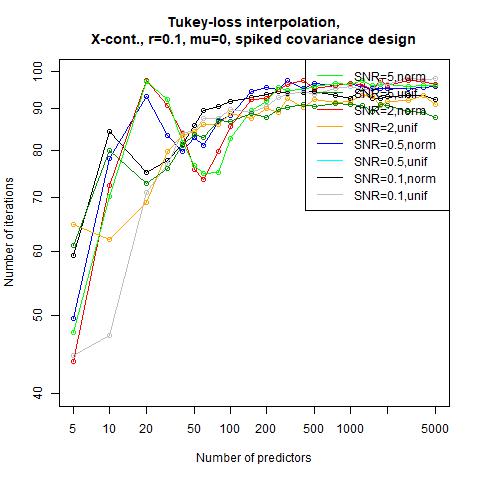} \includegraphics[width=5.25cm]{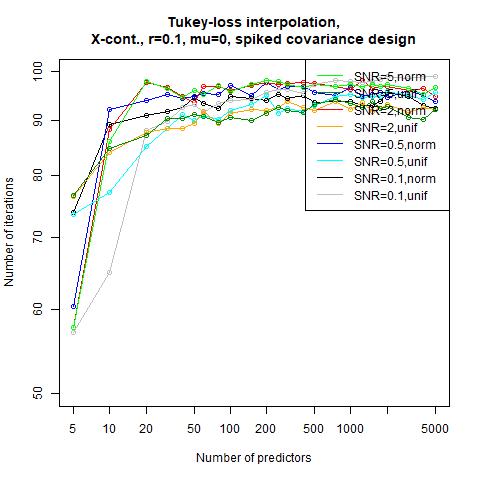} 
\end{center}
\caption{Mean number of iterations of Tukey-loss interpolation when trained on contaminated training data.}\label{fig:tukeymu0spikedItercont}
\end{figure}

The curves, depicted in Fig. \ref{fig:hubermu0spikedIter}, Fig. \ref{fig:tukeymu0spikedIter}, Fig. \ref{fig:hubermu0spikedItercont} and Fig. \ref{fig:tukeymu0spikedItercont} resemble those from the independent design.

\newpage

\subsection{$n=200$} \

\subsubsection{Huber-loss interpolation}

\begin{figure}[H]
\begin{center}
\includegraphics[width=5.25cm]{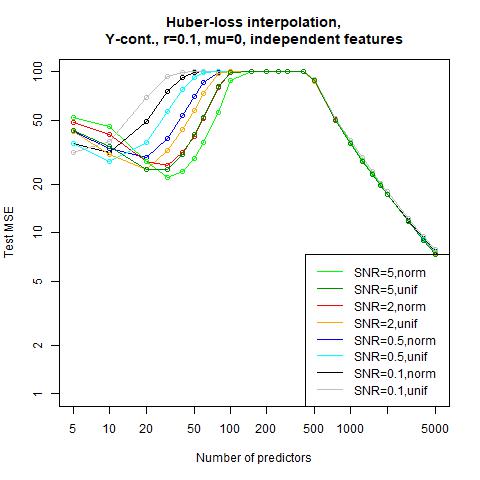} 
\includegraphics[width=5.25cm]{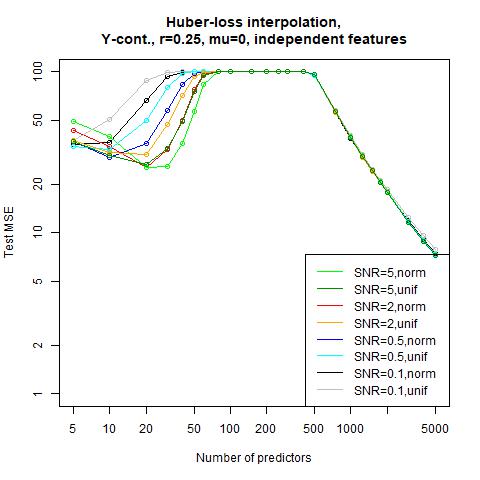} \\
 \includegraphics[width=5.25cm]{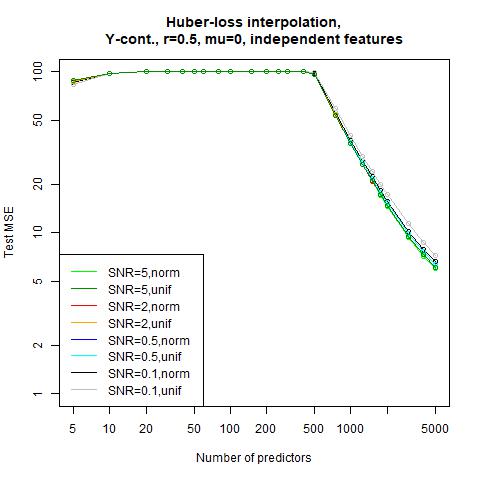} 
\includegraphics[width=5.25cm]{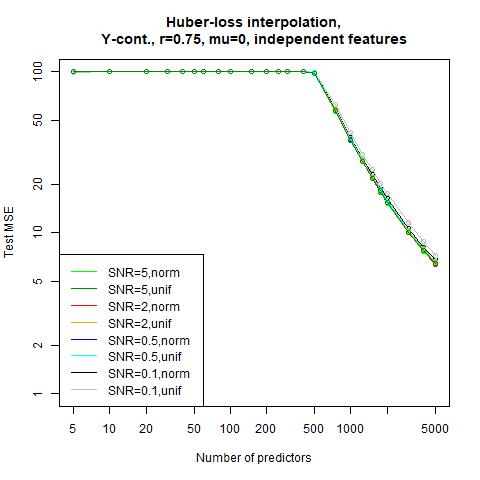}\\
 \includegraphics[width=5.25cm]{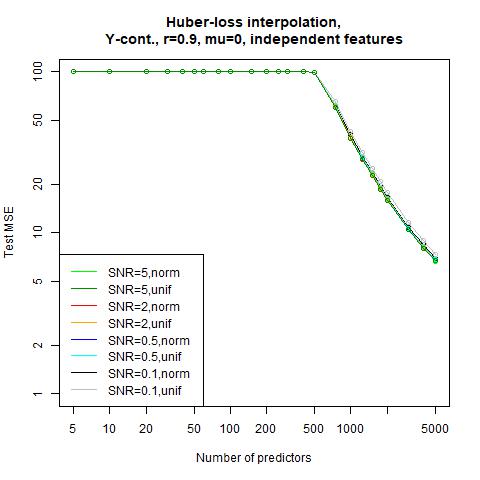} 
\end{center}
\caption{Mean number of iterations of Huber-loss interpolation when trained on $Y$-contaminated training data.}\label{fig:hubermu0indepcontn200iter}
\end{figure}

\newpage

\subsection{$n=200$, $c_{out}=10000$} \

\subsubsection{Huber-loss interpolation}

\begin{figure}[H]
\begin{center}
\includegraphics[width=5.25cm]{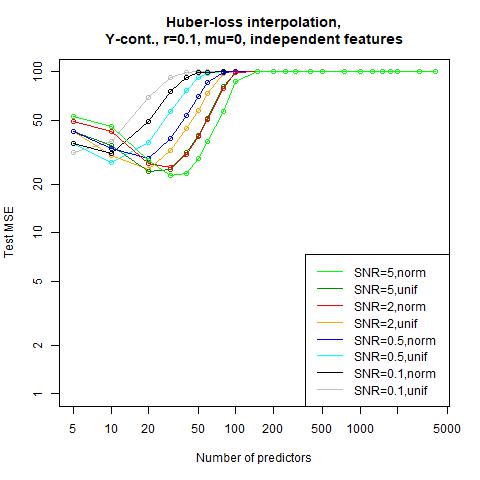} 
\includegraphics[width=5.25cm]{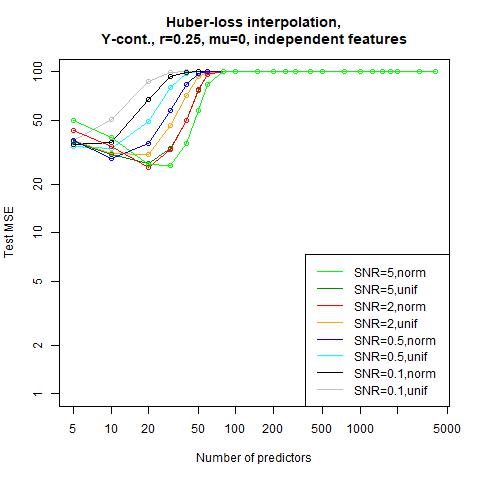} \\
 \includegraphics[width=5.25cm]{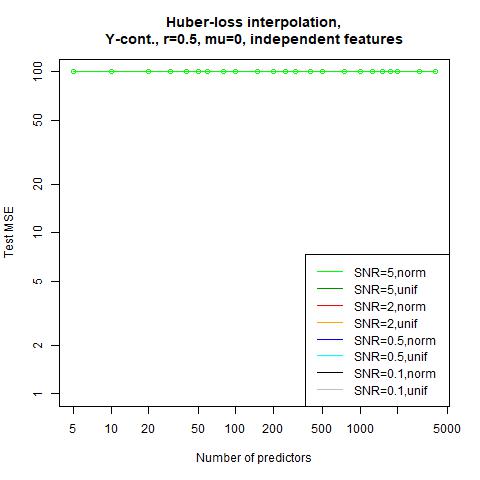} 
\includegraphics[width=5.25cm]{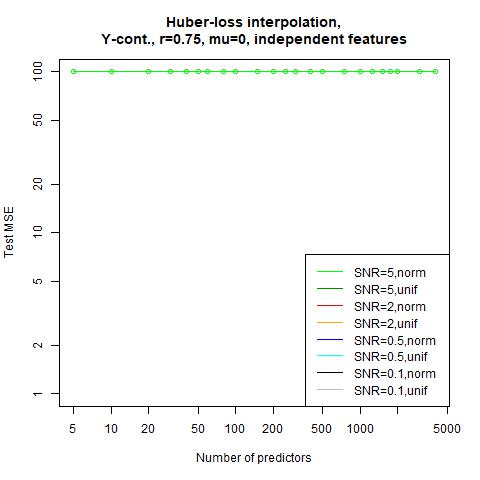}\\
 \includegraphics[width=5.25cm]{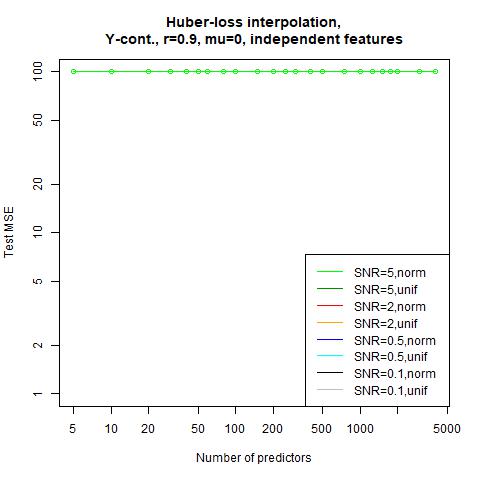} 
\end{center}
\caption{Mean number of iterations of Huber-loss interpolation when trained on $Y$-contaminated training data.}\label{fig:hubermu0indepcontn200r10000iter}
\end{figure}

\section{Discussion and conclusion} \label{sec:concl}

\subsection{Discussion of the results}

The evaluation of the test MSEs in Sec. \ref{sec:testerr} reveals that the minimum $l_2$-norm interpolator indeed shows the double descent behavior, as the MSE drops after the peak at $p=n$, provided a sufficiently high SNR (at least 2 in the experiments). Although, as expected, the MSE corresponding to models trained on contaminated data is higher than for models trained on clean data, it can be observed that the MSE attains smaller values for large $p$ than for low $p$, indicating that the interpolating regime has been reached. One should note that on $Y$-contaminated data, there is no double descent, since there is no descent before the peak but only a descent after the peak. $X$-contamination in contrast seems to only marginally affect the performance of the minimum $l_2$-norm interpolator. Huber-loss interpolation also leads to a double descent on clean data, $X$-contaminated and $Y$-contaminated data with small contamination radius, provided a sufficiently large SNR, but the minimmu $l_2$-norm interpolator surpasses it for large $p$. A similar behavior can be observed for RRBoosting-based interpolation. In contrast, Tukey-loss interpolation and SLTS-based interpolation as well as SLTS and RRBoost do not allow for the double descent phenomenon in our experiments, and even for $Y$-contaminated or clean data, the MSE not necessarily decreases. 

The behavior is nearly unaffected by the underlying covariance structure, i.e., whether the predictors are independent or whether they are distributed according to a spiked covariance scheme. However, the performance of Huber-based interpolation degrades once the predictors are not centered, and once the contamination magnitude $c_{out}$ becomes large, while the shape of the generalization error corresponding to minimum $l_2$-norm interpolation remains unaffected. 

As for the correspondence of the training (Sec. \ref{sec:trainerr}) and test errors, one can observe that for Huber-loss interpolation, the (second) descent starts roughly once the training error vanishes. This could be explained by the fact that for small absolute residuals, the Huber loss equals the squared loss, so that Huber-loss interpolation coincides with minimum $l_2$-norm interpolation in this case. For $\mu=5$, i.e., non-centered predictors, one can only observe that the test MSE no longer increases once the training MSE vanishes, but the interpolation does not corresponds to a decrease of the test MSE here. 

There seems to be no correspondence between the shape of the curves depicting the differences $||\beta-\hat \beta||_1$ (Sec. \ref{sec:l1diff}), as they always decrease for growing $p$.  We also evaluated the differences $||\beta-\hat \beta||_2$ and $||\beta-\hat \beta||_{\infty}$, but their shape is similar.

The number of iterations for Huber- and Tukey-loss interpolation does not seem to correspond to the generalization errors (Sec. \ref{sec:iter}). For $Y$-contamination and centered predictors, the number of iterations for both algorithms drops quickly after a peak around $p=n$, but the test MSE behaves completely differently for Huber- and Tukey-loss interpolation. Moreover, although the test MSE decreases for Huber-loss interpolation for large $p$, the number of iterations remains at its maximum. For non-centered predictors, the number of iteration eventually decreases for very large $p$, which coincides with the starting point where the test MSE decreases for $Y$-contamination and $r\in \{0.1,0.25\}$, but it does neither correspond to the test MSE for $r=0.5$ nor for $X$-contamination.

\subsection{Conclusion and outlook}

In this work, we experimentally studied overparametrized regression on contaminated data. We compared the performance of the minimum $l_2$-norm interpolator, Huber-loss and Tukey-loss interpolation as well as SLTS and RRBoost. We also proposed an interpolation variant on clean subsets, where first SLTS or RRBoost is applied in order to identify a clean subset, on which the minimum $l_2$-norm interpolator is computed. The contamination also includes gross outliers and considers both $X$- and $Y$-contamination.

The results reveal a surprising robustness of the minimum $l_2$-norm interpolator, whose generalization performance that of any robust counterpart, disregarding the covariance structure of the predictors, whether the predictors are centered, the contamination radius or the contamination magnitude. In particular, provided that the SNR is sufficiently large, a double descent phenomenon can be observed, where the test MSE first decreases until it attains its minimum at $p=s$, increases until a peak at $p=n$, and decreases again for $p>n$. For small SNR, there is no minimum at $p=s$, but the test MSE decreases as well for $p>n$. For centered predictors and moderate contamination magnitudes, the Huber-loss interpolator leads to similar test MSE shapes, however, it decreases later than for the minimum $l_2$-norm interpolator. 

It should be a topic for future work to assess whether the theoretical results for the double descent behavior of the minimum $l_2$-norm interpolator can be extended to contaminated data, which, to the best of our knowledge, have not yet been provided. In particular, proving a double descent for the minimum $l_2$-norm interpolator would be important for practical applications where the data can be assumed to be contaminated. For $p>>n$, it would imply large computational advantages to apply minimum $l_2$-norm interpolation instead of a \enquote{robust} counterpart, which, due to a non-convex objective, would necessitate an iterative optimization scheme.

\section{Acknowledgements}

The simulations were conducted on the HPC cluster ROSA, located at the University of Oldenburg (Germany). ROSA was funded by the German Research Foundation (DFG) through its Major Research Instrumentation Programme (INST 184/225-1 FUGG) and the Ministry of Science and Culture (MWK) of Lower Saxony.

\bibliography{Biblio}
\bibliographystyle{abbrvnat}
\setcitestyle{authoryear,open={((},close={))}}

\end{document}